\documentclass [a4paper, hidelinks, 12pt]{book}
\usepackage[applemac]{inputenc}
\usepackage[english]{babel}
\usepackage{graphicx}
\usepackage{bookmark,hyperref}

\usepackage{afterpage}
\usepackage{natbib}

\usepackage{footmisc}
\usepackage{algpseudocode}
\usepackage{algorithm}
\usepackage{multicol}
\usepackage{ragged2e}
\usepackage{qtree}
\usepackage{enumitem} 
\usepackage{color}
\usepackage{lmodern}

\usepackage{times}
\usepackage{url}
\usepackage{latexsym}
\usepackage{subcaption}

\usepackage[T1]{fontenc}

\usepackage{color, colortbl}
\definecolor{Gray}{gray}{0.8}

\usepackage{tabularx}

\usepackage{comment}
\usepackage{booktabs}
\usepackage{todonotes} 

\usepackage{amssymb}
\usepackage{pifont}
\usepackage{makecell}
\usepackage{caption}
\usepackage{subcaption}
\usepackage{scrextend}

\newcommand{\cmark}{\ding{51}}
\newcommand{\xmark}{\ding{55}}

\newcommand*{\affaddr}[1]{#1} 
\newcommand*{\affmark}[1][*]{\textsuperscript{#1}}

\newcommand*{\authnote}[1]{#1} 
\newcommand*{\authmark}[1][*]{\textsuperscript{#1}}

\newcommand{\cc}[1]{\cellcolor{#1}}

\newcommand{\myparagraph}[1]{\paragraph{#1}\mbox{}\\}

\usepackage{float}
\restylefloat{table}

\begin{document}

\begin{titlepage}   
\thispagestyle{empty} 

\vspace*{0.2cm} 

\begin{center}
\vspace*{0.3cm} 
\LARGE {\bfseries \textsc{Paraphrasing, Textual Entailment, and Semantic Similarity \\ Above Word Level}}
\linethickness{1mm}
\\[2cm]
\Large
 \textsc{ \textbf{Venelin Orlinov Kovatchev}} \\[2cm]

\normalsize Tesis presentada para optar \\
 al grado de {\bfseries Doctor en Lingu{\'i}stica} con menci{\'o}n europea \\
en el programa de doctorado \emph{Ciencia Cognitiva y Lenguaje}, \\
Departament de Filologia Catalana i Ling{\"u}{\'i}stica General, \\
 Universidad de Barcelona, \\
 [1cm]

bajo la supervisi{\'o}n de  \\[0.2cm]
{\bfseries Dra. M. Ant{\`o}nia Mart{\'i}} \\
Universidad de Barcelona \\[0.2cm]
{\bfseries Dra. Maria Salam{\'o}} \\
Universidad de Barcelona \\[0.2cm]

\begin{figure}[h]
  \centering
  \includegraphics[width=5cm,clip]{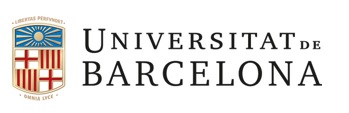} 
    \end{figure}
  
   \vspace{2cm}
 
Mayo de 2020
\end{center}
\end{titlepage} 

\thispagestyle{empty}
\begin{flushright}
\textit{To Mila, who supported me every step of the way.}

\vspace{2cm}

\textit{To Maya and Orlin, for always encouraging my curiosity.}
\end{flushright}

\newpage
\thispagestyle{empty}
\pagenumbering{roman}


\chapter*{Abstract}
\addcontentsline{toc}{chapter}{\numberline{}Abstract}%

This dissertation explores the linguistic and computational aspects of the meaning relations that can
hold between two or more complex linguistic expressions (phrases, clauses, sentences, paragraphs).
In particular, it focuses on Paraphrasing, Textual Entailment, Contradiction, and Semantic Similarity.
This thesis is composed of seven different articles and is divided into three thematic Parts.

In \textit{Part \ref{p:dsm}: ``Similarity at the Level of Words and Phrases''}, I study the 
Distributional Hypothesis (DH). DH is central for most contemporary approaches for automatic 
processing of meaning and meaning relations within Computational Linguistics (CL) and Natural 
Language Processing (NLP). Part \ref{p:dsm} of this thesis explores different methodologies for
quantifying semantic similarity at the levels of words and short phrases. I measure the importance 
of the corpus size and the role of linguistic preprocessing. I also show that (lexical) semantic 
similarity can interact with syntactic-based compositional rules and result in productive patterns at 
the phrase level. The research in Part \ref{p:dsm} resulted in the publication of two articles.

In \textit{Part \ref{p:type}: ``Paraphrase Typology and Paraphrase Identification''}, I focus on
the meaning relation of paraphrasing and the empirical task of automated Paraphrase Identification 
(PI). Paraphrasing is one of the most widely studied meaning relation both in theoretical and 
practical research. PI is among the most popular tasks in CL and NLP. In Part \ref{p:type}
of this thesis I present: 1) EPT: a new typology of the linguistic and reason-based phenomena involved 
in paraphrasing; 2) WARP-Text: a new web-based annotation interface capable of annotating
paraphrase types; 3) ETPC: the largest corpus to date to be annotated with paraphrase types; 
and 4) a qualitative evaluation framework for automated 
PI systems. The findings presented in Part \ref{p:type} provide in-depth knowledge on the nature of the 
paraphrasing relation and improve the evaluation, interpretation, and error analysis in the task of PI.
The research in Part \ref{p:type} resulted in the publication of three articles.

In \textit{Part \ref{p:mrel}:``Paraphrasing, Textual Entailment, and Semantic Similarity''}, I
present a novel direction in the research on textual meaning relations, resulting from joint research carried out on
on paraphrasing, textual entailment, contradiction, and semantic similarity. Traditionally, these meaning 
relations are studied in isolation and the transfer of knowledge and resources between them is limited. 
In Part \ref{p:mrel} of this thesis I present: 1) a methodology for the creation and annotation of corpora 
containing multiple textual meaning relations; 2) the first corpus annotated independently with Paraphrasing, 
Textual Entailment, Contradiction, Textual Specificity, and Semantic Similarity; 3) a statistical corpus-based 
analysis of the interactions, correlations, and overlap between the different meaning relations; 4) SHARel - 
a shared typology of textual meaning relations; 5) a corpus of paraphrasing, textual entailment, and 
contradiction annotated with SHARel. Part \ref{p:mrel} of the thesis gives a new perspective on the 
research of textual meaning relations. I show that a joint study of multiple meaning relations is both possible 
and beneficial for processing and analyzing each individual relation. I provide the first empirical data on the interactions 
between paraphrasing, textual entailment, contradiction, and semantic similarity. The research
in Part \ref{p:mrel} resulted in the publication of two articles.

This thesis has advanced our understanding of important issues associated with the empirical analysis,
corpus annotation, and computational treatment of textual meaning relations. I have addressed existing gaps
in the research field, posed new research questions, and explored novel research directions. The findings
and resources presented in this dissertation have been released to the community to facilitate  further 
research and knowledge transfer.

\chapter*{Resumen}
\addcontentsline{toc}{chapter}{\numberline{}Resumen}%

En esta tesis se exploran los aspectos ling\"{u}\'{i}sticos y computacionales de las relaciones sem\'{a}nticas que puede haber entre dos o m\'{a}s expresiones ling\"{u}\'{i}sticas complejas (sintagmas, cl\'{a}usulas, oraciones, p\'{a}rrafos). En particular, se centra en la par\'{a}frasis, la implicaci\'{o}n, la contradicci\'{o}n y la similitud sem\'{a}ntica. La tesis se compone de siete art\'{i}culos y se estructura en tres partes.

En la \textit{Parte \ref{p:dsm}: ``Similitud de palabras y sintagmas''}, realizo un estudio sobre la Hip\'{o}tesis distribucional (HD). La HD es relevante en muchos de los trabajos actuales sobre el procesamiento del significado y de las relaciones de significado en el \'{a}rea de la Ling\"{u}\'{i}stica Computacional (LC) y el Procesamiento del Lenguaje Natural (PLN). En esta parte se exploran diferentes m\'{e}todos para la cuantificaci\'{o}n de la similitud sem\'{a}ntica de palabras y de sintagmas. He calculado la importancia del tama\~{n}o del corpus y el papel que juega el preprocesado ling\"{u}\'{i}stico. Tambi\'{e}n muestro que la similitud sem\'{a}ntica l\'{e}xica puede interactuar con reglas de composici\'{o}n sint\'{a}ctica lo que da como resultado patrones productivos al nivel de sintagma. La investigaci\'{o}n de esta parte de mi tesis ha dado lugar a la publicaci\'{o}n de dos art\'{i}culos.

En la \textit{Parte \ref{p:type}: ``Tipolog\'{i}a de par\'{a}frasis e identificaci\'{o}n de par\'{a}frasis''} me centro en la relaci\'{o}n sem\'{a}ntica de par\'{a}frasis y en la tarea emp\'{i}rica de la identificaci\'{o}n autom\'{a}tica de par\'{a}frasis (IP). La par\'{a}frasis es una de las relaciones de significado m\'{a}s estudiadas, tanto a nivel te\'{o}rico como aplicado.  La IP es una de las tareas m\'{a}s populares en LC y en el PLN. En la Parte \ref{p:type} de esta tesis presento: 1) EPT, una nueva tipolog\'{i}a de fen\'{o}menos ling\"{u}\'{i}sticos y de fen\'{o}menos basados en el razonamiento implicados en la par\'{a}frasis: 2) WARP-Text, una nueva interfaz web para la anotaci\'{o}n de diferentes tipos de par\'{a}frasis; 3) ETPC: hasta el momento, el corpus de mayor tama\~{n}o anotado con tipos de par\'{a}frasis; y 4) un entorno de evaluaci\'{o}n cualitativa de sistemas autom\'{a}ticos de IP; Los resultados de esta segunda parte proporcionan un conocimiento m\'{a}s a fondo sobre la naturaleza de la relaci\'{o}n de par\'{a}frasis y mejoran la evaluaci\'{o}n, interpretaci\'{o}n y an\'{a}lisis de errores referentes a la tarea de IP. La investigaci\'{o}n de esta segunda parte ha dado lugar a tres publicaciones.

En la \textit{Parte \ref{p:mrel}: ``Par\'{a}frasis, Implicaci\'{o}n textual y Similitud sem\'{a}ntica''}, presento una nueva l\'{i}nea en la investigaci\'{o}n sobre las relaciones de significado. Llevo a cabo una investigaci\'{o}n conjunta sobre par\'{a}frasis, implicaci\'{o}n textual, contradicci\'{o}n y similitud sem\'{a}ntica. Tradicionalmente, estas relaciones se han estudiado separadamente y la transferencia de conocimiento entre ellas ha sido muy limitado. En esta tercera parte de la tesis presento: 1) una metodolog\'{i}a para la creaci\'{o}n y anotaci\'{o}n de corpus que contienen diversas relaciones de significado; 2) el primer corpus anotado independientemente con Par\'{a}frasis, Implicaci\'{o}n textual, Contradicci\'{o}n, Especificidad y Similitud sem\'{a}ntica; 3) un an\'{a}lisis estad\'{i}stico de las interacciones, correlaciones y coincidencias entre las diferentes relaciones de significado; 4) SHARel, una tipolog\'{i}a compartida para las relaciones sem\'{a}nticas textuales; 5) un corpus de par\'{a}frasis , implicaci\'{o}n textual y contradicci\'{o}n anotado con SHARel. Esta tercera parte de la tesis da una nueva perspectiva sobre la investigaci\'{o}n en las relaciones de significado a nivel textual. Pongo de manifiesto que es posible el estudio conjunto de diversas relaciones de significado y tambi\'{e}n que repercute positivamente para cada una de las relaciones en particular. Proporciono por primera vez un conjunto de datos emp\'{i}ricos sobre la integraci\'{o}n de par\'{a}frasis, implicaci\'{o}n textual, contradicci\'{o}n y similitud sem\'{a}ntica. La investigaci\'{o}n de esta tercera parte ha dado lugar a dos art\'{i}culos.

Esta tesis ha permitido avanzar en la comprensi\'{o}n de aspectos importantes relacionados con el an\'{a}lisis emp\'{i}rico, la anotaci\'{o}n de corpus, y el tratamiento computacional de las relaciones de significado a nivel textual. He tratado diversas \'{a}reas de conocimiento poco atendidas hasta ahora, he planteado nuevas preguntas para la investigaci\'{o}n posterior y he explorado en nuevas directrices. Los resultados y recursos presentados en esta tesis son de libre disposici\'{o}n para el colectivo que investiga en LC y PLN con el fin de facilitar la investigaci\'{o}n futura y la transferencia de conocimiento.

\chapter*{Acknowledgments}
\addcontentsline{toc}{chapter}{\numberline{}Acknowledgments}%

First of all, I'd like to thank my partner, Mila. She has been with me every step of the way,
through deadlines, submissions, acceptances, and rejections. She encouraged me and
supported me throughout the whole process, she changed countries and jobs and never
lost faith in me. She has also endured countless hours of talks about language, cognition, 
and machine learning. 

I would also like to thank my supervisors, Toni Mart\'{i} and Maria Salam\'{o}. It has been a great
privilege for me to work with them over the past five years. They have helped me and guided
me through my Master and PhD and have taught me everything I know about Computational
Linguistics and Natural Language Processing. They have given me so much of their time,
energy, and knowledge and have been incredibly patient with my stubbornness. I could not
have asked for better supervisors.

I would also want to thank Torsten Zesch for hosting me for a research stay at the University 
of Duisburg-Essen. His feedback on my work has given me a different perspective on the
problem of processing textual meaning relations and has helped me improve as a researcher. 

Horacio Rodr\'{i}guez Hontoria proposed the topic of this thesis and  has been very helpful
with ideas, references and feedback during my PhD. It was a pleasure to work with him and
I admire his productivity and competence.

I have been very lucky to meet and collaborate with fantastic colleagues at the University of
Barcelona and the University of Duisburg-Essen. I'm very grateful to Mariona Taul\'{e}, 
Darina Gold, Javier Beltran, Eloi Puertas, Montse Nofre and David Bridgewater.

Many colleagues and friends have played an important role in the success of my PhD. Irina
Temnikova was one of the first people that I met during my research. She has been very kind
and helpful to me and introduced me to the CL and NLP community and has always been
a good friend. I would also like to thank Ahmed AbuRa'ed, Amir Hazem, Sanja \v{S}tajner,
Tobias Horsmann, Michael Wojatzki, Alejandro Ariza Casabona, Jeremy Barnes, Alexander
Popov, Thomas O'Rourke, Kristen Schroeder, Elisabet Vila Borrellas, Ruslan Mitkov, Galia
Angelova, Nadezhda Stoyanova, Nikolay Metev, Maria Marinova-Panova, Penko Kirov,
Stefan Lekov, Irina Ivanova, Anna Ignatova, Chris Childress, Theresia Scholten, Andhi Tang, 
Gabriel Sevilla, Flavia Felletti, and the whole team behind LxMLS.

Last but not least I am grateful to my family. My parents Maya and Orlin, my grandparents
Anka, Emilia, and Venelin, my sisters Alissa and Emily, and my aunt Anissia. They have all been very 
supportive and have always been excited to learn more about my research.

\vspace{2cm}

This work has been funded by the APIF grant awarded to me by the University of Barcelona;
by the Spanish Ministery of Economy Project TIN2015-71147-C2-2; by the Spanish Ministery of 
Science, Innovation, and Universities Project PGC2018-096212-B-C33; and by the CLiC research
group (2017 SGR 341).

\tableofcontents

\listoffigures
\addcontentsline{toc}{chapter}{\numberline{}List of Figures}%

\listoftables
\addcontentsline{toc}{chapter}{\numberline{}List of Tables}%

\chapter{Introduction}\label{ch:intro}

\pagenumbering{arabic}

	This thesis is about the meaning relations that can hold between language expressions
(words, phrases, clauses, and sentences). In particular, it focuses on the meaning relations
of paraphrasing, textual entailment, and semantic similarity. The automatic processing of
these meaning relations is an unsolved problem in Computational Linguistics (CL) and
Natural Language Processing (NLP) and has attracted the attention of many researchers.
This thesis explores two different directions within the research on meaning relations:

\begin{enumerate}

	\item Incorporating linguistic knowledge in the empirical tasks of processing meaning relations. 
	In particular, I focus on the paraphrasing meaning relation and the empirical task of Paraphrase 
	Identification (PI). By combining PI with Paraphrase Typology (PT) I aim: 
	\begin{itemize}
		\item[a)] to improve the evaluation and interpretation of automated PI systems.
		\item[b)] to empirically validate PT.
	\end{itemize}

	\item Analyzing and processing multiple meaning relations together. I contrast previous work
	and propose a novel research approach that does not focus on a single meaning relation. I
	present a joint study on Paraphrasing, Textual Entailment, Contradiction, and Semantic 
	Similarity: 
	\begin{itemize}
		\item[a)] to compare the different meaning relations empirically.
		\item[b)] to create a shared typology for textual meaning relations.
	\end{itemize}

\end{enumerate}

My work offers a valuable insight into the nature and interactions of the different meaning 
relations and also aims to improve the automated systems for processing meaning relations. 
I also release to the community three 
new corpora, two new typologies of meaning relations, a new web-based annotation tool, and a 
new software program for a qualitative evaluation of automated paraphrase identification systems. 

%

The structure of this thesis is intentionally chronological\footnote{Articles are presented
in the order in which they were written, which does not necessarily correspond to the
order in which they were published.} in order to capture the four year development of the ideas 
and arguments behind the thesis. The thesis consists of nine Chapters, organized as
follows:

\begin{itemize} 
	\item Chapter \ref{ch:intro} is the Introduction.
	\item Chapters \ref{ch:sepln} to \ref{ch:sharel} correspond to seven published 
	articles. They are grouped in three thematically organized parts. 
	\item Chapter \ref{ch:conc} presents the contributions, the discussion of the results,
	the conclusions, and the directions for future work.
\end{itemize}


The rest of this Introduction chapter is organized as follows. In Section \ref{ch:related},
I familiarize readers with the related work in the research on 
meaning relations. In Section \ref{ch:objectives}, I present my main objectives and 
justify them in the context of the preexisting research. In Section \ref{ch:development},
I describe the development of this thesis and the connecting thread that runs between the 
individual articles. Finally in Section \ref{ch:outline}, I present the outline and structure
of the whole dissertation.


	\section{Related Work}\label{ch:related}

	This section is meant to provide the reader with a compact overview of the previous 
and latest research related to this thesis in order to supplement
and bind together the ``background'' sections in each paper.
From a thematic perspective, the subject matter can be broken down into the following 
research areas:

\begin{itemize}
	\item[(i)]  Textual Meaning Relations. Empirical Tasks. (\textbf{Section \ref{ch:rel:tasks}})
	\item[(ii)] Typologies of Textual Meaning Relations  (\textbf{Section \ref{ch:rel:type}})
	\item[(iii)] Joint Research on Textual Meaning Relations  (\textbf{Section \ref{ch:rel:joint}})
	\item[(iv)] Other Related Work  (\textbf{Section \ref{ch:rel:oth}})
\end{itemize}

I will deal with each of the areas in turn, highlighting the main trends and milestones.
The reader is referred to the original papers for details.


\subsection{Textual Meaning Relations. Empirical Tasks}\label{ch:rel:tasks}

Meaning relations between complex language expressions (e.g.: clauses, sentences, paragraphs),
henceforth ``textual meaning relations'' are the object of study of this thesis. Research on 
textual meaning relations has to account not only for the meaning of a single word or a phrase, but 
also for the compositionality of meaning. In this thesis, I focus on the textual meaning relations of 
Paraphrasing, Textual Entailment, Contradiction\footnote{Contradiction is typically studied jointly with 
Textual Entailment.}, and Semantic Similarity.  It is important to note that the interactions between 
the different relations are non-trivial. In some cases they can overlap (e.g.: two texts that are 
paraphrases often also hold an entailment relation) and in some cases the negative examples for one 
relation can be positive examples for another (e.g.: two texts that are not paraphrases can sometimes 
hold an entailment relation or a contradiction relation).

\myparagraph{Empirical Tasks on Textual Meaning Relations}
\cite{androutsopoulos2010survey} distinguish three types of empirical tasks that are focused on
 processing meaning relations: recognition, generation, and extraction. Their definitions for these
paraphrasing and textual entailment tasks are as follows:

\begin{quote}
\textbf{Recognition}: \textit{``The main input to a paraphrase or textual entailment 
recognizer is a pair of language expressions (or templates), possibly in particular context.
The output is a judgment, possibly probabilistic, indicating whether or not the members
of the input pair are paraphrases or a correct textual entailment pair; the judgments must
agree as much as possible with those of humans.''}

\textbf{Generation}: \textit{``The main input to a paraphrase or textual entailment generator
is a single language expression (or template) at a time, possibly in a particular context. The
output is a set of paraphrases of the input or a set of language expressions that entail or
are entailed by the input; the output set must be as large as possible, but including as few
errors as possible.''}

\textbf{Extraction}: \textit{``The main input to a paraphrase or textual entailment extractor
is a corpus, for example a monolingual corpus of parallel or comparable texts. The system
outputs pairs of paraphrases (possibly templates) or pairs of language expressions (or
templates) that constitute correct textual entailment pairs, based on the evidence of the
corpus; the goal is again to produce as many output pairs as possible, with as few errors
as possible.''}

\end{quote}

The empirical tasks focused on the automatic processing of meaning relations are inspired by human 
capabilities. We, as competent language users, can quickly and unconsciously determine the 
meaning relation that holds between two simple or complex language expressions. We can 
successfully recognize, generate, and extract paraphrases, entailment pairs, and contradiction 
pairs. The empirical tasks focused on textual meaning relations in CL and NLP
aim to produce 
automated systems that can achieve on-task performance comparable with that of 
humans. Human judgments are typically taken as a gold standard for evaluation.

In this thesis, I focus on recognition tasks. In particular, I study \textbf{Paraphrase Identification}, 
\textbf{Recognizing Textual Entailment}, and \textbf{Semantic Textual Similarity}. 
In the rest of this section I present the definition, corpora, and state-of-the-art for each
of these three tasks.

\myparagraph{Paraphrase Identification} 

\textbf{Task format and definition:} Paraphrase Identification (PI) is framed as a binary
classification task. In PI, a human or an automated system needs to determine whether or
not a paraphrasing relation holds between two given texts. The definition of ``paraphrasing''
provided by \cite{dolan2005} is \textit{``whether two sentences at the high level ``mean 
the same thing'' /.../ despite obvious differences in information content.''}. 

\begin{itemize}
	\item[(1)] \textbf{Sentence 1:} The genome of the fungal pathogen that causes 
	Sudden Oak Death has been sequenced by US scientists.\\
	\textbf{Sentence 2:} Researchers announced Thursday they've completed the genetic
	blueprint of the blight-causing culprit responsible for sudden oak death.\\
\end{itemize}

Two sentences that are connected with a paraphrasing relation can be seen in Example 1\footnote{The example is 
taken from \cite{dolan2005}.}. While the two sentences are not completely equivalent, in
the context of PI they are considered paraphrases. \cite{dolan2005} argue that if human
annotators are required to only mark full equivalence of meaning, only identical sentences
are considered paraphrases. Therefore, in the practical setting of PI, they propose a less strict
definition of paraphrasing and allow for some difference in the information content. 

\textbf{PI Corpora:} The task of (PI) was first popularized with the creation of the Microsoft 
Research Paraphrase Corpus (MRPC), presented in \cite{mrpc} and \cite{dolan2005}. The 
MRPC corpus is semi-automatically created from the articles in the news domain and consists 
of 5,801 text pairs, annotated as ``paraphrase'' or ``non-paraphrase''. To date, MRPC is still 
used for the evaluation of automated PI systems despite, its relatively small size.

The Paraphrase Database (PPDB) \citep{ppdb} (and later on its second version PPDB2
 \citep{ppdb2}) was the first large scale paraphrase corpus. It is an automatically 
constructed collection of over 100 million paraphrases at different granularity. While 
the MRPC only contains sentences and longer chunks of text, the PPDB also contains
``paraphrases'' of words and short phrases. The second version of PPDB also includes 
the entailment relation. PPDB and PPDB2 are collections of paraphrases, rather than corpora
specifically created for the task of PI. However, they can be adapted for use in PI tasks.

The Quora Question Pair Dataset \citep{quora} is a semi-automatically collected 
corpus of 400,000 question pairs marked as ``duplicate'' or ``non-duplicate'' by
Quora users. The corpus was used in an online competition\footnote{\url{https://www.kaggle.com/c/quora-question-pairs}} 
and facilitated the use of Deep Learning based systems for the task of PI. Due to
its size, the Quora corpus is very popular for training state-of-the-art PI systems.

The Language-Net corpus \citep{twitt} is the largest PI dataset to date. It was
extracted from Twitter and contains over 51,000 human-annotated sentence 
pairs and over 2.8 million automatically extracted candidate paraphrases.

MRPC, PPDB, Quora, and Language-net are all created for the English language.
The work on PI for languages other than English is very limited. We can mention
the work of \cite{creutz-2018-open} on the creation of paraphrase corpus in
six languages using open subtitles dataset.

\textbf{State-of-the-art in PI:}
The first automated PI systems were based on manually engineered features 
\citep{finch-etal-2005-using,Kozareva} or on a combination of lexical similarity metrics 
and cosine similarity \citep{mihalcea}. Word2Vec \citep{word2vec} and Glove 
\citep{glove} introduced a new paradigm in PI, but also in CL and NLP in general. The 
systems based on Word2Vec and Glove outperformed previous unsupervised systems 
and pushed the state-of-the-art further. 
Deep Learning based systems using autoencoders \citep{socher}, Long Short Term 
Memory Networks (LSTM) \citep{He2016}, and Convolutional Neural Networks (CNN) 
\citep{He2015} set the new state-of-the-art for the Supervised PI systems. More
recently, Transformer based architectures \citep{bert} have made a considerable 
improvement to automated PI systems, approaching human level performance on
the datasets\footnote{The official ACL page for PI (\url{https://aclweb.org/aclwiki/Paraphrase_Identification_(State_of_the_art)})
and  the GLUE benchmark page (\url{https://gluebenchmark.com/leaderboard}) contain 
the full leaderboard of PI systems for a variety of corpora.}.

\myparagraph{Recognizing Textual Entailment}

\textbf{Task format and definition:} Recognizing Textual Entailment (RTE), also known
as Natural Language Inference (NLI), has two different formats. The original RTE was 
framed as a binary classification task. In RTE, a human or an automated system needs to 
determine whether or not a paraphrasing relation holds between two given texts.
The practical definition of Textual Entailment in RTE is \textit{``a directional relationship 
between pairs of text expressions, denoted by T - the entailing ``Text'', and H - the 
entailed ``Hypothesis''. We say that T entails H if the meaning of H can be inferred from 
the meaning of T, as would typically be interpreted by people.''}. An example of
textual entailment relation can be seen in 2. In the example given the Text entails Hyp 1, 
but not Hyp 2, or Hyp 3. 

\begin{itemize}
	\item[(2)] \textbf{Text:} The purchase of Houston-based LexCorp by BMI for \$2Bn
	prompted widespread sell-offs by traders as they sought to minimize exposure. LexCorp 
	had been an employee-owned concern since 2008. \\
	\textbf{Hyp 1:} BMI acquired an American company.\\
	\textbf{Hyp 2:} BMI bought employee-owned LexCorp for \$3.4Bn.\\
	\textbf{Hyp 3:} BMI is an employee-owned concern.\\
\end{itemize}

The second format of the RTE was introduced in \citep{rte4} and the task was 
reformulated as a three class classification between ``entailment'', ``contradiction'', and 
``neutral'' text pairs. In example 2, the Text entails Hyp 1, contradicts Hyp 2, and is
neutral with respect to Hyp 3.

\textbf{RTE Corpora:} The task of RTE was popularized with the introduction of the yearly 
Recognizing Textual Entailment challenge in \cite{rte}.The first three editions of the RTE
challenge were called the Pascal RTE challenge \citep{rte,rte2,rte3} and were framed as a binary
classification between ``entailment'' and ``non entailment'' text pairs.
In the fourth edition of the challenge \citep{rte4}, the Pascal RTE challenge became 
the Text Analysis Conference (TAC) RTE challenge. The task was reformulated as 
a three class classification between ``entailment'', ``contradiction'', and ``neutral''
text pairs. The TAC RTE challenge ran for four years: \cite{rte4}, \cite{rte5}, \cite{rte6}, 
and \cite{rte7}. Like the MRPC corpus, the RTE datasets are not very large in
size, however due to the high quality of the annotation they are still used as an 
evaluation benchmark for state-of-the-art systems.

%

The increasing popularity of Deep Learning systems and the need for more training
data led to the creation of the Stanford Natural Language Inference corpus (SNLI) 
\citep{snli} and later on the Multi-Genre Natural Language Inference corpus (MultiNLI) 
\citep{mnli}. The SNLI contains 570,000 human-written English sentences, while the
MultiNLI contains 433,000 sentences but covers a more diverse range of texts. SNLI
and MultiNLI are currently the most popular corpora for training automated RTE/NLI
systems. Both SNLI and MultiNLI use the three-way classification format of the
task.

As with PI, the work on RTE and NLI is mostly for English. 
Notable exceptions are the XNLI corpus \citep{conneau2018xnli}, a 
machine-translated portion of MultiNLI and the SPARTE corpus \citep{sparte} for RTE in 
Spanish, created from question-answering corpora.

\textbf{State-of-the-art in RTE:}
The development of the automated RTE/NLI systems follows a similar trend as the
development of the automated PI systems. The first RTE systems used manually engineered 
features and simple similarity metrics.
Then, there was a paradigm shift towards various Deep Learning architectures, such as 
autoencoders, LSTMs, and CNNs. And finally, the current state-of-the-art are Transformer
based architectures \footnote{The official ACL page for the RTE Challenge (\url{https://aclweb.org/aclwiki/Recognizing_Textual_Entailment}),
the official SNLI corpus page (\url{https://nlp.stanford.edu/projects/snli/}), the
official MultiNLI corpus page (\url{https://www.nyu.edu/projects/bowman/multinli/}),
and the GLUE benchmark page (\url{https://gluebenchmark.com/leaderboard}) contain 
the full leaderboard of RTE systems for a variety of corpora.}.

With the state-of-the-art systems approaching human level performance on the datasets,
many researchers have tried to \textbf{analyze the workings} of the different RTE and NLI 
systems. \cite{gururangan-etal-2018-annotation} discovered the presence of annotation 
artifacts that enable models that take into account only one of the texts (the hypothesis) 
to achieve 67\% (SNLI) and 52.3-53.9\% (MultiNLI) accuracy, which is substantially higher than the 
majority baselines of 34-35\%. \cite{glockner-etal-2018-breaking} showed that models 
trained with SNLI fail to resolve new pairs that require simple lexical substitution. For 
example the models have problems determining that  \textit{``holding a saxophone''} 
contradicts \textit{``holding an electric guitar''}. The human annotators indicate a 
contradiction in this example, as the annotation guidelines instruct them to assume that
the same event is referred to by both texts. \cite{naik-etal-2018-stress} created label-preserving 
adversarial examples and concluded that automated NLI models are not robust.
\cite{wallace-etal-2019-universal} introduced universal triggers, that is, sequences of tokens 
that fool models when concatenated to any input.

All of these findings indicate that the existing RTE and NLI datasets are much simpler than
what native speakers are capable of. Furthermore, the datasets contain many annotation
artifacts and the systems trained on them are not robust to adversarial examples. Therefore,
despite the high performance achieved on the datasets, the general problem of RTE and NLI is far
from resolved. 


\myparagraph{Semantic Textual Similarity}

\textbf{Task format and definition:} Semantic Textual Similarity (STS) is framed as a
regression task. In STS, a human or an automated system needs to determine the degree
of similarity between two given texts on a continuous scale from 0 to 5. The practical 
definition for Semantic Similarity in STS is \textit{``how similar two sentences are to each 
other according to the following scale:\\
\textbf{[5] Completely equivalent}, as they mean the same thing.\\
\textbf{[4] Mostly equivalent}, but some unimportant details differ.\\
\textbf{[3] Roughly equivalent}, but some important information differs/missing.\\
\textbf{[2] Not equivalent}, but share some details. \\
\textbf{[1] Not equivalent}, but are on the same topic \\
\textbf{[0] On different topics.} \\
}Examples for each semantic similarity from 0 to 5 can be seen in 3.

\begin{itemize}
	\item[(3)] \textbf{Similarity 5:} \\
	The bird is bathing in the sink.  \\
	Birdie is washing itself in the water basin. \\
	\textbf{Similarity 4:} \\
      In May 2010, the troops attempted to invade Kabul. \\
      The US army invaded Kabul on May 7th last year, 2010.\\
	\textbf{Similarity 3:} \\
      John said he is considered a witness but not a suspect.\\
      ``He is not a suspect anymore.'' John said.\\
	\textbf{Similarity 2:} \\
      They flew out of the nest in groups.\\
      They flew into the nest together.\\
	\textbf{Similarity 1:} \\
      The woman is playing the violin.\\
      The young lady enjoys listening to the guitar.\\
	\textbf{Similarity 0:} \\
      John went horse back riding at dawn with a whole group of friends.\\
      Sunrise at dawn is a magnificent view to take in if you wake up
      early enough for it.\\
\end{itemize}

\textbf{STS Corpora:}
The most popular corpora for the STS task are the datasets from the yearly STS 
competition \citep{sts}. While the STS corpora are not large in size, they come from a variety 
of domains and their coverage is extended every year. Unlike the tasks of PI and RTE, the 
competition in STS includes non-English texts (Arabic, Spanish, Turkish). Also unlike PI and 
RTE, at the time this dissertation was begun there were no large scale corpora explicitly designed 
for STS.

\textbf{State-of-the-art in STS:}
The development of the automated STS systems is similar to that of the automated
systems for PI and RTE. The system architecture transitions from feature based through
Deep Learning based systems, and finally to the current state of the art, which are transformer 
based systems\footnote{The official page for the STS challenge\footnote{\url{http://ixa2.si.ehu.es/stswiki/index.php/Main_Page}}
and the GLUE benchmark page (\url{https://gluebenchmark.com/leaderboard}) contain 
the full leaderboard of STS systems for a variety of corpora.}.

\subsection{Typologies of Textual Meaning Relations}\label{ch:rel:type}

In the context of the empirical tasks of Paraphrase Identification (PI), Recognizing Textual
Entailment (RTE), and Semantic Textual Similarity (STS), the corresponding meaning relations 
are typically considered atomic. That is, the researchers in these areas make several 
assumptions about the data and the task:

\begin{itemize}
	\item Each pair of texts has a single label corresponding to it. The label is one of a
	pre-defined set.
	\item The label applies to the whole text pair and cannot be expressed (decomposed)
	as a combination of more simple phenomena.
	\item Each pair of texts is processed the same way by the human annotators and the
	automated systems. It has the same complexity as any other pair in the dataset and
	it contributes the same weight to the evaluation of the model.
\end{itemize}

These assumptions are made to facilitate the definition and evaluation of the empirical tasks.
However, several researchers working on Paraphrasing, Textual Entailment, and Semantic
Similarity have questioned the applicability of these simplifications and have provided counter
examples, such as Examples 4 and 5:

\begin{itemize}
	\item[(4)] \textbf{Sentence 1:} All \textbf{kids} receive the same education .\\
	\textbf{Sentence 2:} All \textbf{children} receive the same education .\\
	\item[(5)] \textbf{Sentence 1:} All \textbf{kids} \underline{receive} the same education .\\
	\textbf{Sentence 2:} The same education \underline{is provided to} all \textbf{children} .\\
\end{itemize}

In both Examples 4 and 5, the two texts have approximately the same meaning and they can be 
labeled as ``paraphrases''. In the context of PI, these two examples have the same label, the same 
degree of complexity, and the same weight in the final evaluation of the system. However, when 
looking at the examples, the human intuition would suggest that: 

\begin{itemize}
	\item Processing Examples 4 and 5 requires different (linguistic) capabilities and follows 
	different (linguistic) strategies. 
	\item Example 5 is arguably harder than Example 4. 
\end{itemize}

These intuitions contradict the empirical assumptions concerning the atomic nature of the data. If 
the meaning relations are indeed atomic and non-decomposable, then Examples 4 and 5 should 
have approximately the same degree of complexity and determining the correct label should 
require similar linguistic capacities and strategies.

Starting from linguistic theory and from examples like 4 and 5, several researchers have
questioned the atomic nature of the Paraphrasing, Textual Entailment, and Semantic
Similarity meaning relations. The "non-atomic" 
approach of studying meaning relations historically began in the field of Textual 
Entailment with the works of \cite{Garoufi} and \cite{Sammons}. Later on \cite{CabrioMagnini}
carried out a large theoretical and empirical study on the nature of the phenomena involved
in entailment. Independently from the research on textual entailment, \cite{Vila} and
\cite{BhagatHovy} proposed different ways to decompose and characterize the paraphrasing 
relation. In the area of semantic similarity, \cite{agirre-etal-2016-semeval-2016} proposed a 
new task of ``interpretable semantic textual similarity''.

In the context of this thesis, there are two important hypotheses, shared by the majority
of the authors working on decomposing meaning relations.


\textbf{The first hypothesis} argues that in order to determine the meaning relation that holds 
between two texts, a human or an automated system needs to make one (or more) simple 
``inference steps''. In Example 4, such inference steps would be: 
\begin{itemize}
	\item[1)] determining that ``kids'' in Example 4.1 means the same as ``children'' in 
	Example 4.2 within the given context.
	\item[2)] determining that all of the linguistic units in the two sentences in Example 4 are the 
	same, except for ``kids'' - ``children''. 
\end{itemize}

Based on 1) and 2), a human or an automated system can determine that in Example 4, the 
two texts have approximately the same meaning and therefore the correct label is ``paraphrases''. 
The hypothesis argues that to correctly predict the textual meaning relation in Example 4, a 
human or an automated system needs to have the capabilities and the background knowledge 
to process each individual ``inference step''. 
%

\textbf{The second hypothesis} argues out that a single example can contain various numbers 
of ``inference steps''. Example 4 has two inference steps. Example 5 has one additional step: 
the substitution of  ``receive'' with ``is provided to'' and the corresponding change in the 
syntactic structure of the two sentences. Following from this hypothesis, the different number and 
nature of inference steps would result in different strategies for processing the examples and 
different degrees of complexity.

All of the authors working on decomposing meaning relations propose a list of linguistic 
phenomena that can be considered to be inference steps. In the rest of this dissertation these
lists are called ``typologies''. In Table \ref{t:cmp_type} I compare the different typologies. I 
also include the data for the two typologies proposed in this thesis: EPT and SHARel, presented 
in Chapters \ref{ch:etpc} and \ref{ch:sharel}. 

\begin{table}[H]
\begin{tabular}{l l l l l r }
\hline
\textbf{Typology} & \textbf{Relation}  & \textbf{Types}  & \textbf{Lvls}  & \textbf{Neg-Ex} & \textbf{Corpus} \\ \hline \hline
\cite{Garoufi}  & TE  & 28 & Yes & Yes & 500 pairs\\ \hline
\cite{Sammons} & TE, CNT  & 22 & No & Yes & 210 pairs\\ \hline
\cite{CabrioMagnini} & TE, CNT & 36 & Yes & Yes & 500 pairs\\ \hline
\cite{BhagatHovy} & PP & 25 & No & No & 355 pairs \\ \hline
\cite{Vila} & PP & 23 & Yes & No & 3900 pairs \\ \hline
\cite{agirre-etal-2016-semeval-2016} & STS & 9 & No & Yes & 3000 pairs \\ \hline \hline
\textit{EPT (Chapter \ref{ch:etpc})} & PP & 27 & Yes & Yes & 5801 pairs\\ \hline
\textit{SHARel (Chapter \ref{ch:sharel})} & \shortstack{PP, STS, \\ TE, CNT} & 34 & Yes & Yes & 520 pairs \\ \hline
\end{tabular}
\caption{Typologies of textual meaning relations}
\label{t:cmp_type}
\end{table}

Table \ref{t:cmp_type} compares typologies of textual meaning relations in terms of:

\begin{itemize}
	\item[] \textbf{Relation}: The textual meaning relation (or relations) that can be decomposed using
	the typology. TE - ``Textual Entailment''; CNT - ``Contradiction''; PP - ``Paraphrasing''; STS
	- ``Semantic Textual Similarity''.
	\item[] \textbf{Types}: The number of phenomena in the typology.
	\item[] \textbf{Lvls}: Whether or not the typology is organized in hierarchical levels. For example,
	some typologies distinguish between morphological, lexical, syntactic, etc. phenomena, while
	others have no explicit structure.
	\item[] \textbf{Neg-Ex}: Whether the typology can be used to decompose and analyze 
	negative examples (i.e.: ``non-paraphrases'', ``non-entailment'', ``0 semantic similarity'') or
	if it is only applicable to positive examples.
	\item[] \textbf{Corpus}: The size of the available corpora annotated with the typology.
\end{itemize}


With respect to the \textbf{relation}, each 
typology is built around a single empirical task. The typologies of \cite{Garoufi}, \cite{BhagatHovy}, 
\cite{Vila}, and \cite{agirre-etal-2016-semeval-2016} are all built around a single
textual meaning relation. The typologies of \cite{Sammons} and \cite{CabrioMagnini} can be
applied to two textual meaning relations: Textual Entailment and Contradiction.

Considering the number of \textbf{types}, 
most of the typologies contain between 23 and 28 phenomena. The majority of these
phenomena are in fact shared across the typologies of paraphrasing and textual entailment. 
The typology for semantic textual similarity is much more simple and task specific.

Taking into account the \textbf{levels} of hierarchical structure, three of the typologies
\citep{Garoufi,CabrioMagnini,Vila} organize the types in terms of the linguistic level of the
phenomena (morphological, lexical, lexico-syntactic, syntactic, discourse, reasoning). The
remaining typologies propose a list of phenomena without trying to organize them.

Looking at the decomposition of  \textbf{negative examples}, the typologies for textual
entailment and semantic similarity can be applied to both positive and negative 
examples. The typologies for paraphrasing \citep{BhagatHovy,Vila} can only decompose
pairs of text that hold a ``paraphrasing'' relation. They cannot be applied to 
``non-paraphrases''.

Finally, with respect to the size of the available \textbf{corpora}, most typologies have been used
to annotate only a small corpus (200-500 text pairs). \cite{Vila} and \cite{agirre-etal-2016-semeval-2016} 
are the only authors that provide corpora of a size sufficient for machine learning experiments.

Table \ref{t:cmp_type} demonstrates some clear tendencies across the different typologies.
It also illustrates some important gaps in the research field. First, at the time of beginning this dissertation
each of the typologies was built around a single task and focused on one (or two) textual
meaning relations. There was no typology that could be applied to multiple textual meaning
relations without adaptation. Second, at the time of beginning this dissertation there was no corpus of
paraphrasing or textual entailment, annotated with a typology and suitable for ``recognition'' 
machine learning experiments. The corpora of \cite{Garoufi}, \cite{Sammons}, \cite{BhagatHovy},
and \cite{CabrioMagnini} are too small in size and the corpus of \cite{Vila} contains only
``paraphrases'', without negative examples. With the creation of EPT and SHARel, I
aimed to address these gaps in the field, as shown in the last two rows of Table \ref{t:cmp_type}.


\subsection{Joint Research on Textual Meaning Relations}\label{ch:rel:joint}

Despite the obvious similarities and interactions between the textual meaning relations,
the joint research on them has been very limited, both in theoretical and in empirical
aspects. Table \ref{t:rel_corp} shows some of the most popular corpora explicitly 
annotated with textual meaning relations. Most of the corpora comes from the empirical
tasks of PI, RTE, and STS. I also include the data for the corpus I present in Chapter 
\ref{ch:law} of this thesis.

\begin{table}[H]
\begin{tabular}{l l l l l }
\hline
\textbf{Corpus} & \textbf{Paraph.} & \textbf{Entailment} & \textbf{Contradiction} & \textbf{Similarity} \\ \hline \hline
MRPC & Yes & No & No & No \\ \hline
Quora & Yes & No & No & No \\ \hline
Language-Net & Yes & No & No & No \\ \hline \hline
RTE (1-3) & No & Yes & No & No \\ \hline
RTE (4-6) & No & Yes & Yes & No \\ \hline
SNLI & No & Yes & Yes & No \\ \hline
MultiNLI & No & Yes & Yes & No \\ \hline \hline
STS (all) & No & No & No & Yes \\ \hline \hline
SICK & No & Yes & Yes & Yes \\ \hline
\cite{sukhareva2016crowdsourcing} & Yes * & Yes & No & No \\ \hline
Chapter \ref{ch:law} (this thesis)  & Yes & Yes & Yes & Yes \\ \hline
\end{tabular}
\caption{Popular corpora for textual meaning relations}
\label{t:rel_corp}
\end{table}

Table \ref{t:rel_corp} clearly demonstrates the separation between the different meaning 
relations in existing corpora. Each corpus is typically built around one single relation, or two 
in the case of textual entailment. At the time of beginning this thesis the only corpora that contained 
multiple textual meaning relations were: 
\begin{itemize}
	\item the SICK corpus \citep{marelli2014sick}, which is annotated for textual entailment, 
	contradiction, and semantic similarity.
	\item the corpus of \cite{sukhareva2016crowdsourcing} who annotate paraphrasing as 
	a specific sub-class of entailment.
\end{itemize}

The corpus presented in Chapter \ref{ch:law} addresses this gap in the existing resources
and is the first corpus annotated with the four most popular textual meaning relations: 
Paraphrasing, Textual Entailment, Contradiction, and Semantic Similarity.

In a more theoretical setting, \cite{madnani2010generating} and \cite{androutsopoulos2010survey} 
discuss and compare different aspects of paraphrasing and textual entailment. They 
argue that paraphrasing is typically a bi-directional entailment. \cite{CabrioMagnini}
and \cite{sukhareva2016crowdsourcing} also suggest that paraphrasing is a sub-class
of textual entailment.

However, \cite{dolan2005} point out that if they enforced a strict bi-directional entailment
and full equivalence of the information content, the annotators would only mark 
identical texts as paraphrases, which would make the Paraphrase Identification task
trivial.  Therefore in their annotation setup they also allow for a limited difference in 
the information content in the two texts. As a result, the equivalence between 
bi-directional entailment and paraphrasing does not hold in their corpus (MRPC).
A similar approach to annotating the paraphrasing relation has also been adopted in
the rest of the PI corpora. Therefore, the relation between entailment and paraphrasing
is non-trivial to define in an empirical setting. However, the lack of corpora annotated for 
multiple textual meaning relations has limited the possibilities for empirical 
data-driven research on the interactions between paraphrasing, textual entailment,
and contradiction.

There has also been some research on using one textual meaning relation to predict 
another and for the transfer of knowledge across tasks. 
\cite{cer2017semeval} argue that to find paraphrases or entailment, some level of
semantic similarity must be given. 
\cite{bosma2006paraphrase} use techniques from Paraphrase Identification in order to
solve textual entailment. \cite{castillo2010using} and \cite{yokote2011effects} use
semantic similarity to solve entailment.

The recent work by several authors is indicative of an increasing interest towards the 
joint study of meaning relations. In particular, the topic is interesting within the context
of transfer learning in NLP and CL. \cite{lan} and \cite{hanan} 
demonstrate the transfer learning capabilities of different systems in the tasks of PI and 
RTE. They cover a wide range of supervised and unsupervised machine learning architectures
and demonstrate promising results.

The interest and success of the transfer learning techniques have also resulted in the creation 
of the GLUE \citep{wang-etal-2018-glue} and SuperGLUE \citep{NIPS2019_8589} 
benchmarks. GLUE and Super GLUE are a collection of multiple datasets for several tasks, 
including PI, RTE and STS. The authors of those benchmarks argue that systems working 
on Natural Language Understanding (NLU) should be able to perform well on all of the 
tasks, and not just on one. The GLUE and SuperGLUE are now the most popular 
benchmarks for evaluating NLU systems and general purpose meaning representation
models.

However, I would argue that a benchmark of multiple datasets is not a replacement for a
single dataset annotated with multiple textual meaning relations. Similarly, a transfer
learning experiment is not a replacement for a single task of multi-class classification.
At the time of beginning this thesis there was an apparent gap in the field - a lack of resources 
(annotation guidelines and corpora) that would enable the joint theoretical and empirical
research of multiple textual meaning relations.

\subsection{Other Related Work}\label{ch:rel:oth}

\textbf{Distributional Semantics} (DS) is the predominant framework for representing and 
comparing the meaning of linguistic units in contemporary Computational Linguistics (CL) 
and Natural Language Processing (NLP). DS has an important role both in theoretical 
research and in developing practical applications. The core hypothesis in DS is the 
Distributional Hypothesis (DH), as formulated by different authors:

\begin{quote}
\textit{``Difference in meaning correlated with difference in distribution''} \\ \citep{Harris}

\textit{``You shall know a word by the company it keeps''} \\ \citep{Firth}

\textit{``The meaning of a word is its use in the language''} \\ \citep{wittgenstien53english}
\end{quote}

While these authors formulate DH in slightly different ways, the central assumption remains
the same and can be stated as follows: \textit{``similar (or semantically related) linguistic 
units appear in similar contexts''}. This assumption allows for a radical empirical
approach towards formalizing the meaning of linguistic units. There exist many Distributional
Semantic Models (DSM) for representing the meaning of words or complex language 
expressions. \cite{BaroniLenci}, \cite{TurneyPantel}, and \cite{TACL457} compare different 
DSMs. More recently, the popular DSMs are based on neural network architectures (Word2Vec 
\citep{word2vec}, Glove \citep{glove}, Skip-Thought \citep{skipt}, InferSent \citep{infersent},
and ELMO \citep{peters2018contextualized}.
DSMs are used in many practical applications. They are also very popular for empirical tasks focused
on textual meaning relations.  Paraphrasing, Textual Entailment, and Semantic Textual Similarity
are often considered evaluation benchmarks for the quality of DSMs.

Within CL and NLP there are many empirical tasks focused on meaning relations at the 
level of tokens (i.e.: words and multi-word expressions), henceforth \textbf{``lexical meaning 
relations''}. \cite{hill2015simlex999} and 
\cite{DBLP:journals/jair/BruniTB14} propose datasets for out-of-context lexical similarity, while
\cite{10.5555/2390524.2390645} and \cite{levy-etal-2015-tr9856} propose datasets for
context-sensitive lexical similarity. \cite{kremer-etal-2014-substitutes} present a dataset for the
``lexical substitution'' task. \cite{10.5555/1859664.1859670} propose the task of ``relation
classification'' at the lexical level.

There are many manually created resources for studying and processing lexical meaning
 relations. These resources include, for example, lists of words with a particular relation, 
morphological rules for creating a particular relation (e.g.: ``happy'' - ``unhappy'', ``agree''- 
``disagree'') and  knowledge bases such as WordNet \citep{wordnet}, WikiData 
\citep{10.1145/2187980.2188242}, DBPedia \citep{dbpedia_iswc} and ConceptNet 
\citep{speer-havasi-2012-representing}. The tasks and resources for lexical meaning relations
are also relevant for the research on textual meaning relations.


Textual Meaning Relations also have an impact on \textbf{Other Areas of CL and NLP}
Systems that can successfully process meaning relations can also be used in other tasks in CL and
NLP, such as text summarization \citep{lloret2008text,harabagiu2010using}, text 
simplification \citep{yimam-biemann-2018-par4sim}, plagiarism detection \citep{Barron}, 
question answering \citep{harabagiu2006methods},  and machine translation evaluation 
\citep{pado2009textual}, among others.

	\section{Motivation and Objectives of the Thesis}\label{ch:objectives}

	This thesis arose from an interest in applying linguistic knowledge to the empirical studies 
of textual meaning relations. My research was motivated by two gaps in the research field:

\begin{itemize}
	\item a lack of large-scale corpora for research on decomposing textual meaning 
relations and, as a consequence, a lack of machine learning experiments.
	\item insufficient resources (annotation guidelines and corpora) and a lack of empirical studies 
on multiple textual meaning relations. 
\end{itemize}

I address both these gaps in turn by combining theoretical knowledge and empirical, 
data-driven approaches (human judgments, statistical analysis, and machine learning
experiments). First, I bring together paraphrase typology and the task of 
Paraphrase Identification. Second, I present a joint study on the textual meaning relations
of Paraphrasing, Textual Entailment, and Semantic Similarity. In the rest of this section
I present in more detail the objectives behind each of the two research directions of
my thesis.

\subsection{Paraphrase Typology and Paraphrase Identification}
The work on Paraphrase Typology (PT) uses knowledge from theoretical linguistics to 
understand the Paraphrasing phenomenon. Paraphrase Identification (PI) 
is an empirical task that aims to produce systems capable of recognizing paraphrasing in
an automatic manner. However, at the time of beginning this dissertation, there had been almost 
no interaction or intersection between these two areas of Paraphrasing research. PT 
research, prior to this dissertation, was mostly theoretical, with very limited practical 
implications and applications. PI research in the era of deep learning is radically 
empirical, focused on quantitative performance, with little to no interpretability and theoretical 
justification. My intuition was that these two research areas are not mutually exclusive, 
however there was no previous work trying to combine them. My \textbf{objectives} in 
combining PT and PI were twofold:

\begin{itemize}
	\item[Obj1] To use linguistic knowledge and paraphrase typology in order to improve the
	evaluation and interpretation of automated PI systems.
	\item[Obj2] To empirically validate and quantify the difference between the various 
	linguistic and reason-based phenomena involved in paraphrasing.
\end{itemize}


%


\subsection{Joint Study on Meaning Relations}
Meaning relations, such as Paraphrasing, Textual Entailment, and Semantic Similarity, have
attracted a lot of attention from the researchers in Computational Linguistics (CL) and Natural 
Language Processing (NLP). There is a substantial amount of theoretical and empirical 
research on these meaning relations and many resources, datasets, and automated systems. 
Traditionally, these relations have been studied in isolation and the transfer of knowledge and 
resources between them has been very limited. My intuition was that these textual meaning
relations can be brought together in a single corpus and compared empirically. My 
\textbf{objectives} in this part were twofold:

\begin{itemize}
	\item[Obj3] To empirically determine the interactions between Paraphrasing, Textual 
	Entailment, Contradiction, and Semantic Similarity in a corpus of multiple textual meaning
	relations.
	\item[Obj4] To propose and evaluate a novel shared typology of meaning relations. The 
	shared typology would then be used as a conceptual framework for joint research on 
	meaning relations.
\end{itemize}


	\section{Thesis Development}\label{ch:development}

My research has three separate phases, described in parts \ref{p:dsm}, \ref{p:type}, and 
\ref{p:mrel} of this thesis. 
First, I explore the basic concepts of Distributional Semantics and the notion of Semantic 
Similarity at the level of words and short phrases in Part \ref{p:dsm}. 
Second, I present my empirical research on bringing together Paraphrase Typology and 
Paraphrase Identification in Part \ref{p:type}. 
Finally, I describe the setup and results of my joint study on multiple textual meaning 
relations in Part \ref{p:mrel}.

The order of the chapters follows the chronological order in which the articles were written. 
At the same time, the order of the chapters follows the logical progression of my dissertation.
Each of the articles is self sufficient: it poses its own research questions, presents related 
work, proposes a methodology, and describes the experimental results. However,
there is also a clear thread that connects all the articles. When brought together, the articles
tell a coherent story about the linguistic phenomena involved in textual meaning relations and
how these phenomena can be used to improve the evaluation and interpretation of 
automated systems and bring together multiple textual meaning relations.

In the rest of this section I briefly present the main motivation, research questions and findings for
each of the three parts and the logical progression of the thesis. I also discuss how each article 
fits within the more general objectives and how the the different articles interact with each
other.

\myparagraph{Part \ref{p:dsm}: Lexical Relations and Distributional Semantics}

The two articles presented in this part of the thesis serve as an introduction to the research on meaning 
relations and aim to familiarize the reader with the core concepts and theories used in the whole 
thesis. 

In the article \textit{``Comparing Distributional Semantics Models for identifying groups of 
semantically related words''} (Chapter \ref{ch:sepln}), I explore the theoretical concepts 
and the empirical tools within the framework of Distributional Semantics (DS). DS is the most 
popular framework in contemporary Natural Language Processing (NLP) and Computational 
Linguistics (CL) and in the research on lexical and textual meaning relations. I experiment 
with different methodologies for representing the meaning of individual words, different ways 
to quantitatively compare meaning representations, and different approaches to measuring 
semantic similarity at the level of words. Lexical similarity is the most ``atomic'' form of 
semantic similarity. Many aspects of lexical similarity are also important for semantic similarity 
at the level of longer pieces of texts. 

In the article \textit{`` DISCOver: DIStributional approach based on syntactic dependencies 
for discovering COnstructions''} (Chapter \ref{ch:discover}), I present a successful data-driven 
methodology that can compose individual words into short phrases. The methodology is based 
on Distributional Semantics, lexical semantic similarity, and syntactic similarity between words. 
Many of the resulting short phrases are novel and have never been observed in the training
data, indicating that the system is composing as opposed to memorizing. This article demonstrates 
the importance of lexical similarity in the context of complex language expressions and the 
compositionality of meaning.

\myparagraph{Part \ref{p:type}: Paraphrase Typology and Paraphrase Identification}

The three articles presented in this part of the thesis tell a coherent story of how the theoretical concepts
of Paraphrase Typology (PT) research can be validated empirically and, at the same time, can 
be used to improve the evaluation, interpretation, and, indirectly, the performance of the 
automated Paraphrase Identification (PI) systems. 

In the article \textit{``WARP-Text: a Web-Based Tool for Annotating Relationships between 
Pairs of Texts.''} (Chapter \ref{ch:warp}), I describe the workings of a novel web-based 
annotation interface. The annotation interfaces that existed at the beginning of this thesis 
were not capable of performing a simultaneous annotation of multiple texts with fine-grained linguistic 
phenomena. WARP-Text fills this gap in the CL and NLP toolbox and creates new opportunities 
for researchers.

In the article \textit{``ETPC - a paraphrase identification corpus annotated with extended 
paraphrase typology and negation''} (Chapter \ref{ch:etpc}), I present the first PI corpus 
annotated with paraphrase types. I also propose a new extended typology for the 
paraphrasing relation, enriching the existing work in the area. I analyze the distribution of 
different linguistic phenomena in the corpus, and I identify general tendencies and potential 
biases in the data. This corpus-based study is the first large-scale empirical research on 
Paraphrase Typology within Paraphrase Identification. It contrasts with the pre-existing work in 
the field, in which researchers typically annotate a small number of examples. This article is 
also the first work on paraphrase typology that analyzes both positive examples (paraphrases) 
and negative examples (non-paraphrases). It contrasts with the pre-existing work in the field, 
which focuses only on the positive examples. The ETPC corpus makes further 
machine learning based studies on Paraphrase Typology possible. 

In the article \textit{``A Qualitative Evaluation Framework for Paraphrase Identification''}
(Chapter \ref{ch:peval}), I perform multiple machine learning experiments on the ETPC
corpus. I re-implement 11 different machine learning systems and create an ``evaluation
framework'' - a software package that can quantify and compare the paraphrase types involved 
in the correct and incorrect prediction of each PI system. I empirically demonstrate that
1) the different paraphrase types are processed differently by the different state-of-the-art
automated PI systems; and 
2) some paraphrase types are easier or harder for all evaluated systems. Furthermore,
I demonstrate that the ``qualitative evaluation framework'' provides much more 
information when comparing automated systems and facilitates error analysis.

\myparagraph{Part \ref{p:mrel}: A Joint Study of Meaning Relations}

The two articles presented in this part of the thesis demonstrate that multiple textual meaning relations can 
co-exist in the same corpus and can be expressed using the same ``atomic'' linguistic 
phenomena. 

In the article \textit{``Annotating and analyzing the interactions between meaning relations''}, I
present the first corpus to explicitly annotate the meaning relations of  Paraphrasing,
Textual Entailment, Contradiction, Semantic Similarity, and Textual Specificity. I propose a
methodology for corpus creation and annotation that guarantees that all relations are 
presented with a sufficient frequency. I compare the reliability of the annotation and the
inter-annotator agreement across all relations. Finally, I perform an empirical analysis of
the frequency, correlation, and overlap between the different meaning relations.

In the article \textit{``Decomposing and Comparing Meaning Relations: Paraphrasing, Textual 
Entailment, Contradiction, and Specificity''} I propose SHARel - a shared typology for 
Paraphrasing, Textual Entailment, Contradiction, Semantic Similarity, and Textual 
Specificity. I demonstrate that a single typology can successfully be applied to all textual
meaning relations. I analyze the distribution of the types across all relations and I outline 
common tendencies and differences.

	\section{Thesis Outline}\label{ch:outline}

	This thesis consists of a collection of seven papers, complemented by an 
introductory and a concluding chapter that provide the necessary context
to make the thesis a coherent story. The seven papers are the following:

\begin{center}
\textbf{Part \ref{p:dsm}: Similarity at the Level of Words and Phrases}
\end{center}

\begin{itemize}
	\item[1.] Venelin Kovatchev, M. Ant{\`o}nia Mart{\'i}, and Maria Salam{\'o}. 2016. Comparing Distributional Semantics Models for identifying groups of 	semantically related words. \textit{Procesamiento del Lenguaje Natural} vol. 57, pp.: 109-116

	\item[2.] M. Ant{\`o}nia Mart{\'i}, Mariona Taul{\'e}, Venelin Kovatchev,  and Maria Salam{\'o}. 2019. DISCOver: DIStributional approach based on syntactic dependencies for discovering COnstructions. \textit{Corpus Linguistics and Linguistic Theory}

\end{itemize}

\begin{center}
\textbf{Part \ref{p:type}: Paraphrase Typology and Paraphrase Identification}
\end{center}

\begin{itemize}

	\item[3.] Venelin Kovatchev, M. Ant{\`o}nia Mart{\'i}, and Maria Salam{\'o}. 2018. WARP-Text: a Web-Based Tool for Annotating Relationships between Pairs of Texts. 	\textit{Proceedings of the 27th International Conference on Computational Linguistics: System Demonstrations}, pp.: 132-136

	\item[4.] Venelin Kovatchev, M. Ant{\`o}nia Mart{\'i}, and Maria Salam{\'o}. 2018. ETPC - a paraphrase identification corpus  annotated with extended paraphrase typology and negation. \textit{Proceedings of the Eleventh International Conference on Language Resources and Evaluation}, pp.: 1384-1392

	\item[5.] Venelin Kovatchev, M. Ant{\`o}nia Mart{\'i}, Maria Salam{\'o}, and Javier Beltran. 2019. A Qualitative Evaluation Framework for Paraphrase Identification. \textit{Proceedings of the Twelfth Recent Advances in Natural Language Processing Conference}, pp.: 569-579

\end{itemize}

\begin{center}
\textbf{Part \ref{p:mrel}: Paraphrasing, Textual Entailment, and Semantic Similarity}
\end{center}

\begin{itemize}

	\item[6.] Darina Gold, Venelin Kovatchev, Torsten Zesch. 2019. Annotating and analyzing the interactions between meaning relations \textit{Proceedings of the Thirteenth Language Annotation Workshop}, pp.: 26-36

	\item[7. ] Venelin Kovatchev, Darina Gold, M. Ant{\`o}nia Mart{\'i}, Maria Salam{\'o}, and Torsten Zesch. 2020. Decomposing and Comparing Meaning Relations: Paraphrasing, Textual Entailment, Contradiction, and Specificity. To appear in \textit{Proceedings of the Twelfth International Conference on Language Resources and Evaluation}, 2020

\end{itemize}

All seven papers have been accepted and published in peer reviewed journals or 
conference proceedings. They 
are co-authored by both my advisors, with one exception: Paper 6 was
written during my research stay at the University of Duisburg-Essen and is
co-authored with the Language Technology Group at that university. In
all papers, except paper 2, I am listed as the first author. In paper 2, I was 
responsible for the evaluation section of the article and part of the 
experimental setup. In paper 6, the first two authors (Darina Gold and
myself) contributed equally to the article and the names are in
alphabetical order.

The papers reprinted here have been reformatted to make the typography of the
thesis consistent, and all of the references and appendices have been integrated in a single
bibliography and appendix section at the end. The thesis also includes some
additional material, such as annotation guidelines, which were included in
the original papers as external web links.

\part{Similarity at the Level of Words and Phrases}\label{p:dsm}


\chapter[Comparing Distributional Semantics Models \\ for Identifying Groups of Semantically Related Words]{\centering Comparing Distributional Semantics Models for Identifying Groups of Semantically Related Words}\label{ch:sepln}
\chaptermark{Comparing DSM}

\begin{center}
Venelin Kovatchev, M. Ant{\`o}nia Mart{\'i}, and Maria Salam{\'o}

University of Barcelona

\vspace{10mm}

Published at \\ \textit{Procesamiento del Lenguaje Natural}, 2016 \\ vol. 57, pp.: 109-116
\end{center}


\paragraph{Abstract} Distributional Semantic Models (DSM) are growing in popularity in Computational Linguistics. DSM use corpora of language use to automatically induce formal representations of word meaning. This article focuses on one of the applications of DSM: identifying groups of semantically related words. We compare two models for obtaining formal representations: a well known approach (\textsc{CLUTO}) and a more recently introduced one (Word2Vec). We compare the two models with respect to the PoS coherence and the semantic relatedness of the words within the obtained groups. We also proposed a way to improve the results obtained by Word2Vec through corpus preprocessing. The results show that: a) \textsc{CLUTO} outperforms Word2Vec in both criteria for corpora of medium size; b) The preprocessing largely improves the results for Word2Vec with respect to both criteria.

\paragraph{Keywords} DSM, Word2Vec, CLUTO, semantic grouping

\section{Introduction}
\label{1:s:intro}

In recent years, the availability of large corpora and the constantly increasing computational power of the modern computers have led to a growing interest in linguistic approaches that are automated and data-driven \citep{arppe}. Distributional semantic models (DSM) \citep{TurneyPantel,BaroniLenci} and the vector representations (VR) they generate fit very well within this framework: the process of extracting vector representations is mostly automated and the content of the representations is data-driven.

The format of the vector is suitable for carrying out different mathematical manipulations. Vectors can be compared directly through an objective mathematical function. They can also be used as a dataset for various Machine Learning algorithms. VR are more often used on tasks related to lexical similarity and relational similarity \citep{TurneyPantel}. In such tasks, the emphasis is on pairwise comparisons between vectors.

This article focuses on another use of the Vector Representations: the grouping of vectors, based on their similarity in the Distributional space. This grouping can be used, among other things, as a methodology for identifying groups of semantically related words. High quality groupings can serve for many purposes: they are a semantic resource on their own, but can also be applied for syntactic disambiguation or pattern identification and generation \citep{DISCOver}, for example.

We compare two different methodologies for obtaining groupings of semantically related words in English - a well known approach (\textsc{CLUTO}) and a more recently introduced one (Word2Vec). The two methodologies are evaluated in terms of the quality of the obtained groups. We consider two criteria: 1) the semantic relatedness between the words in the group; and 2) the PoS coherence of the group. We evaluate the role of the corpus size with both methodologies and in the case of Word2Vec, the role of the linguistic preprocessing (lemmatization and PoS tagging).

The rest of this paper is organized as follows: Section \ref{1:s:related} presents the general framework and related work. Section \ref{1:s:data} describes the available data and tools.  Section \ref{1:s:experiments} presents the experiments and the results obtained. Finally Section \ref{1:s:conclusions} gives conclusions and identifies directions for future work.

\section{Related Work}
\label{1:s:related}

Distributional Semantics Models (DSM) are based on the Distributional Hypothesis, which states that the meaning of a word can be represented in terms of the contexts in which it appears \citep{Harris,Firth}. As opposed to semantic approaches based on primitives \citep{BoledaErk}, approaches based on distributional semantics can obtain formal representations of word meaning from actual linguistic productions. Additionally, this data-driven process for semantic representation can mostly be automated.

Within the framework of DSM, one of the most common ways to formalize the word meaning is a vector in a multi-dimensional distributional space \citep{Lenci}. For this purpose, a matrix with size \textbf{m} by \textbf{n} is extracted from the corpus, representing the distribution of \textbf{m} words over \textbf{n} contexts. The format of a vector allows for direct quantitative comparison between words using the apparatus of linear algebra. At the same time it is a format preferred by many Machine Learning algorithms.

The choice of the matrix is central for the implementation of a particular DSM. \citet{TurneyPantel} suggest a classification of the DSM based on the matrix used. They analyze three different matrices: term-document, word-context, and pair-pattern. The different matrices represent different types of relations in the corpus and the choice of the matrix depends on the goals of the particular research.

\citet{BaroniLenci} present a different, sophisticated approach for extracting information from the corpus. They organize the information as a third order tensor,  with the dimensions representing \textless `word', `link', `word' \textgreater. This third order tensor can then be used to generate different matrices, without the need of going back to the original corpus.

In this paper we focus on one of the classical vector representations - the one based on word-context relation. It measures what \citet{TurneyPantel} call ``attributional similarity''. In particular, we are interested in the possibility to group vectors together, based on their relations in the distributional space.

\citet{Erk} offers a survey of possible applications of different DSM. She lists clustering as an approach that can be used with vectors, for word sense disambiguation. \citet{Moisl} presents a theoretical analysis on the usage of clustering in computational linguistics and identifies key aspects of the mathematical and linguistic argumentation behind it.

Here we analyze and compare two approaches that induce vector representations from a corpus and apply algorithms to identify sets of semantically related words. We are interested in the quality of the obtained groups, as we believe that they can be a useful, empirical, linguistic resource.

\citet{DISCOver} present a methodology named DISCOveR for identifying candidates to be constructions from a corpus. As part of this methodology they use \textsc{CLUTO} \citep{Karypis} for clustering words based on their vector representations. Their approach uses a word-context matrix where the context is defined by combining a syntactic dependency with a lemma. After all the vectors are extracted, \textsc{CLUTO} is used in order to obtain clusters of semantically related words. Later on these clusters are used to generate a list of the candidates to be constructions.

\citet{Mikolov} suggest a different approach towards extracting vector representations and grouping. Their methodology is based on deep learning and is intended for quick processing of very large corpora. Word2Vec\footnote{Available at:  \url{https://code.google.com/archive/p/word2vec/}}, the tool they present, includes an integrated algorithm for grouping words based on proximity in space. The context they use for vector extraction is simple co-occurrence within a specified window of tokens. Originally, they make no use of linguistic preprocessing such as lemmatization, part of speech tagging or syntactic tagging. As part of this paper we evaluate the effect of linguistic preprocessing on the obtained vectors and groups.

\section{Data and Tools}
\label{1:s:data}

In this section we present the corpus that we use in the evaluation (Section \ref{1:ss:d:corpus}) and the two methodologies ({Section \ref{1:ss:d:cluto} and Section \ref{1:ss:d:w2v}).

\subsection{The Corpus}
\label{1:ss:d:corpus}

For all of the experiments described in this paper, we use PukWaC \citep{Wacky}\footnote{Available at: \url{http://wacky.sslmit.unibo.it}}. It is a 2 billion word corpus of English, built up from sites in the .uk domain. It is available online and is already preprocessed: XML tags and other non-linguistic information have been removed, it is lemmatized, PoS tagged and syntactically parsed. The PoS tagset is an extended version of the Penn Treebank tagset. The syntactic dependencies follow the CONLL-2008 shared task format.

\subsection{Grouping with \textsc{CLUTO}}
\label{1:ss:d:cluto}

DISCOveR \citep{DISCOver} is a methodology for identifying candidates to be construction from a corpus. It uses vector representations, extracted from a corpus. \textsc{CLUTO} \citep{Karypis} is used on these representations in order to obtain clusters of semantically related words. \textsc{CLUTO} is a software package for clustering low and high dimensional data sets and for analysis of the characteristics of the various clusters. \textsc{CLUTO} provides three different classes of clustering algorithms, based on partitional, agglomerative and graph-partitioning paradigms. It computes clustering solution based on one of the different approaches.

For this article, we are interested only in the first three steps of the DISCOveR process. Step 1 is the linguistic preprocessing of the corpus. The raw text is cleared from non-linguistic data, it is PoS tagged and syntactically parsed. In Step 2, the DSM matrix is constructed. The rows of the matrix correspond to lemmas  and the columns correspond to contexts. Contexts in this approach are defined as a triple of syntactic relation, direction of the relation and lemma in [direction:relation:lemma] format\footnote{For example, from the sentence ``El barbero afeita la larga barba de Jaime'', three different contexts of the noun lemma barba are generated: [\textless :dobj:afeitar\_v] , [\textgreater :mod:largo\_a] and [\textgreater :de\_sp:pn\_n]. The example is from \citet{DISCOver} }. This matrix is used to generate vector representations for the 10,000 most frequent words in the corpus. Next, Step 3 uses \textsc{CLUTO} to create clusters of semantically related lemmas from the DSM matrix and the corresponding vectors. The clusters are created based on shared contexts.

\citet{DISCOver} start from a raw, unprocessed corpus and in Step 1 they clear the corpus and tag it with the linguistic data relevant to the matrix extraction. The format they use is shown in Table \ref{1:data:DAf}.

\begin{table}[h]
    \begin{center}
        \begin{tabular}{ |c|c| }

            \hline
            Token & sanitarios \\
            \hline
            Lemma & sanitario \\
            \hline
            PoS & NCMP \\
            \hline
            Short PoS & n \\
            \hline
            Sent ID & 000 \\
            \hline
            Token ID & 0 \\
            \hline
            Dep ID & 2 \\
            \hline
            Dep Type & suj \\
            \hline

        \end{tabular}
    \end{center}
    \caption{Diana-Araknion Format}
    \label{1:data:DAf}
\end{table}
\vspace{4mm}

The original DISCOveR experiment is done with the Diana-Araknion corpus of Spanish. For the purpose of this article, we replicated the process for English, using the PukWaC corpus.  For step 1 we had to make sure that our preprocessing is equivalent to the one of Diana-Araknion. The corpus PukWaC is already preprocessed and the format is similar to the one of Diana-Araknion. However, in order to make it fully compatible, we had to make several modifications of the format and linguistic decisions. Regarding the format, we removed any remaining XML tags, enumerated the sentences in the corpus, and generated ``short PoS''\footnote{short PoS is a one letter tag representing the generic PoS tag of the lemma. In this experiment, short PoS is the first letter of the full PoS}. From the linguistic side, we had to decide whether all PoS and Dependencies were relevant for the vector generation or some of them could be merged together or even discarded in order to optimize and speed up the process.

The process of generating vectors and clusters is based on analyzing the contexts where each word appears in. A word is identified by its lemma and its PoS tag. However, in the PukWac tagset there are many PoS tags which specify not only the PoS of the token, but also contain information about other grammatical features, such as person, number, and tense. If these tags are kept unchanged, a separate vector will be generated for different forms of the same word, based on different PoS tag. To avoid this problem and to generate only one vector for all of the different word forms, we have decided to merge certain PoS tags under one category.

We decided to simplify the POS tagset further. It is a common practice in DSM to focus the experiment on the relations between content words. Function words and punctuation are usually not considered relevant contexts. Because of that, we have put them under the common tag ``other''. All of the changes on the PoS tagset are summarized in Table \ref{1:data:Plist}.

\begin{table}[h]
    \begin{center}
        \begin{tabular}{ |c|c|c| }
        \hline
            \textbf{Tag} & Original tag & Description\\
            \hline\hline
            \textbf{J} & JJ JJR JJS & Adjective\\
            \hline
            \textbf{M} & MD & Modal verb\\
            \hline
            \textbf{N} & NN NNS & Noun (common)\\
            \hline
            \textbf{NP} & NP NPS & Noun (personal)\\
            \hline
            \textbf{R} & \begin{tabular}{@{}c@{}}RB RBR \\RBS RP\end{tabular} & Adverb\\
            \hline
            \textbf{S} & IN & Preposition \\
            \hline
            \textbf{V} & VB* VH* VV* & Verb (all) \\
            \hline
            \textbf{O} & \begin{tabular}{@{}c@{}}CC CD DT\\ PDT EX FW\\ LS POS PP* \\ SYM TO UH\\ W* punctuation \end{tabular}& Rest \\
        \hline
        \end{tabular}
    \end{center}
    \caption{PoS tagset modifications}
    \label{1:data:Plist}
\end{table}

The list of syntactic dependencies in PukWaC is also not fully relevant to the task of vector generation. While the unnecessary PoS tags may lead to multiple vectors for the same word, unnecessary dependencies generate additional contexts, increasing the dimensionality of the vectors and leading to a more complicated computational process. Therefore the modification of the dependencies is mostly related to the optimization of the computational process. After analyzing the tagset, we have decided to merge the \textbf{OBJ} and \textbf{IOBJ} tags due to some inconsistencies of their usage. We have also decided to discard the following relations: \textbf{CC} (conjunction), \textbf{CLF} (be/have in a complex tense), \textbf{COORD} (coordination), \textbf{DEP} (unclassified relation), \textbf{EXP} (experiencer in few very specific cases), \textbf{P} (punctuation), \textbf{PRN} (parenthetical), \textbf{PRT} (particle), \textbf{ROOT} (root clause).The final list of dependencies is shown in Table \ref{1:data:Dlist}.

\begin{table}[h]
    \begin{center}
        \begin{tabular}{ |c|c| }
        \hline
            \textbf{Dependency} & Description \\
            \hline\hline
            \textbf{ADV} & Unclassified adv \\
            \hline
            \textbf{AMOD} & Modifier of adj or adv \\
            \hline
            \textbf{LGS} & Logical subj \\
            \hline
            \textbf{NMOD} & Modifier of nom \\
            \hline
            \textbf{OBJ} & Direct or indirect obj \\
            \hline
            \textbf{PMOD} & Preposition\\
            \hline
            \textbf{PRD} & Predicative compl \\
            \hline
            \textbf{SBJ} & Subject \\
            \hline
            \textbf{VC} & Verb chain \\
            \hline
            \textbf{VMOD} & Modifier of verb \\
            \hline
            \textbf{empty} & No dependency\\
        \hline
        \end{tabular}
    \end{center}
    \caption{Syntactic Dependencies}
    \label{1:data:Dlist}
\end{table}

Once the corpus is preprocessed, the process of matrix extraction is mostly automated. For the matrix, we have only generated vectors for words that appear at least 5 times in the corpus. Out of them we have used only the vectors of the 10,000 most frequent words for the clustering process.

For the clustering process, we configure \textsc{CLUTO} to use direct clustering, based on the H2 criterion function, with 25 features per cluster. We have ran the clusterization multiple times, ranging from 100 to 1,000 clusters. We then used \textsc{CLUTO}'s H2 metric to determine the optimal number of clusters, which has been 800 for all of the experiments.

\subsection{Grouping with Word2Vec}
\label{1:ss:d:w2v}

Word2Vec is based on the methodology proposed by \citet{Mikolov}. It takes a raw corpus and a set of parameters and generates vectors and groups. The algorithm of Word2Vec is based on a two layer neural network that are trained to reconstruct linguistic context of words. Word2Vec includes two different algorithms - Continuous Bag-of-Words (CBOW) and Skip-Gram. CBOW learns representations based on the context as a whole - all of the words that co-occur with the target word in a specific window. Skip-Gram learns representation based on each single other word within a specified window. When using Word2Vec usually the emphasis is put on the choice of the paremeters for the algorithm, and not on the specifications of corpus. However, we consider that the specifications of the corpus (size and linguistic preprocessing) can largely affect the quality of the obtained results.

By default Word2Vec works with a raw corpus. Neither of the two models makes explicit use of morpho-syntactic information. However, by modifying the corpus, some morphological information can be used implicitly. If the token is replaced by its corresponding lemma or by the lemma and part of speech tag in a ``lemma\_pos'' format, the resulting vectors would be different: using the lemma would generate only one vector for the word as opposed to separate vector for every word form; using PoS can make a distinction between homonyms with same spelling and different PoS. As part of our work we wanted to examine how linguistic preprocessing can affect the quality of the vectors. For that reason we created three separate corpus samples - one raw corpus, one where each token was replaced by its lemma, and one where each token was replaced by ``lemma\_pos''. We generated vectors separately for each of the corpora. Unfortunately, there was no trivial way to introduce syntactic information implicitly in the models of Word2Vec.

\section{Experiments}
\label{1:s:experiments}

In this section we present the setup for the different experiments (Section \ref{1:ss:e:setup}), the evaluation criteria (Section \ref{1:ss:e:eval}), and the obtained results (Section \ref{1:ss:e:results}).

\subsection{Setup}
\label{1:ss:e:setup}

We carried out a total of 15 experiments - 3 experiments using \textsc{CLUTO} and 12 experiments using Word2Vec.
For the experiments with \textsc{CLUTO}, the only variation between the experiments was the size of the corpus: 4M tokens, 20M tokens, and 40M tokens\footnote{The 40M corpus contains in itself the 20M corpus. The 20M corpus contains in itself the 4M corpus. The same corpora has been used for the experiments with both \textsc{CLUTO} and with Word2Vec.}. In all the experiments we used the preprocessing described at Section \ref{1:ss:d:cluto}, we generated vectors for the 10,000 most frequent words and we split them into 800 clusters.
For the experiments with Word2Vec, we changed three parameters of the experiments: (1) the algorithm (CBOW and Skip-Gram), (2) the linguistic preprocessing of the corpus (raw, lemma, lemma and PoS), and (3) the size of the corpus (4M, 20M, and 40M). We carried out 9 experiments with CBOW (all size and preprocessing combinations) and 3 experiments with Skip-Gram (the three variants of the 40M corpus). \citet{Mikolov} identify two important parameters to be set up when using Word2Vec: the vector size and the window size. For the window size, we used 8, which is the recommended value. For the vector size, \citet{Mikolov} show that increasing vector size from 100 to 300 leads to significant improvement of the results, however further increase does not have big impact. For that reason we have chosen vector size of 400, which is above the recommended minimum. For the number of groups we used 800: the same number that was determined optimal for \textsc{CLUTO}. For the number of lemmas, we used the 10,000 most frequent ones, the same setup as with \textsc{CLUTO}.

\subsection{Evaluation}
\label{1:ss:e:eval}

The two methodologies and all of the different setups are evaluated based on the quality of the obtained groups. We consider two criteria: 1) The semantic relatedness between the words in each group; and 2) The PoS coherence of the groups. The PoS coherence is a secondary criterion which should be considered in addition to the semantic relatedness. Our intuition is that groups that are semantically related and PoS coherent are a better resource than groups that are only semantically related. For evaluating the semantic relations of the words in the groups, we present two methodologies - an automated method based on WordNet distances and a manual evaluation done by experts on a subset of the groups in each experiment. The PoS coherence is calculated automatically.

There is no universal widely accepted criteria for determining the semantic relations between two words. Two of the most common approaches are calculating WordNet distances and expert intuitions. We used both when evaluating the quality of the obtained groups.

For the WordNet similarity evaluation, we use the WordNet interface built in NLTK \citep{nltk}. We calculate the Leacock-Chodorow Similarity\footnote{It calculates word similarity, based on the shortest path that connects the senses and the maximum depth of the taxonomy in which the senses occur.} between each two words\footnote{The calculation is based on the first sense of every word} in every group. We then sum all the obtained scores and divide them by the number of pairs to obtain average WordNet similarity for each method.

For the expert evaluation, we selected a subset of groups, generated in each experiment\footnote{We selected the groups based on a word they contain - three verb groups (the ones that contain ``say'', ``see'', ``want''), 3 noun groups (``person'', ``year'', ``hand''), 1 adjective group(``good''), 1 adverb group(``well). All of the selected words are among the 100 most commonly used words of English.)}. Three experts were asked to rate each group on a scale from 1 (unrelated) to 4 (strongly related)\footnote{In the detailed description of the scale given to the experts: 1 corresponds to ``no semantic relation''; 2 corresponds to ``semantic relation between some words (less than 50\% of the group); 3 corresponds to ``semantic relation between most of the words in the corpus (more than 50\%), but with multiple unrelated words''; 4 corresponds to ``semantic relation between most of the words in the corpus, without many unrelated words''}. We calculate the average between all of the scores they gave on the groups of each experiment.

We define PoS coherence as the percent of words that belong to the most common PoS tag in each group. In order to calculate it, all obtained groups are automatically PoS tagged\footnote{We use only the short PoS tag for this evaluation}. Then for each group, we count the percent of words that belong to each PoS and identify the most common tag.

\subsection{Results}
\label{1:ss:e:results}

Table \ref{1:results:wordnet} shows the WordNet similarity evaluation. The average similarity score obtained by \textsc{CLUTO} is higher than the score obtained by Word2Vec (0.81-0.96 against 0.67-0.81). This indicates that the distances between the words in the \textsc{CLUTO} groups are shorter and the semantic relations are stronger. Increasing the corpus size improves the results for both \textsc{CLUTO} and Word2Vec. Preprocessing (specifically PoS tagging) improves the obtained results for all of the Word2Vec experiments. The groups obtained using Skip-Gram get lower scores in the evaluation compared with the groups obtained using CBOW.

\begin{table}[h]
    \begin{center}
        \begin{tabular}{ |c|c|c| }
        \hline
            \textbf{Methodology} & Corpus & Similarity \\
            \hline\hline
            \textbf{W2V-CBOW} & 4M (raw) & 0.67 \\
            \hline
            \textbf{W2V-CBOW} & 4M (lemma) & 0.67 \\
            \hline
            \textbf{W2V-CBOW} & 4M (pos) & 0.72 \\
            \hline
            \textbf{W2V-CBOW} & 20M (raw) & 0.74 \\
            \hline
            \textbf{W2V-CBOW} & 20M (lemma) & 0.75 \\
            \hline
            \textbf{W2V-CBOW} & 20M (pos) & 0.77 \\
            \hline
            \textbf{W2V-CBOW} & 40M (raw) & 0.77 \\
            \hline
            \textbf{W2V-CBOW} & 40M (lemma) & 0.78 \\
            \hline
            \textbf{W2V-CBOW} & 40M (pos) & 0.81 \\
            \hline
            \textbf{W2V-SG} & 40M (raw) & 0.69 \\
            \hline
            \textbf{W2V-SG} & 40M (lemma) & 73 \\
            \hline
            \textbf{W2V-SG} & 40M (pos) & 0.74 \\
            \hline
            \textbf{CLUTO} & 4M & 0.81 \\
            \hline
            \textbf{CLUTO} & 20M & 0.92 \\
            \hline
            \textbf{CLUTO} & 40M & 0.96 \\
            \hline
        \hline
        \end{tabular}
    \end{center}
    \caption{Wordnet Similarity}
    \label{1:results:wordnet}
\end{table}

Table \ref{1:results:sem} shows the results from the expert evaluation of the semantic relations in the groups. The data is similar to the results with WordNet distances. The groups obtained by \textsc{CLUTO} show higher degree of semantic relatedness (2.8-3.4) compared to the groups obtained by Word2Vec (1.6-2.7). The \textsc{CLUTO} groups at 20M and 40M obtain average above 3, meaning that the experts consider all of the groups to be strongly related. For the experiments with Word2Vec, linguistic preprocessing improves the results, especially at biger corpus size (2.5 against 1.8 for 20M and 2.7 against 2 for 40M). The groups obtained using Skip-Gram algorithm are rated lower than the groups obtained using CBOW. The preprocessed corpus obtains better groups, but the difference is smaller than the one observed with CBOW.

\begin{table}[h]
    \begin{center}
        \begin{tabular}{ |c|c|c| }
        \hline
            \textbf{Methodology} & Corpus & Score \\
            \hline\hline
            \textbf{W2V-CBOW} & 4M (raw) & 1.6 \\
            \hline
            \textbf{W2V-CBOW} & 4M (lemma) & 1.4 \\
            \hline
            \textbf{W2V-CBOW} & 4M (pos) & 1.8 \\
            \hline
            \textbf{W2V-CBOW} & 20M (raw) & 1.8 \\
            \hline
            \textbf{W2V-CBOW} & 20M (lemma) & 2.4 \\
            \hline
            \textbf{W2V-CBOW} & 20M (pos) &  2.5 \\
            \hline
            \textbf{W2V-CBOW} & 40M (raw) & 2 \\
            \hline
            \textbf{W2V-CBOW} & 40M (lemma) & 2.1 \\
            \hline
            \textbf{W2V-CBOW} & 40M (pos) & 2.7  \\
            \hline
            \textbf{W2V-SG} & 40M (raw) & 1.7 \\
            \hline
            \textbf{W2V-SG} & 40M (lemma) & 1.8 \\
            \hline
            \textbf{W2V-SG} & 40M (pos) & 2 \\
            \hline
            \textbf{CLUTO} & 4M & 2.8 \\
            \hline
            \textbf{CLUTO} & 20M & 3.2 \\
            \hline
            \textbf{CLUTO} & 40M & 3.4 \\
            \hline
        \hline
        \end{tabular}
    \end{center}
    \caption{Expert evaluation}
    \label{1:results:sem}
\end{table}

Table \ref{1:results:pos} shows the results for the PoS coherence evaluation. The data shows that the groups obtained from \textsc{CLUTO} are more PoS coherent, compared with the groups obtained by Word2Vec (90-98\% against 69-81\%). For the corpora of size 20M and above, the groups obtained by \textsc{CLUTO} have almost 100\% PoS coherence, meaning that all of the lemmas belong to the same PoS. Both \textsc{CLUTO} and Word2Vec show improved results with the increase of corpus size. The results with Word2Vec indicate that corpus preprocessing largely improves the obtained results (69\%-73\% against 75\%-81\%). In fact, for this experiment the corpus preprocessing have bigger impact than the corpus size: a preprocessed corpus with a size of 4M generates more PoS coherent groups than raw 40M corpus (74-75\% against 73\%). The experiments with Skip-Gram obtain similar results for raw corpus. For Skip-Gram the preprocessed corpus also obtains better overall results, however lemmatized corpus obtains better results than the PoS tagged corpus.

\begin{table}[h]
    \begin{center}
        \begin{tabular}{ |c|c|c| }
        \hline
            \textbf{Methodology} & Corpus & PoS \\
            \hline\hline
            \textbf{W2V-CBOW} & 4M (raw) & 69\% \\
            \hline
            \textbf{W2V-CBOW} & 4M (lemma) & 74\% \\
            \hline
            \textbf{W2V-CBOW} & 4M (pos) &  75\%\\
            \hline
            \textbf{W2V-CBOW} & 20M (raw) &  72\% \\
            \hline
            \textbf{W2V-CBOW} & 20M (lemma) & 77\% \\
            \hline
            \textbf{W2V-CBOW} & 20M (pos) & 80\% \\
            \hline
            \textbf{W2V-CBOW} & 40M (raw) & 73\% \\
            \hline
            \textbf{W2V-CBOW} & 40M (lemma) & 78\% \\
            \hline
            \textbf{W2V-CBOW} & 40M (pos) & 81\% \\
            \hline
            \textbf{W2V-SG} & 40M (raw) & 73\% \\
            \hline
            \textbf{W2V-SG} & 40M (lemma) & 80 \% \\
            \hline
            \textbf{W2V-SG} & 40M (pos) & 77\% \\
            \hline
            \textbf{CLUTO} & 4M & 90\% \\
            \hline
            \textbf{CLUTO} & 20M & 97\% \\
            \hline
            \textbf{CLUTO} & 40M & 98\%  \\
            \hline
        \hline
        \end{tabular}
    \end{center}
    \caption{PoS coherence}
    \label{1:results:pos}
\end{table}

Overall, all three evaluations identify similar patterns in the obtained clusters: (1) the groups obtained by \textsc{CLUTO} perform better than the groups obtained by Word2Vec; (2) Increasing the corpus size improves the quality of the results for both methodologies. This is true for semantic relatedness as well as for PoS coherence. The tendency to obtain more PoS coherent groups justifies the usage of PoS coherence as evaluation criteria; (3) Linguistic preprocessing improves the quality of the groups obtained by Word2Vec (with both algorithms).

\section{Conclusions and Future Work}
\label{1:s:conclusions}

This article compares two methodologies for identifying groups of semantically related words based on Distributional Semantic Models and vector representations. We applied the methodologies to a corpus of English and compared the quality of the obtained groups in terms of semantic relatedness and PoS coherence. We also analyzed the role of different factors, such as corpus size and linguistic preprocessing.

In the comparison of the two methodologies, the results show that \textsc{CLUTO} outperforms Word2Vec with respect to grouping, using corpora of medium size (20M - 40M). However, the quality of the results does depend on the size of the corpus. At 40M \textsc{CLUTO} already obtains very high quality results (98\% PoS coherence and 3.4/4 strength of semantic relationships in the evaluation of the experts) so further increase of the corpus is not likely to show large improvement. On the contrary at 40M Word2Vec still has room for improvement and we expect to narrow the difference between the two methodologies using much larger corpora (1B and above).

In the comparison of the different preprocessing corpora (i.e., raw, lemma, and PoS) in Word2Vec, the results show that lemmatization and PoS tagging largely improve the quality of the groups in both CBOW and Skip-Gram algorithms. This observation is consistent throughout all of the experiments and with respect to all of the evaluation criteria.

The presented comparison opens several lines of future research. First, the evaluation can be extended to bigger corpora, bigger number of vectors, and other languages. Second, the information provided and the suggested criteria for evaluation can be applied to other approaches to DSM and grouping. Finally, the different methodologies and preprocessing options can be evaluated in as part of more complex systems.


\chapter[DISCOver: DIStributional Approach Based on \\ Syntactic Dependencies for Discovering COnstructions]{\centering DISCOver: DIStributional Approach Based on Syntactic Dependencies for Discovering COnstructions}\label{ch:discover}

\chaptermark{DISCOver}

\begin{center}
M. Ant{\`o}nia Mart{\'i}, Mariona Taul{\'e}, Venelin Kovatchev,  and Maria Salam{\'o}

University of Barcelona

\vspace{10mm}

Published at \\ \textit{Corpus Linguistics and Linguistic Theory}, 2019

\end{center}

\paragraph{Abstract} One of the goals in Cognitive Linguistics is the automatic identification and analysis of constructions, since they are fundamental linguistic units for understanding language. This article presents DISCOver, an unsupervised methodology for the automatic discovery of lexico-syntactic patterns that can be considered as candidates for constructions. This methodology follows a distributional semantic approach. Concretely, it is based on our proposed pattern-construction hypothesis: those contexts that are relevant to the definition of a cluster of semantically related words tend to be (part of) lexico-syntactic constructions. Our proposal uses Distributional Semantic Models (DSM) for modeling the context taking into account syntactic dependencies. After a clustering process, we linked all those clusters with strong relationships and we use them as a source of information for deriving lexico-syntactic patterns, obtaining a total number of 220,732 candidates from a 100 million token corpus of Spanish. We evaluated  the patterns obtained intrinsically, applying statistical association measures and they were also evaluated qualitativaly by experts. Our results were superior to the baseline in both quality and quantity in all cases. While our experiments have been carried out using a Spanish corpus, this methodology is language independent and only requires a large corpus annotated with the parts of speech and dependencies to be applied.

\paragraph{Keywords} Constructions, Semantics, Distributional Semantic Models

\section{Introduction}

In cognitive models of language \citep{Croft:2004}, a construction is a conventional symbolic unit that involves a pairing of form and meaning that occurs with a certain frequency. Constructions can be of different types depending on their complexity --morphemes, words, compound words, collocates, idioms and more schematic patterns ~\citep{Goldberg:1995,Goldberg:2006}. 
Cognitive Linguistics assumes the hypothesis that these constructions are learned from usage and stored in the human memory~\citep{Tomasello:2000}, where they are accessed during both the production and comprehension of language. Therefore, constructions are fundamental linguistic units for inferring the structure of language and their identification is crucial for understanding language.

Although a broad range of these linguistic structures have been subjected to linguistic analysis ~\citep{nunberg1994idioms,Wray:2000,fillmore2012framenet}, we assume that there exist a huge number of constructions that are as yet undiscovered. There are very different approaches to the task of identifying and discovering them, depending on the type of construction we are looking for or dealing with. This fact allows for the use of a wide range of methods and approaches aiming at the treatment of this kind of linguistic units.
We distinguish between two different approaches, those that have been guided by previously gathered empirical data\footnote{ See~\cite{Goldberg:1995}.}, and those approaches that apply methods oriented to discovering new constructions from scratch (see Section~\ref{2:sec:relatedwork}).

Following the latter approach, this article presents DISCOver, an unsupervised methodology for the automatic identification and extraction of lexico-syntactic patterns that are candidates for consideration as constructions (see Section~\ref{2:sec:methodology}). It is based on the Harris distributional hypothesis~\citep{Harris} \footnote{This idea was also developed by Firth [1957] and Wittgenstein [1953].}, which states that semantically related words (or other linguistic units) will share the same context.\footnote{Related hypotheses, such as the extended distributional hypotheses, which states that ``patterns that co-occur with similar pairs tend to have similar meanings''~\citep{lin2001dirt}, and latent relation hypotheses~\citep{turney2008latent}, which states that ``pairs of words that co-occur in similar patterns tend to have similar relations'' survived in~\cite{TurneyPantel} have also influenced this work.} We propose the pattern-construction hypothesis, which states that those contexts that are relevant to the definition of a cluster of semantically related words tend to be (part of) lexico-syntactic constructions. What is new in our hypothesis is that we consider all the contexts that are relevant to define a cluster of semantically related words to be part of a construction. In these approaches, Distributional Space Models (DSMs) are used to represent the semantics of words on the basis of the contexts they share. This is in line with the idea proposed by \cite{landauer2007handbook}, who states that DSMs are plausible models of some aspects of human cognition~\citep{BaroniLenci}.

In our methodology, the DSM consists of a frequency lemma-context matrix, in which the context is modeled taking into account syntactic dependency relations. Then, we build up clusters of semantically related words that share the same context and link them using the information present in their contexts. We automatically calculate a threshold in order to determine which clusters are more strongly related. We filter out those related clusters that do not reach the determined threshold and derive lexico-syntactic patterns that are candidates to be considered as constructions. These candidates are tuples involving two lexical items (lemmas) related both by a dependency direction and a dependency label (examples in (1))\footnote{The symbols `$<$' and `$>$' indicate the dependency direction and \textit{mod}, \textit{subj} and \textit{dobj} are dependency labels (where \textit{mod} stands for modifier, and \textit{subj} and \textit{dobj} stand for subject and direct object respectively).}:

\begin{enumerate}
\item a. accidente\_n $[>$:mod:mortal\_a]\footnote{accident\_n$[>$:mod:mortal\_a]}\par
b. aterrizar\_v $[>$:dobj:avioneta\_n]\footnote{to\_land$[>$:dobj:small\_plane\_a]}\par
\end{enumerate}


The tuples correspond to different kinds of linguistic constructions, ranging from collocates (1a) to (parts of) verbal argument structures (1b). All the lexico-syntactic patterns obtained are instances of one of the syntactic dependencies present in the source corpus. We applied this methodology to the Diana-Araknion corpus, obtaining 220,732 patterns that are good candidates to be constructions\footnote{All patterns obtained are available at \url{http://clic.ub.edu/corpus/}}.

Finally, we evaluated the quality of these patterns in two ways: applying statistical association measures and by manual revision by human experts. The results show significant improvement with respect to several baselines (see Section~\ref{2:sec:evaluation}).

Although this method has been applied to the obtention of Spanish constructions, it is language independent and only requires a large corpus annotated with part-of-speech (POS) and syntactic dependencies.

The article is structured as follows. After presenting the related work in Section~\ref{2:sec:relatedwork}, the methodology applied for obtaining the constructions is described in Section~\ref{2:sec:methodology}. The evaluation of our methodology is presented in Section~\ref{2:sec:evaluation} and, finally, the conclusions and future work are drawn in  Section~\ref{2:sec:conclusions}.

\section{Related Work} \label{2:sec:relatedwork}

The boundaries of what a construction is are fuzzy: constructions can be lexical, syntactic, lexico-syntactic, morphological and can combine different levels of abstraction from concrete forms to abstract categories, including the possibility of using variables, so they cover a wide range of linguistic constructs. For more examples, see~\cite{goldberg2013argument}. 


As a consequence, there is no one accepted typology of this kind of linguistic units~\citep{Wray:2000}. There is, therefore, a broad field of research in which to explore the characteristics, the limits and the properties of constructions. In this context, an important task is to acquire the maximum amount of empirically grounded data concerning this kind of units. Thus, when approaching the task of attempting to identify the possible constructions that constitute the core of languages, it is difficult to decide what to look at or where to start ~\citep{sag:2002multiword}. For this reason, constructions are a challenge for Linguistics and Natural Language Processing (NLP), where we find statistical and symbolic approaches to deal with them.

Several linguistic traditions converge when we are trying to define the diverse form that a construction can take. From one side, there is an (almost total) overlapping between constructions and argument structure~\citep{Goldberg:1995} and diatheses alternations~\citep{levin1993english}; from another side, in the lexicographic tradition, constructions also overlap with idioms and collocates. In the field of Computational Linguistics, these linguistic units tend to be grouped under the umbrella term MultiWord Expressions (MWE). \cite{baldwin2010multiword}  define MWE as those lexical items that are decomposable into multiple lexemes and present idiomatic behaviour at some level of linguistic analysis, as a consequence they should be considered as a unit at some level of computational processing. Also in the Computational Linguistics field, \cite{StefanowitschGries:2003} propose the term ``collostruction'' to refer to the wide range of complex linguistic  units as defined in theoretical proposals of Cognitive Grammar. In our approach we consider as constructions those syntactic units consisting of two or more lexical items with internal semantic coherence. These constructions are compositional and appear with a frequency higher than expected.

From the NLP perspective, most approaches for dealing with constructions tend to apply methods that use previously defined empirical knowledge to find instances and variants of specific types of constructions in corpora. This approach allows us to obtain preidentified units and their variations at different degrees of complexity, but does not allow for the identification of as yet unidentified constructions. In order to discover new knowledge, we need an open and flexible method that give us usable and interpretable results.  We organised this overview taking into consideration those approaches that try to find or discover constructions. 

A frequent approach to gathering empirical data about constructions using NLP techniques is to look for well-known, highly conventionalized and previously defined constructions (see the works of \cite{hwang2010towards,Muischnek:2009,kebelmeier:2009,ODonnell:2010,duffield2010identifying}).

Very tied to Construction Grammar theory and in the framework of the methodologies based on statistical metrics, it is worth noting the works of 
\cite{StefanowitschGries:2003}, \cite{StefanowitschGries:2008}, and \cite{gries2005converging}. Their research always focuses  on specific types of constructions, on the analysis of their variants and on the degree of entrenchment between their elements. \cite{griesellis:2015} summarize different statistical measures applied to the analysis of constructions and evaluate their linguistic interpretation and impact.

From the perspective of methods oriented to the discovery of new constructions, we should distinguish between those approaches that include some kind of linguistic filtering of the type of constructions to be dealt with and those that do not apply any kind of restriction. All these methods are strongly grounded on statistical measures: in \cite{evert2008corpora} and \cite{pecina:2010} there is an exhaustive summary and criticism of statistical measures that calculate the degree of association between words.\footnote{The works referred to this section use the term collocate in a very weak sense, roughly equivalent to what is known as MWE in NLP.}  

Looking for ways to identify potential collocations in corpora using statistical measures, \cite{bartsch:2004} explores certain types of collocations involving verbs of verbal communication. Her approach is semiautomatic and involves a manual revision of the results. We also highlight the work of \cite{pecina:2010}, based on fully statistical methods.
However, supervised machine learning requires annotated data, which creates a bottleneck in the absence of large corpora annotated for collocation extraction. A solution to this problem is presented by \cite{dubremetz2014extraction} who propose the use of the MWEtoolkit~\citep{ramisch2010multiword}  to automatically extract candidates that fit a certain POS pattern. See also the work of \cite{forsberg:2014,farahmand2014supervised,tutubalina2015clustering}. 

From a different perspective, based on the calculation of \textit{n}-grams, we also consider the results of the  StringNet project~\citep{Wible:2010}, a knowledge base (KB) which contains candidates to be constructions. In this case, no filters are applied to the lexico-syntactic patterns obtained. As a result, StringNet is a lexicogrammatical KB automatically extracted from the British National Corpus (BNC)\footnote{www.natcorp.ox.ac.uk} consisting of a massive archive of hybrid \textit{n}-grams of co-occurring combinations of POS tags, lexemes and specific word forms.

We also want to highlight the approaches that use syntactic information for obtaining constructions, such as the work of \cite{zuidema2006productive,sangati2015multiword}, based on the framework of Tree Substitution Grammar (TSG).

Harris distributional hypothesis has a great acceptance in the treatment of linguistic semantics to overcome traditional symbolic representations. Relying on this hypothesis, \cite{gamallo2005clustering} developed an unsupervised strategy to  acquire syntactico-semantic restrictions for nouns, verbs and adjectives from partially parsed corpora. Although the resulting data could be used for deriving lexico-syntactic patterns their objective was to capture semantic generalizations, both for the predicates and their arguments.

Currently, there is an increasing interest in the use of distributional models for representing semantics, such as DSMs~\citep{TurneyPantel,Baroni:2013}  or word embeddings~\citep{mikolov2013linguistic}. These models derive word-representations in an unsupervised way from very large corpora. All of them rely on co-ocurrence patterns but differ in the way they reduce dimensionality. As pointed out in \cite{murphy2012learning}, the representations they derive from corpora are lacking in cognitive plausibility, with exceptions such as those defined in \cite{baroni2010strudel}. Our proposal shares with these authors the same semantic approach (distributional hypothesis), because we consider that these models are a good option in which to frame our methodology. In concrete, we used DSMs because they are highly linguistically interpretable and allow us to modelize the context, a key point in our methodology.

DSMs have been applied successfully in linguistic research~\citep{shutova2010metaphor}, in different NLP tasks and applications~\citep{BaroniLenci}  and, especially, in tasks related with measuring different kinds of semantic similarity between words~\citep{TurneyPantel}. Like us, \cite{shutova2016multilingual} use distributional clustering techniques, though they use DSMs to investigate how to find metaphorical expressions. Recently, DSMs have been extended to phrases and sentences by means of composition operations deriving meaning representations for phrases and sentences from their parts (see \cite{Baroni:2013} and \cite{Mitchell:Lapata:2010} for an overview). Nevertheless, DSMs have rarely focused on the discovery of constructions. In this line, it is worth noting the papers presented in the shared task of the Workshop on Distributional Semantics and Compositionality~\citep{biemann2011distributional}. This workshop focused on the extraction of non-compositional phrases from large corpora by applying distributional models that assign a graded compositional score to a phrase. This score denotes the extent to which compositionality holds for a given expression. The participants applied a variety of approaches that can be classified into lexical association measures and Word Space Models. It is also worth noting that approaches based on Word Space Models performed slightly better than methods relying solely on statistical association measures.

In the next section, we describe in depth the DISCOver methodology that we developed to discover lexico-syntactic constructions.


\section{Methodology for Discovering Constructions}\label{2:sec:methodology}

Following a distributional semantic approach, we developed an unsupervised bottom-up method for obtaining the lexico-syntactic patterns that can be considered candidates for constructions. This method uses a medium-sized corpus (100 million tokens) to obtain the distributional properties of words and to stablish similarity relations among them from their contexts. The representation of the contexts is based on syntactic dependencies. 


Figure~\ref{2:fig:process} depicts the five main steps involved in obtaining the lexico-syntactic patterns, the processess involved, and the input and output of each process. Briefly, the first step is the linguistic processing of the Diana-Araknion corpus (See Section~\ref{2:subsec:corpus}). In the next step, a DSM matrix is constructed with the frequencies of the lemmas in each one of the contexts (see Section~\ref{2:subsec:matrices}). Step 3 focuses on clustering semantically related lemmas, that is, those lemmas that share a set of contexts (see Section~\ref{2:subsec:clustering}). In the fourth step, we applied a generalization process by linking all clusters taking into account the information contained in the contexts and then filtering only those links that mantain the strongest relationships (See Section~\ref{2:subsec:generalisation}). Finally, we generate the lexico-syntactic patterns to be considered as candidates to be constructions from the related clusters selected in the previous step (See Section~\ref{2:subsec:patterngeneration}). 

\begin{figure}[h]
  \centering
    \includegraphics[width=1.0\linewidth]{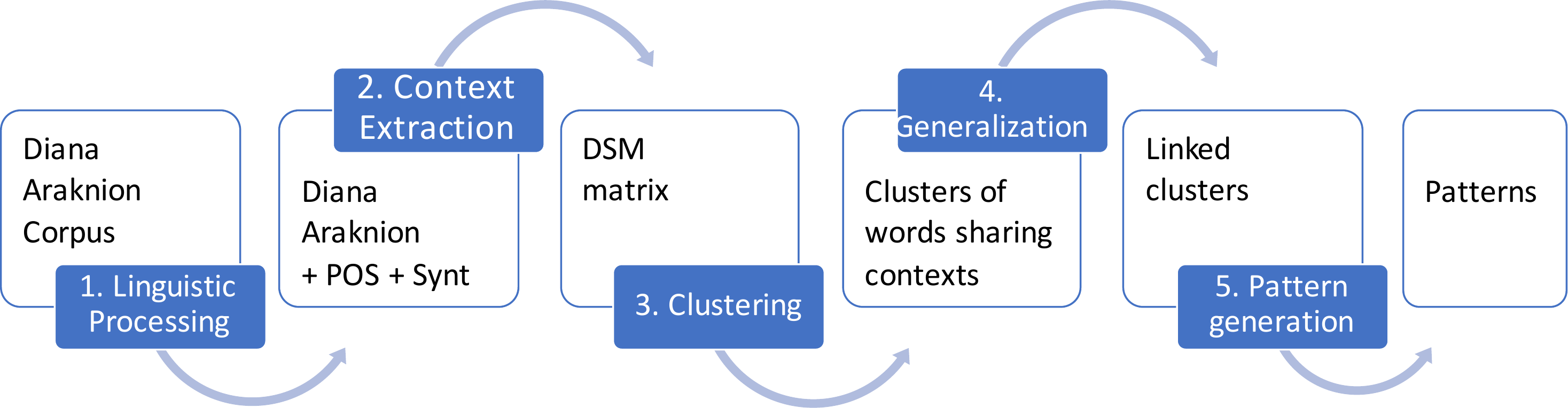}
  \caption{Main steps in DISCOver methodology}
  \label{2:fig:process}
\end{figure}

\subsection{Description of the Task} \label{2:subsec:hypotheses}

Our methodology is based on the pattern-construction hypothesis, which states that those contexts that are relevant to the definition of a cluster of semantically related words tend to be (part of) lexico-syntactic constructions. In our experiments, ``lexico-syntactic constructions'' are patterns in the form of [\textit{lemma, dependency\_direction (\textit{dep\_dir}), dependency\_label (\textit{dep\_lab}), context\_lemma}] (for instance, [despeinar\_v, $>$: dobj, cabellera\_n]\footnote{[to\_tussle\_v, $>$: dobj, one's\_hair\_n]}). \textit{Dependency\_label} is a type of syntactic relation between \textit{lemma} and \textit{context\_lemma}, while \textit{dependency\_direction} is the direction of the \textit{dependency\_label}. To be considered candidates to be constructions patterns must have the following properties:
 
\begin{itemize}
\justifying
\item \textit {Syntactic-semantic coherence}: We expect the two lemmas in each pattern candidate to be syntactically and semantically related.
\item \textit {Generalizability}: The patterns can be generalized and/or derived from other patterns through generalization.
\end{itemize}

Based on these properties of constructions and the initial pattern-construction hypothesis, the main aims of the DISCOver methodology are the following:

\begin{itemize}
\justifying
\item [1.] To identify the contexts that are relevant for the definition of a cluster of semantically related words. Each of these contexts is part of a pattern candidate to be construction attested in the corpus (henceforth Attested-Patterns).
\item [2.]To use the previous contexts in a generalization process in order to identify unseen, but possible candidates to be constructions (henceforth Unattested-Patterns).
\end{itemize}

As a result we obtain two sets of qualitatively different patterns that are candidates to be constructions: attested and unattested patterns. We then proceed to evaluate the internal syntactic-semantic coherence of these patterns.

\subsection{The Corpus} \label{2:subsec:corpus}

As shown in Figure~\ref{2:fig:process}, corpus creation is the first
step in the process of obtaining lexico-syntactic patterns.
Specifically, we built the
Diana-Araknion\footnote{\label{2:footnote:diana} All corpora are available at \url{http://clic.ub.edu/corpus/} or per-request} corpus, a Spanish corpus
which consists of approximately 100 million
tokens\footnote{Concretely, the Diana-Araknion has 93,987,098 tokens
and 1,321,174 types.} (corresponding to 3 million sentences)
gathered mainly from the Spanish Wikipedia (2009), literary works
and texts from Spanish parliamentary discussions, news reports, news
agency documents, and Spanish Royal Family speeches.

The corpus was automatically tokenized and linguistically processed
with POS and lemma tagging, and syntactic dependency parsing. We
used the Spanish analyzers available in the Freeling\footnote{
\url{http://nlp.lsi.upc.edu/freeling}.} open source language-processing
library  \citep{Padro:2012}.

For the purpose of evaluation, we built Diana-Araknion++, a new corpus gathered 
from web-pages in Spanish. It includes Spanish Wikipedia (2015), articles from online newspapers, 
speeches from the European Parliament, university articles and sites from the Spanish 
webspace. This corpus was automatically tokenized and POS tagged and consists of 600M 
tokens.

\subsection{Matrix} \label{2:subsec:matrices}

To generate the frequency matrix (see Step 2 in Figure~\ref{2:fig:process}), we used only the 15,000 most frequent lemmas extracted from the Diana-Araknion corpus including nouns (\textit{N}), verbs (\textit{V}), adjectives (\textit{A}) and adverbs (\textit{R}). We modeled the context in which the words occur giving rise to a \textit{lemma-dep} matrix. This matrix corresponds to the type of \textit{word-context} matrix defined in \cite{TurneyPantel} and in \cite{BaroniLenci}. 
In the \textit{lemma-dep} matrix, the context is based on parsed texts in which both dependency directions and dependency labels are taken into account. Each context is a triple of [\textit{dependency\_direction, dependency\_label, context\_lemma\_POS}].

In what follows, we introduce how this lemma-context matrix is formally represented (see
Section~\ref{2:subsubsec:lemma-context-matrices}) and then we describe the matrix in more detail (see
Section~\ref{2:subsubsec:matricesproposal}).

\subsubsection{Formalization of the Lemma-Context Matrix}\label{2:subsubsec:lemma-context-matrices}

Our DSM consists of a lemma-context PPMI matrix  $X$ with $n_r$ rows and $n_c$ columns.
Note that each row vector $i$ corresponds to a lemma, 
each column $j$ corresponds to a co-occurrence context, and each cell in $X$ has a 
numerical weighted value, $x_{ij}$. This weighted value is the result of applying Positive Pointwise Mutual Information (PPMI)~\citep{Niwa:1994}
to a lemma-context frequency matrix $F$ with size $n_r \times n_c $. Each element in this matrix, $f_{ij}$,
is computed as the number of occurrences of lemma $i$ in context $j$ in the whole corpus.
\cite{TACL457} perform a large-scale evaluation of different co-occurrence DSM models over various tasks. 
They show that term weighting through association scores significantly improves the performance of the DSM model. 

\subsubsection{\textit{Lemma-Dep} Matrix}\label{2:subsubsec:matricesproposal}

The matrix proposed in this work is a lemma-context matrix, hereafter \textit{lemma-dep} matrix, based on syntactic dependencies\footnote{We used the Spanish syntactico-semantic analyzer Treeler to analyse the Diana-Araknion corpus: \url{http://devel.cpl.upc.edu/treeler}.}. In this matrix, the context $j$ of a lemma $i$
is a context word $k$ (\textit{context\_lemma}) directly related by a dependency direction (\textit{dep\_dir}) and a dependency label (\textit{dep\_lab}) to the lemma $i$. The words of the lemma $i$ belong to the following POS: \textit{N}, \textit{V}, \textit{A} and \textit{R}. Each lemma is assigned its corresponding POS. Therefore, in the matrix, context $j$ contains three elements as defined in~\ref{2:eq:context_matrix}:

\begin{equation} \label{2:eq:context_matrix}
context = [ dep\_dir:dep\_lab: context\_lemma ]
\end{equation}

\noindent  where:

\begin{itemize}
\justifying
\item  $dep\_dir$: has two possible values `$<$' or `$>$', indicating the direction of the dependency. 
\item  $dep\_lab$: indicates the dependency label of the lemma $i$ and context\_lemma $k$. The possible values are \{\textit{subj, dobj, iobj, creg, cpred, atr, cc, cag, spec, sp} and \textit{mod}\}.
In the case of dependencies between a preposition and a noun, adjective or verb, the dependency label is labeled by the same preposition and its corresponding $dep\_lab$, that is, \textit{dobj}, \textit{iobj}, \textit{creg}, \textit{cag}, \textit{sp} or/and \textit{cc}.
\item  $context\_lemma$ is the lemma of the context word $k$ with its corresponding POS, which can be \textit{N}, \textit{V}, \textit{A}, \textit{R}, preposition\textit{(P)}, number\textit{(Z)} and date\textit{(W)}. In the case of proper nouns, they are replaced by the \textit{pn\_n} (proper noun) POS.

\end{itemize}
\vspace{-0.3cm}
\begin{figure}[h]
  \centering
    \includegraphics[width=0.7\linewidth]{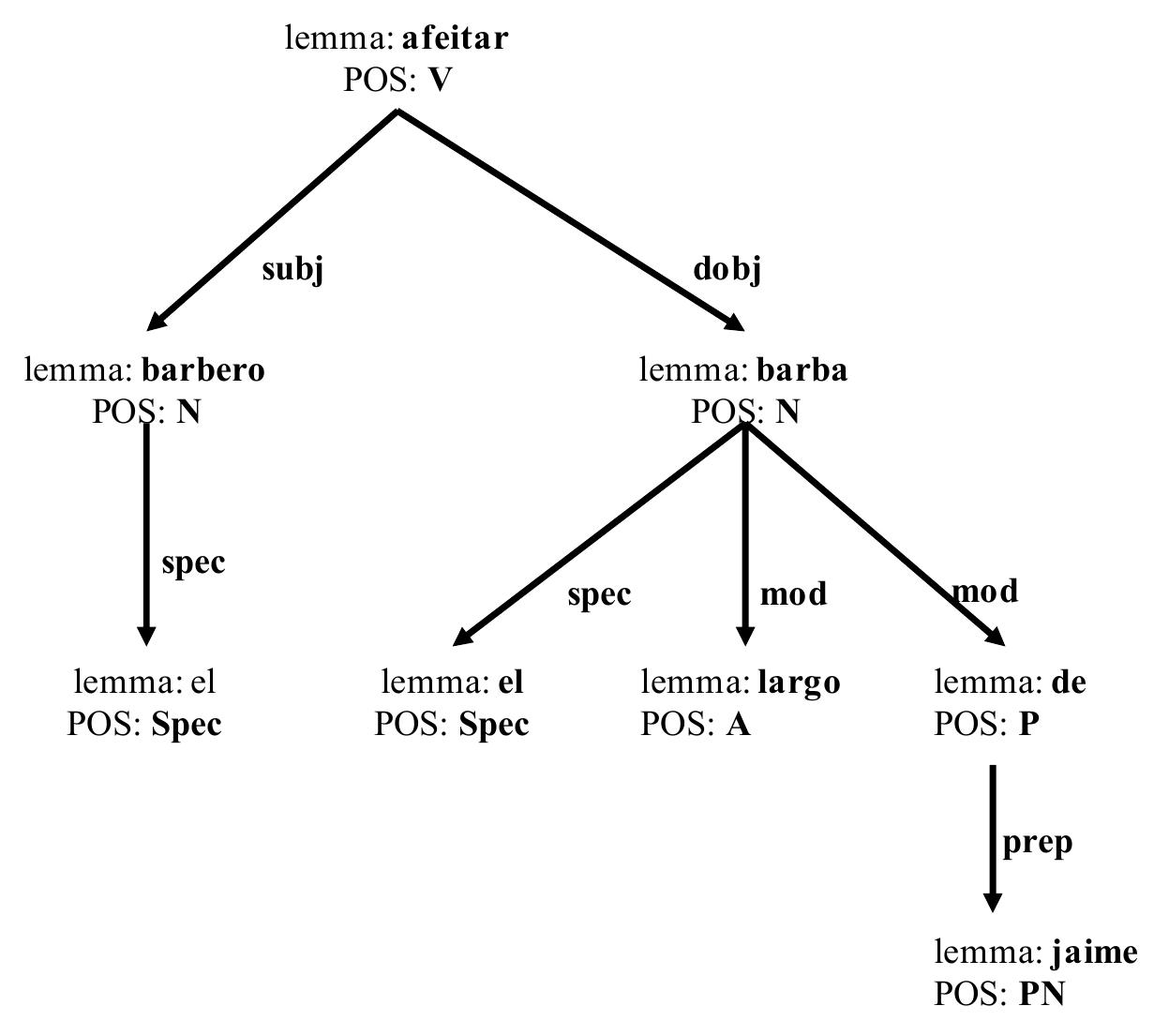}
  \caption{Dependency parsed sentence: \textit{El barbero afeita la larga barba de Jaime} (`The barber shaves off James's long beard')}
  \label{2:fig:parsed_sentence1}
\end{figure}

Figure 2 shows an example of a dependency parsed sentence from which, for instance, three different contexts of the noun lemma \textit{barba\_n} \footnote{'beard'} are generated: [$<$:dobj:afeitar\_v], [$>$:mod:largo\_a] and  [$>$:de\_sp:pn\_n]\footnote{This context is the result of substituting the proper name ``Jaime'' by ``pn\_n''.}. These contexts are represented in the \textit{lemma-dep} matrix.

In [$<$:dobj:afeitar\_v], `$<$' indicates that the verb \textit{afeitar\_v} \footnote{'to shave off'} maintains a parent dependency relation with \textit{barba\_n}, \textit{dobj} indicates that \textit{barba\_n} is the direct object of \textit{afeitar\_v}, and \textit{afeitar\_v} is the context word (lemma $k$) related to \textit{barba\_n} (lemma $i$). In [$>$:mod:largo], \textit{mod} indicates that the adjective \textit{largo\_a} \footnote{'long'} is a modifier of \textit{barba\_n}, and in [$>$:de\_sp:pn\_n] the proper noun  (\textit{Jaime} in Figure 2) is replaced by the \textit{pn\_n} POS tag\footnote{Since the POS tagger does not distinguish between subclasses of proper names (person, organization, place, etc.), the grouping of all  with the \textit{pn\_n} tag gives better results. We used proper nouns in the \textit{context\_lemma} configuration, but not as words in the lemma \textit{i}. Similarly, stopwords are not included in lemma \textit{i}.}.

For each context obtained from the dependency structure, three different dependency contexts are generated: one that makes  all the elements of the context explicit, that is, the \textit{dep\_dir}, \textit{dep\_lab} and \textit{context\_lemma} (for example, [$<$:dobj: afeitar\_v]); another in which the \textit{dep\_lab} is generalized by the variable `oth' (for example, [$<$:oth:afeitar\_v])\footnote{The tag `oth' (\textit{other}) means that the dependency label is not specified.} and, finally, one context that generalizes the \textit{context\_lemma} by substituting it for the variable `*' (for example, [$<$:dobj:*\_v])\footnote{The symbol `*\_v' means that a verb occurs in this position, but we do not specify which one it is.}. The three lemmas represented in example (2) do not share any context, therefore they could not be semantically related in our model. Instead, applying the generalization of contexts, we obtained a relationship between lemma$_{1}$ and lemma$_{2}$ in example (3), and between lemma$_{1}$ and lemma$_{3}$ in example (4). In example (3), the \textit{dep\_lab} is generalized, whereas in example (4) the \textit{context\_lemma} is generalized.

\begin{enumerate}[resume]
\item lemma$_{1}$ $[<:subj:robar\_v\footnote{`to\_rob'}]$\par
lemma$_{2}$ $[<:dobj:robar\_v]$\par
lemma$_{3}$ $[<:subj:hurtar\_v\footnote{`to\_steal'}]$\\
\item lemma$_{1}$ $[<:oth:robar\_v]$\par
lemma$_{2}$ $[<:oth:robar\_v]$\par
lemma$_{3}$ $[<:oth:hurtar\_v]$\\
\item lemma$_{1}$ $[<:subj:*\_v]$\par
lemma$_{2}$ $[<:dobj:*\_v]$\par
lemma$_{3}$ $[<:subj:*\_v]$
\end{enumerate}

In this way, the generalization of contexts allows us to take into account contexts that are similar (they share two, but not all of the elements, of their context), but not identical. Therefore, we can distinguish between those lemmas that share the same or similar context, and those that have a completly different context. By adding these contexts that are similar but not identical we add new knowledge, that is, knowledge not directly present in the corpus. This new knowledge is used to generate the Unattested-Patterns.

\subsection{Clustering} \label{2:subsec:clustering}

Once we described the $X$ matrix, we proceeded to the third step detailed in Figure~\ref{2:fig:process} that is devoted to the clustering of this matrix. 
The motivation of the clustering process is to find, for each lemma in the matrix, all semantically related words (lemmas). This will allow us to create new Unattested-Patterns  after the linking and filtering cluster processes. To perform this clustering step, we used the \textsc{Cluto} toolkit~\citep{Karypis}\footnote{We use {\sc{vcluster}} program provided in the toolkit, which computes the clustering using one of five different approaches. Four of these approaches are partitional, whereas the fifth approach is agglomerative.},  
which is used to cluster a collection of objects (in our case, lemmas) into a predetermined number of clusters labeled $k$. We applied a methodology based on \cite{Calinski:1974} and using cosine similarity and \textsc{Cluto}'s $\mathcal{H}_2$ metric to estimate the optimal amount of clusters. 

We experimented with a number of different clustering configurations. The variables we took into account were: a) the number of most frequent lemmas, with the 10,000 to 15,000 most frequent lemmas giving the best results; b) the inclusion of proper nouns or their substitution for their POS; and c) considering the lemmas with and without their POS.

We evaluated the results of these configurations manually and opted for 15,000 lemmas with proper nouns grouped according to their POS tag (pn\_n) and with the POS tag assigned to the lemmas. This configuration gave an optimal $k$ of 1,500 clusters applying the \cite{Calinski:1974} method and the $\mathcal{H}_2$ metric.

The inclusion of POS improves the internal consistency of the clusters. Since the POS tagger does not distinguish between subclasses of proper names (person, organization, place, etc.), grouping them according to the \textit{pn\_n} tag also gives better results. Regarding the number of lemmas, all results obtained using between 10,000 and 15,000 lemmas gave satisfactory results. The choice of the number of lemmas determines the number and the content of the clusters. In all cases, the quality of clusters obtained was acceptable. We consider a cluster as acceptable when all or almost all words contained in it share one of the following relations: synonymy, hypernymy, or hyponymy. This would allow for the use of one or more configurations for the obtention of the final lexico-syntactic patterns (see Section~\ref{2:subsec:patterngeneration}).

Using \textsc{Cluto} with the selected configuration, we obtained a set of clusters $C=\{c_i: 1\leq i \leq k\}$ from matrix $X$. Formally, the content of each cluster $c_i \in C$ is defined in~\ref{2:eq:content_cluster}, where $le$ is a set of related lemmas and $ctx$ is a set of contexts. Each lemma\_pos only belongs to one cluster (i.e., it can only be defined in one $le$), whereas a context\_lemma can be in several contexts ($ctx$) of different clusters.

\begin{equation} \label{2:eq:content_cluster}
c_i = <le, ctx>
\end{equation} 

Formally, a context (called $context\_cluster$) in $ctx$ is described as follows:

\begin{equation} \label{2:eq:context_cluster}
context\_cluster = <[dep\_dir : dep\_lab : context\_lemma], score>
\end{equation} 

\noindent where $dep\_dir, dep\_lab, context\_lemma$ corresponds to the definition of a context as shown in Section~\ref{2:subsubsec:matricesproposal}. The $score$ is the sum of the different scores given by \textsc{Cluto}\footnote{The sum of the twenty-five most descriptive and discriminative scores given automatically by \textsc{Cluto}.}.

For example, Table~\ref{2:tab:cluster421_description}\footnote{\label{2:foot:ref} The translation to English of Tables 1 and 2, as well as additional examples and clusters are available at \url{http://clic.ub.edu/corpus/}} 
describes the lemmas, $le$, and the most scored contexts, $ctx$, in cluster number 421\_n (one of the clusters obtained in the corpus analyzed).

\begin{table}[ht]
\caption{Example of a real cluster (421\_n) in the Diana-Araknion corpus in Spanish}
\begin{center}
  \begin{small}
\begin{tabular}{c l l l  }
    \multicolumn{4}{ l }{Cluster: 421\_n} \\ \hline 
 \multicolumn{1}{ p{1.0cm} } { Lemmas ($c_{421\_le}$)} &     \multicolumn{3}{p{9.8cm}}{barba\_n, bigote\_n, cabellera\_n, cabello\_n, ceja\_n, crin\_n, melena\_n, mostacho\_n, patilla\_n, pelaje\_n, pelo\_n, perilla\_n, vello\_n} \\
     \hline
      & [$<$ : dobj : erizar\_v],11 &  [$<$ : oth : erizar\_v],11 & [$<$ : oth : rizar\_v],10  \\ 
      &  [$<$ : subj : erizar\_v],10 & [$>$ : mod : espeso\_a],9 & [$>$ : oth : espeso\_a],9 \\   
      & [$>$ : mod : negro\_a],7 & [$<$ : oth : negro\_a],5 & [$>$ : mod : gris\_a],8 \\   
      & [$<$ : dobj : rizar\_v],8 & [$>$ : oth : gris\_a],7 & [$<$ : oth : pelo\_n],6\\  
      Contexts &  [$>$ : mod : rubio\_a],7 & [$>$ : mod : barba\_n],7 & [$<$ : oth : atusar\_v],7 \\ 
     ($c_{421\_ctx}$) & [$>$ : mod : largo\_a],4 & [$>$ : oth : rubio\_a],6 & [$<$ : mod : pelo\_n],2 \\ 
     & [$>$ : mod : rojizo\_a],4 & [$>$ : oth : rojizo\_a],6 & [$>$ : oth : largo\_a],3  \\ 
     & [$<$ : oth : bigote\_n],3 & [$>$ : mod : blanco\_a],3  & [$>$ : mod : cano\_a],5 \\ 
     & [$>$ : mod : hirsuto\_a],5  & [$>$ : oth : hirsuto\_a],2& [$>$ : oth : largo\_a],3\\ 
     & [$>$ : oth : negro\_a],2  & [$>$ : mod : rojizo\_a],2& \\ 
\end{tabular}
  \end{small}
\end{center}
\label{2:tab:cluster421_description}
\end{table}

\subsubsection{Results of the Clustering Process} \label{2:subsec:resultsclustering}

Following our configuration, we obtained a total of 1,500 clusters in the clustering process ($k$=1500). It is worth noting that the clusters are highly morpho-syntactically and semantically cohesive.

The clusters contain lemmas belonging mostly to the same POS. 
It is worth mentioning that more than half of the clusters are nouns (54.20\%), followed by verbs (25.80\%) and adjectives (16.67\%). Clusters of adverbs make up only 3.33\% of the total.
%

Clusters contain relevant implicit information, in the sense that their lemmas belong to well-defined semantic categories, often at a very fine-grained level. For instance, we obtained clusters of adjectives with a \textit{Positive Polarity} (5) and with a \textit{Negative Polarity} (6)\footref{2:foot:ref}. 
These results encourage us to tag all the clusters with one or more semantic labels. That will enrich the obtained patterns.

\begin{enumerate} [resume]
\item \{$c_{111}$, \textit{Positive\_Polarity} adjectives: admirable\_a, asombroso\_a, genial\_a...\}\footnote{'admirable, amazing, great'}\\

\item\{$c_{38}$, \textit{Negative\_Polarity} adjectives: atroz\_a, aterrador\_a, espantoso\_a...\}\footnote{'atrocious, scary, frightening'}\\





\end{enumerate}


\subsection{Generalization: Linking and Filtering Clusters} \label{2:subsec:generalisation}

The process of generalization by linking clusters (see Step 4 in Figure~\ref{2:fig:process}) is based on the set of clusters and contexts obtained using \textsc{Cluto}. The processes of linking clusters and pattern generation detailed in Section~\ref{2:subsec:patterngeneration} are the core steps of the DISCOver methodology. The process of linking clusters uses the set of the twenty-five highest scored contexts in each cluster. According to our pattern-construction hypothesis (see Section \ref{2:subsec:hypotheses}), the goal of the linking of clusters is to establish the relationships between clusters using their contexts,  as defined in~(\ref{2:eq:context_cluster}), obtaining as a result a matrix of all possible contextual relations between clusters (see Section~\ref{2:subsubsec:buildinggraph}). Next, we apply a filtering process in order to select strongly related links taking into account different criteria (see Section~\ref{2:subsubsec:filteringlinks}).

\subsubsection{Linking Clusters and Building the Matrix of Related Clusters} \label{2:subsubsec:buildinggraph}

Basically, the aim of the cluster linking process 
is to establish the relationships between clusters and to store them in a matrix, $R\_clusters$, with $k$ rows and $k$ columns. The $k$-value corresponds to the number of clusters 
obtained in the clustering step.  

For building the matrix, for each origin cluster ($x$) each $dep\_dir$ and  $dep\_lab$ of the $context\_cluster$ (defined in Equation~\ref{2:eq:context_cluster}) are converted into a $contextual\_$ $relation$ (see Equation~\ref{2:eq:context_generalisation}), while the $context\_lemma$ of the  $context\_cluster$ is used to locate the cluster ($y$) in which it occurs. We obtain as a result a matrix, $R\_clusters$, in which clusters are related according to a set of contextual relations  stored in a $relation\_set$. The sum of the scores of the $context\_clusters$ in \ref{2:eq:context_cluster} are added together in a matrix, $R\_scores$. The $R\_scores$ matrix is later used in the process for determining filtering thresholds.


\begin{equation} \label{2:eq:context_generalisation}
contextual\_relation = <dep\_dir, dep\_lab>
\end{equation}

For the contextual relation, defined in \ref{2:eq:context_generalisation}, $dep\_dir$  and  $dep\_lab$ are the dependency direction and the dependency label
defined in a context of cluster $i$ related to cluster $j$. Note that the $relation\_set$ of a cluster in itself is empty as $R\_clusters[i][i]=\emptyset$ and  $R\_clusters[i][j] \neq R\_clusters[j][i]$.

Following the example of cluster 421\_n, described in Table~\ref{2:tab:cluster421_description}
,  the result of
the cluster linking process for this particular cluster ($i=421\_n$) is shown in Table~\ref{2:tab:cluster421_linking}\footnote{For the sake of simplicity, the contexts are not included in the Table 2 and we only show a relation of each type. }. The first column in this table shows the related clusters, $j$, the second column shows the relation\_type that relates cluster 421\_n to the related clusters $j$ (i.e. \textsc{strong}, \textsc{semi} or \textsc{weak}, See~\ref{2:subsubsec:filteringlinks}), and finally the last column describes the lemmas in the related clusters. 

\begin{table}[htp]
\caption{Some examples of cluster linking process in cluster $i$=421\_n (described in Table~\ref{2:tab:cluster421_description}).}
\begin{center}
\begin{tabular}{p{1.8cm}|p{1.75cm}|p{7.9cm}}
\textbf{Related}  &\textbf{Relation\_} &  \textbf{Lemmas}\\
\textbf{clusters}($j$) & \textbf{type} & ($c_{j.le}$, where $c_j$ refers to the related cluster, $j$)\\\hline
1223\_a & \textsc{strong}  & azabache\_a, bermejo\_a, \textbf{cano\_a}, canoso\_a, \textbf{hirsuto\_a}, lacio\_a, lustroso\_a, ondulante\_a, sedoso\_a...\\\hline 
932\_v & \textsc{Semi} & afeitar\_v, \textbf{atusar\_v}, cepillar\_v, empolvar\_v, enguantar\_v, peinar\_v, rasurar\_v...\\\hline 
405\_n & \textsc{weak}  & contario\_n, final\_n, largo\_n, menudo\_n... \\ 
\end{tabular}
\label{2:tab:cluster421_linking}
\end{center}
\end{table}



\subsubsection{Filtering Related Clusters} \label{2:subsubsec:filteringlinks}



In the $R\_clusters$ matrix, not all contextual relationships between clusters are accepted since they have a low $R\_scores$. For this reason, we established two criteria to automatically determine which relationships will be maintained and which ones are filtered out in the pattern generation process. For each criterion only those relations higher than a predetermined score value will be considered. The criteria are the following:

\begin{itemize}
\justifying
\item \textbf{Criterion 1}: For each pair of clusters $i$ and $j$,  
we take into account those relations that in each of their directions (i.e., $R\_scores[i][j]$ or $R\_scores[j][i]$) have a score above a minimum predetermined value, that is, $threshold_1$. This $threshold_1$ is automatically determined by finding a score value that allows for the grouping of 30\% of the clusters. The relations that fulfill criterion 1 are called \textsc{Strong} relations. 

\item \textbf{Criterion 2}: For each pair of clusters $i$ and $j$, we take into account those relations in which the sum of scores in both directions  (i.e., $R\_scores[i][j] + R\_scores[j][i]$) is higher than a predetermined value, that is, $threshold_2$, which is determined by finding a value that allows for the grouping of 50\% of the clusters. The relations that fulfill criterion 2 are called \textsc{Semi} relations. 
\end{itemize}

Considering the example of cluster 421\_n, the result of the filtering process is that, out of the three clusters linked to cluster 421\_n in our example\footref{2:foot:ref} (1223\_a, 932\_v, and 405\_n), we will only select those with \textsc{strong} and \textsc{semi} relations, that is, 1223\_a, and 932\_v. Those labelled as \textsc{weak} (e.g.,  405\_n shown in Table~\ref{2:tab:cluster421_linking}) are filtered out because they do not reach the established thresholds.

\subsection{Pattern Generation} \label{2:subsec:patterngeneration}

Once the process for automatically linking and filtering clusters was carried out, we proceeded to generate the lexico-syntactic patterns to be considered as candidates for constructions (see Step 5 in Figure~\ref{2:fig:process}). 
Each generated pattern is defined as follows:

\begin{equation} \label{2:eq:pattern}
pattern = <lemma_i, dep\_dir, dep\_lab, lemma_j>
\end{equation}
\noindent
where $lemma_i$ and $lemma_j$ are the lemmas contained in the related clusters ($i$ and $j$), $dep\_dir$ and $dep\_lab$ are the dependency direction and the dependency label between the related clusters. So, there is a pattern for each $lemma_i$ and $lemma_j$ pair.

As we mentioned in Section~\ref{2:subsec:clustering}, all possible configurations using between 10,000 and 15,000 lemmas gave acceptable related clusters. In order to increase the number of patterns generated we carried out the same process with a configuration using 10,000 lemmas. We combined the patterns obtained using the 10,000 and 15,000 lemmas together and removed those that were shared by both configurations. In Tables~\ref{2:tau:clustersrelated},~\ref{2:tau:clustersrelatedpos} and~\ref{2:tau:generatedclusters}, we show 
the number of resulting clusters and patterns, after removing the overlapping patterns, for the two configurations.

\begin{table}[htp]
\caption{Distribution of the number of related and unrelated clusters and their percentage}
\centering
\begin{tabular}{lrr}
 &\textbf{10,000 lemmas} & \textbf{15,000 lemmas}\\
\cline{1-3} 
\textbf{Relation}& \textbf{Clusters (\%)} &\textbf{Clusters (\%)}\\ 
\hline
\textsc{Strong} &  441 (31.50\%) & 461 (30.73\%)\\
\textsc{Semi} & 339 (24.21\%) & 396 (26.40\%) \\
\textbf{Total} & \textbf{780} (55.71\%) & \textbf{857} (57.13\%) \\
\noalign{ \vspace {0.5cm}}
\textsc{Weak} & 589 (42.07\%) & 636 (42.40\%) \\
Unrelated & 31 (2.21\%) & 7 (0.47\%) \\
\end{tabular}
\label{2:tau:clustersrelated}
\end{table}

As shown in Table~\ref{2:tau:clustersrelated} (second and third columns), more than 55\% of the linked clusters maintain \textsc{Strong} and \textsc{Semi} relationships, whereas only the 2.68\% of the clusters remain unrelated. Table~\ref{2:tau:clustersrelatedpos} (second and third columns) shows the distribution of linked clusters by POS in both configurations.

\begin{table}[htp]
\caption{Distribution of the number of related clusters and their percentage by POS}
\centering
\begin{tabular}{lrr}
&\textbf{10,000 lemmas} & \textbf{15,000 lemmas}\\
\hline
\textbf{POS} & \textbf{Clusters (\%)} &\textbf{Clusters (\%)}\\ 
\hline
N & 415 (53.21\%) & 464 (54.14\%)\\
V & 197 (25.26\%) & 182 (12.24\%)\\
A & 142 (18.21\%) & 173 (20.19\%)\\
R & 26 (3.30\%) & 38 (4.43\%)\\
\noalign{ \vspace {0.5cm}}
\textbf{Total}&\textbf{780} (100\%) & \textbf{857} (100\%)\\
\end{tabular}
\label{2:tau:clustersrelatedpos}
\end{table}

The total number of lexico-syntactic patterns obtained from the two configurations of clusters (780 and 857 \textsc{Strong} and \textsc{Semi} related clusters) is 237,444. For the purpose of pattern generation, \textsc{Strong} and \textsc{Semi} clusters have been treated equally. From these patterns, we removed 16,712 patterns, those that were present in both sets of generated patterns, given as a result the total number of 220,732 patterns (See Table~\ref{2:tau:generatedclusters}). 

\begin{table}[htp]
\caption{Distribution of the generated patterns }
\centering
\begin{tabular}{lrrr}
 \textbf{Lemmas}&\textbf{Attested-Patterns} & \textbf{Unattested-Patterns}&\textbf{Total} \\
\hline
10,000 & 23,980 & 48,147 & 72,127 \\
15,000 & 37,840 & 127,477 & 165,317 \\
\noalign{ \vspace {0.3cm}}
10,000 + 15,000 & 61,820 & 175,624 & 237,444 \\
\textbf{Overlapping} & \textbf{8,531} & \textbf{8,181} & \textbf{16,712}\\
\textbf{Sum (no overlap)} & \textbf{53,289} & \textbf{167,443} &\textbf{220,732}\\
\end{tabular}
\label{2:tau:generatedclusters}
\end{table}

The DISCOver methodology allows for the generation of patterns that actually occur in the corpus (Attested-Patterns), but also of lexico-syntactic patterns that are not present in the corpus but which are highly plausible in Spanish (Unattested-Patterns), since the components of the clusters are closely semantically related. As a result, we are able to enlarge the descriptive power of the source corpus. 
Among the patterns we generated, 61,820 were Attested-Patterns, that is, patterns that are present in the source corpus, and 175,624 were Unattested-Patterns, that is, new patterns (see Table~\ref{2:tau:generatedclusters}).

Retaking the example of cluster 421\_n and its related clusters we obtain patterns such as those shown in (7)\footnote{$<$moustache$_{c\_421}$ $<$:dobj: to\_brush$_{c\_932\_v}$$>$; $<$mane$_{c\_421}$ $<$:dobj: to\_smooth$_{c\_1267\_v}$$>$; $<$fur$_{c\_421}$ $>$:mod: silky$_{c\_1223\_a}$$>$; $<$goetee$_{c\_421}$ $>$:mod: grey$_{c\_149\_a}$$>$}:

\begin{enumerate} [resume]
\item $<$bigote$_{c\_421}$ $<$:dobj: cepillar$_{c\_932\_v}$$>$\par
$<$melena$_{c\_421}$ $<$:dobj: alisar$_{c\_1267\_v}$$>$\par
$<$pelaje$_{c\_421}$ $>$:mod: sedoso$_{c\_1223\_a}$$>$\par
$<$perilla$_{c\_421}$ $>$:mod: gris$_{c\_149\_a}$$>$\par
\end{enumerate}

All of these patterns are Unattested-Patterns, that is,  they do not occur in the Diana-Araknion corpus but are generated applying our methodology and are perfectly acceptable in Spanish. These patterns would not have been extracted using, for example, a \textit{n}-gram based method or plain statistical methods.

It is worth noting the high degree of semantic cohesion between the lemmas of the same cluster and between the lemmas of the related clusters ((8)\footnote{$<$accident$_{c\_470}$  $<$:dobj to\_cause$_{c\_560}$$>$; $<$fire$_{c\_470}$ $<$:dobj to\_avoid$_{c\_560}$$>$; $<$sinister$_{c\_470}$ $<$:dobj to\_produce$_{c\_560}$$>$.}, (9)\footnote{$<$accident$_{c\_470}$  $<$:subj to\_trigger$_{c\_560}$$>$, $<$ravage$_{c\_470}$ $<$:subj to\_produce$_{c\_560}$$>$.}, (10)\footnote{$<$chancellor$_{c\_70}$ $>$:mod argentinian$_{c\_1}$$>$; $<$ambassador$_{c\_70}$ $>$:mod belgian$_{c\_1}$$>$; $<$representative$_{c\_70}$ $>$:mod chilian$_{c\_1}$$>$} and (11)\footnote{$<$singer $_{c\_155}$ $>$:mod belgian$_{c\_1}$$>$; $<$song-writer$_{c\_155}$ $>$:mod canadian$_{c\_1}$$>$; $<$pianist$_{c\_155}$ $>$:mod american$_{c\_1}$$>$}).

\begin{enumerate} [resume]
\item $<$accidente $_{c\_470}$  $<$:dobj causar$_{c\_560}$$>$\par
$<$fuego $_{c\_470}$ $<$:dobj evitar$_{c\_560}$$>$\par
$<$siniestro $_{c\_470}$ $<$:dobj producir$_{c\_560}$$>$\par

\item $<$accidente $_{c\_470}$  $<$:subj desencadenar$_{c\_560}$$>$\par
$<$destrozo $_{c\_470}$ $<$:subj producir$_{c\_560}$$>$\par
$<$incendio $_{c\_470}$ $<$:subj originar$_{c\_560}$$>$\par

\item $<$canciller $_{c\_70}$ $>$:mod argentino$_{c\_1}$$>$\par
$<$embajador $_{c\_70}$ $>$:mod belga$_{c\_1}$$>$\par
$<$mandatario $_{c\_70}$ $>$:mod chileno$_{c\_1}$$>$\par

\item $<$cantante $_{c\_155}$ $>$:mod belga$_{c\_1}$$>$\par
$<$compositor $_{c\_155}$ $>$:mod canadiense$_{c\_1}$$>$\par
$<$pianista $_{c\_155}$ $>$:mod estadounidense$_{c\_1}$$>$\par
\end{enumerate}

This strong cohesion allows for a semantic annotation of the clusters to obtain more abstract syntactico-semantic constructions that combine semantic categories (12) and (13). The semantic labels associated with each cluster have been manually added, taking into account the WordNet \citep{wordnet} upper ontologies.

\begin{enumerate} [resume]
\item $<$\textit{Event}-n$_{c\_470}$  $<$:dobj \textit{Causative}-v $_{c\_560}$$>$\par
 $<$\textit{Event}-n $_{c\_470}$ $<$:subj \textit{Causative}-v $_{c\_560}$$>$\par
\item $<$\textit{Person/Politician}-n $_{c\_70}$ $>$:mod \textit{Nationality}-a $_{c\_1}$$>$\par
 $<$\textit{Person/Musician}-n $_{c\_155}$ $>$:mod \textit{Nationality}-a $_{c\_1}$$>$\par
\end{enumerate}

In the end, we could obtain a hierarchy of candidates to be considered as different types of  constructions, ranging from the most abstract syntactico-semantic constructions combining different semantic classes (12-13) to the most concrete lexico-syntactic constructions (i.e., lemma combinations) (8-11).


\section{Evaluation}\label{2:sec:evaluation}

In this section we evaluate the quality of the results obtained through the DISCOver methodology: 
the clusters obtained (see Section 
\ref{2:subsec:clusterevaluation}) and the lexico-syntactic patterns (see Section 
\ref{2:subsec:patternevaluation}).

\subsection{Clustering Evaluation}\label{2:subsec:clusterevaluation}

DISCOver is a methodology for discovering lexico-syntactic patterns.The clusters of 
semantically related words are a by-product that we obtain as part of the process.
Since the focus of this work is the methodology used and the patterns obtained, the 
evaluation of all possible representation and clustering algorithms is outside the scope
of this article. Nevertheless, we prepared a cluster evaluation experiment in order to
justify our choice and show that the quality of the obtained
vectors and clusters is at least comparable with other state-of-the-art methods. As
a baseline, we use standard Word2Vec \citep{mikolov2013linguistic},
representations with the recommended built-in k-means clustering algorithm. We
evaluated the resulting clusters with respect to two criteria: a) the POS purity of 
each cluster, calculated automatically; and b) the semantic coherence of the lemmas 
in each cluster, evaluated manually by experts. The criterium applied to
determine the coherence of cluster was to check if the 
words within the cluster held one of the following semantic relations: synonymy, hypernymy or hyponymy.

CLUTO obtained much higher results in terms of both evaluation criteria. The POS 
coherence of the obtained clusters was 98\%, compared to 70\% obtained by Word2Vec.
Manual evaluation shows that 99\% of the clusters obtained by CLUTO were more 
semantically coherent than the corresponding ones obtained by Word2Vec. These results
justify the representations and parameters as adequate for the task
and as comparable with the state of the art. \cite{Kovatchev:2016}
present a more in-depth comparison of the clustering algorithms using corpora of different
sizes.

\subsection{Pattern Evaluation}\label{2:subsec:patternevaluation}

Obtaining high quality lexico-syntactic patterns is the main objective of the DISCOver
methodology. In this section, we present two different evaluations of the obtained 
patterns: (1) an automatic evaluation, applying statistical association measures; 
and (2) a manual evaluation by expert linguists\footnote{An extrinsic evaluation has 
also been carried out in a text classification task (See Section \ref{2:sec:conclusions}). }. 
For these evaluations, we used the sum of the patterns of both the 15,000 and 10,000 word
configurations.

First, we evaluated the patterns automatically using statistical association measures and a
different, much larger, corpus (Diana-Araknion++). In Section~\ref{2:subsec:hypotheses}, 
we define two main properties of constructions: 1) Syntactic-semantic coherence and 
2) Generalizability. ``Syntactic-semantic coherence'' entails that the words in each
pattern need to be syntactically and semantically related. The ``syntactic coherence'' of
the patterns is not evaluated explicitly, as it is considered to be a by-product of the 
methodology: all linked clusters from which the patterns are derived have a plausible 
syntactic relationship and a high connectivity score  (see Section \ref{2:subsubsec:buildinggraph}).
However, we need to evaluate the semantic coherence of the patterns, that is, whether 
there is a semantic relation between the two lemmas. Defining and evaluating ``semantic
relatedness'' is a non-trivial task, which often requires the use of external resources, such as
WordNet and BabelNet \citep{babelnet}. However, these resources are built considering
the paradigmatic relationship between words (such as synonymy, hypernymy, and hyponymy),
while we are interested in evaluating syntagmatic relationships.

\cite{evert2008corpora} and \cite{pecina:2010} discuss the use of association measures for 
identifying collocations. They define collocations as ``the empirical concept of recurrent 
and predictable word combinations, which are a directly observable property of natural 
language''. In the context of distributional semantics, this definition corresponds to ``semantic coherence''.

In the DISCOver process, we obtained two qualitatively different types of candidates-to-be-constructions: 
Attested-Patterns, which are observed in the corpus and Unattested-Patterns, which are obtained as
a result of a generalization process that includes clustering, linking and filtering. In order to
evaluate the quality of these candidates-to-be-constructions, we formulate two hypotheses
and disprove their corresponding null hypotheses.

\begin{itemize}
	\item{\textbf{Hypothesis 1}:} \textit{The two lemmas in each construction are semantically related}. 

	\vspace{5mm}
	\underline{Null hypothesis 1 (henceforth $H_{0}$1):}
	The degree of statistical association between the two lemmas in each of the Attested-Patterns,
	measured in a corpus other than the one they were extracted from, is equal to statistical chance.

	\vspace{5mm}
	\item{\textbf{Hypothesis 2}:} \textit{Constructions can be generalized and/or derived from other 
	constructions through generalization}. Unattested-Patterns (derived through a generalization process) 
	should be possible language expressions and have the property of semantic coherence.

	\vspace{5mm}
	\underline{Null hypothesis 2.1 (henceforth $H_{0}$2.1):}
	Unattested-Patterns are not possible language expressions. They cannot appear in a corpus.

	\vspace{5mm}
	\underline{Null hypothesis 2.2 (henceforth $H_{0}$2.2):}
	If Unattested-Patterns appear in a corpus, they will not have the property of semantic coherence. 
	That is, they will have association scores equal to statistical chance.

\end{itemize}

In order to prove the two main hypotheses we needed to disprove the three null hypotheses. 

For a baseline of $H_{0}$1, we extracted a list of all bigrams (BI-Patterns) from the 
original Diana-Araknion corpus. Each bigram contains at least one of the 15,000 most frequent words. 
We removed all bigrams containing non-content words. All of the Attested-Patterns and the 
BI-Patterns were found and extracted from the Diana-Araknion 100M token corpus. 

For a baseline of $H_{0}$2.1, we generated patterns by combining frequent lemmas (FL-Patterns): 
FL-Patterns-15 contain all combinations of the most
frequent 15,000 lemmas found in the Diana-Arakion corpus; FL-Patterns-30 contain all
combinations in which one lemma is among the 15,000 most frequent lemmas and the other
among the 30,000 most frequent ones; FL-Patterns-all contain all word combinations
which contain at least one of the 15,000 most frequent lemmas\footnote{The total number
of lemmas used in the FL-Patterns (all) is 422,000.}.

We use two different statistical methods \citep{evert2008corpora}: simple Mutual Information 
(MI), which is an effect size measure, and the Z-score (Z-sc), which is an evidence-based 
measure. Effect-size measures and evidence-based measures are qualitatively different, and
for evaluation can be used complementarily. Our final experimental setup includes 
the following: 

\begin{itemize}
	\item Attested-Patterns, in five different test groups, based on their observed frequency
	in the Diana-Araknion corpus: 

	\begin{itemize}
		\item Att-Patterns-all with an original frequency of 1 or more
		\item Att-Patterns-2 with an original frequency of 2 or more
		\item Att-Patterns-3 with an original frequency of 3 or more
		\item Att-Patterns-4 with an original frequency of 4 or more
		\item Att-Patterns-5 with an original frequency of 5 or more
	\end{itemize}

	\item BI-Patterns, with an original frequency of 5 or more\footnote{5,285 of the
	BI-patterns coincide with Attested-Patterns.}

	\item Unattested-Patterns

	\item FL-Patterns-15, FL-Patterns-30, FL-Patterns-all

\end{itemize}

\textbf{Evaluating $H_{0}$1}:

We calculated the MI and Z-sc association scores of the two words 
in each of the Attested-Patterns and BI-Patterns in the Diana-Araknion++ 600M token corpus. The association 
score was calculated based on the sentential co-occurrence of the two words. Patterns that co-occurred 
less than 5 times obtained a score of 0. First, we compared the obtained association with standard 
thresholds, representing statistical chance: 0, 0.5, and 1 for MI; 0, 1.96, and 3.29 for Z-sc. Second, 
we compared the average association score of the Attested Patterns with those of the BI-Patterns.

Table \ref{2:tau:assocTR} shows what percentage of the Attested-Patterns in each group obtains scores
higher than statistical chance. Overall, the majority of the Attested-Patterns outperform the statistical 
chance baseline. The results are consistent for both the measures and 
their thresholds, even though they measure the association in a qualitatively different manner. It is 
important to note that filtering out the Attested-Patterns with a frequency of 1 significantly improves the 
results. We believe this factor should be taken into consideration in future experiments.

\begin{table}[h]
\caption{Association score of Attested-Patterns compared with statistical chance}
\centering
\begin{tabular}{lcccccc}
& \multicolumn{3}{c}{MI} & \multicolumn{3}{c}{Z-sc} \\
\hline \hline
\textbf{Patterns} & \textbf{\textgreater0} & \textbf{\textgreater0.5} & \textbf{\textgreater1} & \textbf{\textgreater0} & \textbf{\textgreater1.96} & \textbf{\textgreater3.29} \\
\hline 
Att-Patterns-5 &	85\% & 83\% & 80\% & 85\% & 83\% & 82\%\\
Att-Patterns-4 &	84\% & 82\% & 79\% & 84\% & 82\% & 80\%\\
Att-Patterns-3 &	82\% & 80\% & 77\% & 82\% & 80\% & 78\%\\
Att-Patterns-2 &	78\% & 76\% & 72\% & 78\% & 76\% & 73\%\\
Att-Patterns-all & 68\% & 66\% & 62\% & 68\% & 65\% & 62\%\\
\end{tabular}
\label{2:tau:assocTR}
\end{table}

As a complementary evaluation, we directly compared the association scores of the Attested-Patterns with 
those of the BI-Patterns. Table \ref{2:tau:assocAVG} shows the average association scores for the two 
types of patterns\footnote{The average is calculated as a simple average of all patterns of the
corresponding type.}. The Attested-Patterns have a much higher degree of association 
than the BI-Patterns. In the case of MI, the Attested-Patterns obtain scores more than two times higher than 
the BI-Patterns. In the case of Z-sc, the Attested-Patterns obtain scores between 30\% and 100\% higher than 
the BI-Patterns.

\begin{table}[h]
\caption{Average association score of Attested-Patterns and BI-patterns}
\centering
\begin{tabular}{lcc}
\hline \hline
\textbf{Patterns}& \textbf{Average MI} &\textbf{Average Z-sc} \\
\hline
Attested-Patterns-5 	& 3.90 & 52\\
Attested-Patterns-4 	& 3.86 & 49\\
Attested-Patterns-3 	& 3.80 & 46\\
Attested-Patterns-2 	& 3.70 & 42\\
Attested-Patterns-all 	& 3.50 & 35\\
\noalign{\vspace {.5cm}}
BI-Patterns 		& 1.72 & 27\\
\hline \hline
\end{tabular}
\label{2:tau:assocAVG}
\end{table}

The obtained results disprove $H_{0}$1 and confirm Hypothesis 1. That is, we can conclude that the 
Attested-Patterns are semantically coherent.

\textbf{Evaluating $H_{0}$2.1}:

We checked how many of the Unattested-Patterns were present in Diana-Araknion++. As a baseline we used
the FL-Patterns. Both Unattested-Patterns 
and FL-Patterns are not directly obtained, but are rather a result of generalization and generation using different 
methodologies. For each group, we calculated the percentage of the patterns that appear once and the 
percentage of the patterns that appear at least five times. Table \ref{2:tau:upocc} shows the results obtained. 

\begin{table}[h]
\caption{Occurrence of Unttested-Patterns and FL-Patterns}
\centering
\begin{tabular}{lcc}
\hline \hline
\textbf{Patterns} & \textbf{Occurred Once} & \textbf{Occurred Five Times}  \\
\hline 
Unattested-Patterns &	54\% & 24\%\\
\noalign{\vspace {.5cm}}
FL-Patterns-15 & 24\% & 9\% \\
FL-Patterns-30 & 11\% & 4\% \\
FL-Patterns-all & 4\% & 0.6\% \\
\hline \hline
\end{tabular}
\label{2:tau:upocc}
\end{table}

Unattested-Patterns 
appear much more frequently than the patterns generated by simply combining frequent lemmas. 
56\% of the Unattested-Patterns were observed in Diana-Araknion++. This is more than double
the observance rate of the FL-Patterns-15 and five times higher than for FL-Patterns-30. 24\% of the
Unattested-Patterns appear in Diana-Araknion++ with a frequency of 5 or more. This is almost three
times higher than FL-Patterns-15 and six times higher than FL-Patterns-30. The results of 
FL-Patterns-all are much lower, showing that unfiltered pattern generation is not effective. 
Unattested-Patterns are linguistic patterns given that they appear in a corpus with
a much higher probability than patterns generated using a simpler frequency based methodology.
These results disprove $H_{0}$2.1.

\textbf{Evaluating $H_{0}$2.2}:

We calculated the association score (MI and Z-sc) between 
the lemmas in each of the Unattested-Patterns that occurred at least 5 
times\footnote{Calculating this score for patterns with lower frequency is unreliable 
due to the low-frequency bias in some of the measures.} in Diana-Araknion++. We
compared the scores with the same thresholds we used when evaluating $H_{0}$1. Table
\ref{2:tau:upassoc} shows the percentage of patterns with a score higher than the
statistical chance thresholds. 

\begin{table}[h]
\caption{Association scores of Unttested-Patterns}
\centering
\begin{tabular}{lcccccc}
& \multicolumn{3}{c}{MI} & \multicolumn{3}{c}{Z-sc} \\
\hline \hline
\textbf{Patterns} & \textbf{\textgreater0} & \textbf{\textgreater0.5} & \textbf{\textgreater1} & \textbf{\textgreater0} & \textbf{\textgreater1.96} & \textbf{\textgreater3.29} \\
\hline 
Unattested-Patterns &	93\% & 86\% & 76\% & 93\% & 80\% & 70\%\\
\end{tabular}
\label{2:tau:upassoc}
\end{table}

The observed degree of association is very high. Over 90\% of the observed
Unattested-Patterns obtained a positive association score with respect to both 
measures. When comparing them with the statistical chance thresholds, the 
obtained results are similar to those obtained by Attested-Patterns in $H_{0}$1. 
The Unattested-Patterns, when observed in a different corpus, are semantically coherent. This 
disproves $H_{0}$2.2.

In conclusion, the automated statistical evaluation of the patterns 
obtained by DISCOver shows that: (1) Attested-Patterns are semantically coherent, as they outperform two baselines: statistical chance thresholds
and BI-Patterns. These results disprove $H_{0}$1.; (2) A significant percentage 
(56\%) of the Unattested-Patterns can be found in Diana-Araknion++, which 
is much higher than the occurrence of FL-Patterns. These results disprove $H_{0}$2.1; 
(3) Whenever Unattested-Patterns occur in Diana-Araknion++, the statistical 
association between the lemmas in the patterns is much higher than the 
statistical chance baseline. This disproves $H_{0}$2.2.

As we have disproved all 3 of the null hypotheses, we can conclude that
the patterns obtained by the DISCOver methodology have both properties of
constructions: syntactic and semantic coherence and generalizability. Therefore
they are good candidates-to-be-constructions.

We also performed a manual evaluation of the lexico-syntactic patterns. This 
complementary validation reinforces the results obtained in the two statistical evaluations. 
We prepared a dataset of 600 patterns for the manual evaluation: 300 patterns obtained 
by applying the DISCOver methodology (the patterns were randomly selected from all Attested
and Unattested Patterns) and 300 of the FL-Patterns-15. Three experts were asked to classify each pattern as a 
correct or incorrect construction. The instructions given to them were: a) evaluate whether the
pattern is a possible Spanish pattern in your judgement as a native speaker; b) in case of doubt, 
consult the Google Search engine to check whether it is used by users. Our research questions 
in this evaluation were: 1) How do the experts evaluate the patterns obtained by DISCOver?;
2) Are experts more likely to accept patterns obtained by DISCOver than random patterns 
of frequent words?

The average percentage of agreement between the three annotators was 81.67\% (see
Table \ref{2:tau:iaatest}), which is considered high for a semantic evaluation task. The 
corresponding Fleiss Kappa score is 0.602 with expected agreement of 0.539, which is 
statistically significant.

\begin{table}[h]
\caption{Interannotator agreement test}
\centering
\begin{tabular}{lc}
\textbf{Annotators (A)}& \textbf{\%Agreement}\\
\hline
A1 and A2 & 85\%\\
A1 and A3 & 80.17\%\\
A2 and A3 & 79.83\%\\
\noalign{\vspace {.5cm}}
A1, A2 and A3& 81.67\%\\
\end{tabular}
\label{2:tau:iaatest}
\end{table}

The results of the evaluation are shown in Table \ref{2:tau:manresults}. We use three pattern quality categories. ``Strict Positive'' includes patterns that were
annotated as positive by all three annotators, ``Positive'' includes patterns that were
annotated as positive by at least two annotators and ``Negative'' groups together patterns that were
annotated as positive by one or none of the annotators. The experts
accepted the majority of the DISCOver patterns as constructions. At the same time
they rejected the majority of the FL-Patterns. We also want to highlight that the percentage of ``Strict Positive"
patterns is very similar to the percentage of patterns that obtain a high association
score. These findings confirm the results that we obtained in the automatic
evaluation (See Tables \ref{2:tau:assocTR} and \ref{2:tau:upassoc}).

\begin{table}[h]
\caption{Expert evaluation}
\centering
\begin{tabular}{lcc}
\textbf{}& \textbf{DISCOver} &\textbf{FL-Patterns}\\
\textbf{Strict Positive} & 84\% & 14\% \\
\textbf{Positive} & 93\% & 38\% \\
\textbf{Negative} & 7\% & 62\% \\
\end{tabular}
\label{2:tau:manresults}
\end{table}


\section{Conclusions and Future Work} \label{2:sec:conclusions}

This article describes DISCOver, an unsupervised methodology for automatically identifying lexico-syntactic patterns to be considered as constructions. We based this methodology on the pattern-construction hypothesis, which states that the linguistic contexts that are relevant for defining a cluster of semantically related words tend to be (part of) a lexico-syntactic construction.

Following this assumption, we developed a bottom-up language independent methodology to discover lexico-syntactic patterns in corpora. The DSM developed allows us to model the contexts of words (lemmas) taking into account their dependency directions and dependency labels.  We applied a clustering process to the resulting matrix to obtain clusters of semantically related lemmas. Then we linked all the clusters that were strongly semantically related and we used them as a source of information for deriving lexico-syntactic patterns, obtaining a total number of 220,732 candidates to be constructions. We evaluated the DISCOver methodology by applying different evaluations. First, the patterns were automatically evaluated using statistical association measures and a different, much larger, corpus. We evaluated whether the patterns we generated obtained a significantly higher association score than statistical chance. We also compared the asociation scores of the DISCOver patterns with a baseline of bigrams. DISCOver obtained better results with respect to both baselines. The patterns obtained by generalization were additionaly evaluated against a baseline of randomly generated patterns. DISCOver significantly outperforms these baselines. Second, the patterns were manually evaluated by expert linguists obtaining good results (89.33\%).

This methodology only requires having at one's disposal a medium-sized corpus automatically annotated with POS tags and syntactic dependencies. Therefore, our methodology can be easily replicated with other corpora and other languages. For instance, the DISCOver patterns were also used in a text
classification task \citep{francosalvadoretal:2015}. The patterns
obtained using our methodology have been compared to other
representations (i.e., tf-idf, tf-idf \textit{n}-grams, and
enriched graph). The use of these patterns results
in an accuracy of 91.69\%, which outperfoms the representations
based on tf-idf (25.26\%), tf-idf \textit{n}-grams (79.26\%) and
an enriched graph (43.98\%),
proving to be the best option to represent the content of the
corpus.

Furthermore, our methodology increases the descriptive power of the source corpus. First, the lexico-syntactic patterns generated constitute a structured and formalized semantic representation of the corpus. Second, the linking process enlarges the content of the initial data with new relationships not directly present in the corpus (i.e., a total of 167,443 Unattested-Patterns).

The Diana-Araknion-KB\footnote {Available at \url{http://clic.ub.edu/corpus/}} can be used as a source of information to derive relevant linguistic information, such as the selection restrictions of verbs, nouns and adjectives; to disambiguate syntactic analysis in order to discard candidate parse trees; to provide a knowledge base of related words with a high degree of association measures for psycholinguistic research; and, to allow for a fine-grained corpus comparision. 

The methodology presented and the results obtained, which are available in the Diana-Araknion-KB, open several lines of future research. 

First, the Diana-Araknion-KB can be used as a source of information for the development of patterns at different levels of abstraction, in such a way as to obtain a hierarchy of patterns with components belonging to different levels of linguistic knowledge, that is, combining lexical, morpho-syntactic and semantic information.
Second, since the same semantic category can be shared by more than one cluster, we could group them into metaclusters containing all the clusters with the same semantic category. 
Third, a further cluster linking process could be carried out allowing all members of a metacluster to combine with all the target clusters that are related with at least one of the members of the metacluster.
Fourth, constructions could be linked in terms of transitivity to obtain larger structures. That is, if cluster A combines with cluster B, and B combines with cluster C, we have the candidate construction: A+B+C.
Fifth, the methodology can be used to extract and study patterns in corpora from a specific area, such as the Biomedical domain.

To sum up, we consider that this methodology for discovering constructions outperforms the results of other proposals in the sense that it is fully automatic, language independent, and easily replicable in other corpora and languages. The quality of the results obtained and their wide range of possible applications confirm the DISCOver methodology as a promising line of research and DSMs as a good choice for discovering linguistic knowledge.

\part{Paraphrase Typology and \\ Paraphrase Identification}\label{p:type}


\chapter[WARP-Text: a Web-Based Tool for \\Annotating Relationships between Pairs of Texts]{\centering WARP-Text: a Web-Based Tool for Annotating Relationships between Pairs of Texts}\label{ch:warp}

\chaptermark{WARP-Text}

\begin{center}
Venelin Kovatchev, M. Ant{\`o}nia Mart{\'i}, and Maria Salam{\'o}

University of Barcelona

\vspace{10mm}

Published at \\ \textit{Proceedings of the \\27th International Conference on Computational Linguistics: System Demonstrations}, 2018 \\ pp.: 132-136
\end{center}

\paragraph{Abstract}	We present WARP-Text, an open-source web-based tool for annotating relationships between pairs of texts. WARP-Text supports multi-layer annotation and custom definitions of inter-textual and intra-textual relationships. Annotation can be performed at different granularity levels (such as sentences, phrases, or tokens). WARP-Text has an intuitive user-friendly interface both for project managers and annotators. WARP-Text fills a gap in the currently available NLP toolbox, as open-source alternatives for annotation of pairs of text are not readily available. WARP-Text has already been used in several annotation tasks and can be of interest to the researchers working in the areas of Paraphrasing, Entailment, Simplification, and Summarization, among others. 

\section{Introduction}

%
%
    %
    %
    %
    %
    %
    %

	Multiple research fields in NLP have pairs of texts as their object of study:
	Paraphrasing, Textual Entailment, Text Summarization, Text Simplification, Question
	Answering, and Machine Translation, among others. All these fields benefit from high 
	quality corpora, annotated at different granularity levels. However, existing 
	annotation tools have limited capabilities to process and annotate such corpora. The 
	most popular state-of-the-art open source tools do not natively support detailed pairwise 
	annotation and require significant adaptations and modifications of the code for such tasks.

	We present the first version of WARP-Text, an open source\footnote{The code 
	is available at \url{https://github.com/venelink/WARP} under Creative Commons Attribution 4.0 International License.} web-based annotation 
	tool, created and designed
	specifically for the annotation of relationships between pairs of texts at multiple 
	layers and at different granularity levels. Our objective was to create a tool
	that is functional, flexible, intuitive, and easy to use. WARP-Text was built using 
	PHP and MySQL standard implementation.
	
	WARP-Text is highly configurable: the administrator interface manages the number, 
	order, and content of the different annotation layers. The pre-built layers allow for 
	custom definitions of labels and granularity levels. The system architecture is flexible 
	and modular, which allows for the modification of the existing layers and the addition 
	of new ones. 

	The annotator interface is intuitive and easy to use. It does not require previous 
	knowledge or extensive annotator training. The interface has already been 
	used 	in the task of annotating atomic paraphrases \citep{etpc} and is 
	currently being used on two annotation tasks in Text Summarization. The learning process 
	of the annotators was quick and the feedback was overwhelmingly positive. 
	
	The rest of this article is organized as follows. Section \ref{3:rw} presents the Related
	Work. Section \ref{3:sa} describes the architecture of the interface, the annotation
	scheme, the usage cases, and the two interfaces: administrator and annotator. Finally, 
	Section \ref{3:conc} presents the conclusions and the future work.

\section{Related Work}\label{3:rw}

	In the last several years, the NLP community has shown growing interest in tools that
	are web-based, open source, and multi-purpose: WebAnno \citep{webanno}, Inforex 
	\citep{inforex}, and Anafora \citep{anafora}. Other popular non web-based annotation 
	systems include GATE \citep{gate} and AnCoraPipe \citep{ancora}.
	These systems are intended to be feature-rich and multi-purpose. However, in many
	tasks, it is often preferable to create a specialized annotation tool to address problems that are non-trivial to solve
	using the multi-purpose annotation tools. One such problem is working with multiple 
	texts in parallel. While multi-purpose annotation tools can be adapted for such use, this often leads to a more complex annotation scheme,
	complicates the annotation process, requires additional annotator training
	and post-processing of the annotated corpora. \citet{Toledo}
	and more recently \citet{NASTASE18}, \citet{BATAN18}, and \citet{ARASE18} emphasize the 
	lack of a feature-rich open-source tool for annotation of pairs of texts\footnote{See also the discussion
	about looking for tools for annotating pairs of texts in the Corpora Mailing List (May 2017):
	\url{http://mailman.uib.no/public/corpora/2017-May/026526.html} - \url{http://mailman.uib.no/public/corpora/2017-May/026619.html}}. Some of these
	authors develop simple custom-made tools  with limited re-usability, designed for for carrying out one specific annotation task. 
	WARP-Text aims to address this gap in the NLP toolbox by providing a feature rich system
	which could be used in all these annotation scenarios.

	To the best of our knowledge, the only existing multi-purpose tool that is designed to work with pairs
	of text and allows for detailed annotation is CoCo \citep{coco}. It has already been used
	for annotations in paraphrasing \citep{Vila2015} and plagiarism detection \citep{Barron}. 
	However, CoCo is not open source and is currently not being supported or updated. 

\section{WARP-Text}\label{3:sa}

	By addressing various limitations of existing tools, WARP-Text fills a gap in the
	state-of-the-art NLP toolbox. It offers project managers and annotators a rich set
	of functionalities and features: the ability to work with pairs of texts simultaneously; multi-layer
	annotation; annotation at different granularity levels; annotation of discontinuous scope
	and long-distance dependencies; and the custom definition of relationships.
	WARP-Text consists of two separate web interfaces: annotator and administrator. In the \textit{administrator interface} 
	the project manager configures the annotation scheme, defines the relationships and sets 
	all parameters for the annotation process. The annotators work in the \textit{annotator interface}. 

	 WARP-Text is a tool for qualitative document annotation. It provides a wide range of 
	configuration options and can be used for fine-grained annotation. It is best suited to 
	medium sized corpora (containing thousands of small documents) and is not fully 
	optimized for processing, analyzing, searching, and annotating large corpora (containing 
	millions of documents). WARP-Text has full UTF-8 support and is language independent in 
	the sense that it can handle documents in any UTF-8 supported natural language. 
	So far it has been used to annotate texts in English, Bulgarian (Cyrillic), and Arabic.

	WARP-Text is a multi-user system and provides two different forms of interaction 
	between the different annotators. In the \textit{collaborative mode}, multiple annotators work 
	on the same text and each annotator can see and modify the annotations of the others.
	In the \textit{independent mode}, the annotators perform the annotation independently from
	one another. The different annotations can then be compared in order to calculate 
	inter-annotator agreement.

\subsection{Annotation Scheme}

	The atomic units of the annotation scheme in WARP-Text are \textit{relationships}. The 
	properties of the \textit{relationships} are \textit{label} and \textit{scope}. The 
	\textit{scope} of a \textit{relationship} is a list of continuous or discontinuous \textit{elements} in each of the two 
	texts. The granularity level of the scope determines the \textit{element} type. An \textit{element} can be 
	the whole text, a sentence, a phrase, a token, or can be defined manually.
	A \textit{layer} in WARP-Text is a set of relationships, whose scopes belong to the 
	same granularity level\footnote{There is no one-to-one correspondence between
	granularity level and annotation layer. Each annotation layer is a sub-task 
	in the main annotation task. Multiple annotation layers can work at the same granularity 
	level. For example: at layer (1) the annotator annotates the semantic relations between 
	the tokens in the two texts; at layer (2) the annotator annotates the scope of negation and 
	the negation cues in the two texts. Both layer (1) and layer (2) work at the token 
	granularity level.}. The definition of relationships and their grouping into layers is fully 
	configurable through the administrator interface.
	WARP-Text supports multi-layer annotation. That is, the same pair of texts can be 
	annotated multiple times, at different granularity levels and using different sets of 
	relationships.

\subsection{Administrator Interface}\label{3:admin}

	The administrator interface has three main modules: a) the \textit{dataset management module},
	b) the \textit{user management module}, and c) the \textit{layer management module}.
	In the \textit{dataset management module} the project manager can: a) import a corpus, in a
	delimited text format, for annotation; b) monitor the current annotation status and statistics; and 
	c) export the annotated corpus as an SQL file or an XML file.
	In the \textit{user management module} the project manager creates new users and modifies
	existing ones. In this module the project manager also distributes the tasks (pairs) among the 
	annotators.
	In the \textit{layer management module} the project manager configures each of the
	layers and determines the order of the layers in the annotation process. The project
	manager configures for each individual layer: 1) the granularity level; 2) the 
	relationships that belong to the layer; 3) the sub-relationships or properties of 
	the relationships; 4) optional parameters such as ``sentence lock'' and ``display
	previous layers''.

\subsection{Annotator Interface}\label{3:anno}

	The annotator interface has three main modules: a) the \textit{annotation statistics module}, 
	b) the \textit{review annotations module}, and c) the \textit{annotation panel module}. 
	In the \textit{annotation statistics module} the annotator monitors the progress of the 
	annotation and sees statistics such as the number of annotated pairs, and the remaining 
	number of pairs. 
	In the \textit{review annotations module} the annotator reviews the text pairs (s)he 
	already annotated and introduces corrections where necessary.
	The \textit{annotation panel module} is the core of the annotator interface. One of our 
	main objectives in the creation of WARP-Text was to make it easier to use for the 
	annotators and to optimize the annotation time. For that reason we have made the
	\textit{annotator panel module} as automated as possible and have limited the 
	intervention of annotators to a minimum. The \textit{annotation panel module} is 
	generated dynamically, based on the user and project configuration. It loads the first text 
	pair, 	assigned to the current annotator and guides the annotator through the different 
	layers in the order specified by the project manager. Once the text pair has been 
	annotated at all configured layers, the module  updates the database, loads the next 
	pair and repeats the process.

	We illustrate the annotation process with the interface configuration that was
	used in the annotation of the Extended Typology Paraphrase Corpus (ETPC) \citep{etpc}.
	The annotation scheme of ETPC consists of two layers: one layer that is configured for
	annotation at the text granularity level; and one layer that is configured for
	annotation at the token granularity level.

	\begin{figure}[h]
	\begin{center}
		\includegraphics[width=\textwidth]{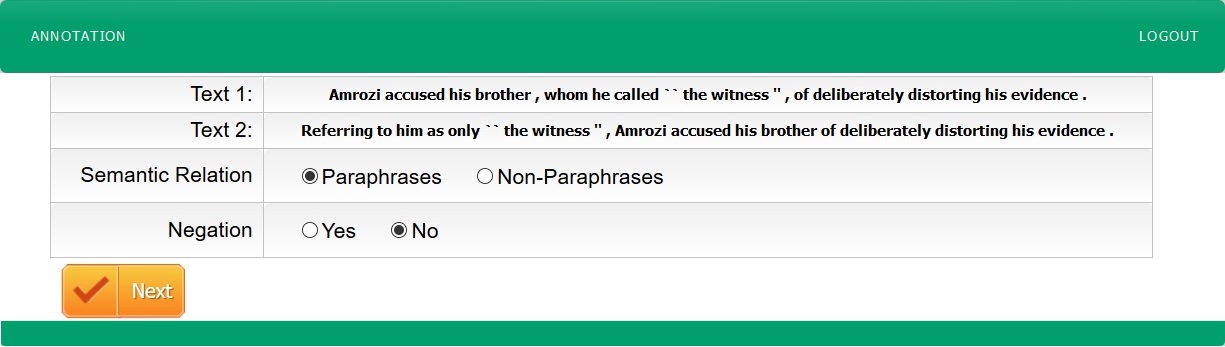}
		\caption{Annotating relationships at textual level.}
		\label{3:an:textual}
	\end{center}
	\end{figure}

	The textual layer (Figure \ref{3:an:textual}) displays the two texts and allows the annotator to select the 
	values for an arbitrary number of relationships between the texts. In the case of 
	ETPC, the two textual relationships that we were interested in were: 1) ``The 
	semantic relationship between the two texts'': ``Paraphrases'' or ``Non-paraphrases''; 
	and 2) ``The presence of negation in either of the two sentences'': ``Yes'' or ``No''. 
	In ETPC, both relationships had two 
	possible options, however WARP-Text supports multiple options for each 
	relationship. In this first layer, the scope of the relationship is the whole text.

	\begin{figure}[h]
	\begin{center}
		\includegraphics[width=\textwidth]{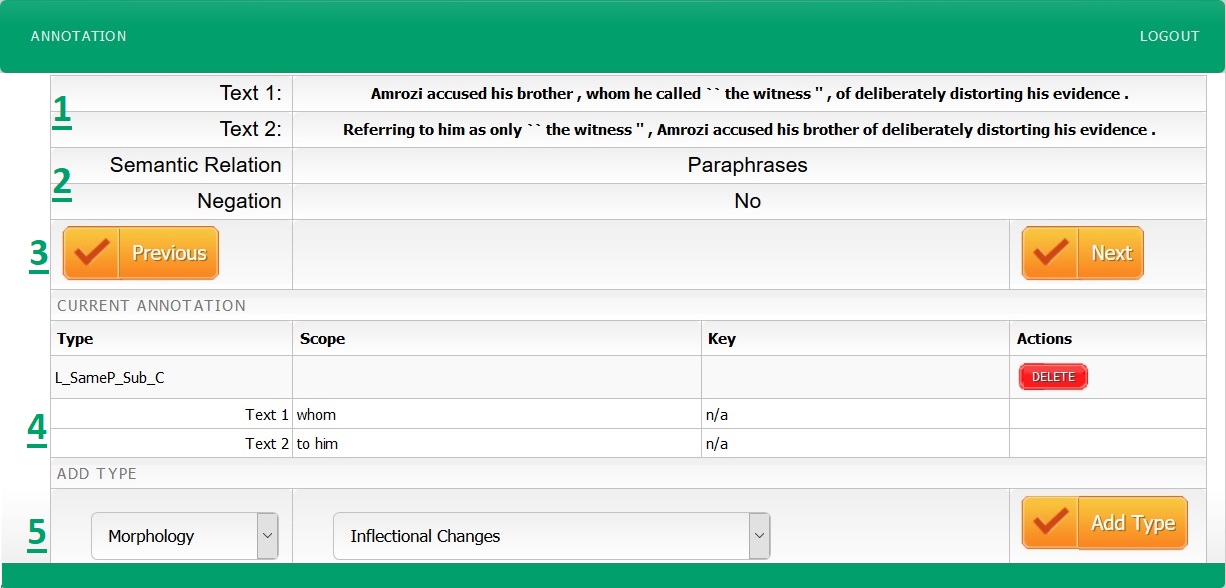}
		\caption{Annotating relationships at token level.}
		\label{3:an:token}
	\end{center}
	\end{figure}

	The second layer (Figure \ref{3:an:token}) has five functional parts, labeled in the figure with numbers from 1 to 5. 
	The annotator can see the two texts 
	in (1), the annotation at the previous layers in (2), and at the annotation at the current layer in (4). (3) is
	the navigation panel between the different layers. Finally, (5) is where the annotator can choose to add a 
	new relationship. The list of possible relationships is defined by the project manager in the administrator
	interface. In the case of ETPC we organized the relationships in a two-level hierarchical system based on
	their linguistic meta-category. The token-layer annotation is more complex than the textual-layer annotation
	as it requires the annotation of scope in addition to the annotation a label\footnote{The token level annotation
	layer is an instance of the more general ``atomic level annotation layer''. The organization and work flow 
	described here are the same when the granularity level is ``paragraph'', ``sentence'', ``phrase'', or 
	custom defined.}. When the annotator chooses a relationship, the "Add Type" button goes to the scope 
	selection page (Figure \ref{3:an:scope}). The 	scope can be discontinuous and can include elements from 
	one of the texts only or from both. In the case 
	of ETPC, the elements that the annotator can select are tokens. In other configurations, they can be 
	phrases or sentences. 

	\begin{figure}[h]
	\begin{center}
		\includegraphics[width=\textwidth]{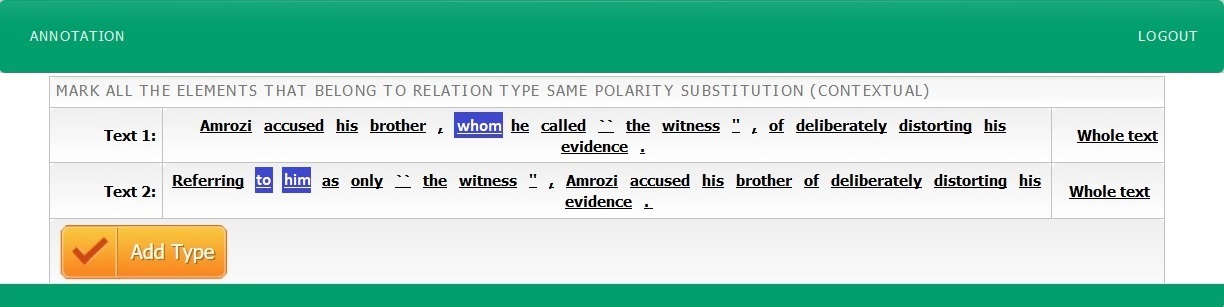}
		\caption{Scope selection page.}
		\label{3:an:scope}
	\end{center}
	\end{figure}

	The flexibility of WARP-Text makes it easy to adapt for multiple tasks. The textual layer can be used in tasks
	such as the annotation of textual paraphrases, textual entailment, or semantic similarity. The atomic level
	annotation layer has even more applications. As we showed in ETPC, it can be used to annotate fine-grained 
	similarities and differences between pairs of texts. It can also be used for tasks such as manual correction of 
	text alignment. Another possible use is, given a summary or a simplified text, to identify in the reference text 
	the exact sentences or phrases which are summarized or simplified.

\section{Conclusions and Future Work}\label{3:conc}

	In this paper we presented WARP-Text, a web-based tool for annotating relationships
	between pairs of texts. Our software fills an important gap as the high quality 
	annotation of pairwise corpora at different granularity levels is needed and can benefit multiple fields in NLP. 
	Previously available tools are not well suited for the task, require substantial 
	modification, or are hard to configure. The main advantages of WARP-Text 
	are that it is feature-rich, open source, highly configurable, 
	and intuitive and easy to use.

	As future work, we plan to add several functionalities to both interfaces. In the 
	administrator interface, we plan to offer project managers tools for 
	visualization and data analysis, and automatic calculation of inter-annotator 
	agreement. In the annotator interface, we plan to fully explore the advantages
	of multi-layer architecture. By design, WARP-Text can support parent-child 
	dependencies between layers. However, the pre-built modules available
	in this first release of the tool use only independent layers. That is, the annotation 
	at one layer does not affect the configuration of the other layers. We also 
	plan to explore the possibility of incorporating external automated pre-processing tools.


\chapter[ETPC - a Paraphrase Identification Corpus \\ Annotated with Extended Paraphrase Typology and Negation]{\centering ETPC - a Paraphrase Identification Corpus Annotated with Extended Paraphrase Typology and Negation}\label{ch:etpc}

\chaptermark{ETPC}

\begin{center}
Venelin Kovatchev, M. Ant{\`o}nia Mart{\'i}, and Maria Salam{\'o}

University of Barcelona

\vspace{10mm}

Published at \\ \textit{Proceedings of the \\Eleventh International Conference on Language Resources and Evaluation}, 2018 \\ pp.: 1384-1392
\end{center}

\paragraph{Abstract} We present the Extended Paraphrase Typology (EPT) and the Extended Typology Paraphrase Corpus (ETPC). 
The EPT typology addresses several practical limitations of existing paraphrase typologies: it is the first typology 
that copes with the non-paraphrase pairs in the paraphrase identification corpora and distinguishes between 
contextual and habitual paraphrase types. ETPC is the largest corpus to date annotated with atomic paraphrase 
types. It is the first corpus with detailed annotation of both the paraphrase and the non-paraphrase pairs and the 
first corpus annotated with paraphrase and negation. Both new resources contribute to better understanding the 
paraphrase phenomenon, and allow for studying the relationship between paraphrasing and negation. To the 
developers of Paraphrase Identification systems ETPC corpus offers better means for evaluation and error analysis. 
Furthermore, the EPT typology and ETPC corpus emphasize the relationship with other areas of NLP such as 
Semantic Similarity, Textual Entailment, Summarization and Simplification.

\paragraph{Keywords} Paraphrasing, Paraphrase Typology, Paraphrase Identification

\section{Introduction}

	The task of Paraphrase Identification (PI) consists of comparing two texts of arbitrary size in order to 
	determine whether they have approximately the same meaning. The most common approach to PI is 
	as a binary classification problem, in which a system learns to make correct binary predictions 
	(paraphrase or non-paraphrase) for a given pair of texts. The task of PI is challenging from more than one 
	point of view. From the resource point of view, defining the task and preparing high quality corpora is 
	a non-trivial problem due to the complex nature of ``paraphrasing''. From the application point of view,
	for a system to perform well on PI often requires a complex ML architecture and/or a large set of
	manually engineered features. From the evaluation point of view, the classical task of PI does not offer 
	many possibilities for detailed error analysis, which in turn limits the reusability and the improvement of 
	PI systems.  

	In the last few years, researchers in the field of paraphrasing have adopted the approach of decomposing 
	the meta phenomenon of \textit{``textual paraphrasing''} into a set of \textit{``atomic paraphrase''} 
	phenomena, which are more strictly defined and easier to work with. \textit{``Atomic paraphrases''}
	are hierarchically organized into a typology, which provides a better means to study and understand
	paraphrasing. While the theoretical advantages of these approaches are clear, their practical implications
	have not been fully explored. The existing corpora annotated with paraphrase typology are limited in size,
	coverage and overall quality. The only corpus of sufficient size to date annotated with paraphrase typology 
	is the corpus by \citet{Vila2015}, which contains 3900 re-annotated \textit{``textual paraphrase''} pairs 
	from the MRPC corpus \citep{mrpc}.

	The use of a paraphrase typology in practical tasks has several advantages. First, \textit{``atomic 
	paraphrases''} are much more strict in their definition, which makes the results more useful and understandable. 
	Second, the more detailed annotation can be useful to (re)balance binary PI corpora in 
	terms of type distribution. Third, annotating a corpus with paraphrase types provides much better feedback 
	to the PI systems and allows for a detailed, per-type error analysis. Fourth, enriching the corpus and 
	improving the evaluation can provide a linguistic insight into the workings of complex machine learning 
	systems (i.e. Deep Learning) that are traditionally hard to interpret. Fifth, corpora annotated with a paraphrase typology 
	open the way for new research and new tasks, such as ``PI by type`` or ``Atomic PI in context''.
	Finally, decomposing \textit{``textual paraphrases''} can help relate the task of PI to tasks such as Recognizing Textual 
	Entailment, Text Summarization, Text Simplification, and Question Answering.
	
	In this paper, we present the Extended Typology Paraphrase Corpus (ETPC), the result of annotating the 
	MRPC \citep{mrpc} corpus with our Extended Paraphrase Typology (EPT). EPT is oriented towards practical 
	applications and takes inspiration from several authors that work on the typology of paraphrasing and 
	textual entailment. To the best of our knowledge, this is the first attempt to make a detailed annotation of the
	linguistic phenomena involved in both the positive (paraphrases) and negative (non-paraphrases) examples
	in the MRPC (for a total size of 5801 textual pairs). The focus on non-paraphrases and the qualitative and 
	quantitative comparison between \textit{``textual paraphrases''} and \textit{``textual non-paraphrases''} 
	provides a different perspective on the PI task and corpora. 

	As a separate layer of annotation, we have identified all pairs of texts that include negation and we have 
	annotated the negation scope. This makes ETPC the first corpus that is annotated both with paraphrasing 
	and with negation.

	The rest of this article is organized as follows. Section \ref{4:rw} is devoted to the Related Work. 
	Section \ref{4:et} describes the proposed Extended Typology, the reasons and the practical considerations behind 
	it. Section \ref{4:ann} explains the annotation process, the annotation scheme and instructions, the tool that we
	used and the corpus preprocessing. Section \ref{4:corp} presents ETPC, with its structure and type distribution.
	It discusses the results of the annotation and outlines some of the practical applications of the corpus. 
	Finally, Section \ref{4:conc} concludes the article and outlines the future work.

\section{Related Work} \label{4:rw}

	The task of PI is one of the classical tasks in NLP. Several corpora can be used in the task for training and/or 
	for evaluation. Traditionally, PI is addressed using the MRPC corpus \citep{mrpc}. The MRPC corpus consists 
	of 5801 pairs, that have been manually annotated as paraphrases or non-paraphrases. 
	More recently, \citet{ppdb} introduce PPDB - a very large automatic collection of paraphrases, which 
	consists of 220 million pairs. The introduction of PPDB allowed for the training of deep learning systems, 
	due to the significant increase of the available data. However, the quality of the PPDB pairs is much lower 
	than those of MRPC, which makes it less reliable for evaluation. A common approach is to work on both 
	datasets simultaneously - using the PPDB for training, and the MRPC for development and evaluation. 

	Closely related to the PI task is the yearly task of Recognizing Textual Entailment (RTE)
	\citep{rte}, which has also produced various datasets and multiple practical systems. The 
	meta-phenomena of paraphrasing and textual entailment are very similar and are often studied together 
	at least from a theoretical point of view. \citet{androutsopoulos2010survey} present a summary of the tasks 
	related to both paraphrasing and textual entailment.

	The idea of decomposing paraphrasing into simpler and easier to define phenomena has been growing 
	in popularity in the last few years. \citet{Bhagat} and later \citet{BhagatHovy} propose a simplified 
	framework that identifies several possible phenomena involved in the paraphrasing relation. \citet{Vila}
	propose a more complex, hierarchically structured typology that studies the different phenomena at the 
	corresponding linguistic levels (lexical, morphological, syntactic, and discourse). More recently, 
	\citet{Benikova} approach the problem by focusing on the paraphrasing at the level of events, understood
	as predicate-argument structure.

	A similar decomposition tendency is noticed in the field of Textual Entailment. \citet{Garoufi}, \citet{Sammons},
	and \citet{CabrioMagnini} propose different frameworks for decomposing the textual ``inference'' into 
	simple, atomic phenomena. It is important to note that the similarity and the relation between paraphrasing
	and textual entailment is even stronger in the context of the decomposition framework and the resulting 
	typologies. The two most exhaustive typologies: \citet{Vila} for paraphrasing and 
	\citet{CabrioMagnini} for textual entailment share the majority of their atomic phenomena as well as the 
	overall structure and organization of the typology. 

	One of the advantages of the decomposition approaches is that naturally they work towards bridging the gap 
	between the research at different granularity levels. A corpora annotated with semantic relations at both the
	 textual and the atomic (morphological, lexical, syntactic, discourse) levels can be a valuable resource for 
	studying the relation between them. In this same line of work, \citet{shwartz} emphasize the importance 
	of studying lexical entailment ``in context'' and the lack of resources that can enable such work. The corpora 
	annotated with atomic paraphrase and atomic entailment phenomena can be used for that purpose without 
	adaptation or additional annotation.

	The application of paraphrase typology for the creation of resources and in practical tasks is still very limited. 
	Most of the authors annotate a very small subsamples of around 100 text pairs to illustrate the proposed
	typology. The largest available corpus annotated with paraphrase types to date is the one of \citet{Vila2015}. 
	\citet{Barron} use this corpus to demonstrate some possible uses of the decomposition approach to 
	paraphrasing.

\section{Extended Paraphrase Typology} \label{4:et}

	We propose the Extended Paraphrase Typology (EPT), which was created to address several of the practical limitations 
	of the existing typologies and to provide better resources to the NLP community. EPT ha better coverage than
	previous typologies, including the annotation of non-paraphrases. This allows for a more in-depth understanding of the meta-phenomena and of the relation 
	between \textit{``textual paraphrases''} and \textit{``atomic paraphrases''}.

	\subsection{Basic Terminology}\label{4:et:def}

		In order to discuss the issues and limitations of existing paraphrase typologies, we first define 
		\textit{``paraphrasing''}, \textit{``textual paraphrase''}, and \textit{``atomic paraphrase''}. 

		We understand \textit{``paraphrasing''} to be a specific semantic relation between two texts of arbitrary length. 
		The two texts that are connected by a paraphrase relation have approximately the same meaning. We call them 
		\textit{``textual paraphrases''}. There is no limitation for \textit{``textual paraphrases''} in terms of 
		the nature 	of the linguistic phenomena involved. The concept of \textit{``textual paraphrases''} is a practical simplification of 
		a complex 	linguistic phenomenon, which is adopted in most paraphrase-related tasks, datasets, and applications. 
		The original annotation of the MRPC and the PPDB corpora is built around the notion of textual paraphrases.
		Another term that we use in the article is \textit{``textual non-paraphrases''}. With this term we refer to
		pairs of texts (of arbitrary length), which are not connected by a paraphrase relation. 

		\textit{``Atomic paraphrases''} are paraphrases of a particular type. They must satisfy specific (linguistic) conditions, 
		defined in the paraphrase typology. \textit{``Atomic paraphrases''} are identified by the linguistic phenomenon 
		which is responsible for the preservation of the meaning between the two texts. \textit{``Atomic paraphrases''} 
		have 	a (linguistically defined) scope, such as a word, a phrase, an event, or a discourse structure. The most 
		complete typologies to date organize \textit{``atomic paraphrases''} hierarchically, in terms of the linguistic level 
		of the involved phenomenon. Unlike \textit{``textual paraphrases''}, \textit{``atomic paraphrases''} cannot 
		be of arbitrary length. Their length is defined and restricted by their scope.

	\subsection{From Atomic to Textual Paraphrases}\label{4:et:comp}

		The relation between textual and atomic paraphrases is not easy to define and explore. It poses many challenges
		to the researchers, annotators, and developers of practical systems. In this section, we illustrate several
		issues that we want to address with the creation of the EPT and the ETPC.

		The first issue to be addressed is that multiple atomic paraphrases can appear in a single textual paraphrase pair.
		The two texts in 1a and 1b are textual paraphrases\footnote{All examples in this subsection are from the MRPC corpus.
		When we say that the texts are textual paraphrases or textual non-paraphrases, we refer to the labels corresponding
		to these pairs in MRPC.}. However, they include more than one atomic paraphrase': \textit{``magistrate''} and 
		\textit{``judge''} are an instance of \textit{``same polarity substitution''}, while \textit{``A federal magistrate ... 	ordered''} 
		and \textit{``Zuccarini was ordered by a federal judge...''} are an instance of \textit{``diathesis alternation''}\footnote{These 
		types and annotation are from \citet{Vila2015}.}.

		\begin{itemize}
		\item[1a]\underline{A federal \textbf{magistrate} in Fort Lauderdale ordered} him held without bail.	
		\item[1b]\underline{Zuccarini was ordered} held without bail Wednesday \underline{by a federal \textbf{judge} in} \underline{Fort Lauderdale, Fla}.
		\end{itemize}

		Second issue is that atomic paraphrases can appear in textual pairs that are not paraphrases.
		The two texts in 2a and 2b as a whole are not textual paraphrases, even if they have a high degree of lexical overlap 
		and a similar syntactic structure. However, \textit{``Microsoft''} and \textit{``shares of Microsoft''} are an instance of \textit{``same 
		polarity substitution''} - both phrases have the same role and meaning in the context of the two sentences. This demonstrates
		the possibility of atomic paraphrases being present in textual non-paraphrases. \footnote{In fact, it is possible to find atomic
		paraphrases within pairs of texts connected by various relations, such as entailment, simplification, summarization,
		contradiction, and question-answering, among others. This is illustrated by the significant overlap of atomic types in 
		Paraphrase Typology research and typology research in Textual Entailment.}

		\begin{itemize}
		\item[2a]\underline{Microsoft} fell 5 percent before the open to \$27.45 from Thursday's close of \$28.91.
		\item[2b]\underline{Shares in Microsoft} slipped 4.7 percent in after-hours trade to \$27.54 from a Nasdaq close of \$28.91.
		\end{itemize}

		Third issue is that in certain cases, the semantic relation between the elements in an atomic paraphrase  
		can only be interpreted within the context  (as shown in the work of \citet{shwartz}). 
		The two texts in 3a and 3b are textual paraphrases. The out-of-context meaning of \textit{``cargo''} and 
		\textit{``explosives''} differs significantly, however within the given context, they are an instance of \textit{``same polarity 
		substitution''}.

		\begin{itemize}
		\item[3a]They had published an advertisement on the Internet on June 10, offering the \underline{cargo} for sale, he added.
		\item[3b]On June 10, the ship's owners had published an advertisement on the Internet, offering the \underline{explosives} for sale.
		\end{itemize}

		And finally, 4a and 4b illustrate an issue that is often overlooked in theoretical paraphrase research: 
		the linguistic phenomena behind certain atomic paraphrases do not always preserve the meaning. The 
		meanings of \textit{``beat''} and \textit{``battled''} are similar, and play the same syntactic and discourse
		role in the structure of the texts. Therefore, the substitution of \textit{``beat''} for \textit{``battled''}
		fulfills the formal requirements of a \textit{``same polarity substitution''}. However, after this substitution, the resulting
		texts are not paraphrases as they differ substantially in meaning.

		\begin{itemize}
		\item[4a]He \underline{beat} testicular cancer that had spread to his lungs and brain.	
		\item[4b]Armstrong, 31, \underline{battled} testicular cancer that spread to his brain.
		\end{itemize}

	\subsection{Objectives of EPT and Research Questions.}\label{4:et:obj}

		We argue that the objectives behind a paraphrase typology are twofold:
		1) to classify and describe the linguistic phenomena involved in paraphrasing (at the atomic level); and
		2) to provide the means to study the function of atomic paraphrases within pairs of texts of arbitrary size and 
		with various semantic relations (such as, textual paraphrases, textual entailment pairs, contradictions, and 
		unrelated texts). 

		Traditionally, the authors of paraphrase typologies have focused on the first objective while the latter is mentioned 
		only briefly or ignored altogether. In our work, we want to extend the existing work on paraphrase typology in the
		direction of Objective 2, as we argue that it is crucial for applications. We pose four research questions, that
		we aim to address with the creation of EPT and ETPC:

		\begin{itemize}
		\item[\textbf{RQ1}] what is the relation between atomic and textual paraphrases considering the distribution of atomic 
		paraphrases in textual paraphrases?
		\item[\textbf{RQ2}] what is the relation between atomic paraphrases and textual non-paraphrases considering the 
		distribution of atomic paraphrases in textual non-paraphrases?
		\item[\textbf{RQ3}] what is the role of the context in atomic paraphrases?
		\item[\textbf{RQ4}] in which cases do the linguistic phenomena behind an atomic paraphrase preserve the meaning and 
		in which they do not?
		\end{itemize}

	\subsection{The Extended Paraphrase Typology} \label{4:et:type}

		The full Extended Paraphrase Typology is shown in Table \ref{4:et:tab}. It is organized in seven meta categories:
		``Morphology'', ``Lexicon'', ``Lexico-syntax'', ``Syntax'', ``Discourse'', ``Other'', and ``Extremes''. Sense Preserving (Sens Pres.) shows 

		\begin{table} [H]
		\begin{center}

		\begin{tabular}{| l | l | c |}
		\hline
		\textbf{ID} & \textbf{Type} & \textbf{\begin{tabular}[x]{@{}c@{}}Sense\\Pres.\end{tabular}} \\
		\hline \hline
		\multicolumn{3}{|c|}{Morphology-based changes} \\
		\hline \hline
		1 & Inflectional changes & + / - \\
		\hline
		2 & Modal verb changes & + \\
		\hline
		3 & Derivational changes & + \\
		\hline \hline
		\multicolumn{3}{|c|}{Lexicon-based changes} \\
		\hline \hline
		4 & Spelling changes & + \\
		\hline
		5 & Same polarity substitution (habitual) & + \\
		\hline
		6 & Same polarity substitution (contextual) & + / - \\
		\hline
		7 & Same polarity sub. (named entity) & + / - \\
		\hline
		8 & Change of format & + \\
		\hline \hline
		\multicolumn{3}{|c|}{Lexico-syntactic based changes} \\
		\hline \hline
		9 & Opposite polarity sub. (habitual) & + / - \\
		\hline
		10 & Opposite polarity sub. (contextual) & + / - \\
		\hline
		11 & Synthetic/analytic substitution & + \\
		\hline
		12 & Converse substitution & + / - \\
		\hline \hline
		\multicolumn{3}{|c|}{Syntax-based changes} \\
		\hline \hline
		13 & Diathesis alternation & + / - \\
		\hline
		14 & Negation switching & + / - \\
		\hline
		15 & Ellipsis & + \\
		\hline
		16 & Coordination changes & + \\
		\hline
		17 & Subordination and nesting changes & + \\
		\hline \hline
		\multicolumn{3}{|c|}{Discourse-based changes} \\
		\hline \hline
		18 & Punctuation changes & + \\
		\hline
		19 & Direct/indirect style alternations & + / - \\
		\hline
		20 & Sentence modality changes & + \\
		\hline
		21 & Syntax/discourse structure changes & + \\
		\hline \hline
		\multicolumn{3}{|c|}{Other changes} \\
		\hline \hline
		22 & Addition/Deletion & + / - \\
		\hline
		23 & Change of order & + \\
		\hline
		24 & Semantic (General Inferences) & + / - \\
		\hline \hline
		\multicolumn{3}{|c|}{Extremes} \\
		\hline \hline
		25 & Identity & + \\
		\hline
		26 & Non-Paraphrase & - \\
		\hline
		27 & Entailment & - \\
		\hline
		
		\end{tabular}

		\end{center}
		\caption{Extended Paraphrase Typology} \label{4:et:tab}
		\end{table}

		whether a certain type can give raise to textual paraphrases (+), to textual non-paraphrases (-), or to both (+ 
		/ -)\footnote{A more detailed table of EPT, with additional examples for each atomic type is available at \url{https://github.com/venelink/ETPC} and in Appendix \ref{a_etpc} of the thesis.}.
		The typology contains 25 atomic paraphrase types (+) and 13 atomic non-paraphrase types (-). It is 	based on 
		the work of \citet{Vila} and aims to extend it in two directions in order to address the four 
		Research Questions.

		First, we have added three new atomic paraphrase types - we split the atomic types \textit{``same polarity substitution''} 
		and \textit{``opposite polarity substitution''} into two separate types based on the nature of the relation between the 
		substituted words: \textit{``habitual''} and \textit{``contextual''}. We have also added the type \textit{``same polarity 
		substitution (named entity)''}. While the principle behind all substitutions is the same, in practice there is a significant 
		difference whether the replaced words are connected in their habitual meaning, contextually, or refer to related named 
		entities in the world. Instances of the new types can be seen in sentence
		pairs 5 (\textit{``same polarity substitution (habitual)''}), 6 (\textit{``same polarity substitution (contextual)''}),
		7 (\textit{``same polarity substitution (named entity''}), 8 (\textit{``opposite polarity substitution (habitual)''}), and
		9 (\textit{``opposite polarity substitution (contextual)''})

		\begin{itemize}
		\item[5a]A federal \underline{magistrate} in Fort Lauderdale ordered him held without bail.
		\item[5b]Zuccarini was ordered held without bail Wednesday by a federal \underline{judge} in Fort Lauderdale, Fla.
		\item[6a]Meanwhile, the global death toll \underline{approached} 770 with more than 8,300 people sickened since the severe acute respiratory syndrome virus first appeared in southern China in November.
		\item[6b]The global death toll from SARS \underline{was} at least 767, with more than 8,300 people sickened since the virus first appeared in southern China in November.
		\item[7a]He told The Sun newspaper that \underline{Mr. Hussein}'s daughters had British schools and hospitals in mind when they decided to ask for asylum.
		\item[7b]``\underline{Saddam}'s daughters had British schools and hospitals in mind when they decided to ask for asylum -- especially the schools,'' he told The Sun.
		\item[8a]Leicester \underline{failed} in both enterprises.
		\item[8b]He \underline{did not succeed} in either case.
		\item[9a]A big surge in consumer confidence has \underline{provided} the only positive economic news in recent weeks.
		\item[9b]Only a big surge in consumer confidence has \underline{interrupted} the bleak economic news.
		\end{itemize}

		Second, we have introduced the \textit{``sense preserving''} feature in 13 of the atomic types. As we have shown in 
		the previous section (examples 4a and 4b), the same atomic linguistic transformation (such as substitution, diathesis 
		alternation, and negation switching)
		can give raise to different semantic relations at textual level: paraphrasing, entailment, and contradiction, among 
		others. This idea has already been expressed by \citet{CabrioMagnini} in the field of Recognizing Textual 
		Entailment. Building on this idea, we identify 13 atomic types that can, in different instances, give rise to both
		paraphrases and non-paraphrases. Sentence pairs 10 and 11 show an example of sense preserving and
		non-sense preserving \textit{''Inflection change''} types. In 10a and 10b, both \textit{``streets''} and \textit{``street''}
		are a generalization with the meaning \textit{``all streets''}. In a similar way, in 11b, \textit{``boats''} has
		the meaning as \textit{``all boats''}. However in 11a, \textit{``boat''} can have the meaning \textit{``one particular boat''},
		thus the inflectional change \textit{``boat - boats''} is not sense-preserving.

		\begin{itemize}
		\item[10a]It was with difficulty that the course of \underline{streets} could be followed.
		\item[10b]You couldn't even follow the path of the \underline{street}.
		\item[11a]You can't travel from Barcelona to Mallorca with the \underline{boat}.
		\item[11b]\underline{Boats} can't travel from Barcelona to Mallorca.
		\end{itemize}

		The changes introduced in EPT allow us to work on all four Research Questions (RQs) defined in Section \ref{4:et:obj}
		This is a clear advantage over the existing paraphrase typologies, which are only suitable for addressing \textbf{RQ1}. For \textbf{RQ1}, 
		we annotated all atomic types in the positive (``paraphrases'') portion of MRPC and measured their distribution. For
		\textbf{RQ2}, we annotated all atomic types in the negative (``non-paraphrases'') portion of MRPC and compared
		the distribution of the types in the positive and negative portions. For \textbf{RQ3}, the two newly added ``contextual''
		types allow us to distinguish and compare context dependent from context independent atomic paraphrases.
		Finally, for \textbf{RQ4}, the addition of ``sense preserving'' allows us to annotate, isolate and compare the sense
		preserving and non-sense preserving instances of the same linguistic phenomena.

\section{Annotation Scheme and Guidelines}\label{4:ann}

	We propose the Extended Paraphrase Typology (EPT) with a clear practical objective in mind:  to create 
	language resources that improve the performance, evaluation, and understanding of the 
	systems competing on the task of PI and to open new research directions. We used the EPT to annotate
	the MRPC corpus with atomic paraphrases. We annotated all 5801 text pairs in the corpus, including both 
	the pairs annotated as paraphrases (3900 pairs) and those annotated as non-paraphrases (1901 pairs).

	As a basis, we used the MRPC-A corpus by \citet{Vila2015}, which already contains some 
	annotated atomic paraphrases. Our annotation consisted of three steps, corresponding to the three 
	different layers of annotation.

	First, we annotated the non-sense preserving atomic phenomena (Section \ref{4:annp}) in the textual 
	non-paraphrases. Second, we annotated the sense preserving atomic paraphrase phenomena (Section 
	\ref{4:anpp}) in both textual paraphrases and textual non-paraphrases. And third, we identified all sentences 
	in the corpus containing negation, and explicitly annotated the negation scope (Section \ref{4:anneg}).

	For the purpose of the annotation, we created a web-based annotation tool, Pair-Anno, capable of
	annotating aligned pairs of discontinuous scopes in two different texts\footnote{Screenshots of Pair-Anno
	can be seen at \url{https://github.com/venelink/ETPC}.}.  As the scope of each atomic phenomena is one or 
	more sets of tokens, prior to the annotation we automatically tokenized the corpus using NLTK \citep{nltk}.

	\subsection{Non-Sense Preserving Atomic Phenomena}\label{4:annp}

		Textual non-paraphrases in the MRPC corpus typically have a very high degree of lexical overlap and 
		a similar syntactic and discourse structure. Normally, they differ only by a few elements (morphological,
		lexical, or structural), but the modification of these few elements leads to a substantial difference in the 
		meaning of the two texts as a whole. The annotation of non-sense preserving phenomena aims to identify
		these key elements and study the linguistic nature of the modification.

		When annotating atomic phenomena, our experts identified and annotated the type, the scope, and in some
		paraphrase types, the key element. Both the scope and the key are kept as a 0-indexed list of tokens. Examples 12a and
		12b show a textual pair, annotated as non-paraphrase in the MRPC corpus. Table \ref{4:negap} 	shows the 
		annotation of non-sense preserving atomic phenomena in 12a and 12b. The key differences are 
		\textit{``opposite polarity substitution (habitual)''} (type id 10) of ``slip'' with ``rise'', and the \textit{``same polarity 
		substitution (named entity)''} (type id 7) of ``Friday'' with ``Thursday''.

		\begin{itemize}
		\item[12a]The loonie , meanwhile , continued to slip in early trading Friday .
		\item[12b]The loonie , meanwhile , was on the rise again early Thursday .
		\end{itemize}

		\begin{table}[h!]
			\begin{center}
			\begin{tabular}{|c|c|c|c|c|c|}
			\hline
			type & pair & s1 scope & s2 scope & s1 text & s2 text \\
			\hline\hline
			7 & 146 & 11 & 11 & Friday & Thursday \\
			\hline
	 		10 & 146 & 7 & 8 & slip & rise \\
			\hline
			\end{tabular}
			\end{center}
			\caption{Non-sense preserving phenomena}
			\label{4:negap}
		\end{table}

		The annotation of 12a and 12b illustrates one of the issues when annotating non-sense preserving
		phenomena. In many textual pairs, there is more than one ``key'' difference. In those cases, all of the
		phenomena were annotated separately. Nevertheless, the annotators were instructed to be
		conservative and only annotate phenomena that carry substantial differences in the meaning of the two
		texts. Determining which differences are substantial, and which are not was the main challenge for the 
		annotators. Due to the difficulty of the task, we selected annotators that were expert linguists with 
		a high proficiency of English\footnote{The full annotation guidelines for both sense preserving and non-sense
		preserving phenomena can be found at \url{https://github.com/venelink/ETPC} and in Appendix \ref{a_etpc} 
		of the thesis.}.

		When the two texts were substantially different and it was not possible to identify the atomic phenomena
		responsible for the difference, the pair was annotated with atomic type \textit{``non-paraphrase''}
		(examples 13a and 13b) or \textit{``entailment''} (examples 14a and 14b).

		\begin{itemize}
		\item[13a]That compared with \$35.18 million, or 24 cents per share, in the year-ago period.
		\item[13b]Earnings were affected by a non-recurring \$8 million tax benefit in the year-ago period.
		\item[14a]The year-ago comparisons were restated to include Compaq results.
		\item[14b]The year-ago numbers do not include figures from Compaq Computer.
		\end{itemize}

	\subsection{Sense Preserving Atomic Phenomena}\label{4:anpp}

		For the annotation of the sense preserving atomic phenomena, we used the same  annotation scheme
		format as the one for the non-sense preserving phenomena. Each phenomenon is identified by a type, a
		scope, and, where applicable, a key. 15a and 15b show a textual pair, annotated as a paraphrase in the MRPC. 
		An example of a single annotated atomic phenomenon can be seen in Table \ref{4:posap}
		 
		\begin{itemize}
		\item[15a]Amrozi accused his brother , whom he called `` the witness '' , of deliberately distorting his evidence .
		\item[15b]Referring to him as only `` the witness '' , Amrozi accused his brother of deliberately distorting his evidence .
		\end{itemize}

		\begin{table}[h!]
			\begin{center}
			\begin{tabular}{|c|c|c|c|c|c|}
			\hline
			type & pair & s1 scope & s2 scope & s1 text & s2 text \\
			\hline\hline
			 6 &  1 & 5 & 1, 2 & whom & to him \\
			\hline
			\end{tabular}
			\end{center}
			\caption{Sense preserving phenomenon}
			\label{4:posap}
		\end{table}

		For the 3900 text pairs already annotated by \citet{Vila2015}, we worked with the existing corpus
		and we only re-annotated the 3 new sense preserving paraphrase types introduced in EPT. For the 1901 
		textual non-paraphrases, which were not annotated in MRPC-A, we performed a full annotation with all
		25 sense preserving atomic types.

	\subsection{Inter-Annotator Agreement}\label{4:iaa}

		In this section, we present the measures for calculating the inter-annotator agreement and the agreement
		score on the first two layers of annotation: non-sense preserving atomic phenomena and sense preserving 
		atomic phenomena.

		The measure that we use is the IAPTA TPO, introduced by \citet{Vila2015}. It is a  fine-grained 
		measure, created specifically for the task of annotating paraphrase types. It takes into account the 
		agreement with respect to both the label and the scope of the phenomena. It is a pairwise agreement 
		measure, obtained by calculating the Precision, Recall and F1 of one of the annotators, while using another 
		annotator as a gold standard. There are two versions of the measure - TPO-partial, which requires that the
		annotators select the same label and that the scopes overlap by at least one token; and TPO-total which
		requires full overlap of label and scope.

		The classical TPO measures are pairwise, they calculate the agreement between two annotators. When
		the annotation process involves more than two annotators, we first calculate the pairwise TPO measure
		between any two annotators and then we use one of three different techniques for calculating the overall
		agreement for the corpus. TPO (avg) is the most simple score, as it is the average of all pairwise TPO 
		scores. TPO (union) is the union of all pairwise TPO agreement tables. That is, any phenomena that is
		annotated with the same label and the same scope by any 2 annotators is part of the TPO (union). Finally,
		TPO (gold) is the average F1 score of the three annotators, when treating TPO (union) as a gold standard.
		TPO (union) and TPO (gold) are two new measures, that we propose as part of this paper. TPO (union)
		represents all the ``high quality'' phenomena (that is, phenomena annotated the same way by multiple
		annotators). TPO (gold) represents the probability that any of our annotators would annotate ``high
		quality'' phenomena.		

		The annotation of the sense preserving atomic paraphrases was carried out by two expert annotators, 
		while the annotation of the non-sense preserving atomic phenomena was carried out by three expert 
		annotators. For the purpose of calculating the inter-annotator agreement, all experts were given the
		same 180 text pairs (roughly 10 \% of all non-paraphrase pairs in the corpus). The pairs were split in
		3 equal parts and given to the annotators in three different stages of the annotation: one at the 
		beginning, one in the middle, and one at the end of the annotation process. Table \ref{4:npagr} shows the 
		obtained scores, where  ETPC (-) stands for the non-sense preserving layer, ETPC (+) stands for 
		the sense-preserving layer of annotation and MRPC-A is the annotation of \citet{Vila2015}. 
		For ETPC (+) we only had two annotators, so we were not able to 	calculate TPO (union) and TPO (gold). 
		Since these measures have been introduced by us in the current paper, the MRPC-A corpus by 
		\citet{Vila2015} does not have values for them either.

		\begin{table}[h!]
			\begin{center}
			\begin{tabular}{|c|c|c|c|}
			\hline
			\textbf{Measure} & \textbf{ETPC (-)} & \textbf{ETPC (+)} &  \textbf{MRPC-A} \\
			\hline \hline
			TPO-partial (avg) & 0.72 & 0.86 & 0.78 \\
			\hline
			TPO-total (avg) & 0.68 & 0.68 & 0.51 \\
			\hline
			TPO (union) & 0.77 & n-a & n-a \\
			\hline
			TPO (gold) & 0.86 & n-a & n-a \\
			\hline

			\end{tabular}
			\end{center}
			\caption{Inter-annotator Agreement}
			\label{4:npagr}
		\end{table}		

		ETPC (+) and MRPC-A are directly comparable as they measure the agreement on the same task 
		(annotation of sense-preserving atomic phenomena). The results show much higher agreement 
		score with respect to both TPO-partial (0.86 against 0.78) and TPO-total (0.68 against 0.51). 
		ETPC (-) measures the agreement on a different task (annotation of non-sense preserving 
		phenomena). The TPO-partial score of ETPC (-) is lower than both ETPC (+) and MRPC-A (0.72 
		against 0.86 and 0.78 respectively), however the TPO-total score is equal to that of ETPC (+) 
		and much higher than that of MRPC-A. It is interesting to note that there is almost no difference 
		between TPO-partial and TPO-total for ETPC (-) (0.72 against 0.68), while for ETPC (+) and MRPC-A, 
		the difference is significant.
		The TPO (union) for ETPC (-) shows that 77\% of all phenomena are annotated the same way by 
		at least 2 of the annotators. The TPO (gold) indicates that the probability of any of our experts 
		annotating a ``gold'' example is 86\%. Considering the difficulty of the task, the obtained results 
		indicate the high quality of the annotated corpus.

	\subsection{Annotation of Negation}\label{4:anneg}

		During the first two steps of the annotation, we identified all sentences that contain negation. For 
		every instance of negation we annotated the negation cues and the scope of negation. 16a and 16b
		illustrate an example of annotated negation.

		\begin{itemize}
			\item[16a](Moore had (\textbf{no} [\underline{negation marker}]) immediate comment Tuesday [\underline{scope}])
			\item[16b](Moore (\textbf{did not} [\underline{negation marker}]) have an immediate response Tuesday [\underline{scope}])
		\end{itemize}

\section{The ETPC Corpus}\label{4:corp}

	This section presents the results of the annotation of the ETPC corpus. Section \ref{4:corp:np} shows
	the results of annotating non-sense preserving phenomena. Section \ref{4:corp:pp} shows the results
	of annotating sense preserving phenomena. Section \ref{4:corp:disc} discusses the results and the 
	Research Questions, and Section \ref{4:corp:app} lists some applications of ETPC.

	\subsection{Non-Sense Preserving Atomic Phenomena} \label{4:corp:np}

	Table \ref{4:corp:npn} shows the distribution of the non-sense preserving phenomena.
	Type Relative Frequency (Type RF) shows the relative distribution of the atomic types. Occurrence 
	Frequency (Type OF) shows  the distribution of phenomena per sentence, that is in how many textual 
	pairs each phenomenon can be found\footnote{The sum of all Occurrence Frequencies exceeds 100, 
	as one sentence often contains more than one atomic phenomenon.}. The total number of non-sense 
	preserving phenomena is 3406 in 1901 text pairs.

	\begin{table}[h!]
		\begin{center}
		\begin{tabular}{|c|c|c|}
		\hline
		\textbf{Type} & \textbf{Type RF} & \textbf{Type OF} \\
		\hline \hline
		Inflectional & 0.02\% & 0.04\% \\
		\hline
		Same Polarity (con) & 9.3\% & 15.5\% \\
		\hline
		Same Polarity (ne) & 27.5\% & 22.5\% \\
		\hline
		Opp Polarity (hab) & 2.7\% & 4.4\% \\
		\hline
		Opp Polarity (con) & 0.01\% & 0.02\% \\
		\hline
		Converse & 0.01\% & 0.02\% \\	
		\hline
		Diathesis & 0.01\% & 0.01\% \\
		\hline
		Negation & 0.02\% & 0.03\%\\
		\hline
		Direct/Indirect & 0\% & 0\% \\
		\hline
		Addition/Deletion & 52\% & 65.5\% \\
		\hline
		Semantic based & 0\% & 0\% \\
		\hline
		Non-paraphrase & 7.6\% & 13.7\%\\
		\hline
		Entailment & 0.02\% & 0.04\%\\
		\hline

		\end{tabular}
		\end{center}
		\caption{Distribution of non-sense preserving phenomena}
		\label{4:corp:npn}
	\end{table}	

	Both Type Relative Frequency (RF) and Occurrence Frequency (OF) indicate that the non-paraphrase portion 
	of the corpus is not well balanced with respect to atomic phenomena. In 260 of the text pairs (13.7\%), the 
	annotators selected \textit{``non-paraphrase''} indicating that the two texts were substantially different. In the rest 
	of the pairs, the most common reason for the ``non-paraphrase'' label at textual level was \textit{``Addition/Deletion''} (52\% RF,
	 65.5\% OF), followed by \textit{``Same polarity substitution (named entity)''} (27\% RF, 22.5\% OF), \textit{``Same 
	polarity substitution (contextual)''} (RF 9,3\%, OF 15.5\%), and \textit{``Opposite polarity substitution (habitual)''} 
	(RF 2.8\%, OF 4.6\%). These are the only types with Type Relative Frequency and Occurrence Frequency 
	above 1\%, and they constitute over 99\% of all non-sense preserving atomic phenomena annotated in 
	the corpus. Six of the atomic phenomena are represented only with a few examples, while two are not 
	represented at all. 

	\subsection{Sense Preserving Atomic Phenomena} \label{4:corp:pp}

	Table \ref{4:corp:npp} shows the distribution of sense preserving atomic phenomena in the textual
	paraphrase and non-paraphrase portions of the corpus\footnote{At the time of the submission of this paper, the
	annotation of the non-paraphrase portion was not finished. The reported results are for 500 annotated pairs 
	(about 30\% of the corpus). The full figures will be made available at \url{https://github.com/venelink/ETPC}}. For the
	textual paraphrase portion, we used the numbers reported by \citet{Vila2015} with partial
	re-annotation to account for the new types in ETPC. For \textit{``same polarity substitution''}, 35\%
	of the phenomena were re-annotated as \textit{``habitual''}, 47\% as \textit{``contextual''}, and
	18\% as \textit{``named entity''}.
	For \textit{``opposite polarity substitution''}  21\% of the phenomena were \textit{``contextual''} and  
	79\% of the phenomena were \textit{``habitual''}.

	\begin{table}[h!]
		\begin{center}
		\begin{tabular}{|c|c|c|}
		\hline
		\textbf{Type} & \begin{tabular}[x]{@{}c@{}}\textbf{Non} \\ \textbf{Paraphrase}\end{tabular} & \textbf{Paraphrase} \\
		\hline \hline
		Inflectional & 2.13\% & 2.78\% \\
		\hline
		Modal verb & 0.59\% & 0.83\% \\
		\hline
		Derivational & 0.35\% & 0.85\% \\
		\hline
		Spelling changes & 1.30\% & 2.91\% \\
		\hline
		Same Polarity (hab) & 10.55\% &  8.68\% \\
		\hline
		Same Polarity (con) & 11.15\% & 11.66\% \\
		\hline
		Same Polarity (ne) & 7.11\% & 5.08\% \\
		\hline
		Format & 1.06\% & 1.1\% \\
		\hline
		Opp Polarity (hab) & 0\% & 0.07\% \\
		\hline
		Opp Polarity (con) & 0\% & 0.02\% \\
		\hline
		Synthetic/analytic & 7.82\% & 3.80\% \\
		\hline
		Converse & 0.12\% & 0.20\% \\
		\hline
		Diathesis & 0.83\% & 0.73\% \\
		\hline
		Negation & 0\% & 0.09\% \\
		\hline
		Ellipsis & 0.47\% & 0.30\% \\
		\hline
		Coordination & 0.24\% & 0.22\% \\
		\hline
		Subord. and nesting & 1.18\% & 2.14\% \\
		\hline
		Punctuation & 2.72\% & 3.77\% \\
		\hline
		Direct/Indirect & 0.24\% & 0.30\% \\
		\hline
		Sentence modality & 0\% & 0\% \\
		\hline
		Synt./Disc. structure & 1.30\% & 1.39\% \\
		\hline
		Addition/Deletion & 20.04\% & 25.94\% \\
		\hline
		Change of order & 3.08\% & 3.89\% \\
		\hline
		Semantic & 0\% & 1.53\% \\
		\hline
		Identity & 25.02\% & 17.54\% \\
		\hline
		Non-Paraphrase & 2.49\% & 3.81\% \\
		\hline
		Entailment & 0.12\% & 0.37\% \\
		\hline
		\end{tabular}
		\end{center}
		\caption{Distribution of Sense preserving phenomena in textual paraphrases and textual non-paraphrases}
		\label{4:corp:npp}
	\end{table}	

	The results show that the distribution of sense-preserving phenomena is relatively consistent between the two
	portions of the corpus. The most notable differences between the two distributions are the frequencies of
	\textit{``same polarity substitution (named entity)''}, \textit{``synthetic/analytic''}, \textit{``addition/deletion''}, 
	and \textit{``identity''}. Both distributions are not well balanced in terms of atomic types, with 8 types 
	(\textit{``addition/deletion''}, \textit{``identity''}, \textit{``same polarity substitution (contextual)''}, 
	\textit{``same polarity substitution (habitual)''}, \textit{``synthetic/analytic''}, \textit{``same polarity substitution 
	(named entity)''}, \textit{``change of order''}, and \textit{``punctuation''}) responsible for over 80\% of the
	phenomena. 

	\subsection{Discussion}\label{4:corp:disc}

	In this section we briefly discuss the annotation results and the Research Questions that we posed in Section 
	\ref{4:et:obj}

	With respect to \textbf{RQ1} and \textbf{RQ2}, we measured the raw frequency distribution of the sense preserving atomic phenomena in both 
	the paraphrase and non-paraphrase portions of the corpus. We make two important observations from the 
	data. First, the corpus is not well balanced in terms of type distribution in either of the portions. It can be seen
	in Table \ref{4:corp:npp} that 8 of the types 
	are overrepresented while the rest are underrepresented. This imbalance is even more significant
	in terms of meta-categories. The structure meta-types ``syntax'' and ``discourse'' account for less than 10 \% of all
	types. Second, the raw frequency distribution of atomic phenomena in textual paraphrases and textual 
	non-paraphrases is very similar. This finding suggests that it is the non-sense preserving phenomena that 
	are mostly responsible for the relation at textual level in this corpus. This makes the annotation of the 
	non-sense preserving phenomena even more important for the PI task.
	
	With respect to \textbf{RQ3}, we annotated the \textit{``same polarity substitution (contextual)''} and \textit{``opposite 
	polarity substitution (contextual) ''} types in all portions of the corpus. For \textit{``same polarity substitution''},
	over 40\% of the sense-preserving and over 25\% of the non-sense preserving instances were contextual. For
	\textit{``opposite polarity substitution''}, 21\% of the sense-preserving instances were annotated as contextual, while in
	the non-sense preserving portion we found almost no contextual instances.  

	With respect to \textbf{RQ4}, we measured the raw frequency distribution of the non-sense preserving phenomena. If 
	we compare it with the distribution of sense preserving phenomena, we can see that the differences are noteworthy
	and we can easily differentiate between the two distributions. Non-sense preserving phenomena are even less 
	balanced than sense preserving phenomena, with just 4 types responsible for almost all instances. The 
	structure types ``syntax'' and ``discourse'' are not represented at all, with all frequent types being either ``lexical'', 
	``lexico-syntactic'', or ``other''.
	
	Finally, it is worht mentioning that 13\% of the sentences in the textual paraphrase portion of the corpus and 12\% of the sentences in the textual 
	non-paraphrase portion contain negation. The relative distribution in the paraphrase and in the non-paraphrase 
	portion of the corpus is consistent. The negation scope for each of these sentences has been annotated in a 
	separate layer.

	\subsection{Applications of ETPC}\label{4:corp:app}

		The ETPC corpus has clear advantages over the currently available PI corpora, and the MRPC in
		particular. It is much more informative and can be used in several ways.

			First, ETPC can be used as a single PI corpus. The annotation with atomic types makes it much more
			informative for evaluation than any other existing PI corpus. PI systems are currently evaluated 
			in terms of binary Precision, Recall, F1 and Accuracy. ETPC provides the developers with much more 
			detailed information, without requiring any additional work on the developers' side. Knowing which 
			atomic types are involved in the correct and incorrect classification helps the error analysis and 
			should lead to an improvement in the these systems' performance. It also promotes reusability.

			Second, ETPC can be used to provide quantitative and qualitative analysis of the MRPC corpus, as we
			have already shown in section \ref{4:corp:disc} By having a detailed statistical analysis of the content 
			of the corpus we can identify possible biases and promote the creation of better and more balanced corpora.

			Third, ETPC can be easily split into various smaller corpora built around a certain atomic type or 
			a class of types. Each of them can be used for a new task of Atomic Paraphrase Identification. It 
			can be used to study the nature of the relation between atomic paraphrases and textual paraphrases.
			
			Finally, ETPC can be used to study the role of negation in PI, a research question that, to date, has received 
			very little attention.

\section{Conclusions and Future Work} \label{4:conc}

	In this paper we presented the ETPC corpus - the largest corpus annotated with detailed paraphrase 
	typology to date. For the annotation we used the new Extended Paraphrase Typology, a practically 
	oriented typology of atomic paraphrases. The annotation process included three expert linguists and
	covered the whole 5801 text pairs from the MRPC corpus. The full corpus is publicly available in two formats:
	SQL and XML\footnote{We have also made publicly 
	available all complementary data, such as annotation guidelines, screenshots of the interface, detailed 
	statistics, as well as the ETPC\_Neg corpus, composed only from the paraphrase and non-paraphrase 
	pairs containing negation ( \url{https://github.com/venelink/ETPC} ).}. 

	ETPC is a high quality resource for paraphrase related research and the task of PI. It provides more
	in-depth analysis of the existing corpora and promotes better understanding of the phenomena, the
	data, and the task. It also identifies several problems, such as the under-representation of structure
	based types and the over-representation of lexical based types. ETPC sets an example for the 
	development of new feature-rich corpora for paraphrasing research. It also promotes collaboration 
	between similar areas, such as PI, RTE and Semantic Similarity.

	Our work opens several lines of future research. First, the ETPC can be used to re-evaluate existing
	state-of-the-art PI systems. This detailed evaluation can lead to improvements of the existing PI systems 
	and the creation of new ones. Second, it can be used to create new corpora for paraphrase research, 
	which will be more balanced in terms of type distribution. Third, it can be used to study the nature of the
	paraphrase phenomenon and the relation between ``atomic'' and ``textual'' paraphrases. Finally, the 
	EPT and ETPC can be extended to other research areas, such as lexical and textual entailment, semantic 
	similarity, simplification, summarization, and question answering, among others.

\chapter[A Qualitative Evaluation Framework for \\Paraphrase Identification]{\centering A Qualitative Evaluation Framework for Paraphrase Identification}\label{ch:peval}

\chaptermark{Paraphrase Evaluation}

\begin{center}
Venelin Kovatchev, M. Ant{\`o}nia Mart{\'i}, Maria Salam{\'o}, and Javier Beltran

University of Barcelona

\vspace{10mm}

Published at \\ \textit{Proceedings of the \\Twelfth Recent Advances in Natural Language Processing Conference}, 2019 \\ pp.: 569-579
\end{center}

\paragraph{Abstract}
	In this paper, we present a new approach for the evaluation, error analysis, 
	and interpretation of supervised and unsupervised Paraphrase Identification 
	(PI) systems. Our evaluation framework makes use of a PI corpus annotated 
	with linguistic phenomena to provide a better understanding and interpretation 
	of the performance of various PI systems.
	Our approach allows for a qualitative evaluation and comparison of the PI 
	models using human interpretable categories. It does not require modification of
	the training objective of the systems and does not place additional burden on 
	the developers.
	We replicate several popular supervised and unsupervised PI  systems. 
	Using our evaluation framework we show that: 1) Each system performs
	differently with respect to a set of linguistic phenomena and makes qualitatively
	different kinds of errors; 2) Some linguistic phenomena are more challenging 
	than others across all 	systems.

\section{Introduction}

	In this paper we propose a new approach to evaluation, error analysis and interpretation in the 
	task of Paraphrase Identification (PI). Typically, PI is defined as comparing two texts
	of arbitrary size in order to determine whether they have approximately the same
	meaning \citep{mrpc}. 
	The two texts in 1a and 1b are considered 
	paraphrases, while the two texts at 2a and 2b are non-paraphrases.\footnote{Examples are
	from the MRPC corpus \citep{mrpc}} In 1a and 1b there is a change in the wording
	(\textit{``magistrate'' - ``judge''}) and the syntactic structure (\textit{``was ordered'' - ``ordered''})
	but the meaning of the sentences is unchanged. In 2a and 2b there are significant 
	differences in the quantities (\textit{``5\%'' - ``4.7\%''} and \textit{``\$27.45'' - ``\$27.54''}).

		\begin{itemize}
		\item[1a]A federal magistrate in Fort Lauderdale ordered him held without bail.	
		\item[1b]He was ordered held without bail Wednesday by a federal judge in Fort Lauderdale, Fla.

		\item[2a]Microsoft fell \textbf{5 percent} before the open to \textbf{\$27.45} from Thursday's close of \$28.91.
		\item[2b]Shares in Microsoft slipped \textbf{4.7 percent} in after-hours trade to \textbf{\$27.54} from a Nasdaq close of \$28.91.
		\end{itemize}

	The task of PI can be framed as a binary classification problem. The performance 
	of the different PI systems is reported using the Accuracy and F1 score measures. 
	However this form of evaluation does not facilitate the interpretation and error 
	analysis of the participating systems. Given the Deep Learning nature of most of the
	state-of-the-art systems and the complexity of the PI task, we argue that better 
	means for evaluation, interpretation, and error analysis are needed.
	We propose a new evaluation methodology to address this gap in the field.
	We demonstrate our methodology on the ETPC corpus \citep{etpc} - a recently
	published corpus, annotated with detailed linguistic phenomena involved in
	paraphrasing. 

	We replicate several popular state-of-the-art Supervised and Unsupervised PI 
	Systems and demonstrate the advantages of our evaluation methodology by 
	analyzing and comparing their performance. We show that while 
	the systems obtain similar quantitative results (Accuracy and F1), they
	perform differently with respect to a set of human interpretable linguistic categories
	and make qualitatively different kinds of errors. We also show that some of the
	categories are more challenging than others across all evaluated	systems. 



\section{Related Work}\label{5:rw}


	The systems that compete on PI range from using
	hand-crafted features and Machine Learning algorithms 
	\citep{fernando2008,Madnani,ji} to end-to-end Deep Learning models
	\citep{He2015,He2016,wang,subw,skipt,infersent}. 
	The 
	PI systems are typically divided in two groups: Supervised PI
	systems and Unsupervised PI systems. 

	``Supervised PI systems'' 
	\citep{He2015,He2016,wang,subw} are explicitly trained for the PI task on a 
	PI corpora. 
	``Unsupervised PI systems'' in the PI field is a term used for systems that use a
	general purpose sentence representations such as \citet{word2vec,glove,skipt,infersent}. 
	To predict the paraphrasing relation, they 
	can compare the sentence representations of the candidate paraphrases directly 
	(ex.: cosine of the angle), and use a PI corpus to learn a threshold. Alternatively
	they can use the representations as features in a classifier.



	The complexity of paraphrasing has been emphasized by many researchers 
	\citep{BhagatHovy,Vila,Benikova}. Similar observations have been made for Textual
	Entailment \citep{Sammons, CabrioMagnini}. \citet{gold-etal-2019-annotating} study
	the interactions between paraphrasing and entailment. 
	
	Despite the complexity of the phenomena, the popular PI corpora \citep{mrpc,ppdb,quora,twitt}
	are annotated in a binary manner. In part it is due to lack of annotation tools
	capable of fine-grained annotation of relations. WARP-Text \citep{kovatchev-etal-2018-warp}
	fills this gap in the NLP toolbox.

	The simplified corpus format poses a problem with respect to the quality of the PI task
	and the ways it can be evaluated.
	The vast majority of the state-of-the-art 
	systems in PI provide no or very little error analysis. This makes it difficult to interpret 
	the actual capabilities of a system and its applicability to other corpora and tasks.

	Some researchers have approached the problem of non-interpretability 
	by evaluating the same architecture on multiple datasets and multiple tasks. \citet{lan} 
	apply this approach to Supervised PI systems, while \citet{hanan} use it for evaluating 
	Unsupervised PI systems and general sentence representation models.

	\citet{Linzen} demonstrate how by modifying the task definition and the evaluation
	the capabilities of a Deep Learning system can be determined implicitly. The main advantage of 
	such an approach is that it only requires modification and additional annotation of the corpus. 
	It does not place any additional burden on the developers of the systems and can be applied to
	multiple systems without additional cost. 

	We follow a similar line of research and propose a new evaluation that uses ETPC \citep{etpc}:
	a PI corpus with a multi-layer annotation of various linguistic phenomena. Our methodology 
	uses the corpus annotation to provide much more feedback to the competing systems and to
	evaluate and compare them qualitatively. 


\section{Qualitative Evaluation Framework}\label{5:setup}

\subsection{The ETPC Corpus}\label{5:corpus}

	ETPC \citep{etpc} is a re-annotated version of the MRPC corpus. It contains 5,801 text 
	pairs. Each text pair in ETPC  has two separate layers of annotation. The first layer 
	contains the traditional binary label (paraphrase or non-paraphrase) of every text pair. 
	The second layer contains the annotation of 27 \textit{``atomic'' }linguistic phenomena 
	involved in paraphrasing, according to the authors of the corpus. 
	All phenomena are linguistically motivated and humanly interpretable.

		\begin{itemize}
		\item[3a]\underline{A federal \textbf{magistrate} in Fort Lauderdale ordered him} held without bail.	
		\item[3b]\underline{He was ordered} held without bail Wednesday \underline{by a federal \textbf{judge} in Fort} \\ \underline{Lauderdale, Fla}.
		\end{itemize}

	We illustrate the annotation with examples 3a and 3b. At the binary level, this pair
	is annotated as ``paraphrases''. At the ``atomic'' level, ETPC contains the annotation of
	multiple phenomena, such as the
	\textit{``same polarity substitution (habitual)''} of ``magistrate'' and ``judge'' 
	(marked \textbf{bold}) or the \textit{``diathesis alternation''} of \textit{``...ordered him 
	held''} and \textit{``he was ordered by...''} (marked \underline{underline}). 

	For the full set of phenomena, the linguistic reasoning behind them, their frequency in the 
	corpus, real examples from the pairs, and the annotation guidelines, please refer to \citet{etpc}.

	\subsection{Evaluation Methodology}\label{5:methodology}

	We use the corpus to evaluate the capabilities of the different PI systems implicitly.
	That means, the training objective of the systems remains unchanged: they are required 
	to correctly predict the value of the binary label at the first annotation layer. However, 
	when we analyze and evaluate the performance of the systems, we make use of both 
	the binary and the atomic annotation layers. Our evaluation framework is
	created to address our main research question (RQ 1): 

\begin{itemize}
	\item[\textbf{RQ 1}] Does the performance of a PI system 
	on each candidate-paraphrase pair depend on the different phenomena involved in that 
	pair?
\end{itemize}



	We evaluate the performance of the systems in
	terms of their \textit{``overall performance''} (Accuracy and F1) and 
	\textit{``phenomena performance''}. 

	\textit{``Phenomena performance''}
	is a novelty of our approach and allows for qualitative analysis and comparison.
	To calculate \textit{``phenomena performance''}, we create 27 subsets of the
	test set, one for each linguistic phenomenon. Each of the subsets consists of all text 
	pairs that contain the corresponding phenomenon\footnote{i.e. The ``diathesis alternation'' 
	subset contains all pairs that contain the  ``diathesis alternation'' phenomenon (such
	as the example pair 3a--3b). Some of the pairs can also contain multiple phenomena: the
	example pair 3a--3b contains both \textit{``same polarity substitution (habitual)''} and 
	\textit{``diathesis alternation''}. Therefore pair 3a--3b will be added both to the 
	\textit{``same polarity substitution (habitual)''} and to the \textit{``diathesis alternation''}
	phenomena subsets. Consequentially, the sum of all subsets exceeds the size
	of the test set.}. 
	Then, we use each of the 27 subsets as a test set and we calculate 
	the binary classification Accuracy (paraphrase or non-paraphrase) for each subset. 
	This score indicates how well the system performs in cases that include one specific 
	phenomenon. We compare the performance of the different phenomena 
	and also compare them with the \textit{``overall performance''}. 

	Prior to running the experiments we verified that: 1) the relative distribution of the phenomena in 
	paraphrases and in non-paraphrases is very similar; and 2) there is no significant correlation 
	(Pearson \textit{r} \textless 0.1) between the distributions of the individual phenomena. These findings show
	that the sub-tasks are non-trivial: 1) the binary labels of the pairs cannot be
	directly inferred by the presence or absence of phenomena; and 2) the different subsets of the
	test set are relatively independent and the performance on them cannot be trivially reduced to
	overlap and phenomena co-occurrence.


	The \textit{``overall performance''} and \textit{``phenomena performance''} of
	a system compose its \textit{``performance profile''}. 
	With it  we aim to address the rest of our research questions (RQs): 

	\begin{itemize}
		\item[\textbf{RQ 2}] Which are the strong and weak sides of each individual system?
		\item[\textbf{RQ 3}] Are there any significant differences between the \textit{``performance profiles''} of the systems?
		\item[\textbf{RQ 4}] Are there phenomena on which all systems perform well (or poorly)?
	\end{itemize}

\section{PI Systems}\label{5:systems}

	To demonstrate the advantages of our evaluation framework, we have replicated
	several popular Supervised and Unsupervised PI systems. 
	We have selected the systems based on three criteria: popularity, architecture,
	and performance. The systems that we chose are popular and widely used not only
	in PI, but also in other tasks. The systems use a wide variety of different ML
	architectures and/or different features. Finally, the systems obtain comparable
	quantitative results on the PI task. They have also been reported to obtain good
	results on the MRPC corpus which is the same size as ETPC. The choice of system 
	allows us to best demonstrate the limitations of the classical quantitative evaluation 
	and the advantages of the proposed qualitative evaluation.

	To ensure comparability, all systems have been trained and evaluated on the 
	same computer and the same corpus. We have used the configurations 
	recommended in the original papers where available. During the replication
	we did not do a full grid-search as we want to replicate and thereby contribute
	to generalizable research and systems. As such, the quantitative results that we obtain may 
	differ from the performance reported in the original papers, especially for the
	Supervised systems. However, the results are sufficient for the objective of 
	this paper: to demonstrate the advantages of the proposed 
	evaluation framework.

	We compare the performance of five Supervised and five Unsupervised systems 
	on the PI task, including one Supervised and one Unsupervised baseline systems.
	We also include Google BERT \citep{bert}
	for reference.

	The \textbf{Supervised PI systems} include: 

	\begin{itemize}

	\item[\textbf{[S1]}] Machine translation evaluation metrics as
	hand-crafted features in a Random Forest classifier. Similar to \citet{Madnani} \textit{(baseline)}
	\item[\textbf{[S2]}] A replication of the convolutional network 
	similarity model of \citet{He2015}
	\item[\textbf{[S3]}] A replication of the lexical composition and 
	decomposition system of \citet{wang}
	\item[\textbf{[S4]}] A replication of the pairwise word interaction modeling 
	with deep neural network system by \citet{He2016}
	\item[\textbf{[S5]}] A character level neural network model by \citet{subw}

	\end{itemize}


	The \textbf{Unsupervised PI systems} include: 
	\begin{itemize}
		\item[\textbf{[S6]}] A binary Bag-of-Word sentence representation (baseline) 
		\item[\textbf{[S7]}] Average over sentence of pre-trained Word2Vec word embeddings \citep{word2vec}
		\item[\textbf{[S8]}] Average over sentence of pre-trained Glove word embeddings \citep{glove}
		\item[\textbf{[S9]}] InferSent sentence embeddings \citep{infersent}
		\item[\textbf{[S10]}] Skip-Thought sentence embeddings \citep{skipt}
	\end{itemize}


	In the unsupervised setup we first represent each of the two sentences under 
	the corresponding model. Then we obtain a feature vector by concatenating 
	the absolute distance and the element-wise multiplication of the two representations. 
	The feature vector is then fed into a logistic regression classifier to predict the textual
	relation. This setup has been used in multiple PI papers, more recently by \citet{hanan}.
	While the vector representations of BERT are unsupervised, they are fine-tuned on
	the dataset. Therefore we put them in a separate category (System \#11).

\section{Results} \label{5:results}

	\subsection{Overall Performance}\label{5:sec:binary}

		\begin{table} [h!]
		\begin{center}

		\begin{tabular}{| l | l | c | c |} 
		\hline
		\textbf{ID} & \textbf{System Description} & \textbf{Acc} & \textbf{F1} \\
		\hline \hline
		\multicolumn{4}{|c|}{SUPERVISED SYSTEMS} \\
		\hline
		1 & MTE features (baseline) 	& .74 & .819 \\
		\hline
		2 & \citet{He2015} 			& .75 & .826 \\
		\hline
		3 & \citet{wang} 			& \textbf{.76} & \textbf{.833} \\
		\hline
		4 & \citet{He2016} 			& .76 & .827 \\
		\hline
		5 & \citet{subw} 			& .70 & .800 \\
		\hline
		\multicolumn{4}{|c|}{UNSUPERVISED SYSTEMS} \\
		\hline
		6 & Bag-of-Words (baseline) 	& .68 & .790 \\
		\hline
		7 & Word2Vec (average) 		& .70 & .805 \\
		\hline
		8 & GLOVE (average) 			& .72 & .808 \\
		\hline
		9 & InferSent 				& \textbf{.75} & \textbf{.826} \\
		\hline
		10 & Skip-Thought 			& .73 & .816 \\
		\hline \hline
		11 & Google BERT 			& .84 & .889 \\
		\hline \hline
		\end{tabular}

		\end{center}
		\caption{Overall Performance of the Evaluated Systems} \label{5:binary}
		\end{table}

	Table \ref{5:binary} shows the \textit{``overall performance''} of the systems on
	the 1725 text pairs in the test set. Looking at the table, we can observe several
	regularities. First, the deep systems outperform the baselines. Second, the
	baselines that we choose are competitive and obtain high results. Since both
	baselines make their predictions based on lexical similarity and overlap, we can
	conclude that the dataset is biased towards those phenomena. Third, the 
	supervised systems generally outperform the unsupervised ones, but without 
	running a full grid-search the difference is relatively small.
	And finally, we can identify the best performing systems: 
	\textbf{S3} \citep{wang} for the supervised and \textbf{S9} \citep{infersent} for the unsupervised.
	BERT largely outperforms all other systems.

	The \textit{``overall performance''} provides a good overview of the task and 
	allows for a quantitative comparison of the different systems. However, it also has several 
	limitations. 
	It does not provide much insight into the workings of the systems and does not
	facilitate error analysis. In order to study and improve the performance of a
	system, a developer has to look at every correct and incorrect predictions
	and search for custom defined patterns. 
	The \textit{``overall performance''} is also not very informative for a comparison 
	between the systems. For example \textbf{S3} \citep{wang} and \textbf{S4} \citep{He2016}
	obtain the same Accuracy score and only differ by 0.06 F1 score. 
	With only
	looking at the quantitative evaluation it is unclear which of these systems would
	generalize better on a new dataset. 

	\subsection{Full Performance Profile}\label{5:sec:sprof}

	Table \ref{5:pr-wang} shows the full \textit{``performance profile''} of \textbf{S3} \citep{wang},
	the supervised system that performed best in terms of \textit{``overall 
	performance''}. 
	Table \ref{5:pr-wang} shows 
	a large variation of the performance of \textbf{S3} on the different 
	phenomena. The accuracy ranges from .33 to 1.0. 
	We also report the statistical significance of
	the difference 
	between the correct and incorrect predictions for each phenomena and the 
	correct and incorrect predictions 
	for the full test set, using the Mann--Whitney U-test\footnote{
	The Mann--Whitney U-test is a non-parametric equivalence of T-test. The U-Test does not
	assume normal distribution of the data and is better suited for small samples.} \citep{mann1947}.

	Ten of the phenomena show significant difference from the overall performance at
	\textit{p} \textless 0.1. Note that eight of them are also significant at \textit{p} \textless 0.05. The statistical significance of
	\textit{``Opposite polarity substitution (habitual)''}, and 
	\textit{``Negation Switching''} cannot be verified due to the 
	relatively low frequency of the phenomena in the test set. 

	The demonstrated
	variance in phenomena performance and its statistical significance address
	\textbf{RQ 1}: 
	we show that the performance of a PI 
	system on each candidate-paraphrase pair depends on the different 
	phenomena involved in that pair or at least there is a strong 
	observable relation between the performance and the phenomena.

	The individual \textit{``performance profile''} also addresses \textbf{RQ 2}.
	The profile is humanly interpretable, and we can clearly see how the 
	system performs on various sub-tasks at different linguistic levels. 
	The qualitative evaluation shows that \textbf{S3} 

		\begin{table} [H]
		\begin{center}

		\begin{tabular}{| l | c | c |}
		\hline
		\multicolumn{3}{|c|}{\textbf{OVERALL PERFORMANCE}} \\
		\hline \hline
		Overall Accuracy 				& .76 & \\
		\hline
		Overall F1 						& .833 & \\
		\hline \hline
		\multicolumn{3}{|c|}{\textbf{PHENOMENA PERFORMANCE}} \\
		\hline \hline
		\textbf{Phenomenon} & \textbf{Acc} &\textbf{p} \\
		\hline \hline
		\multicolumn{3}{|c|}{Morphology-based changes} \\
		\hline \hline
		Inflectional changes				& .79 & .21\\
		\hline
		\textbf{Modal verb changes} 		& \textbf{.90} & \textbf{.01}\\
		\hline
		Derivational changes		 		& .72 & .22\\
		\hline \hline
		\multicolumn{3}{|c|}{Lexicon-based changes} \\
		\hline \hline
		\textbf{Spelling changes} 			& \textbf{.88} & \textbf{.01}\\
		\hline
		Same polarity sub. (habitual) 		& .78 & .18\\
		\hline
		Same polarity sub. (contextual) 		& .75 & .37\\
		\hline
		Same polarity sub. (named ent.) 		& .73 & .14\\
		\hline
		Change of format 				& .75 & .44\\
		\hline \hline
		\multicolumn{3}{|c|}{Lexico-syntactic based changes} \\
		\hline \hline
		Opp. polarity sub. (habitual) 		& 1.0 & na \\
		\hline
		Opp. polarity sub. (context.)	 	& .68 & .14\\
		\hline
		Synthetic/analytic substitution 		& .77 & .39\\
		\hline
		\textbf{Converse substitution}		& \textbf{.92} & \textbf{.07}\\
		\hline \hline
		\multicolumn{3}{|c|}{Syntax-based changes} \\
		\hline \hline
		Diathesis alternation 				& .83 & .12\\
		\hline
		Negation switching				& .33 & na\\
		\hline
		\textbf{Ellipsis}					& \textbf{.64} & \textbf{.07}\\
		\hline
		Coordination changes 				& .77 & .47\\
		\hline
		\textbf{Subordination and nesting} 	& \textbf{.86} & \textbf{.01}\\
		\hline \hline
		\multicolumn{2}{|c|}{Discourse-based changes} \\
		\hline \hline
		\textbf{Punctuation changes} 		& \textbf{.87} & \textbf{.01}\\
		\hline
		Direct/indirect style 				& .76 & .5\\
		\hline
		\textbf{Syntax/discourse structure }	& \textbf{.83} & \textbf{.05}\\
		\hline \hline
		\multicolumn{2}{|c|}{Other changes} \\
		\hline \hline
		\textbf{Addition/Deletion} 			& \textbf{.70} & \textbf{.05}\\
		\hline
		\textbf{Change of order} 			& \textbf{.81} & \textbf{.04}\\
		\hline
		Contains negation 				& .78 & .32\\
		\hline
		Semantic (General Inferences) 		& .80 & .21\\
		\hline \hline
		\multicolumn{2}{|c|}{Extremes} \\
		\hline \hline
		Identity 						& .77 & .29\\
		\hline
		\textbf{Non-Paraphrase}			& \textbf{.81} & \textbf{.04}\\
		\hline
		Entailment 					& .76 & .5 \\
		\hline
		
		\end{tabular}

		\end{center}
		\caption{Performance profile of \citet{wang}} \label{5:pr-wang}
		\end{table}

	performs better when it has to
	deal with: 1) surface phenomena such as \textit{``spelling changes''}, 
	\textit{``punctuation changes''}, and \textit{``change of order''}; 2) dictionary
	related phenomena such as \textit{``opposite polarity substitution (habitual)''},
	\textit{ ``converse substitution''}, and \textit{``modal verb changes''}.
	\textbf{S3} performs worse when facing phenomena such as 
	\textit{``negation switching''}, \textit{``ellipsis''}, \textit{``opposite polarity
	substitution (contextual)''}, and \textit{``addition/deletion''}.


		\begin{table} [H]
		\begin{center}

		\begin{tabular}{| l || c c  c  c  c || c  c  c  c  c || c |  }
		\hline
		\textbf{Phenomenon} & \multicolumn{11}{|c|}{\textbf{Paraphrase Identification Systems}} \\
		\hline
		& \multicolumn{5}{|c|}{Supervised} & \multicolumn{5}{|c|}{Unsupervised} & \\
		\hline \hline
		& S1 & S2 & S3 & S4 & S5 & S6 & S7 & S8 & S9 & S10 & S11 \\
		\hline \hline
		\rowcolor{Gray}	
		OVERALL			& .74 & .75 & .76 & .76 & .70 & .68 & .70 & .72 & .75 & .73 & .84\\
		\hline
		Inflectional 			& .77 & .76 & .79 & .79 & .75 & .79 & .75 & .76 & .78 & .80 & .84\\

		\rowcolor{Gray}
		Modal verb 			& .84 & .89 & .90 & .89 & .91 & .92 & .89 & .84 & .81 & .89 & .92\\

		Derivational 			& .80 & .83 & .72 & .73 & .84 & .80 & .88 & .86 & .80 & .77 & .87\\
		\hline 
		\rowcolor{Gray}
		Spelling 				& .85 & .83 & .88 & .90 & .89 & .85 & .89 & .88 & .85 & .89 & .94\\

		Same pol. (hab.) 	& .74 & .77 & .78 & .76 & .76 & .76 & .76 & .75 & .76 & .76 & .85\\

		\rowcolor{Gray}
		Same pol. (con.) 	& .74 & .74 & .75 & .74 & .70 & .71 & .71 & .71 & .73 & .73 & .81\\

		Same pol. (NE) 		& .74 & .72 & .73 & .75 & .64 & .67 & .65 & .70 & .73 & .66 & .80\\

		\rowcolor{Gray}
		Change Format 		& .80 & .79 & .75 & .84 & .85 & .82 & .81 & .80 & .80  & .71 & .91\\
		\hline 
		Opp. pol. (hab.) 		& 1.0 & 1.0 & 1.0 & .50 & 1.0 & 1.0 & 1.0 & 1.0 & 1.0 & 1.0 & 1.0 \\

		\rowcolor{Gray}
		Opp. pol. (con.) 		& .77 & .84 & .68 & .84 & .52 & .84 & .61 & .77 & .65 & .52 & .71\\

		Synth./analytic	& .73 & .73 & .77 & .77 & .74 & .70 & .72 & .71 & .73 & .74 & .83\\

		\rowcolor{Gray}
		Converse sub.		& .93 & .93 & .92 & .86 & .93 & .86 & .79 & .79 & .93 & .79 & .86\\
		\hline 
		Diathesis altern. 		& .77 & .85 & .83 & .77 & .83 & .89 & .85 & .83 & .84 & .81 & .85\\

		\rowcolor{Gray}
		Negation switc 		& 1.0 & .67 & .33 & .33 & .33 & .67 & .33 & .67 & .33 & .67 & .33\\

		Ellipsis 				& .77 & .71 & .64 & .74 & .80 & .65 & .81 & .74 & .61 & .71 &  .81\\

		\rowcolor{Gray}
		Coordination 			& .92 & .92 & .77 & .92 & .77 & .92 & .85 & .85 & .92 & .92 & .92\\

		Subord. \& Nest.	 	& .83 & .84 & .86 & .84 & .81 & .81 & .85 & .86 & .80 & .85 & .93\\
		\hline 
		\rowcolor{Gray}
		Punctuation 			& .88 & .90 & .87 & .87 & .86 & .87 & .89 & .89 & .89 & .88 & .93\\

		Direct/indirect 		& .84 & .84 & .76 & .80 & .76 & .80 & .80 & .84 & .80 & .80 & .92\\

		\rowcolor{Gray}
		Syntax/Disc.	 	& .80 & .83 & .83 & .81 & .78 & .81 & .80 & .80 & .76 & .78 & .82\\
		\hline 
		Add./Del. 		& .69 & .68 & .70 & .72 & .67 & .64 & .65 & .66 & .70 & .67 & .82\\

		\rowcolor{Gray}
		Change of order 		& .82 & .83 & .81 & .81 & .77 & .82 & .82 & .82 & .83 & .84 & .89\\

		Contains neg. 		& .78 & .74 & .78 & .79 & .78 & .72 & .74 & .78 & .75 & .76 & .85\\

		\rowcolor{Gray}
		Semantic (Inf.) 	& .80 & .89 & .80 & .81 & .88 & .90 & .90 & .92 & .76 & .79 & .90\\
		\hline 
		Identity 				& .74 & .75 & .77 & .77 & .73 & .72 & .73 & .73 & .76 & .74 & .85\\

		\rowcolor{Gray}
		Non-Paraphrase 		& .76 & .77 & .81 & .75 & .71 & .55 & .67 & .68 & .77 & .79 & .88\\

		Entailment 			& .80 & .80 & .76 & .76 & .88 & .80 & .84 & .88 & .92 & .88 & .76\\
		\hline
		
		\end{tabular}

		\end{center}
		\caption{Performance profiles of all systems} \label{5:comparison}
		\end{table}

	\subsection{Comparing Performance Profiles}\label{5:sec:cprof}

	Table \ref{5:comparison} shows the full performance profiles of all systems. 
	The systems are identified by their IDs, as shown in Table \ref{5:binary}. 
	In addition to providing a better error analysis for every individual system, 
	the \textit{``performance profiles''} of the different systems can be used to
	compare them qualitatively. 
	This comparison is much more 
	informative than the \textit{``overall performance''} comparison shown
	in Table \ref{5:binary}. Using the \textit{``performance profile''}, we can 
	quickly compare the strong and weak sides of the different systems. 

	When looking at the \textit{``overall performance''}, we already pointed out
	that \textbf{S3} \citep{wang} and \textbf{S4} \citep{He2016}
	have almost identical quantitative results: 0.76 accuracy, 0.833 F1 for \textbf{S3} 
	against 0.76 accuracy, 0.827 F1 for \textbf{S4}. However, when we 
	compare their \textit{``phenomena performance''} it is evident that, while 
	these systems make approximately the same number of correct and incorrect 
	predictions, the actual predictions and errors can vary. 

	Looking at the accuracy, we can see that \textbf{S3} performs better on 
	phenomena such as \textit{``Converse substitution''}, \textit{``Diathesis alternation''},
	and \textit{``Non-Paraphrase''}, 	while \textbf{S4} performs better on 
	\textit{``Change of format''}, \textit{``Opposite polarity substitution 
	(contextual)''}, and \textit{``Ellipsis''}.
	
	We performed McNemar paired test comparing the errors 
	of the two systems for each phenomena. Table \ref{5:comp-34} shows some
	of the more interesting results. Four of the phenomena with largest difference
	in accuracy show significant difference with \textit{p} \textless 0.1. These 
	differences in performance are substantial, considering that the two 
	systems have nearly identical quantitative performance. 

		\begin{table} [h!]
		\begin{center}

		\begin{tabular}{| l | c | c | c |}	

		\hline
		\textbf{Phenomenon} & \textbf{\#3} & \textbf{\#4} & \textbf{p} \\
		\hline \hline
		Format			& .75 & .84 & .09 \\
		\hline
		Opp. Pol. Sub (con.) 	& .68 & .84 & .06 \\
		\hline
		Ellipsis			& .64 & .74 & .08 \\
		\hline
		Non-Paraphrase		& .81 & .75 & .07 \\
		\hline

		\end{tabular}

		\end{center}
		\caption{Difference in phenomena performance between S3 \citep{wang} and S4 \citep{He2016}} \label{5:comp-34}
		\end{table}

	We performed the same test on systems with a larger quantitative difference.
	Table \ref{5:comp-35} shows the comparison between 
	\textbf{S3} and \textbf{S5} \citep{subw}. Ten of the phenomena show significant 
	difference with \textit{p} \textless 0.1 and seven with \textit{p} \textless 0.05.
	These results answer our \textbf{RQ 3}: we show that there are significant differences
	between the \textit{``performance profiles''} of the different systems.	

		\begin{table} [h!]
		\begin{center}

		\begin{tabular}{| l | c | c | c |}	

		\hline
		\textbf{Phenomenon} & \textbf{\#3} & \textbf{\#5} & \textbf{p} \\
		\hline \hline
		Derivational		& .72 & .84 & .03 \\
		\hline
		Same Pol. Sub (con.) 	& .75 & .70 & .02 \\
		\hline
		Same Pol. Sub (NE)	& .73 & .64 & .01 \\
		\hline
		Format			& .75 & .85 & .03 \\
		\hline
		Opp. Pol. Sub (con.) 	& .68 & .52 & .10 \\
		\hline
		Ellipsis			& .64 & .80 & .10 \\
		\hline
		Addition/Deletion	& .70 & .67 & .02 \\
		\hline
		Identity			& .77 & .73 & .01\\
		\hline
		Non-Paraphrase		& .81 & .71 & .01 \\
		\hline
		Entailment			& .76 & .88 & .08 \\
		\hline

		\end{tabular}

		\end{center}
		\caption{Difference in phenomena performance: S3~\citep{wang} and S5 \citep{subw} } \label{5:comp-35}
		\end{table}


	\subsection{Comparing Performance by Phenomena}

	The \textit{``phenomena performance''} of the individual systems clearly differ 
	among them, but they also show noticeable tendencies. Looking at the performance
	by phenomena, it is evident that certain phenomena consistently obtain lower 
	than average accuracy across multiple systems while other phenomena 
	consistently obtain higher than average accuracy.

	In order to quantify these observations and to confirm that there is a statistical significance
	we performed Friedman-Nemenyi test \citep{Demsar}. For each system, we ranked the 
	performance by phenomena from 1 to 27, accounting for ties. We calculated the
	significance of the difference in ranking between the phenomena using the 
	Friedman test \citep{Friedman:1940} and obtained a Chi-Square value of 198, which rejects the 
	null hypothesis with \textit{p} \textless 0.01. 
	Once we had checked for the non-randomness of our results, we computed 
	the Nemenyi test \citep{Nemenyi:1963} to find out which phenomena were significantly different.
	In our case, we compute the two-tailed Nemenyi test for k = 27 phenomena 
	and N = 11 systems. The Critical Difference (CD) for these values
	is 12.5 at \textit{p} \textless 0.05.

	Figure \ref{5:fr-nem} shows the Nemenyi test with the CD value. Each phenomenon is plotted with
	its average rank across the 11 evaluated systems. The horizontal lines connect phenomena
	which rank is within CD of each other. Phenomena which are not connected by
	a horizontal line have significantly different ranking. We can observe that each phenomenon
	is significantly different from at least half of the other phenomena.  

\begin{figure*}[t]
\includegraphics[width=14cm]{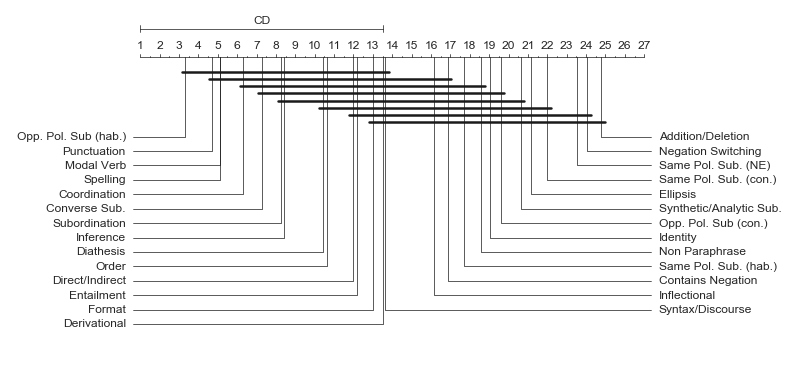}
\caption{Critical Difference diagram of the average ranks by phenomena}\label{5:fr-nem}
\end{figure*}

	We can observe
	that some phenomena, such as \textit{``opposite polarity substitution (habitual)'',
	``punctuation changes'', ``spelling'', ``modal verb changes'',} and \textit{``coordination 
	changes''} 
	are statistically much easier according to our evaluation,
	as they are consistently among the best performing phenomena across all 
	systems. Other phenomena, such as  \textit{``negation switching'', 
	``addition/deletion'', ``same polarity substitution (named entity)'',
	``opposite polarity substitution (contextual)'',} and \textit{``ellipsis''} are
	statistically much harder, as they are consistently among the worst performing
	phenomena across all systems. With the exception of \textit{``negation switching''}
	and \textit{``opposite polarity substitution (habitual)''}, these phenomena occur
	in the corpus with sufficient frequency. These results answer our \textbf{RQ 4}: we show
	that there are phenomena which are easier or harder for the majority of the
	evaluated systems.

	\section{Discussion}

	In Section \ref{5:methodology} we described our evaluation methodology and
	posed four research questions. The experiments that we performed and the
	analysis of the results answered all four of them. We briefly discuss the
	implications of the findings.

	By addressing \textbf{RQ 1}, we showed that the performance of a system can 
	differ significantly based on the phenomena involved in each candidate-paraphrase 
	pair. By addressing \textbf{RQ 4}, we showed that some phenomena are
	consistently easier or harder across the majority of the systems.
	These findings empirically prove the complexity of paraphrasing and
	the task of PI. The results justify the distinction between the
	qualitatively different linguistic phenomena involved in paraphrasing and demonstrate
	that framing PI as a binary classification problem is an oversimplification.

	By addressing \textbf{RQ 2}, we showed that each system has strong and weak 
	sides, which can be identified and interpreted via its \textit{``performance profile''}.
	This information can be very valuable when analyzing the errors made by the
	system or when reusing it on another task. Given the Deep architecture of the
	systems, such a detailed interpretation is hard to obtain via other means and metrics.
	By addressing \textbf{RQ 3}, we showed that two systems can differ significantly in their 
	performance on 	candidate-paraphrase pairs involving particular phenomenon. 
	These differences can be seen even in systems that have almost identical quantitative 
	(Acc and F1) performance on the full test set. 
	These findings justify the need for a qualitative evaluation framework for PI.
	The traditional binary evaluation metrics do not account for the difference in 
	phenomena performance. They do not provide enough information for the analysis
	or for the comparison of different PI systems. Our proposed framework
	shows promising results.
	
	Our findings demonstrate the limitations of the traditional PI task definition and datasets and the 
	way PI systems are typically interpreted and evaluated. We show the advantages 
	of a qualitative evaluation framework and emphasize the need to further research 
	and improve the PI task. 
	The \textit{``performance profile''} also enables the direct
	empirical comparison of related phenomena such as \textit{``same polarity 
	substitution (habitual)''} and \textit{``(contextual)''}
	or \textit{``contains negation''} and \textit{``negation switching''}. These 
	comparisons, however, fall outside of the scope of this paper.

	Our evaluation framework is not specific to the ETPC corpus
	or the typology behind it. The framework can be applied to other corpora
	and tasks, provided they have a similar format. While ETPC is the largest
	corpus annotated with paraphrase types to date, it has its limitations as some
	interesting paraphrase types (ex.: \textit{``negation switching''}) do
	not appear with a sufficient frequency. We release the code for the creation and
	analysis of the \textit{``performance profile''} \footnote{\url{https://github.com/JavierBJ/paraphrase\_eval}}. 

\section{Conclusions and Future Work} \label{5:conclusions}

	We present a new methodology for evaluation, interpretation, and comparison of
	different Paraphrase Identification systems. The methodology only requires at
	evaluation time a corpus annotated with detailed semantic relations. The training
	corpus does not need any additional annotation. The evaluation also does not
	require any additional effort from the systems' developers. Our methodology has
	clear advantages over using simple quantitative measures (Accuracy and F1 Score):
	1) It allows for a better interpretation and error analysis on the individual systems; 2) It allows for a 
	better qualitative comparison between the different systems; and 3) It identifies phenomena 
	which are easy/hard to solve for multiple systems and may require further research.

	We demonstrate the methodology by evaluating and comparing several of the
	state-of-the-art systems in PI. The results show that there is a statistically
	significant relationship between the phenomena involved in each candidate-paraphrase 
	pair and the performance of the different systems. We show the strong and weak sides
	of each system using human-interpretable categories and we also identify phenomena
	which are statistically easier or harder across all systems. 

	As a future work, we 
	intend to study phenomena 
	that are hard for the majority of the systems 
	and proposing ways to improve the performance on those phenomena. 
	We also plan to apply the evaluation methodology to more tasks and systems that require 
	a detailed semantic evaluation, 
	and further test it with transfer learning experiments.

\section*{Acknowledgements}

	We would like to thank Darina Gold, Tobias  Horsmann, Michael  Wojatzki, and Torsten Zesch 
	for their support and suggestions, and the anonymous reviewers for their feedback and 
	comments.

	This work has been funded by Spanish Ministery of Science, Innovation, and Universities Project PGC2018-096212-B-C33, by 
	the CLiC research group (2017 SGR 341), and by the APIF grant of the first author.

\part{Paraphrasing, Textual Entailment, \\and Semantic Similarity}\label{p:mrel}

\chapter[Annotating and Analyzing the \\ Interactions between Meaning Relations]{\centering Annotating and Analyzing the Interactions between Meaning Relations}\label{ch:law}

\chaptermark{Relations}

\begin{center}

	Darina Gold\affmark[1]\authmark[*], Venelin Kovatchev\affmark[2]\affmark[3]\authmark[*], Torsten Zesch\affmark[1]\\
	\affaddr{\affmark[1]Language Technology Lab, University of Duisburg-Essen, Germany}\\
	\affaddr{\affmark[2]Language and Computation Center, Universitat de Barcelona, Spain}\\
	\affaddr{\affmark[3]Institute of Complex Systems, Universitat de Barcelona, Spain}\\	
	\authnote{\authmark[*]Both authors contributed equally to this work}\\

\vspace{10mm}

Published at \\ \textit{Proceedings of the \\Thirteenth Language Annotation Workshop}, 2019 \\ pp.: 26-36
\end{center}

\paragraph{Abstract}
	Pairs of sentences, phrases, or other text pieces can hold semantic relations such as paraphrasing, textual entailment, contradiction, specificity, and semantic similarity.
	These relations are usually studied in isolation and no dataset exists where they can be compared empirically.
	Here we present a corpus annotated with these relations and the analysis of these results.
	The corpus contains 520 sentence pairs, annotated with these relations.
	We measure the annotation reliability of each individual relation and we examine their interactions and correlations.
	Among the unexpected results revealed by our analysis is that the traditionally considered direct relationship between paraphrasing and bi-directional entailment does not hold in our data.

\section{Introduction}
\label{6:sec:intro}
Meaning relations refer to the way in which two sentences can be connected, e.g.\ if they express approximately the same content, they are considered paraphrases.
Other meaning relations we focus on here are textual entailment and contradiction\footnote{Mostly, contradiction is regarded as one of the relations within an entailment annotation.} \citep{rte}, and specificity. 

Meaning relations have applications in many NLP tasks, e.g. recognition of textual entailment is used for summarization \citep{lloret2008text} or machine translation evaluation \citep{pado2009textual}, and paraphrase identification is used in summarization \citep{harabagiu2010using}.

The complex nature of the meaning relations makes it difficult to come up with a precise and widely accepted definition for each of them. Also, there is a difference between theoretical definitions and definitions adopted in practical tasks. In this paper, we follow the approach taken in previous annotation tasks
and we give the annotators generic and practically oriented instructions. 

\textbf{Paraphrases} are differently worded texts with approximately the same content \citep{BhagatHovy,de1981introduction}.
The relation is symmetric. 
In the following example, (a) and (b) are paraphrases.
\begin{itemize}
	\item [(a)] \textit{Education is equal for all children.}
	\item [(b)] \textit{All children get the same education.}
\end{itemize}

\textbf{Textual Entailment} is a directional relation between pieces of text in which the information of the \textit{Text} entails the information of the \textit{Hypothesis} \citep{rte}. 
In the following example, Text (t) entails  Hypothesis (h):
\begin{itemize}
	\item[(t)] \textit{All children get the same education.}
	\item[(h)] \textit{Education exists.}
\end{itemize}

\textbf{Specificity} is a relation between phrases in which one phrase is more precise and the other more vague.
Specificity is mostly regarded between noun phrases \citep{cruse1977pragmatics,encc1991semantics,farkas2002specificity}.
However, there has also been work on specificity on the sentence level \citep{louis2012corpus}.
In the following example, (c) is more specific than (d) as it gives information on who does not get good education:
\begin{itemize}
	\item [(c)] \textit{Girls do not get good education.}
	\item [(d)] \textit{Some children do not get good education.}
\end{itemize}

\textbf{Semantic Similarity} between texts is not a meaning relation in itself, but rather a gradation of meaning similarity.
It has often been used as a proxy for the other relations in applications such as summarization \citep{lloret2008text}, plagiarism detection \citep{alzahrani2010fuzzy,bar2012text}, machine translation \citep{pado2009textual}, question answering \citep{harabagiu2006methods}, and natural language generation \citep{agirre2013sem}.
We use it in this paper to quantify the strength of relationship on a continuous scale.
Given two linguistic expressions, semantic text similarity measures the degree of semantic equivalence \citep{agirre2013sem}.
For example, (a) and (b) have a semantic similarity score of 5 (on a scale from 0-5 as used in the SemEval STS task) \citep{agirre2013sem,agirre2014semeval}.  

\paragraph{Interaction between Relations}
Despite the interactions and close connection of these meaning relations, to our knowledge, there exists neither an empirical analysis of the connection between them nor a corpus enabling it.
We bridge this gap by creating and analyzing a corpus of sentence pairs annotated with all discussed meaning relations.

Our analysis finds that previously made assumptions on some relations (e.g.\ paraphrasing being bi-directional entailment \citep{madnani2010generating,androutsopoulos2010survey,sukhareva2016crowdsourcing}) are not necessarily right in a practical setting. 
Furthermore, we explore the interactions of the meaning relation of specificity, which has not been extensively studied from an empirical point of view. 
We find that it can be found in pairs on all levels of semantic relatedness and does not correlate with entailment.

\section{Related Work}
\label{6:sec:related_work}
To our knowledge, there is no other work where the discussed meaning relations have been annotated separately on the same data, enabling an unbiased analysis of the interactions between them.
There are corpora annotated with multiple semantic phenomena, including meaning relations. 

\subsection{Interactions between Relations}
There has been some work on the interaction between some of the discussed meaning relations, especially on the relation between entailment and paraphrasing, and also on how semantic similarity is connected to the other relations.

\paragraph{Interaction between Entailment and Paraphrases}
According to \citet{madnani2010generating,androutsopoulos2010survey}, bi-directional entailment can be seen as paraphrasing.
Furthermore, according to \citet{androutsopoulos2010survey} both entailment and paraphrasing are intended to capture human intuition.
\citet{etpc} emphasize the similarity between linguistic phenomena underlying paraphrasing and entailment.
There has been practical work on using paraphrasing to solve entailment \citep{bosma2006paraphrase}.

\paragraph{Interaction between Entailment and Specificity}
Specificity was involved in rules for the recognition of textual entailment \citep{bobrow2007precision}.

\paragraph{Interaction with Semantic Similarity}
\citet{cer2017semeval} argue that to find paraphrases or entailment, some level of semantic similarity must be given.
Furthermore, \citet{cer2017semeval} state that although semantic similarity includes both entailment and paraphrasing, it is different, as it has a gradation and not a binary measure of the semantic overlap.
Based on their corpus, \citet{marelli2014sick} state that paraphrases, entailment, and contradiction have a high similarity score; paraphrases having the highest and contradiction the lowest of them.
There also was practical work using the interaction between semantic similarity and entailment: \citet{yokote2011effects} and \citet{castillo2010using} used semantic similarity to solve entailment.




\subsection{Corpora with Multiple Semantic Layers}
There are several works describing the creation, annotation, and subsequent analysis of corpora with multiple parallel phenomena.

\textbf{MASC}
The annotation of corpora with multiple phenomena in parallel has been most notably explored within the Manually Annotated Sub-Corpus (MASC) project\footnote{\url{http://www.anc.org/MASC/About.html}} ---
It is a large-scale, multi-genre corpus manually annotated with multiple semantic layers, including WordNet senses\citep{miller1998wordnet}, Penn Treebank Syntax \citep{marcus1993building}, and opinions.
The multiple layers enable analyses between several phenomena.

\textbf{SICK}
is a corpus of around 10,000 sentence pairs that were 
annotated with semantic similarity and entailment in parallel \citep{marelli2014sick}. As it is the corpus that is the most similar to our work, we will compare some of our annotation decisions and results with theirs. 

\textbf{\citet{sukhareva2016crowdsourcing}} annotated subclasses of entailment, including \textit{paraphrase, forward, revert,} and \textit{null} on propositions extracted from documents on educational topics that were paired according to semantic overlap.
Hence, they implicitly regarded paraphrases as a kind of entailment.

\section{Corpus Creation}
\label{6:sec:corpus_creation}
To analyze the interactions between semantic relations, a corpus annotated with all relations in parallel is needed.
Hence, we develop a new corpus-creation methodology which ensures all relations of interest to be present.
First, we create a pool of potentially related sentences. 
Second, based on the pool of sentences, we create sentence pairs that contain all relations of interest with sufficient frequency. 
This contrasts existing corpora on meaning relations that are tailored towards one relation only.
Finally, we take a portion of the corpus and annotate all relations via crowdsourcing.
This part of our methodology differs significantly from the approach taken in the SICK corpus \citep{marelli2014sick}. 
They don't create new corpora, but rather re-annotate pre-existing corpora, which does not allow them to control for the overall similarity between the pairs.

\subsection{Sentence Pool}

\begin{table}[h]
	\footnotesize
	\begin{tabularx}{\linewidth}{X}
		\toprule
		Getting a high educational degree is important for finding a good job, especially in big cities.                    \\ \midrule 
		In many countries, girls are less likely to get a good school education.                                              \\ \midrule
		Going to school socializes kids through constant interaction with others.                                             \\ \midrule
		One important part of modern education is technology, if not the most important.                                      \\ \midrule
		Modern assistants such Cortana, Alexa, or Siri make our everyday life easier by giving quicker access to information. \\ \midrule
		New technologies lead to asocial behavior by e.g. depriving us from face-to-face social interaction.                  \\ \midrule
		Being able to use modern technologies is obligatory for finding a good job.                                           \\ \midrule
		Self-driving cars are safer than humans as they don't drink.                                                          \\ \midrule
		Machines are good in strategic games such as chess and Go.                                                            \\ \midrule
		Machines are good in communicating with people.                                                                       \\ \midrule
		Learning a second language is beneficial in life.                                                                     \\ \midrule
		Speaking more than one language helps in finding a good job.                                                          \\ \midrule
		Christian clergymen learn Latin to read the bible.                                                                    \\ 
		\bottomrule
	\end{tabularx}
	\caption{List of given source sentences}
	\label{6:tab:list_source_sents}
\end{table}

In the first step, the authors create 13 sentences, henceforth \textit{source sentences}, shown in Table~\ref{6:tab:list_source_sents}.
The sentences are on three topics: \textit{education}, \textit{technology}, and \textit{language}.
We choose sentences that can be understood by a competent speaker without any domain-specific knowledge and which due to their complexity potentially give rise to a variety of lexically differing sentences in the next step.
Then, a group of 15 people, further on called \textit{sentence generators}, is asked to generate \textit{true} and \textit{false} sentences that vary lexically from the source sentence.\footnote{The full instructions given to the sentence generators is included with the corpus data.}
Overall, 780 sentences are generated.
The 13 \textit{source sentences} are not considered in the further procedure.

For creating the \textit{true} sentences, we ask each sentence generator to create two sentences that are true and for the \textit{false} sentences, two sentences that are false given one source sentence. 
This way of generating a sentence pool is similar to that of the textual entailment SNLI corpus \citep{snli}, where the generators were asked to create true and false captions for given images.
The following are exemplary true and false sentences created from one source sentence.


	\begin{itemize}
		\item [] \textbf{Source:} \textit{Getting a high educational degree is important for finding a good job, especially in big cities.}
		\item [] \textbf{True:} \textit{Good education helps to get a good job.}
		\item [] \textbf{False:} \textit{There are no good or bad jobs.}
	\end{itemize}



\subsection{Pair Generation}
\label{6:subsec:pair_generation}
We combine individual sentences from the sentence pool into pairs, as meaning relations are present between pairs and not individual sentences.
To obtain a corpus that contains all discussed meaning relation with sufficient frequency, we use four pair combinations: 
\begin{itemize}

	\item[1)] a pair of two sentences that are true given the same source sentence \\ (\textit{true-true})
	\item[2)] a pair of two sentences that are false given the same source sentence \\ (\textit{false-false})
	\item[3)] a pair of one sentence that is true and one sentence that is false given the same source sentence (\textit{true-false})
	\item[4)] a pair of randomly matched sentences from the whole sentence pool and all source sentences (\textit{random})
\end{itemize}

From the 780 sentences in the sentence pool, we created a corpus of 11,310 pairs, with a pair distribution as follows: 5,655 (50\%) \textit{true-true}; 2,262 (20\%) \textit{false-false}, 2,262 (20\%) \textit{true-false}, and 1,131 (10\%) \textit{random}. We include all possible 5,655 \textit{true-true} combinations of 30 true sentences for each of the 13 source sentences. For \textit{false-false}, \textit{true-false}, and \textit{random} we downsample the full set of pairs to obtain the desired number, keeping an equal number of samples per source sentence.
We chose this distribution because we are mainly interested in paraphrases and entailment, as well as their relation to specificity. 
We hypothesize that pairs of sentences that are both true have the highest potential to contain these relations.

From the 11,310 pairs, we randomly selected 520 (5\%) for annotation, with the same 50-20-20-10 distribution as the full corpus.  
We select an equal number of pairs from each source sentence.
We hypothesize that length strongly correlates with specificity, as there is potentially more information in a longer sentence that in a shorter one.
Hence, for half of the pairs, we made sure that the difference in length between the two sentences is not more than 1 token.

\subsection{Relation Annotation}
\label{6:subsec:relation_crowdsourcing}

We annotate all the relations in the corpus of 520 sentence pairs using Amazon Turk. 
We select 10 crowdworkers per task, as this gives us the possibility to measure how well the tasks has been understood overall, but especially how easy or difficult individual pairs are in the annotation of a specific relation.
In the SICK corpus, the same platform and number of annotators were used.

We chose to annotate the relations separately to avoid biasing the crowdworkers who might learn heuristic shortcuts when seeing the same relations together too often.
We launched the tasks consecutively to have the annotations as independent as possible.
This differs from the SICK corpus annotation setting, where entailment, contradiction, and semantic similarity were annotated together.



The complex nature of the meaning relations makes it difficult to come up with a precise and widely accepted definition and annotation instructions for each of them. 
This problem has already been emphasized in previous annotation tasks and theoretical settings \citep{BhagatHovy}. The standard approach in most of the existing paraphrasing and entailment datasets is to use a more generic and less strict definitions. 
For example, pairs annotated as ``paraphrases'' in MRPC \citep{mrpc} can have ``obvious differences in information content''. 
This ``relatively loose definition of semantic equivalence'' is adopted in most empirically oriented paraphrasing corpora.

We take the same approach towards the task of annotating semantic relations: we provide the annotators with simplified guidelines, as well as with few positive and negative examples. 
In this way, we believe that annotation is more generic, reproducible, and applicable to any kind of data. It also relies more on the intuitions of a competent speaker than on understanding complex linguistic concepts. 
Prior to the full annotation, we performed several pilot studies on a sample of the corpus in order to improve instructions and examples given to the annotators.
In the following, we will shortly outline the instructions for each task.

\textbf{Paraphrasing}
In Paraphrasing (PP), we ask the crowdworkers whether the two sentences have approximately the same meaning or not, which is similar to the definition of \citet{BhagatHovy} and \citet{de1981introduction}.

\textbf{Textual Entailment}
In Textual Entailment (TE), we ask whether the first sentence makes the second sentence true.
Similar to RTE Tasks \citep{rte} - \citep{rte7}, we only annotate for forward entailment (FTE). 
Hence, we use the pairs twice: in the order we ask for all other tasks and in reversed order, to get the entailment for both directions.
Backward Entailment is referred to as \textit{BTE}.
If a pair contains only backward or forward entailment, it is uni-directional (UTE). 
If a pair contains both forward and backward entailment, it is bi-directional (BiTE).
Our annotation instructions and the way we interpret directionality is similar to other crowdworking tasks for textual entailment \citep{marelli2014sick,snli}.

\textbf{Contradiction}
In Contradiction (Cont), we ask the annotators whether the sentences contradict each other.
Here, our instructions are different from the typical approach in RTE \citep{rte}, where contradiction is often understood as the absence of entailment. 

\textbf{Specificity}
In Specificity (Spec), we ask whether the first sentence is more specific than the second.
To annotate specificity in a comparative way is new \footnote{\citet{louis2012corpus} labelled individual sentences as \textit{specific}, \textit{general}, or \textit{cannot decide}.}.
Like in textual entailment, we pose the task only in one direction.
If the originally first sentence is more specific, it is forward specificity (FSpec), whereas if the originally second sentence is more specific than the first, it is backward specificity (BSpec).

\textbf{Semantic Similarity}
For semantic similarity (Sim), we do not only ask whether the pair is related, but rate the similarity on a scale 0-5. Unlike previous studies \citep{agirre2014semeval}, we decided not to provide explicit definitions for every point on the scale. 

\textbf{Annotation Quality} To ensure the quality of the annotations, we include 10 control pairs, which are hand-picked and slightly modified pairs from the original corpus, in each task.\footnote{The control pairs are also available online at \url{https://github.com/MeDarina/meaning_relations_interaction}} 
We discard workers who perform bad on the control pairs.
\footnote{Only 2 annotators were discarded across all tasks. To have an equal number of annotations for each task, we re-annotated these cases with other crowdworkers.}

\subsection{Final Corpus}
For each sentence pair, we get 10 annotations for each relation, namely paraphrasing, entailment, contradiction, specificity, and semantic similarity. 
Each sentence pair is assigned a binary label for each relation, except for similarity.
We decide that if the majority (at least 60\% of the annotators) voted for a relation, it gets the label for this relation.

Table~\ref{6:tab:ex_anno} shows exemplary annotation outputs of sentence pairs taken from our corpus.
For instance, sentence pair \#4 contains two relations: forward entailment and forward specificity. 
This means that it has uni-directional entailment and the first sentence is more specific than the second.
The semantic similarity of this pair is 2.7.


\paragraph{Inter-Annotator Agreement}

We evaluate the agreement on each task separately.
For semantic similarity, we determine the average similarity score and the standard deviation for each pair.
We also calculate the Pearson correlation between each annotator and the average score for their pairs. 
We report the average correlation, as suggested by SemEval \citep{agirre2014semeval} and SICK.

For all nominal classification tasks we determine the majority vote and calculate the \% of agreement between the annotators. 
This is the same measure used in the SICK corpus.
Following the approach used with semantic similarity, we also calculated Cohen's $kappa$ between each annotator and the majority vote for their pairs. 
We report the average $kappa$ for each task.\footnote{We are aware that $\kappa$ does not fit the restrictions of our task very well and also that it is usually not averaged. However, we wanted to report a chance corrected measure, which is non-trivial in a crowd-sourcing setting, where each pair is annotated by a different set of annotators.}

\begin{table}[h]
	\small
	\centering
	\begin{tabular}{l||l|l|l|l|l|}
		& \% & $\kappa$ &  \%\cmark & \%\xmark   & control         \\
		\toprule
		PP         & .87 & .67 & .83 & .90 & .98\\
		TE          & .83 & .61 &     .75               &       .89 & .89\\
		Cont       & .94 & .71 &   .84                 &       .95 & .95\\
		Spec         & .80 & .56 & .81               &        .82 & .89\\
		\bottomrule
	\end{tabular}
	\caption{Inter-annotator agreement for binary relations
	\\ \cmark denotes a relation being there 
	\\ \xmark denotes a relation not being there}
	\label{6:tab:iaa}
\end{table}

Table~\ref{6:tab:iaa} shows the overall inter-annotator agreement for the binary tasks. We report: 1) the average \%-agreement for the whole corpus; 2) the average $\kappa$ score; 3) the average \%-agreement for the pairs where the majority label is \textit{``yes''}; 4) the average \%-agreement for the pairs where the majority label is \textit{``no''}; 5) the average \% agreement between the annotators and the expert-provided ``control labels'' on the control questions.


The overall agreement for all tasks is between .80 - .94, which is quite good given the difficulty of the tasks.
Contradiction has the highest agreement with .94. 
It is followed by the paraphrase relation, which has an agreement of .87.
The agreements of the entailment and specificity relations are slightly lower, which reflects that the tasks are more complex. 
SICK report agreement of .84 on entailment, which is consistent with our result.

The agreement is higher on the control questions than on the rest of the corpus. We consider it the upper boundary of agreement.
The agreement on the individual binary classes shows that, except for the specificity relation, annotators have a higher agreement on the absence of relation.

\begin{table}[h]
	\small
	\centering
	\begin{tabular}{l||l|l|l|l|l|l|}
		& 50\% & 60\% & 70\% &  80\% & 90\%   & 100\%         \\
		\toprule
		PP         & .11 & .12 & .13 & .20 & .24 & .20\\
		TE         & .17 & .19 & .17 & .16 & .19 & .10\\
		Cont       & .04 & .07 & .18 & .23 & .23 & .25\\
		Spec       & .22 & .18 & .21 & .13 & .13 & .12\\
		\bottomrule
	\end{tabular}
	\caption{Distribution of Inter-annotator agreement}
	\label{6:tab:iaa-dist}
\end{table}

Table \ref{6:tab:iaa-dist} shows the distribution of agreement for the different relations. We take all pairs for which at least 50\% of the annotators found the relation and shows what percentage of these pairs have inter-annotator agreement of 50\%, 60\%, 70\%, 80\%, 90\%, and 100\%. We can observe that, with the exception of contradiction, the distribution of agreement is relatively equal. For our initial corpus analysis, we discarded the pairs with 50\% agreement and we only considered pairs where the majority (60\% or more) of the annotators voted for the relation. However, the choice of agreement threshold an empirical question and the threshold can be adjusted based on particular objectives and research needs.

The average standard deviation for semantic similarity is 1.05. SICK report average deviation of .76, which is comparable to our result, considering that they use a 5 point scale (1-5), and we use a 6 point one (0-5). 
Pearson's r between annotators and the average similarity score is 0.69 which is statistically significant at $\alpha=0.05$. 

\paragraph{Distribution of Meaning Relations}
Table~\ref{6:tab:pair_generation} shows that all meaning relations are represented in our dataset. 
We have 160 paraphrase pairs, 195 textual entailment pairs, 68 contradiction pairs, and 381 specificity pairs.
There is only a small number of contradictions, but this was already anticipated by the different pairings.
The distribution is similar to \citet{marelli2014sick} in that the set is slightly leaning towards entailment\footnote{As opposed to contradiction. However, as contradiction and entailment were annotated exclusively, it is not directly comparable.}.
Furthermore, the distribution of uni- and bi-directional entailment with our and the SICK corpus are similar: they are nearly equally represented.\footnote{In SICK 53\% of the entailment is uni-directional and 46\% are bi-directional, whereas we have 44\% uni-directional and 55\% bi-directional.} 

\paragraph{Distribution of Meaning Relations with Different Generation Pairings}
Table~\ref{6:tab:pair_generation} shows the distribution of meaning relations and the average similarity score in the differently generated sentence pairings.
In the true/true pairs, we have the highest percentage of paraphrase (49\%), entailment (60\%), and specificity (79\%).
In the false/false pairs, all relations of interest are present: paraphrases (27\%), entailment (36\%), and specificity (72\%).
Unlike in true/true pairs, false/false ones include contradictions (10\%).
True/false pairs contain the highest percentage of contradiction (85\%).
There were also few entailment and paraphrase relations in true/false pairs.
In the random pairs, there were only few relations of any kind.
The proportion of specificity is high in all pairs.

This different distribution of phenomena based on the source sentences can be used in further corpus creation when determining the best way to combine sentences in pairs. 
In our corpus, the balanced distribution of phenomena we obtain justifies our pairing choice of 50-20-20-10.


\begin{table}[]
	\small
	\centering
	\begin{tabular}{l|r|r|r|r|r}
		& all  & T/T  & F/F & T/F  & rand. \\
		
		\toprule
		PP      &  31\% &   49\%&    27\% &  2\% &  6\%      \\
		TE        &  38\%  &  60\%     &  36\% &  2\%&     2\%         \\
		Cont.   & 13\%  &    0\%    &      10 \%    &  56\%&0\%       \\
		Spec      &  73\% &    79\%    &    72\%     &  66\%   &   63\%            \\
		\midrule
		$\varnothing$Sim &   2.27&  2.90 &    2.39   & 1.32   & 0.77 \\
		\bottomrule
	\end{tabular}
	\caption{Distribution of meaning relations within different pair generation patterns}
	\label{6:tab:pair_generation}
\end{table}

\paragraph{Lexical Overlap within Sentence Pairs}

As discussed by \citet{joao2007new}, a potential flaw of most existing relation corpora is the high lexical overlap between the pairs.
They show that simple lexical overlap metrics pose a competitive baseline for paraphrase identification.
Due to our creation procedure, we reduce this problem.
In Table~\ref{6:tab:comp_bleu}, we quantified it by calculating unigram and bigram BLEU score between the two
texts in each pair for our corpus, MRPC and SNLI, which are the two most used corpora for paraphrasing and textual entailment. 
The BLEU score is much lower for our corpus that for MRPC and SNLI.

\begin{table}[h]
	\small
	\centering
	\begin{tabular}{l|r|r|r}
	& MRPC & SNLI & Our corpus \\
	\toprule
	unigram & 61 & 24 &18 \\
	bigram & 50 & 12 & 6 \\
	\bottomrule
	\end{tabular}
	\caption{Comparison of BLEU scores between the sentence pairs in different corpora}
	\label{6:tab:comp_bleu}
\end{table}

\paragraph{Relations and Negation}
Our corpus also contains multiple instances of relations that involve negations and also double negations. Those examples could pose difficulties to automatic systems and could be of interest to researchers that study the interaction between inference and negation. 
Pairs \#1, \#2, and \#9 in Table~\ref{6:tab:ex_anno} are examples for pairs containing negation in our corpus.

\section{Interactions between Relations}
\label{6:sec:corpus_analysis}
We analyze the interactions between the relations in our corpus in two ways. First, we calculate the correlation between the binary relations and the interaction between them and similarity. 
Second, we analyze the overlap between the different binary relations and discuss interesting examples. 


\subsection{Correlations between Relations}
We calculate correlations between the binary relations using the Pearson correlation. 
For the correlations of the binary relations with semantic similarity, we discuss the average similarity and the similarity score scales of each binary relation.
\subsubsection{Correlation of Binary Meaning Relations}
In Table~\ref{6:tab:correlation_rest}, we show the Pearson correlation between the meaning relations.
For entailment, we show the correlation for uni-directional (UTE), bi-directional (BTE), and any-directional (TE).

Paraphrases and any-directional entailment are highly similar with a correlation of $.75$.
Paraphrases have a much higher correlation with bi-directional entailment (.70) than with uni-directional entailment (.20).
Prototypical examples of pairs that are both paraphrases and textual entailment are pairs \#1 and \#2 in Table~\ref{6:tab:ex_anno}.
Furthermore, both paraphrases and entailment have a negative correlation with contradiction, which is expected and confirms the quality of our data.

Specificity does not have any strong correlation with any of the other relations, showing that it is independent of those in our corpus.

\begin{table}[h]
	\centering
	\small
	\begin{tabular}{l|rrrrr|r}
		& TE  & UTE & BiTE & Cont & Spec & $\varnothing$ Sim\\
		\toprule
		PP     & \cc{green!75} .75 & \cc{green!20}.20 & \cc{green!70}.70 & \cc{red!25}-.25       &  \cc{red!1}-.01  & 3.77  \\
		TE    &    &\cc{green!57}.57&\cc{green!66}.66& \cc{red!30}-.30      &   \cc{red!1}-.01  & 3.59 \\
		UTE     & & &\cc{red!23}-.23&\cc{red!17}-.17&\cc{red!4}-.04 & 3.21\\
		BiTE     & & & &\cc{red!20}-.20&-.01 & 3.89\\
		Cont     &    & &  &        &   \cc{red!9}-.09  & 1.45 \\
		Spec   &  & &    &         &  & 2.27  \\ 
		\bottomrule
	\end{tabular}
	\caption{Correlation between all relations}
	\label{6:tab:correlation_rest}
\end{table}

\subsubsection{Binary Relations and Semantic Similarity}

\begin{figure}[h!]
	\begin{center}
	\includegraphics[width=8cm]{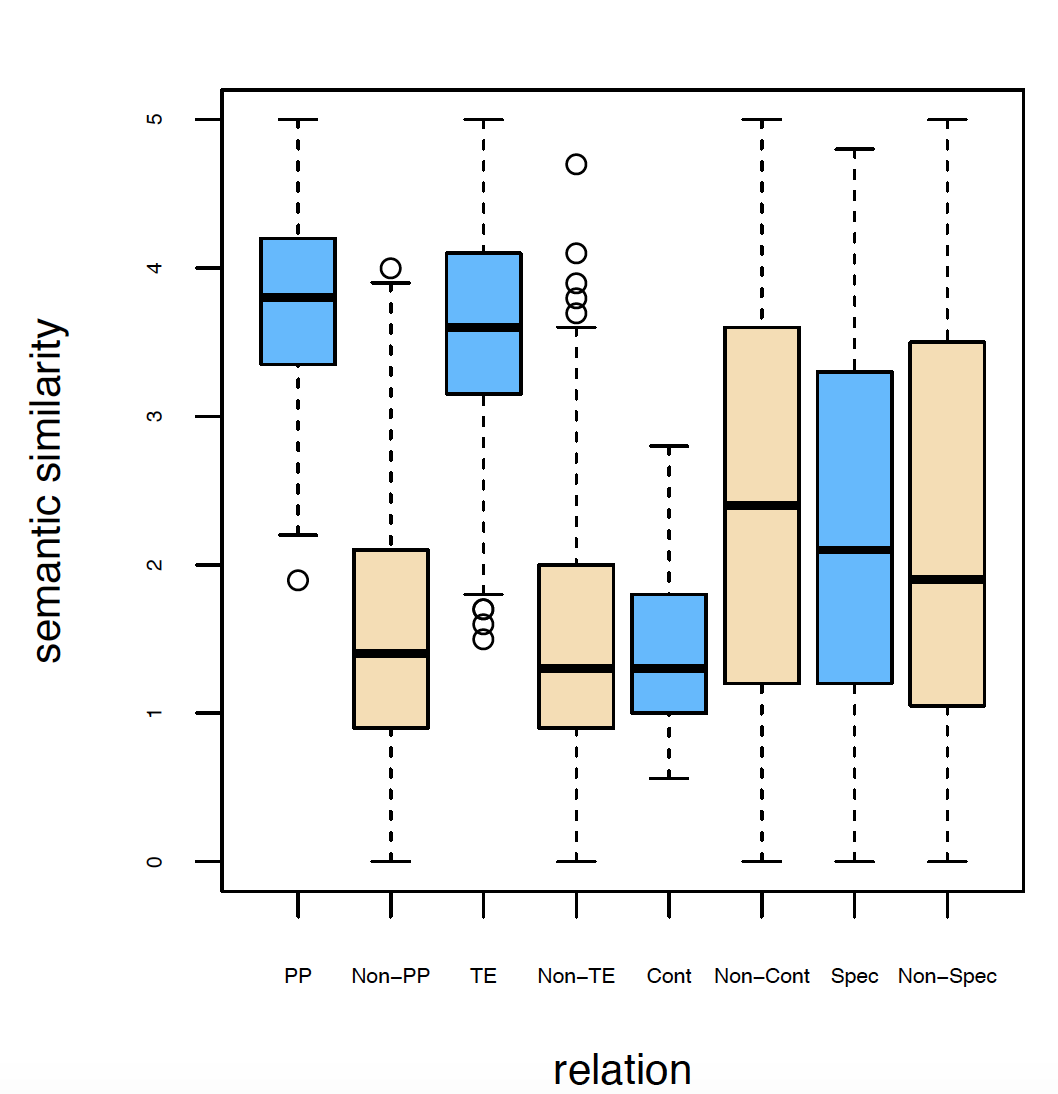}
	\caption{Similarity scores of sentences annotated with different relations}
	\label{6:fig:similarity_vs_binary}
	\end{center}
\end{figure}

We look at the average similarity for each relation (see Table~\ref{6:tab:correlation_rest}) and show boxplots between relation labels and similarity ratings (see Figure~\ref{6:fig:similarity_vs_binary}).
Table~\ref{6:tab:correlation_rest} shows that bi-directional entailment has the highest average similarity, followed by paraphrasing, while contradiction has the lowest.

Figure~\ref{6:fig:similarity_vs_binary} shows plots of the semantic similarity for all pairs where each relation is present and all pairs where it is absent.
The paraphrase pairs have much higher similarity scores than the non-paraphrase pairs.
The same observation can be made for entailment. The contradiction pairs have a low similarity score, whereas the non-contradiction pairs do not have a clear tendency with respect to similarity score.
In contrast to the other relations, pairs with and without specificity do not have any consistent similarity score.

\subsection{Overlap of Relation Labels}

Table~\ref{6:tab:rel_in_rel} shows the overlap between the different binary labels. Unlike Pearson correlation, the overlap is asymmetric - the \% of paraphrases that are also entailment (UTE in PP) is different from the \% of entailment pairs that are also paraphrases (PP in UTE).
Using the overlap measure, we can identify interesting interactions between phenomena and take a closer look at some examples.

\begin{table}[h]
	\small
	\centering
	\begin{tabular}{l|rrrrr}
		& PP    & UTE & BiTE  & Contra & Spec  \\
		\toprule
		In PP     &    & 28 \%    & 64 \%   & 0      & 73 \%   \\
		In UTE    & 52 \% &       & -    & 0      & 73 \%   \\
		In BiTE    & 94 \% & -     &      & 0      & 72 \%   \\
		In Contra & 0  & 0     & 0    &        & 63 \%   \\
		In Spec   & 30 \% & 17 \%    & 21 \%   & 11 \%     & \\
		\bottomrule
	\end{tabular}
	\caption{Distribution of overlap within relations}
	\label{6:tab:rel_in_rel}
\end{table}

\subsubsection{Entailment and Paraphrasing Overlap}
In a more theoretical setting, bi-directional entailment is often defined as being paraphrases \citep{madnani2010generating,androutsopoulos2010survey,sukhareva2016crowdsourcing}.
This implies that paraphrases equal bi-directional entailment. In our corpus, we can see that only 64\% of the paraphrases are also annotated as bi-directional entailment. An example of a pair that is annotated both as paraphrase and as bi-directional entailment is pair \#10 in Table~\ref{6:tab:ex_anno}. 
However, in the corpus we also found that 28 \% of the paraphrases are only uni-directional entailment, while in 8\% annotators did not find any entailment.
An example of a pair where our annotators found paraphrasing, but not entailment is sentence pair \#5 in Table~\ref{6:tab:ex_anno}. The agreement on the paraphrasing for this pair was 80\%, the agreement on (lack of) forward and backward entailment was 80\% and 70\% respectively.
Although the information in both sentences is nearly identical, there is no entailment, as ``having a higher chance of getting smth'' does not entail ``getting smth'' and vice versa.

\begin{table}[h]
	\scriptsize
	\centering
	\begin{tabularx}{\textwidth}{l|X|X||lllllll}
		\# & \makecell[c]{Sentence 1}  & \makecell[c]{Sentence 2} & PP & FTE & BTE & Cont & FSpec & BSpec & Sim \\ 
		\toprule
		1 & The importance of technology in modern education is overrated.	& Technology is not mandatory to improve education & \cmark & \cmark & \cmark & & & & 2.8 \\ \hline
		2 & Machines cannot interact with humans. &	No machine can communicate with a person. & \cmark & \cmark & \cmark & & & & 4.9 \\ \hline
		3 & The modern assistants make finding data slower.	& Today's information flow is greatly facilitated by digital assistants. & & & & \cmark & &\cmark & 1.9 \\ \hline
		4  & The bible is in Hebrew. & Bible is not in Latin. &   & \cmark       &        &     & \cmark        &         & 2.7 \\ \hline
		5  &  All around the world, girls have higher chance of getting a good school education. & Girls get a good school education everywhere. & \cmark  &       &         &      &          & \cmark        & 4.7 \\ \hline
		6  & Reading the Bible requires studying Latin.  & The Bible is written in Latin.                &    & \cmark       & \cmark       &      &          & \cmark        & 3.6 \\ \hline
		7  & Speaking more than one language can be useful.                                        & Languages are beneficial in life.             & \cmark  & \cmark       & \cmark       &      &          & \cmark        & 4.4 \\ \hline
		8 & You can find a good job if you only speak one language. & People who speak more than one language could only land pretty bad jobs. &   &   & \cmark &  &  &   &  2.3 \\ \hline
		9  & All Christian priests need to study Persian, as the Bible is written in Ancient Greek. & Christian clergymen don't read the bible.    &    &         &         &      &          & \cmark        & 0.9 \\ \hline
		10 & School makes students antisocial. & School usually prevents children from socializing properly. & \cmark & \cmark & \cmark & & & \cmark & 3.9 \\ 
		\bottomrule
	\end{tabularx}
	\caption{Annotations of sentence pairs on all meaning relations taken from our corpus}
	\label{6:tab:ex_anno}
\end{table}

If we look at the opposite direction of the overlap, we can see that 52\% of the uni-directional and 94\% of the bi-directional entailment pairs are also paraphrases. 
This finding confirms the statement that bi-directional entailment is paraphrasing (but not vice versa).

There is also a small portion (6\%) of bi-directional entailments that were not annotated as paraphrases.
An example of this is pair \#6 in Table~\ref{6:tab:ex_anno}.
Although both sentences make each other true, they do not have the same content.

Neither paraphrasing nor entailment had any overlap with contradiction, which further verifies our annotation scheme and quality.

These findings are partly due to the more ``relaxed'' definition of paraphrasing adopted here. 
Our definition is consistent with other authors that work on paraphrasing and the task of paraphrase identification, so we argue that our findings are valid with respect to the practical applications of paraphrasing and entailment and their interactions.

\subsubsection{Overlap with Specificity}
Specificity has a nearly equal overlap within all the other relations. 
In the pairs annotated with paraphrase or entailment, 73\% are also annotated with specificity.
The high number of pairs that are in a paraphrase relation, but also have a difference in specificity is interesting, as it seems more natural for paraphrases to be on the same specificity level.
One example of this is pair \#7 in Table~\ref{6:tab:ex_anno}.
Although they are paraphrases (with 100\% agreement), the first one is more specific, as it 1) specifies the ability of speaking a language and 2) says ``more than one language''.

There are also 27\% of uni-directional entailment relation pairs that are not in any specificity relation.
One example of this is  pair \#8 in Table~\ref{6:tab:ex_anno}.
Although the pair contains uni-directional entailment (backward entailment), none of the sentences is more specific than the other.

If we look at the other direction of the overlap, we can observe that in 62\% of the cases involving difference in specificity, there is no uni-directional nor bi-directional entailment.
An example of such a relation pair is pair \#9 in Table~\ref{6:tab:ex_anno}.
The two sentences are on the same topic and thus can be compared on their specificity.
The first sentence is clearly more specific, as it gives information on what needs to be learned and where the Bible was written, whereas the second one just gives an information on what Christian clergymen do.
These findings indicate that entailment is not specificity.

\subsection{Discussion}
Our methodology for generating text pairs has proven successful in creating a corpus that contains all relations of interest. By selecting different sentence pairings, we have obtained a balance between the relations that best suit our needs. 

The inter-annotator agreement was good for all relations. 
The resulting corpus can be used to study individual relations and their interactions.
It should be emphasized that our findings strongly depend on our decisions concerning the annotations setup, the guidelines in particular.
When examining the interactions between the different relations, we found several interesting tendencies.

\paragraph{Findings on the Interaction between Entailment and Paraphrases}
We showed that paraphrases and any-directional entailment had a high correlation, high overlap, and a similarly high semantic similarity.
Almost all bi-directional entailment pairs are paraphrases. 
However, only 64\% of the paraphrases are bi-directional entailment, indicating that paraphrasing is the more general phenomena, at least in practical tasks.

\paragraph{Findings on Specificity}
With respect to specificity, we found that it does not correlate with other relations, showing that it is independent of those in our corpus. 
It also shows no clear trend on the similarity scale and no correlation with the difference in word length between the sentences. 
This indicates that specificity cannot be automatically predicted using the other meaning relations and requires further study. 

In the examples that we discuss, we focus on interesting cases, which are complicated and unexpected  (ex.: paraphrases that are not entailment or entailment pairs that do not differ in specificity). 
However, the full corpus also contains many conventional and non-controversial examples.

\section{Conclusion and Further Work}
\label{6:sec:conclusion}
In this paper, we made an empirical, corpus-based study on interactions between various semantic relations. 
We provided empirical evidence that supports or rejects previously hypothesized connections in practical settings. 
We release a new corpus that contains all relations of interest and the corpus creation methodology to the community. 
The corpus can be used to further study relation interactions or as a more challenging dataset for detecting the different relations automatically\footnote{The full corpus, the annotation guidelines, and the control examples can be found at \url{https://github.com/MeDarina/meaning_relations_interaction}. The annotation guidelines are also available in Appendix \ref{a_law} of the thesis.}.

Some of our most important findings are: 

\begin{itemize}
\item[1)] there is a strong correlation between paraphrasing and entailment and most paraphrases include at least uni-directional entailment; 
\item[2)] paraphrases and bi-directional entailment are not equivalent in practical settings; 
\item[3)] specificity relation does not correlate strongly with the other relations and requires further study; 
\item[4)] contradictions (in our dataset) are perceived as dis-similar.
\end{itemize}

As a future work, we plan to: 1) study the specificity relation in a different setting; 2) use a linguistic annotation to determine more fine-grained distinctions between the relations; 3) and annotate the rest of the 11,000 sentences in a semi-automated way.

\section*{Acknowledgements}
We would like to thank Tobias Horsmann and Michael Wojatzki and the anonymous reviewers for their suggestions and comments.
Furthermore, we would like to thank the sentence generators for their time and creativity.
This work has been partially funded by Deutsche Forschungsgemeinschaft within the project ASSURE.
This work has been partially funded by Spanish Ministery of Economy Project TIN2015-71147-C2-2, by the
CLiC research group (2017 SGR 341), and by the APIF grant of the second author.

\chapter{Decomposing and Comparing Meaning Relations: \\ Paraphrasing, Textual Entailment, Contradiction, and Specificity}\label{ch:sharel}

\chaptermark{SHARel}

\begin{center}

	Venelin Kovatchev\affmark[1]\affmark[2], Darina Gold\affmark[3], \\ M. Ant{\`o}nia Mart{\'i}\affmark[1]\affmark[2], Maria Salam{\'o}\affmark[1]\affmark[2], Torsten Zesch\affmark[3]\\
	\affaddr{\affmark[1]Language and Computation Center, Universitat de Barcelona, Spain}\\
	\affaddr{\affmark[2]Institute of Complex Systems, Universitat de Barcelona, Spain}\\	
	\affaddr{\affmark[3]Language Technology Lab, University of Duisburg-Essen, Germany}\\

\vspace{10mm}

Accepted for publication at \\ \textit{Proceedings of the \\ Twelfth International Conference on Language Resources and Evaluation}, 2020
\end{center}

\paragraph{Abstract}
	In this paper, we present a methodology for decomposing and comparing multiple
	meaning relations (paraphrasing, textual entailment, contradiction, and specificity).
	The methodology includes SHARel - a new typology that consists of 26 linguistic and 8 reason-based 
	categories. We use the typology to annotate a corpus of 520 sentence pairs in English
	and we demonstrate that unlike previous typologies, SHARel can be applied to all 
	relations of interest with a high inter-annotator agreement.
	We analyze and compare the frequency and distribution of the linguistic and 
	reason-based phenomena involved in paraphrasing, textual entailment, contradiction, 
	and specificity. This comparison allows for a much more in-depth analysis of the
	workings of the individual relations and the way they interact and compare with
	each other.
	We release all resources (typology, annotation guidelines, and annotated corpus) 
	to the community.

\section{Introduction}

	This paper proposes a new approach for the decomposition of textual meaning 
	relations. Instead of focusing on a single meaning relation we demonstrate that
	Paraphrasing, Textual Entailment, Contradiction, and Specificity can all be 
	decomposed to a set of simpler and easier-to-define linguistic and reason-based
	phenomena. The set of ``atomic'' phenomena is shared across all relations. 

	In this paper, we adopt the definitions of meaning relations used by \citet{gold-etal-2019-annotating}.
	\textbf{Paraphrasing} is a symmetrical relation between two differently worded texts 
	with approximately the same content (1a and 1b). 
	\textbf{Textual Entailment} is a directional relation between two texts in which 
	the information of the \textit{Premise} (2a) entails the information of the 
	\textit{Hypothesis} (2b). 
	\textbf{Contradiction} is a symmetrical relation between two texts that cannot be
	true at the same time (3a and 3b)\footnote{In the Recognizing Textual Entailment 
	(RTE) literature, contradiction is often understood as the lack of entailment. However
	we adopt a more strict definition of the phenomenon.}.
	\textbf{Specificity} is a directional relation between two texts in which one text 
	 is more precise (4a) and the other is more vague (4b).

\begin{itemize}
	\item [1]\textbf{a)} \textit{Education is equal for all children.} \\
	\textbf{b)} \textit{All children get the same education.}

	\item[2] \textbf{a)} \textit{All children get the same education.} \\
	\textbf{b)} \textit{Education exists.}

	\item[3] \textbf{a)} \textit{All children get the same education.} \\
	\textbf{b)} \textit{Some children get better education.}

	\item [4] \textbf{a)} \textit{Girls do not get good education.} \\
	\textbf{b)} \textit{Some children do not get good education.}
\end{itemize}	

	The detection, extraction, and generation of pairs of texts with a particular meaning 
	relation are popular and non-trivial tasks within Computational Linguistics (CL) and 
	Natural Language Processing (NLP). Multiple datasets exist for each of 
	these tasks \citep{mrpc,rte,sts,ppdb,snli,quora,twitt,etpc}. These tasks are also 
	related to the more general problem of Natural Language Understanding (NLU) and are 
	part of the General Language Understanding Evaluation (GLUE) benchmark \citep{wang-etal-2018-glue}.

	Recently, several researchers have argued that a single label such as ``paraphrasing'', 
	``textual entailment'', or ``similarity'' is not enough to characterize and understand the 
	meaning relation \citep{Sammons,BhagatHovy,Vila,CabrioMagnini,agirre-etal-2016-semeval-2016,Benikova,etpc}.
	These authors demonstrate that the different instances of meaning relations require
	different capabilities and linguistic knowledge. 
	For example, the pairs 5 and 6 are both examples of a ``paraphrasing''
	relation. However determining the relation in 5a--5b only requires lexical knowledge, 
	while syntactic knowledge is also needed for correctly predicting the relation in  
	6a--6b. This distinction cannot be captured by a single ``paraphrasing'' label. The lack
	of distinction between such examples can be a problem in error analysis and in
	downstream applications.

\begin{itemize}
	\item[5] \textbf{a)} \textit{Education is equal for all \underline{children}.} \\
	\textbf{b)} \textit{Education is equal for all \underline{kids}.}

	\item[6] \textbf{a)} \textit{\underline{All children receive} the same education.} \\
	\textbf{b)} \textit{The same education \underline{is provided to all children}.}
\end{itemize}	

	A richer set of labels is needed to better characterize the complexity of meaning 
	relations.  We believe that a typology of ``paraphrasing'', ``textual entailment'', and 
	``semantic similarity'' would capture the distinctions between the different instances of
	each relation. 
	\citet{kovatchev-etal-2019-qualitative} empirically demonstrate that in the case of Paraphrase Identification (PI),
	the different ``paraphrase types'' are processed in a different way 
	by automated PI systems. 


	In this paper, we demonstrate that multiple meaning relations can be decomposed using 
	a shared typology. This is the first step towards building a single framework 
	for analyzing, comparing, and evaluating multiple meaning relations.
	Such a framework has not only theoretical importance, but also clear practical 
	implications. Representing every meaning relation with the same set of linguistic and
	reason-based phenomena allows for a better understanding of the nature of the relations 
	and facilitates the transfer of knowledge (resources, features, and systems) between them.

	For the purpose of decomposing the meaning relations we propose \textbf{S}ingle 
	\textbf{H}uman-Interpretable Typology for \textbf{A}nnotating Meaning \textbf{Rel}ations 
	(SHARel). With the goal of showing the applicability of the new typology, we also perform 
	an annotation experiment using the SHARel typology. We annotate a corpus of 520 text pairs
	in English, 	containing paraphrasing, textual entailment, contradiction, and textual specificity. 
	The quality of the typology and of the annotation is evident from the high inter-annotator 
	agreement.

	Finally, we present a novel, quantitative comparison between the different meaning 
	relations in terms of the types involved in each of them.

	The rest of this article is organized as follows. Section \ref{7:mt:rl} lists the
	Related Work. Section \ref{7:mt:type} presents the typology, the objectives behind
	it and the process of selection of the types. Section \ref{7:mt:anno} describes the
	annotation process - the corpus, the annotation guidelines, and the annotation
	interface. Section \ref{7:mt:results} shows the results of the annotation. Section
	\ref{7:mt:diss} discusses the implications of the findings and the way our results
	relate to our objectives and research questions. Finally, Section \ref{7:mt:conc}
	concludes the paper and addresses the future work.

\section{Related Work}\label{7:mt:rl}

	The last several years have seen an increasing interest towards the 
	decomposition of paraphrasing \citep{BhagatHovy,Vila,Benikova,etpc}, textual 
	entailment \citep{Sammons,LoBue,CabrioMagnini}, and textual similarity 
	\citep{agirre-etal-2016-semeval-2016}. 

	\citet{Sammons} argue that in order to process a complex meaning relation 
	such as textual entailment a competent speaker has to take several ``inference 
	steps''. This means that a meta-relation such as paraphrasing, textual entailment,
	or semantic similarity can be ``decomposed'' or broken down into such ``inference 
	steps''. These ``inference steps'', traditionally called ``types'' can be either 
	linguistic or reason-based in their nature. The linguistic types require certain 
	linguistic capabilities from the speaker, while the reason-based types require
	common-sense reasoning and world knowledge.

	The different authors working on decomposing meaning relations all follow a similar 
	approach. 
	First, they propose a typology - a set of ``atomic'' linguistic and/or reasoning  types 
	involved in the inference process of the particular meta-relation (paraphrasing, 
	entailment, or similarity), Then, they use the ``atomic'' types in a corpus annotation 
	and finally, they analyze the distribution and correlation of the types. The corpus
	based studies have demonstrated that different atomic types can be found in various 
	corpora for paraphrasing, textual entailment, and semantic similarity research. 

	\citet{kovatchev-etal-2019-qualitative} empirically demonstrated that the performance of a 
	Paraphrase Identification (PI) system on each candidate-paraphrase pair depends on 
	the ``atomic types'' involved in that pair. That is, they showed that state-of-the-art 
	automatic PI systems process ``atomic paraphrases'' in a different manner and
	with a statistically significant difference in quantitative performance (Accuracy and F1). They show that more
	frequent and relatively simple types like ``lexical substitution'', ``punctuation changes''
	and ``modal verb changes'' are easier
	across multiple automated PI systems, while other types like ``negation switching'',
	``ellipsis'' and ``named entity reasoning'' are much more challenging. 

	Similar observations have been made in the field of Textual Entailment. 
	\citet{gururangan-etal-2018-annotation} discovered the presence of annotation 
	artifacts that enable models that take into account only one of the texts (the hypothesis) 
	to achieve performance substantially higher than the majority baselines in SNLI and 
	MNLI. \citet{glockner-etal-2018-breaking} showed that models trained with SNLI fail 
	to resolve new pairs that require simple lexical substitution. \citet{naik-etal-2018-stress} 
	create label-preserving adversarial examples and conclude that automated NLI models 
	are not robust. \citet{wallace-etal-2019-universal} introduce universal triggers, that is, 
	sequences of tokens that fool models when concatenated to any input. All these 
	authors identify different problems and biases in the datasets and the systems trained
	on them. However they focus on a single phenomenon and/or a specific linguistic 
	construction. A typology-based approach can evaluate the performance and robustness 
	of automated systems on a large variety of tasks. 
	
	One limitation of the different decompositional approaches is that there exist many
	different typologies and each typology is created considering only one meaning relation 
	(paraphrasing, textual entailment, textual similarity). This follows the traditional 
	approach in the research on meaning relations: each relation is studied in isolation, 
	with its own theoretical concepts, datasets, and practical tasks. 

	In recent years, the "single relation" approach has been questioned by several
	authors. \citet{androutsopoulos2010survey} analyze the relations between
	paraphrasing and textual entailment. \citet{marelli2014sick} present SICK: a corpus
	that studies entailment, contradiction, and semantic similarity. \citet{lan} and
	\citet{hanan} explore the transfer learning capabilities between paraphrasing
	and textual entailment. \citet{gold-etal-2019-annotating} present a corpus that is annotated for
	paraphrasing, textual entailment, contradiction, specificity, and textual similarity.
	These works demonstrate that the different meaning relations can be studied
	together and can benefit from one another.

	However, to date, the joint research of meaning relations is limited only to the
	binary textual labels. There has been no work on comparing the different typologies
	and the way different relations can be decomposed. None of the existing typologies
	is fully compatible with multiple meaning relations, which further restricts the research
	in this area. We aim to address this research gap in this paper.
	




\section{Shared Typology for Meaning Relations} \label{7:mt:type}

	This section is organized as follows.
	Section \ref{7:mt:obj} presents the problem of decomposing meaning relations.
	Section \ref{7:mt:typ} describes our proposed typology and the rationale behind it. 
	Section \ref{7:mt:rq} formulates our research questions.

	\subsection{Decomposing Meaning Relations}\label{7:mt:obj}

	The goal behind the \textbf{S}ingle \textbf{H}uman-Interpretable Typology for 
	\textbf{A}nnotating Meaning \textbf{Rel}ations (SHARel) is to come up with a unified 
	list of linguistic and reason-based phenomena that are required in order to 
	determine the meaning relations that hold between two texts. The list of types should 
	not be limited to texts that hold a specific single textual relation, such as paraphrasing, 
	textual entailment, contradiction, and textual specificity. Rather, the types should be 
	applicable to texts holding multiple different relations.

	\begin{itemize}
		\item[7]\textbf{a} All \underline{children} \textit{receive} the same education. \\
		\textbf{b} The same education \textit{is received} by all \underline{kids}.
		\item[8]\textbf{a} All \underline{children} \textit{receive} the same education. \\
		\textbf{b} The same education \textit{is} \textbf{not} \textit{received} by all \underline{kids}.
	\end{itemize}

	In 7a and 7b, the meaning relation at a textual level is paraphrasing, while
	in 8a and 8b, the textual relation is contradiction.
	In order to determine the meaning relation for both 7 and 8, a competent speaker or 
	an automated system needs to make several inference steps. First, they have to
	determine that ``kids'' and ``children'' have the same 
	meaning and the same syntactic and semantic role in the texts.
	Second, they need to account for the change in grammatical voice. 
	In terms of typology, these inference steps involve two different
	types - ``same polarity substitution'' ( ``kids'' - ``children'') and ``diathesis 
	alternation'' (``receive'' - ``is received'').
	In addition, in example 8b, the human or the automated system needs to determine the presence
	and the function of ``negation'' (\textbf{not}).

	By successfully performing all necessary inference steps, the human (or the automated system)
	is able to determine that in the pair 7a-7b there is equivalence of the expressed meaning,
	while in the pair 8a-8b there is a logical contradiction. The required inference steps in
	the two examples are not specific to the textual label (paraphrasing or contradiction).
	The ``types'' are general linguistic or reason-based phenomena. 



	With the goal of addressing such situations, we propose a list of types that, 
	following the existing theoretical research, can be applied to multiple meaning 
	relations. We justify the choice of types for SHARel in the context of existing 
	typologies. 

	\subsection{The SHARel Typology}\label{7:mt:typ}

	Table \ref{7:l-types} shows the SHARel Typology and its 34 different types, organized
	in 8 categories. The first 6 categories (morphology, lexicon, lexico-syntactic, syntax, 

		\begin{table} [H]
		\footnotesize
		\begin{center}

		\begin{tabular}{| l | l |}
		\hline
		\textbf{ID} & \textbf{Type} \\
		\hline \hline
		\multicolumn{2}{|c|}{Morphology-based changes} \\
		\hline \hline
		1 & Inflectional changes \\
		\hline
		2 & Modal verb changes \\
		\hline
		3 & Derivational changes \\
		\hline \hline
		\multicolumn{2}{|c|}{Lexicon-based changes} \\
		\hline \hline
		4 & Spelling changes \\
		\hline
		5 & Same polarity substitution (habitual) \\
		\hline
		6 & Same polarity substitution (contextual) \\
		\hline
		7 & Same polarity sub. (named entity) \\
		\hline
		8 & Change of format \\
		\hline \hline
		\multicolumn{2}{|c|}{Lexico-syntactic based changes} \\
		\hline \hline
		9 & Opposite polarity sub. (habitual) \\
		\hline
		10 & Opposite polarity sub. (contextual) \\
		\hline
		11 & Synthetic/analytic substitution \\
		\hline
		12 & Converse substitution \\
		\hline \hline
		\multicolumn{2}{|c|}{Syntax-based changes} \\
		\hline \hline
		13 & Diathesis alternation \\
		\hline
		14 & Negation switching \\
		\hline
		15 & Ellipsis \\
		\hline
		16 & Anaphora \\
		\hline
		17 & Coordination changes  \\
		\hline
		18 & Subordination and nesting changes  \\
		\hline \hline
		\multicolumn{2}{|c|}{Discourse-based changes} \\
		\hline \hline
		18 & Punctuation changes \\
		\hline
		20 & Direct/indirect style alternations \\
		\hline
		21 & Sentence modality changes \\
		\hline
		22 & Syntax/discourse structure changes \\
		\hline \hline
		\multicolumn{2}{|c|}{Other changes} \\
		\hline \hline
		23 & Addition/Deletion \\
		\hline
		24 & Change of order \\
		\hline \hline
		\multicolumn{2}{|c|}{Extremes} \\
		\hline \hline
		25 & Identity \\
		\hline
		26 & Unrelated \\
		\hline \hline
		\multicolumn{2}{|c|}{Reason-based changes} \\
		\hline \hline
		27 & Cause and Effect \\
		\hline
		28 & Conditions and Properties \\
		\hline
		29 & Functionality and Mutual Exclusivity \\
		\hline
		30 & Named Entity Reasoning \\
		\hline
		31 & Numerical Reasoning \\
		\hline
		32 & Temporal and Spatial Reasoning \\
		\hline
		33 & Transitivity \\
		\hline
		34 & Other (General Inference)\\
		\hline		
		\end{tabular}

		\end{center}
		\caption{The SHARel Typology} \label{7:l-types}
		\end{table}

	discourse, other) consist of the 24 ``linguistic'' types. The two types in the ``extremes'' 
	category (``identity'' and ``unrelated'') are neither linguistic, nor reason-based. The last 
	category consists of the 8 ``reason-based'' types.

	The distinction between linguistic and reason-based types is introduced by 
	\citet{Sammons} and \citet{CabrioMagnini} for textual entailment. The linguistic
	phenomena require certain linguistic capabilities from the human speaker or the 
	automated system. The reason-based phenomena require world knowledge and
	common-sense reasoning. 

	For the linguistic types, we compared the existing typologies and decided to use the
	Extended Paraphrase Typology (EPT) \citep{etpc} as a starting point. The authors of EPT 
	have already combined 
	various linguistic types from the fields of Paraphrasing and Textual Entailment and have
	taken into account the work of \citet{Sammons}, \citet{Vila}, \citet{CabrioMagnini}. 
	As such, the majority of the linguistic types that they propose are in principle applicable 
	to both Paraphrasing and Textual Entailment. 

	We examined the types from EPT and made several adjustments in order to make the 
	linguistic types fully independent of the textual relation.
	\begin{itemize}
	\item EPT contains ``entailment'' and ``non-paraphrase'' types  in the category 
	``extremes''. These types were created specifically for the task of Paraphrase 
	Identification (PI). We removed these types from the list.
	\item We added ``unrelated'' type (\#26) to the category ``extremes'' to capture
	information which is not related at all to the other sentence in the pair.
	\item We added ``anaphora'' type (\#16) in the syntax category. This change was 
	suggested 	by our annotators during the process of corpus annotation.
	\end{itemize}


	For the reason-based types we studied the typologies of \citet{Sammons}, 
	\citet{LoBue} and \citet{CabrioMagnini}. While these typologies have a lot
	of similarities and shared types, they are not fully compatible. We analyzed the
	type of common-sense reasoning and background knowledge that is required for each of the
	types in these three typologies. We combined similar types into more general
	types and reduced the original list of over 30 reason-based types to 8. For example,
	the ``named entity reasoning'' (\#30) includes both reasoning about geographical
	entities and publicly known persons (those two were originally separated types).
	\footnote{The annotation guidelines and examples for all types can be seen at
	\url{https://github.com/venelink/sharel} and in Appendix \ref{a_sharel} of the thesis.}

	With respect to specificity, we propose a fine-grained token level annotation, which
	allows us to determine the particular 
	elements in one sentence that are more (or less) specific than their 
	counterpart in the other sentence. 
	\citet{ko2019domain} demonstrated that specificity needs to be more linguistically and 
	informational theoretically based to be more semantically plausible. This could partially
	be solved through a more fine-grained annotation of specificity, as it is performed 
	in this study.

\begin{table}[H]
\begin{center}
\begin{tabular}{l l l l l l }
\hline
\textbf{Typology} & \textbf{Relation}  & \textbf{All} & \textbf{Ling.}  &  \textbf{Reason.} & \textbf{Hierarchy}  \\ \hline \hline
\citet{Sammons} & TE, CNT  & 22 & 13 & 9 & No \\ \hline
\citet{LoBue} & TE, CNT & 20 & 0 & 20 & No \\ \hline
\citet{CabrioMagnini} & TE, CNT & 36 & 24 & 12 & Yes \\ \hline
\citet{BhagatHovy} & PP & 25 & 22 & 3 & No \\ \hline
\citet{Vila} & PP & 23 & 19 & 1 & Yes  \\ \hline
\citet{etpc} & PP & 27 & 23 & 1 & Yes\\ \hline
\textit{SHARel} & \shortstack{TE, CNT \\ PP, SP, TS} & 34 & 24 & 8 & Yes \\ \hline
\end{tabular}
\caption{Comparing typologies of textual meaning relations}
\label{7:l-cmp-types}
\end{center}
\end{table}

	Table \ref{7:l-cmp-types} lists some properties of the existing typologies of meaning relations.
	All typologies before SHARel were created only for one (or two) meaning relations.
	SHARel contains general types that are not specific to any particular meaning relation
	and can be applied to pairs holding Textual Entailment, Contradiction, Paraphrasing,
	Textual Specificity, or Semantic Textual Similarity meaning relation. SHARel follows
	the good practices of typology research and organizes the types in a hierarchical
	structure of 8 categories and has a good balance between linguistic and reasoning
	types.


%



\subsection{Research Questions}\label{7:mt:rq}

	There are two main objectives that motivated this paper: \\1) To demonstrate that 
	multiple meaning relations can be decomposed using a single, shared typology; 
	\\ 2) To demonstrate some of the advantages of a shared typology of meaning relations.
	\\ Based on our objectives, we pose two research questions (RQs) that we want to address in 
	this article.

	\begin{itemize}
		\item[] \textbf{RQ1:} Is it possible to use a single typology for the decomposition of multiple
		(textual) meaning relations?
		\item[] \textbf{RQ2:} What are the similarities and the differences between the (textual) 
		meaning relations in terms of types? 
	\end{itemize}

	We address these research questions in a corpus annotation study. For the first
	research question we evaluate the quality of the corpus annotation by measuring the 
	inter-annotator agreement. For the second research question we measure the relative
	frequencies of the types in sentence pairs with each textual meaning relation.

\section{Corpus Annotation}\label{7:mt:anno}

	This section is organized as follows:
	Section \ref{7:mt:rcorp} describes the corpus that we chose to use in the 
	annotation.
	Section \ref{7:mt:corp} presents the annotation setup. 	
	Finally, in Section \ref{7:mt:agr} we report the annotation agreement. 

	\subsection{Choice of Corpus}\label{7:mt:rcorp}

	In order to determine the applicability of SHARel to all relations of interest, 
	we carried out a corpus annotation. We used the publicly available corpus 
	of \citet{gold-etal-2019-annotating}. It consists of 520 text pairs and is already annotated at 
	sentence level for paraphrasing, entailment, contradiction, specificity and 
	semantic similarity. \citet{gold-etal-2019-annotating} performed the annotation for each relation 
	independently. That is, for each pair of sentences 10 annotators were asked 
	whether a particular relation (paraphrasing, entailment, contradiction, 
	specificity) held or not.

	The corpus of \citet{gold-etal-2019-annotating} contains 160 pairs annotated as paraphrases, 
	195 pairs annotated as textual entailment (in one direction or in both) and 
	68 pairs annotated as contradiction. As the annotation for the different 
	relations was carried out independently, there is an overlap between the 
	relations. For example 52\% of the pairs annotated as 
	entailment were also annotated as paraphrases. The total number of pairs
	annotated with at least one relation among paraphrasing, entailment, and
	contradiction is 256. The remaining 244 pairs were annotated as unrelated.
	In 381 of the pairs, one of the sentences was marked as more specific
	than the other. 

	The corpus of \citet{gold-etal-2019-annotating} is the only corpus to date which contains
	all relations of interest. All text pairs are in 
	the same domain and topic, they have similar syntactic structure and 
	vocabulary. The lexical overlap between the two sentences in each pair
	is much lower than in corpora such as MRPC \citep{mrpc} or SNLI
	\citep{snli}. This means that even though the two sentences in a pair 
	are in a meaning relation such as paraphrasing or textual entailment, there 
	are very few words that are directly repeated. All these properties of the
	corpus were taken into consideration when we chose it for our annotation.

	\subsection{Annotation Setup}\label{7:mt:corp}



	We performed an annotation with the SHARel 
	typology on all pairs from \citet{gold-etal-2019-annotating} that have at least one of the 
	following relations: paraphrasing, forward entailment, backwards 
	entailment, and  contradiction. We discarded pairs that are annotated as 
	"unrelated". 
	This is a typical approach when decomposing meaning relations. 
	\citet{Sammons,CabrioMagnini,Vila} only decompose pairs
	with a particular relation (entailment, contradiction, or paraphrasing). 

	After discarding the unrelated portion, the total number of pairs that we 
	annotated with SHARel was 276.
	Prior to the annotation we tokenized each sentence using the NLTK 
	python library. 

	During the annotation process, our annotators go through each pair
	in the corpus. For each linguistic and reason-based phenomenon that
	they encounter, they annotate the type and the scope (the specific 
	tokens affected by the type). We used an open source web-based 
	annotation interface, called WARP-Text \citep{kovatchev-etal-2018-warp}.

	We prepared extended guidelines with examples for each type.
	Each pair of texts was annotated independently by two trained expert 
	annotators. In the cases where there were disagreements, the annotators
	discussed their differences in order to obtain the best possible 
	annotation for the example pair \footnote{The annotation guidelines 
	and the annotated corpus are available at \url{https://github.com/venelink/sharel}}.

	\subsection{Agreement}\label{7:mt:agr}	

	For calculating inter-annotator agreement, we use the two different versions 
	of the IAPTA-TPO measures. The IAPTA-TPO measures was proposed by 
	\citet{Vila2015} specifically for the task of annotating paraphrase types. They 
	were later on refined by \citet{etpc}. IAPTA-TPO measure the agreement 
	on both the label (the annotated phenomenon) and the scope, which is non-trivial
	to capture using traditional measures such as Kappa. IAPTA-TPO (Total) measures 
	the cases where the annotators fully agree on both label and scope. IAPTA-TPO 
	(Partial) measures the cases where the annotators agree on the label, but the
	scope overlaps only partially.

	The agreement of our annotation can be seen in Table \ref{7:npagr}. We 
	calculate the agreement on all pairs (all), and we also report the agreement for 
	the pairs with textual label paraphrases (pp), entailment (ent), and 
	contradiction (cnt). 

		\begin{table}[H]
			\begin{center}
			\begin{tabular}{|c|c|c|}
			\hline
			& \textbf{TPO-Partial} & \textbf{TPO-Total} \\
			\hline \hline
			This corpus (all) & .78 & .52 \\
			\hline
			This corpus (pp) & .77 & .51 \\
			\hline
			This corpus (ent) & .77 & .52 \\
			\hline
			This corpus (cnt) & .75 & .50 \\
			\hline
			MRPC-A & .78 & .51 \\
			\hline
			ETPC (non-pp) & .72 & .68 \\
			\hline
			ETPC (pp) & .86 & .68 \\
			\hline

			\end{tabular}
			\end{center}
			\caption{Inter-annotator Agreement}
			\label{7:npagr}
		\end{table}	

	To put our results in perspective, we compare our agreement with the
	one reported in MRPC-A \citep{Vila2015} and ETPC \citep{etpc}. For ETPC the authors
	report both the agreement on the pairs annotated as paraphrases (pp) and
	as non-paraphrases (non-pp). To date, MRPC-A and ETPC 
	are the only two corpora of sufficient size annotated with a typology of
	meaning relations. They also use the same inter-annotation measure to report
	agreement, so we can compare with them directly.

	The overall agreement that we obtain (.52 Total and .78 Partial) is almost
	identical to the agreement reported for MRPC-A (.51 Total and .78 Partial)
	and slightly lower than the agreement reported for ETPC (.68 Total and .86
	Partial).

	\citet{etpc} detected a significant difference in the agreement between
	paraphrase and non-paraphrase pairs. In their annotation, the 
	``non-paraphrase'' includes mostly entailment and contradiction pairs and
	the lower agreement indicates that their typology is not well equipped for
	dealing with those cases. However in our corpus, we don't observe such a difference. 
	Our annotation agreement is very consistent across all pairs indicating
	that SHARel is successfully applied to all relations of interest.

	The consistently high agreement score indicates the high quality of the 
	annotation. Even though our task and our typology are much more complex
	than those of \citet{Vila} and \citet{etpc}, we still obtain comparable
	results.

	In addition to calculating the inter-annotation agreement, we also
	asked the annotators to mark and indicate any examples and/or
	phenomena not covered by the typology. Based on their ongoing feedback 
	during the 	annotation, we decided to introduce the ``anaphora'' type. 
	We re-annotated the portion of the corpus that was already annotated at the
	time when we introduced the new type.

	Arriving at this point, we have demonstrated that it is possible to 
	successfully use a single typology for the decomposition of multiple (textual)
	meaning relations. This answers our first research question (\textbf{RQ1}).

\section{Analysis of the Results}\label{7:mt:results}	

	Before this paper, the comparison between textual 
	meaning relations was limited to measuring the overlap and correlation
	between the binary label of the pairs. \citet{gold-etal-2019-annotating} present such an analysis.
	They find some expected results such as the high correlation and overlap between
	paraphrasing and (uni-directional) entailment and the negative correlation
	between paraphrasing and contradiction or entailment and contradiction. They
	also report some interesting and unexpected results. They point that in practical setting
	paraphrasing does not equal bi-directional entailment. With respect to specificity
	they find that it does not correlate with other textual meaning relations, and does 
	not overlap with textual entailment.

	In this section, we go further than the binary labels of the textual meaning
	relations and compare the distribution of types across all relations. A 
	typological comparison can be much more informative about the interactions
	between the different relations.

	This section is organized as follows. Section	\ref{7:mt:freq} analyzes and 
	compares the frequency distribution of the different types in pairs 
	with the following textual relations: Paraphrasing, Textual Entailment,
	and Contradiction. 
	Section \ref{7:mt:spec} discusses the Specificity relation and
	the types involved in it. 

	\subsection{Type Frequency}\label{7:mt:freq}

	To determine the similarities and differences between the textual
	meaning relations in terms of types, we measured the relative type 
	frequencies for pairs that have the corresponding label.
	Table \ref{7:t-freq} shows the relative frequencies in pairs that have 
	paraphrasing, entailment, or contradiction relations at textual level. 
	For the entailment relation we consider only the 
	pairs marked as ``uni-directional entailment''. That is, pairs that have 
	entailment only in one of the directions. We discard the pairs that have
	bi-directional entailment to reduce the overlap with paraphrases
	(94 \% of the bi-directional entailment 	pairs are also paraphrases).

	For reference, we have also included the type frequencies for the 
	paraphrase portion of the ETPC \citep{etpc} corpus. ETPC is the largest
	corpus to date annotated with paraphrase types. The EPT typology
	used to annotate the ETPC also shares the majority of the
	linguistic types with SHARel. This allows us to put our results in perspective and to
	determine to what extent are they consistent with previous findings.

		\begin{table}[H]
		\footnotesize
		\begin{center}

		\begin{tabular}{| l | l | r | r | r | r |}
		\hline
		\textbf{ID} & \textbf{Type} & \textbf{Paraph.} & \textbf{Entailment} & \textbf{Contradiction} & \textbf{ETPC} \\
		\hline \hline
		\multicolumn{6}{|c|}{Morphology-based changes} \\
		\hline \hline
		1 & Inflectional changes 				& 4 \%	& 4 \%	& 1.9 \%	& 2.78 \%\\
		\hline
		2 & Modal verb changes 				& 0.25 \% & 1 \%	& 0 		& 0.83 \%\\
		\hline
		3 & Derivational changes 				& 2 \%	& 0 		& 0.6 \%	& 0.85 \%\\
		\hline \hline
		\multicolumn{6}{|c|}{Lexicon-based changes} \\
		\hline \hline
		4 & Spelling changes 					& 0.25 \%	& 0.4 \%	& 0 		& 2.91 \%\\
		\hline
		5 & Same pol. sub. (habitual)			& 25.2 \% & 17 \%	& 26 \%	& 8.68 \% \\
		\hline
		6 & Same pol. sub. (contextual) 			& 9.7 \%	& 6.3 \%	& 5.5 \%	& 11.66 \%\\
		\hline
		7 & Same pol. sub. (named ent.) 		& 0.7 \%	& 0.4 \%	& 1.2 \%	& 5.08 \%\\
		\hline
		8 & Change of format 				& 0.7 \%	& 0.9 \%	& 0 & 1.1 \%\\
		\hline \hline
		\multicolumn{6}{|c|}{Lexico-syntactic based changes} \\
		\hline \hline
		9 & Opposite pol. sub. (habitual) 			& 2.7 \%	& 3.5 \%	& 7.5\%	& 0.07 \% \\
		\hline
		10 & Opposite pol. sub. (context.) 		& 0.5 \%	& 0.9 \%	& 1.2 \%	& 0.02 \%\\
		\hline
		11 & Synthetic/analytic sub.				& 6.7 \%	& 6.8 \%	& 3.7 \%	& 3.80 \%\\
		\hline
		12 & Converse substitution 				& 2.5 \%	& 3.2 \%	& 3.1 \%	& 0.20 \%\\
		\hline \hline
		\multicolumn{6}{|c|}{Syntax-based changes} \\
		\hline \hline
		13 & Diathesis alternation 				& 1.5 \%	& 2.2\%	& 1.9 \%	& 0.73 \%\\
		\hline
		14 & Negation switching 				& 4 \%	& 4 \%	& 11.2 \%	& 0.09 \%\\
		\hline
		15 & Ellipsis 						& 0  & 0 & 0 & 0.30 \%\\
		\hline
		16 & Anaphora 						& 1.7 \%	& 2.7 \%	& 0.6 \%	& 0\\
		\hline
		17 & Coordination changes  			& 0 & 0 & 0 & 0.22 \%\\
		\hline
		18 & Subordination and nesting 	 	 	& 0.25 \%	& 0 & 0 & 2.14 \%\\
		\hline \hline
		\multicolumn{6}{|c|}{Discourse-based changes} \\
		\hline \hline
		18 & Punctuation changes 				& 0 & 0 & 0 & 3.77 \%\\
		\hline
		20 & Direct/indirect style altern.	 		& 0 & 0 & 0 & 0.30 \%\\
		\hline
		21 & Sentence modality changes 			& 0 & 0 & 0 & 0\\
		\hline
		22 & Syntax/discourse structure		 	& 0 & 0 & 0 & 1.39 \%\\
		\hline \hline
		\multicolumn{6}{|c|}{Other changes} \\
		\hline \hline
		23 & Addition/Deletion 				& 16.25 \%& 16.4 \%	& 16.2 \%	& 25.94 \%\\
		\hline
		24 & Change of order 				& 0.5 \%	& 0.9 \%	& 0.6 \%	& 3.89 \%\\
		\hline \hline
		\multicolumn{6}{|c|}{Extremes} \\
		\hline \hline
		25 & Identity 						& 12.5 \%	& 14.5 \%	& 11.8 \%	& 17.5 \% \\
		\hline
		26 & Unrelated 						& 0	& 0 & 0 & 3.81 \%\\
		\hline \hline
		\multicolumn{2}{|c|}{Reasoning} \\
		\hline \hline
		27 & Cause and Effect 				& 4.7 \%	& 5.4 \%	& 5 \%	& n/a \\
		\hline
		28 & Conditions and Properties 			& 2 \%	& 6 \%	& 0.6 \%	& n/a \\
		\hline
		29 & Funct. and Mutual Exclus.	 		& 0 & 0.4 \% & 0 & n/a \\
		\hline
		30 & Named Ent. Reasoning 			& 0 & 0 & 0 & n/a \\
		\hline
		31 & Numerical Reasoning 				& 0 & 0 & 0 & n/a \\
		\hline
		32 & Temp. and Spatial Reasoning 		& 0 & 0 & 0 & n/a \\
		\hline
		33 & Transitivity 					& 0.25 \% & 0.9 \% & 0 & n/a \\
		\hline
		34 & Other (General Inference)			& 0.5 \%	& 0.4 \% & 0.6 \% & 1.53 \%\\
		\hline
		\end{tabular}

		\end{center}
		\caption{Type Frequencies} \label{7:t-freq}
		\end{table}

	We can observe that the distribution of types is not balanced for any
	of the portions. Some types are over-represented, while others are
	under-represented or not represented at all. We focus our analysis
	on four different tendencies: 1) linguistic types that are frequent 
	across all relations; 2) types whose frequency changes across the 
	different relations; 3) the frequency of reason-based types;
	and 4) types that are infrequent or not represented at all.


	\paragraph{Frequent linguistic types across all relations}

	\paragraph{} The most frequent types across all relations are 
	\textit{same polarity substitution (habitual)} (\#5), \textit{same polarity 
	substitution (contextual)} (\#6), \textit{same polarity substitution 
	(named entity)} (\#7), \textit{addition/deletion} (\#23), and 
	\textit{identity} (\#25). These phenomena account for more than 50\% 
	of the types in the corpus. This finding is also consistent with the results 
	reported in the ETPC. It is worth noting that in the ETPC, the distribution 
	within the different \textit{same polarity substitution} types (\#5, \#6, \#7)
	 differs from our annotation. The frequency of \textit{same polarity 
	substitution (habitual)} (\#5) is lower, while \textit{same polarity 
	substitution (contextual)} (\#6) and \textit{same polarity substitution 
	(named entity)} (\#7) have a much higher frequency.

	Other frequent types shared across all relations are \textit{inflectional} (\#1), 
	\textit{opposite polarity substitution (habitual)} (\#9), 
	\textit{synthetic/analytic substitution} (\#11), \textit{converse substitution} 
	(\#12), \textit{diathesis alternation} (\#13), and \textit{negation switching}
	(\#14). For all of these types, the frequency that we obtain is substantially 
	higher than in the ETPC corpus. 
	
	\paragraph{Differences in type frequencies across relations}

	\paragraph{}
	We can observe that paraphrasing has the highest frequency of 
	\textit{Same Polarity Substitution}, both habitual (\#5) and contextual (\#6).
	This is a tendency that can also be observed in ETPC.

	Entailment is the relation with the highest relative frequency of
	phenomena in the reason-based category. The reason-based 
	phenomena (\#27-\#34) account for 13.1\% of all phenomena within entailment, 
	doubling the frequency of these phenomena in paraphrasing (5.65\%) and 
	contradiction (6.2\%). Most of that difference comes
	from the "conditions/properties" (\#28) type. The entailment relation also
	has the lowest frequency of same polarity substitutions (\#5, \#6, and \#7).

	Contradiction has the highest frequency of opposite polarity
	substitution (\#9 and \#10) and negation switching (\#14), doubling 
	the frequency of these phenomena in paraphrasing and entailment
	pairs. Interestingly, contradictions have a comparable frequency
	of same polarity substitution (\#5, \#6, and \#7) and identity (\#25) 
	to paraphrases. This suggests that contradictions are more similar to 
	paraphrases than to entailment, at least in terms of the phenomena 
	involved.

	\paragraph{Frequency of reason-based types}

	\paragraph{}
	We can observe that reason-based types (\#27-\#34) are much less frequent 
	than linguistic types. Reasoning accounts for less than 14\% 
	of the examples across all relations. That means that in the majority 
	of the cases, the textual relation can be determined via linguistic 
	means and does not require reasoning or world knowledge. The most 
	frequent reasoning type across all relations is \textit{cause/effect}.

	It is important to note that the frequency of reasoning 
	phenomena in our annotation is much higher than the 1.5\% reported 
	in ETPC. In ETPC all reason based phenomena were annotated with a
	single label - \textit{Other (General Inferences)} (\#34) so the frequency
	of this type corresponds to the sum of all types from \#27 to \#34 in our
	annotation. 
	These findings indicate that the methodology of \citet{gold-etal-2019-annotating} successfully
	addresses one of the problems in the ETPC corpus, already emphasized by
	other researchers - the lack of reason-based types.

	\paragraph{Low frequency types and missing types}

	\paragraph{}
	In our annotation, there are several linguistic and reason-based types
	that are not represented at all.
	Regarding the linguistic types, there are no discourse based types, no 
	\textit{ellipsis} (\#15), no \textit{coordination changes} (\#17), and almost no 
	\textit{subordination and nesting changes} (\#18). Regarding the reason-based
	types, there are no \textit{Named Entity Reasoning} (\#30), 
	\textit{Numerical Reasoning} (\#31), and no \textit{Temporal and
	Spatial Reasoning} (\#32).

	We argue that the absence of these types in our annotation is due 
	to the way in which the \citet{gold-etal-2019-annotating} corpus was created. The authors of that 
	corpus aimed at obtaining simple, one-verb sentences. The 
	average length of a sentence is 10.5 tokens, which is much lower 
	than the length of sentences in other corpora (ex.: 22 average length
	for ETPC). The corpus contains almost no Named Entities (proper 
	names, locations, or quantities). These characteristics of the 
	corpus do not facilitate transformations at the syntactic and discourse
	levels or Named Entity Reasoning. 

	Our intuition that the lack of these types is due to the corpus 
	creation is further reinforced by the fact that these types are missing 
	across all meaning relations. However, these missing 
	types can be observed in other paraphrasing and entailment 
	corpora, such  as \citet{Sammons}, \citet{CabrioMagnini}, 
	and \citet{etpc}. For these reasons we decided to keep them as 
	part of the ShaRel typology. It would, nevertheless, 
	require a further research and richer corpora to empirically 
	determine the importance of these phenomena for the different 
	meaning relations.

	\paragraph{Summary}

	The similarities and common tendencies between paraphrases,
	entailment, and contradiction clearly indicate that these relations
	belong within the same conceptual framework and should be studied and
	compared together. The results also suggest the possibility of the
	transfer of knowledge and technologies between these relations.


	The differences between the textual meaning relations in terms
	of the involved types can help us to understand each of the individual
	relations better. This information can also be useful in the 
	automatic classification of the different relations in a practical
	task.


	\subsection{Decomposing Specificity}\label{7:mt:spec}

	We define specificity as the opposite of generality or 
	fuzziness. \citet{yager92} defines specificity as the degree to 
	which a fuzzy subset points to one element as its member. This
	meaning relation has not been studied extensively. It has also not
	been decomposed. To the best of our knowledge this is the first work
	to do so. \citet{gold-etal-2019-annotating} 
	show that there is no direct correlation between specificity and the 
	other textual meaning relations, including textual entailment. For that reason, we took a different 
	approach to the decomposition of specificity and treat it
	separately from the other relations. We added one extra step in 
	the annotation process, focused on the specificity relation. 

	The corpus of \citet{gold-etal-2019-annotating} is annotated for specificity at the textual 
	level. That is, the crowd workers identified which of the two given sentences 
	is more specific. In 9, the annotators would indicate that
	\textbf{b} is more specific than \textbf{a}.

	\begin{itemize}
		\item[9]\textbf{a} All children receive the same education. \\
		\textbf{b} The same education is received by all girls.
	\end{itemize}

	In our annotation, we 	performed an additional annotation of the 
	specificity and we identified the particular elements (words, phrases, 
	clauses) in one sentence that were more specific than their 
	counterpart. In example 9, we can identify that ``girls'' is more 
	specific than ``children''. The difference in the specificity of ``girls'' 
	and ``children'' is the reason why \textbf{b} is annotated as more 
	specific than \textbf{a}. We called that ``scope of specificity''.
	
	In 80\% of the pairs with specificity at textual level, our 
	annotators were able to point at one or more particular elements
	that are responsible for the difference in specificity. In 20\% of 
	the pairs, the specificity was not decomposable. This finding also 
	confirms \citet{ko2019domain}'s findings, who showed that 
	frequency-based features are well-suited for automatic specificity 
	detection.

	In our analysis on the nature of the specificity relation, we combined 
	the annotation of ``scope of specificity'' and the traditional annotation
	of linguistic and reason-based types discussed in the previous 
	sections. In particular, we looked for overlap between the ``scope of
	specificity'' and the scope of linguistic and reason-based types.	
	Example 10 shows the two separate annotations side by side. In 
	\textbf{a} and \textbf{b}, we show the annotation of the linguistic and 
	reason-based types: \textit{``same polarity substitution (habitual)''} of
	``children'' and ``girls'', and \textit{``diathesis alternation''} of
	``receive'' and ``is received by''. In \textbf{c} and \textbf{d} we show 
	the annotation of the specificity: ``children'' - ``girls''. 
	When we compare the two annotations we can observe that the ``scope 
	of specificity'' overlaps with the scope of \textit{``same
	polarity substitution (habitual)''}. 

	\begin{itemize}
		\item[10]\textbf{a} All \underline{children} \textit{receive} the same education. \\
		\textbf{b} The same education \textit{is received} by all \underline{girls}. \\
		\textbf{c} All \textbf{children} receive the same education. \\
		\textbf{d} The same education is received by all \textbf{girls}.
	\end{itemize}

	We argue that when there is an 
	overlap between the ``scope of specificity'' and a linguistic or a 
	reason-based type, it is the linguistic or reason-based phenomenon
	that is responsible for the difference in specificity. In example 10
	we can say that the substitution of ``children'' and ``girls'' is responsible
	for the difference of specificity.


		\begin{table} [h!]
		\begin{center}

		\begin{tabular}{| l | l | r |}
		\hline
		\textbf{ID} & \textbf{Type} & \textbf{Freq.} \\
		\hline \hline

		3 & Derivational Changes & 1 \% \\
		\hline
		5 & Same Pol. Sub. (habitual) & 17 \% \\
		\hline
		6 & Same Pol. Sub. (contextual) & 9 \% \\
		\hline
		7 & Same Pol. Sub. (named entity) & 2 \% \\
		\hline
		9 & Opp. Pol. Sub (habitual) & 2 \% \\
		\hline
		11 & Synthetic / Analytic sub. & 9 \% \\
		\hline
		14 & Negation Switching & 1 \% \\
		\hline
		16 & Anaphora & 1\% \\
		\hline
		23 & Addition / Deletion & 50 \% \\
		\hline
		27 & Cause and Effect & 7 \% \\
		\hline
		28 & Condition / Property & 1 \% \\
		\hline
		33 & Transitivity & 1 \% \\
		\hline
		34 & Other (General Inferences) & 1 \%\\
		\hline
		
		\end{tabular}

		\end{center}
		\caption{Decomposition of Specificity} \label{7:st-freq}
		\end{table}	

	Table \ref{7:st-freq}
	shows the overlap between ``scope of specificity'' and ``atomic
	types''. In 97 \% of the cases where specificity was decomposable
	the more/less specific elements overlapped with an atomic type.
	In 50 \% of the cases the specificity was due to additional
	information (\#23). The other frequent cases include \textit{same
	polarity substitution} (\#5, \#6, and \#7), \textit{synthetic/analytic substitution} (\#11),
	and \textit{cause and effect} (\#27) reasoning. While the overall 
	tendencies are similar to the other meaning 	relations, specificity
	also has its unique characteristics. We found almost no specificity at
	morphological level and the frequency of \textit{Same polarity 
	substitution} (\#5, \#6, and \#7), while still high, was lower than that of paraphrasing and
	contradiction pairs. The relative frequency of \textit{Synthetic/analytic
	substitution} (\#11) was the highest of all relations and the reasoning types
	were almost as frequent as in entailment pairs, although the type
	distribution is different. We found no syntactic or discourse driven
	specificity changes.


\section{Discussion}\label{7:mt:diss}

	In Section \ref{7:mt:type}, we posed two Research Questions that
	we wanted to address within this paper. We answered both of them
	in sections \ref{7:mt:anno} and \ref{7:mt:results}.
	Our annotation demonstrated that a shared typology 
	can be successfully applied to multiple relations. The quality of the
	annotation is attested by the high inter-annotator agreement.
	We also demonstrated that a shared typology, such as SHARel,
	is useful to compare different meaning relations in a 
	quantitative and human interpretable way.



	In this paper we provide a new perspective on the joint research 
	into multiple meaning relations. Traditionally, the meaning 
	relations have been studied in isolation. Only recently researchers 
	have started to explore the possibility of a joint research and 
	a transfer of knowledge. We propose a new framework for a joint
	research on meaning relations via a shared typology. This framework
	has clear advantages: it is intuitive to use and interpret; it is easy to adapt
	in practical setting - both in corpora creation and in empirical tasks;
	it is based on solid linguistic theory.
	We believe that our approach can lead to a better understanding of
	the workings of the meaning relations, but also to improvements in
	the performance of automated systems.
	
	The biggest challenge in the
	joint study of meaning relations is the limited availability of 
	corpora annotated with multiple relations. The corpus that we 
	used for our study is relatively small in size. It also has 
	restrictions in terms of sentence length and the frequency of Named
	Entities. However, it is the only corpus to date annotated with
	all relations of interest. 

	Despite the limitations of the chosen corpus, the obtained results
	are promising. We provide interesting insights into the workings
	of the different relations, and also outline various practical 
	implications. \citet{kovatchev-etal-2019-qualitative} demonstrated that a corpus with a size
	of a few thousand sentence pairs can be successfully used as
	a qualitative evaluation benchmark. SHARel and the annotation
	methodology we used easily scale to such size of corpora. 
	This opens up the possibility
	for a qualitative evaluation of multiple meaning relations as well
	as for easier transfer of knowledge based on the particular types
	involved in the relations.

\section{Conclusions and Future Work}\label{7:mt:conc}

	In this paper we presented the first attempt towards 
	decomposing multiple meaning relations using a
	shared typology.
	For this purpose we used SHARel - a typology that is not restricted to
	a single meaning relation. We applied the SHaRel typology in 
	an annotation study and demonstrated its applicability. We
	analyzed the shared tendencies and the key differences 
	between paraphrasing, textual entailment, contradiction, 
	and specificity at the level of linguistic and reason-based types. 

	Our work is the first successful step towards building a 
	framework for studying and processing multiple meaning 
	relations. We demonstrate that the linguistic and reasoning
	phenomena underlying the meaning relations are very
	similar and can be captured by a shared typology.
	A single framework for meaning relations can facilitate the
	analysis and comparison of the different relations and 
	improve the transfer of knowledge between them.

	As future work, we aim to use the findings and 
	resources of this study in practical applications such
	as the development and evaluation of systems for 
	automatic detection of paraphrases, entailment,
	contradiction, and specificity. We plan to use the SHARel
	typology for a general-purpose qualitative evaluation
	framework for meaning relations.

\afterpage{\null\newpage}

\bookmarksetup{startatroot}
\addtocontents{toc}{\bigskip}

\chapter{Conclusions}\label{ch:conc}


	In this final chapter, I look back at what has been done in this thesis. In Section 
	\ref{ch:contrib}, I summarize the main contributions of my research. I discuss the 
	importance of my findings in the context of the research on textual meaning relations.
	In Section \ref{ch:materials}, I present a description of the different resources, created
	as part of this thesis and released to the community.
	In Section \ref{ch:future}, 	I outline some open issues for future research on textual 
	meaning relations and the way forward to addressing them.

	\section{Contributions and Discussion of the Results}\label{ch:contrib}

The contributions of this thesis can be grouped into four thematic categories, 
corresponding to the four thesis objectives formulated in Section \ref{ch:objectives}.

\myparagraph{Empirical applications of paraphrase typology}

The first \textbf{objective} of this thesis is \textit{To use linguistic knowledge and 
paraphrase typology in order to improve the evaluation and interpretation of the 
automated paraphrase identification systems.}
This objective has been addressed in the articles \citet{etpc}, 
\citet{kovatchev-etal-2018-warp}, and \citet{kovatchev-etal-2019-qualitative}, presented 
in Chapters \ref{ch:warp}, \ref{ch:etpc}, and \ref{ch:peval}. 

Traditionally, the task of Paraphrase Identification is framed as a binary classification 
problem. It requires manually or semi-automatically annotated data for training and 
testing. The performance of the automated systems is evaluated using Accuracy and 
F1 measures. The state-of-the-art systems that work in Paraphrase Identification are 
mostly based on complex deep learning architectures and trained on large amounts of 
data. The linguistic intuitions and resources are relatively less important for these 
systems.

In this thesis I found evidence that the linguistic intuitions from the theoretical research on 
Paraphrase Typology can successfully be incorporated in the empirical task of Paraphrase 
Identification. The \textbf{contributions} of this part of the thesis are two empirical
applications of paraphrase typology:

\begin{itemize}

	\item \underline{A statistical corpus-based analysis presented in Chapter \ref{ch:etpc}.} 
	I measured and compared the distribution of the different paraphrase types in the 
	MRPC corpus. The analysis of the results shows that the type distribution is severely 
	imbalanced: some paraphrase types appear in the majority of the examples, while other 
	types are underrepresented. I hypothesize that this imbalance introduces a potential bias 
	in the datasets. 

	\item \underline{A ``qualitative evaluation framework'' presented in Chapter \ref{ch:peval}.} 
	I evaluated and compared the performance of 11 different automated paraphrase 
	identification systems on each of the paraphrase types. The results depicted that Accuracy 
	and F1 measures fail to capture important aspects of the performance of the automated 
	systems. I showed that the performance of the evaluated systems varies significantly based
	on the types involved in each candidate-paraphrase pair. I also showed that systems with 
	quantitatively similar performance can make qualitatively different predictions and errors.


\end{itemize} 

%
%
%

This work gave a new perspective on the task of Paraphrase Identification. I argued that  
the ``binary'' definition of the task is oversimplified as it does not account for the different 
linguistic and reason-based phenomena involved in paraphrasing. The experiments showed 
that linguistic knowledge, in particular Paraphrase Typology, can be beneficial when analyzing 
the quality of the corpora and the performance of the automated systems. 

\myparagraph{In-depth knowledge about paraphrasing}

The second \textbf{objective} of this thesis is \textit{To empirically validate and quantify the 
difference between the various linguistic and reason-based phenomena involved in paraphrasing.}
This objective has been addressed in the articles \citet{etpc} and 
\citet{kovatchev-etal-2019-qualitative}, presented in Chapters \ref{ch:etpc} and \ref{ch:peval}.

At the beginning of this dissertation, the research on Paraphrase Typology was
predominantly theoretical. The different authors proposed lists of phenomena involved 
in the textual meaning relations and provided examples for each different type. However,
there were no empirical experiments that could measure the practical implications
of the theoretical difference between the paraphrase types.

In this thesis I prepared and carried out the first empirical experiment aimed at
validating the theoretical concepts of the research on Paraphrase Typology. The
following \textbf{contributions} advance the research on Paraphrase Typology and
provide novel, more in-depth knowledge about paraphrasing:

\begin{itemize}

	\item \underline{A new paraphrase typology, presented in Chapter \ref{ch:etpc}.} 
	I extended the existing typologies in such a way so that they could be
	applied both to paraphrase and non-paraphrase pairs. All paraphrase
	typologies prior to this thesis were only focused on texts that hold a
	paraphrasing relation. Extending the coverage of the typology to 
	non-paraphrasing pairs was crucial for the empirical evaluation.
	\item \underline{A statistical corpus-based analysis presented in Chapter 
	\ref{ch:etpc}.} I measured and compared the frequency distribution of 
	paraphrase types in the paraphrase and non-paraphrase pairs in the ETPC 
	corpus.
	\item \underline{An analysis of machine learning experiments presented in 
	Chapter \ref{ch:peval}.} I analyzed the performance of 11 different 
	automated paraphrase identification systems. The data showed that the 
	performance of the automated systems varies significantly based on the 
	paraphrase types involved in each candidate paraphrase pair. These 
	results suggest that paraphrase types are processed differently by 
	automated paraphrase identification systems. 

\end{itemize} 

This thesis explored novel directions within the research on Paraphrase Typology.  
I presented the first empirical experiment that quantifies the difference in 
processing paraphrase types. The data allows to identify types that are 
easier or harder for the automated paraphrase identification systems. The proposed
methodology is not limited to the paraphrasing textual meaning relation. It can 
easily be extended to other relations such as textual entailment or semantic textual 
similarity.



\myparagraph{An empirical study on multiple textual meaning relations}

The third \textbf{objective} of this thesis is \textit{To empirically determine the 
interactions between Paraphrasing, Textual Entailment, Contradiction, and 
Semantic Similarity in a corpus of multiple textual meaning relations.}. This 
objective has been addressed in the article \citet{gold-etal-2019-annotating}, 
presented in Chapter \ref{ch:law}.

Textual meaning relations, such as Paraphrasing, Textual Entailment, and
Semantic Textual Similarity are a popular topic within Natural Language 
Processing and Computational Linguistics. Traditionally, these meaning 
relations are studied in isolation. The research on them and the analysis of the 
interactions between them was very limited at the start of this dissertation. 

In this thesis, I took a new look of the problem and broadened the area of 
study. I carried out a joint study on multiple textual meaning relations.
The results showed that it is possible to address the analysis of several 
meaning relations at the same time. The findings of the thesis emphasize 
that such analysis can benefit each individual relation. The following 
\textbf{contributions} enable the research in a novel direction, focused on
multiple textual meaning relations: 

\begin{itemize}

	\item \underline{A new corpus creation methodology presented in Chapter 
	\ref{ch:law}.} I proposed a novel methodology for creating a corpus that 
	contains multiple textual meaning relations: paraphrasing, textual entailment, 
	contradiction, textual similarity, and textual specificity.
	\item \underline{A new corpus presented in Chapter \ref{ch:law}.} To the
	best of my knowledge this is the first corpus containing pairs independently
	annotated for paraphrasing, textual entailment, contradiction, textual similarity, 
	and textual specificity. Each meaning relation was annotated independently by 
	10 different annotators, to ensure the quality of the corpus.
	\item \underline{A statistical corpus analysis presented in Chapter \ref{ch:law}.} 
	I measured and compared the frequency of each meaning relation in the corpus.
	I also analyzed the interactions, correlations, and overlap between the different 
	textual meaning relations in the corpus. To the best of my knowledge, this is the 
	first empirical comparison between paraphrasing, textual entailment, contradiction, 
	textual similarity, and textual specificity. 

\end{itemize}

The findings of this thesis have improved the understanding on important issues 
associated with each individual textual meaning relation and the way they interact
with each other. Thanks to this study, some theoretical hypotheses and assumptions
that exist in the literature have been empirically confirmed. I also reported some 
unexpected results:

\begin{itemize}
	\item There is a negative statistical correlation between contradiction and 
	textual entailment; and between contradiction and paraphrasing.
	\item There is a strong positive statistical correlation between uni-directional textual 
	entailment and paraphrasing.
	\item In the corpus of study, paraphrasing is not equal to 
	bi-directional textual entailment. This finding contradicts pre-existing
	theoretical hypotheses and assumptions.
	\item The data indicates that there is no statistical correlation between textual
	specificity and the other textual meaning relations. This also contradicts
	pre-existing hypothesis claiming that specificity should be strongly correlated
	with textual entailment.
	\item The analysis showed that paraphrasing, textual entailment, and contradiction 
	have a strong statistical correlation with the degree of textual semantic similarity.
	Contrary to some previous studies, in my experiments pairs that contradict
	each other are perceived as similar.
\end{itemize}

This thesis emphasized the importance of a joint study on multiple textual meaning
relations. The proposed methodology for corpus creation and analysis and the new
corpus open new directions for future research.


\myparagraph{A shared typology of textual meaning relations}

The fourth \textbf{objective} of this thesis is \textit{To propose and evaluate a novel shared 
typology of meaning relations. The shared typology would then be used as a 
conceptual framework for a joint research on meaning relations.}. This objective
has been addressed in the article \citet{mtype_lrec}, presented in Chapter 
\ref{ch:sharel}.

In the recent years, several of the researchers working on paraphrasing, textual
entailment, and semantic textual similarity independently argued that a
single label is not sufficient to express a complex textual meaning relation.
To address this problem they proposed various typologies, that is, lists of 
linguistic and reasoning phenomena involved in each textual meaning relation.
At the beginning of this dissertation, each typology was focused on a single 
textual meaning relation and was not applicable to other relations.

This thesis showed that it is possible to have a single typology for multiple 
meaning relations. It also emphasized the advantages of a shared typology. 
The following \textbf{contributions} facilitate the further the research on a 
shared typology of textual meaning relations:

\begin{itemize}
	\item \underline{The SHARel typology presented in Chapter \ref{ch:sharel}.} I
	propose a new typology, that is applicable to multiple textual meaning 
	relations: paraphrasing, textual entailment, contradiction, textual 
	specificity, and semantic similarity.
	\item \underline{A corpus based study presented in Chapter \ref{ch:sharel}.} I
	empirically validated the applicability of SHARel in a corpus annotation.
	The different meaning relations were compared in terms of the 
	phenomena involved in each one of them. This comparison is more
	informative than measuring binary correlation or overlap.
\end{itemize}

This thesis has expanded the research on textual meaning relations. The 
SHARel typology is a step forward from the existing typologies - it is
linguistically motivated and hierarchically organized. It contains both
linguistic and reason-based types and has a wider coverage than any
other typology.
A shared typology of textual meaning relations provides some valuable insight
into the workings of each individual relation. Furthermore the shared
typology can be used as a conceptual framework for in-depth comparison 
between the different meaning relations. It also greatly facilitates the transfer 
of knowledge and resources between paraphrasing, textual entailment, and 
semantic similarity research. 
 
%

	\section{Resources}\label{ch:materials}

	Throughout my research, I have created several language resources that I
have made available to the Natural Language Processing and Computational
Linguistics  community. These resources are also part of the contributions
of this thesis:

\begin{itemize}

	\item The Extended Paraphrase Typology (EPT) and annotation guidelines with
	examples for creating a corpus annotated with EPT.
	\item The Single Human-Interpretable Typology for Annotating Meaning Relations 
	(SHARel) and  annotation guidelines with examples for creating a corpus 
	annotated with SHARel.
	\item The WARP-Text web-based annotation interface for a fine-grained annotation
	of pairs of text.
	\item The Extended Typology Paraphrase Corpus (ETPC) - the first Paraphrase
	Identification corpus annotated with paraphrase types.
	\item The first corpus explicitly annotated with Paraphrasing, Textual Entailment,
	Contradiction, Semantic Textual Similarity, and Textual Specificity.
	\item The first corpus of multiple meaning relations, annotated with SHARel.

\end{itemize}

During my research and in collaboration with the Language Technology Lab at the
University of Duisburg-Essen, I co-organized the first RELATIONS workshop\footnote{\url{https://sites.google.com/view/relations-2019}} \citep{ws-2019-relations}, bringing
together researchers on textual meaning relations. The workshop was collocated with
the 13th International Conference on Computational Semantics (IWCS) in Gothenburg, 
Sweden, May 23 2019.


	\section{Future Research Directions}\label{ch:future}

	This thesis gives two new perspectives on the research of textual meaning relations within
Natural Language Processing and Computational Linguistics. 

\begin{itemize}
	\item The importance of linguistic knowledge in the automatic processing 
	of meaning relations. 
	\item The presentation of the first empirical research on multiple textual meaning relations. 
\end{itemize}

My work opens several new directions for future research.

In Part \ref{p:type} of this thesis I applied paraphrase typology to the task of Paraphrase 
Identification. One of my findings was that the existing corpora for Paraphrase Identification 
is not well balanced in terms of paraphrase types. I argued that this imbalance can introduce 
a bias in the task and decrease the generalizability of automated PI systems. A promising research
direction in this area is to work towards creating better corpora for the recognition tasks on 
textual meaning relations. Following what I did in my dissertation, I am currently working on
a new corpus for Recognizing Textual Entailment for Spanish. The objective behind the 
corpus creation is to obtain a corpus more balanced in terms of certain under-represented
phenomena, such as negation and named entity reasoning. My work would also help expand the 
research on textual meaning relations to languages other than English.

%

In Chapter \ref{ch:peval} I presented the advantages of a qualitative evaluation framework
over traditional measures such as Accuracy and F1. An important future research in this
area would be to extend the qualitative evaluation framework to other empirical tasks on 
textual meaning relations, such as recognizing textual entailment or semantic textual similarity.
My work in Part \ref{p:mrel} of this thesis and in particular the SHARel typology is a first step 
towards extending the coverage of the qualitative evaluation. I showed that the typology 
can be applied to multiple textual meaning relations. The next step in this research direction
would be the creation of a larger corpus and the corresponding software package for a general
qualitative evaluation framework for textual meaning relations. 


The qualitative evaluation of Paraphrase Identification systems showed that some phenomena, 
such as negation, ellipsis, and named entity reasoning are challenging across all evaluated
systems. As a future research, I believe that each of these phenomena has to be analyzed
in more detail in the context of the importance it has for the automatic processing of textual
meaning relations. In a continuation of my thesis, I am currently investigating the role of negation
for paraphrase identification, recognizing textual entailment, semantic textual similarity, and
question answering. The preliminary results indicate that negation is extremely challenging
across multiple automated systems.

%

Finally, my work in Part \ref{p:mrel} of this thesis opens several new directions on the
joint research of multiple textual meaning relations. One potential area is the creation
of datasets and automated systems for the simultaneous processing of multiple textual 
meaning relations. Some preliminary experiments that I carried out on the corpus presented in 
Chapter \ref{ch:law} indicate it is possible to use one meaning relation to predict the 
others. The next step in this research direction would be the creation of larger datasets
with multiple textual meaning relations and the creation of more sophisticated automated
systems.

%
%

\bibliographystyle{plainnat}
\bibliography{vk_thesis}

\begin{thebibliography}{180}
\providecommand{\natexlab}[1]{#1}
\providecommand{\url}[1]{\texttt{#1}}
\expandafter\ifx\csname urlstyle\endcsname\relax
  \providecommand{\doi}[1]{doi: #1}\else
  \providecommand{\doi}{doi: \begingroup \urlstyle{rm}\Url}\fi

\bibitem[Agirre et~al.(2012)Agirre, Diab, Cer, and Gonzalez-Agirre]{sts}
Eneko Agirre, Mona Diab, Daniel Cer, and Aitor Gonzalez-Agirre.
\newblock Semeval-2012 task 6: A pilot on semantic textual similarity.
\newblock In \emph{Proceedings of the First Joint Conference on Lexical and
  Computational Semantics - Volume 1: Proceedings of the Main Conference and
  the Shared Task, and Volume 2: Proceedings of the Sixth International
  Workshop on Semantic Evaluation}, SemEval '12, pages 385--393, Stroudsburg,
  PA, USA, 2012. Association for Computational Linguistics.
\newblock URL \url{http://dl.acm.org/citation.cfm?id=2387636.2387697}.

\bibitem[Agirre et~al.(2013)Agirre, Cer, Diab, Gonzalez-Agirre, and
  Guo]{agirre2013sem}
Eneko Agirre, Daniel Cer, Mona Diab, Aitor Gonzalez-Agirre, and Weiwei Guo.
\newblock {* SEM 2013 shared task: Semantic textual similarity}.
\newblock In \emph{Second Joint Conference on Lexical and Computational
  Semantics (* SEM), Volume 1: Proceedings of the Main Conference and the
  Shared Task: Semantic Textual Similarity}, volume~1, pages 32--43, 2013.

\bibitem[Agirre et~al.(2014)Agirre, Banea, Cardie, Cer, Diab, Gonzalez-Agirre,
  Guo, Mihalcea, Rigau, and Wiebe]{agirre2014semeval}
Eneko Agirre, Carmen Banea, Claire Cardie, Daniel Cer, Mona Diab, Aitor
  Gonzalez-Agirre, Weiwei Guo, Rada Mihalcea, German Rigau, and Janyce Wiebe.
\newblock {Semeval-2014 task 10: Multilingual semantic textual similarity}.
\newblock In \emph{Proceedings of the 8th international workshop on semantic
  evaluation (SemEval 2014)}, pages 81--91, 2014.

\bibitem[Agirre et~al.(2016)Agirre, Gonzalez-Agirre, Lopez-Gazpio, Maritxalar,
  Rigau, and Uria]{agirre-etal-2016-semeval-2016}
Eneko Agirre, Aitor Gonzalez-Agirre, I{\~n}igo Lopez-Gazpio, Montse Maritxalar,
  German Rigau, and Larraitz Uria.
\newblock {S}em{E}val-2016 task 2: Interpretable semantic textual similarity.
\newblock In \emph{Proceedings of the 10th International Workshop on Semantic
  Evaluation ({S}em{E}val-2016)}, pages 512--524, San Diego, California, June
  2016. Association for Computational Linguistics.
\newblock \doi{10.18653/v1/S16-1082}.
\newblock URL \url{https://www.aclweb.org/anthology/S16-1082}.

\bibitem[Aldarmaki and Diab(2018)]{hanan}
Hanan Aldarmaki and Mona Diab.
\newblock Evaluation of unsupervised compositional representations.
\newblock In \emph{Proceedings of COLING 2018}, 2018.

\bibitem[Alzahrani and Salim(2010)]{alzahrani2010fuzzy}
Salha Alzahrani and Naomie Salim.
\newblock Fuzzy semantic-based string similarity for extrinsic plagiarism
  detection.
\newblock \emph{Braschler and Harman}, 1176:\penalty0 1--8, 2010.

\bibitem[Androutsopoulos and Malakasiotis(2010)]{androutsopoulos2010survey}
Ion Androutsopoulos and Prodromos Malakasiotis.
\newblock A survey of paraphrasing and textual entailment methods.
\newblock \emph{Journal of Artificial Intelligence Research}, 38:\penalty0
  135--187, 2010.

\bibitem[Arase and Tsujii(2018)]{ARASE18}
Yuki Arase and Jun'ichi Tsujii.
\newblock Spade: Evaluation dataset for monolingual phrase alignment.
\newblock In \emph{Proceedings of LREC-2018}, 2018.

\bibitem[Arppe et~al.(2010)Arppe, Gilquin, Glynn, Hilpert, and Zeschel]{arppe}
Antti Arppe, Gaetanelle Gilquin, Dylan Glynn, Martin Hilpert, and Arne Zeschel.
\newblock Cognitive corpus linguistics: five points of debate on current theory
  and methodology.
\newblock \emph{Corpora}, 5\penalty0 (1):\penalty0 1--27, 2010.

\bibitem[Auer et~al.(2008)Auer, Bizer, Kobilarov, Lehmann, Cyganiak, and
  Ives]{dbpedia_iswc}
S{\"o}ren Auer, Chris Bizer, Georgi Kobilarov, Jens Lehmann, Richard Cyganiak,
  and Zachary Ives.
\newblock {DB}pedia: A nucleus for a web of open data.
\newblock In \emph{Proceedings of the 6th International Semantic Web Conference
  (ISWC)}, volume 4825 of \emph{Lecture Notes in Computer Science}, pages
  722--735. Springer, 2008.
\newblock \doi{doi:10.1007/978-3-540-76298-0_52}.

\bibitem[Baldwin and Kim(2010)]{baldwin2010multiword}
Timothy Baldwin and Su~Nam Kim.
\newblock {Multiword Expressions}.
\newblock \emph{Handbook of natural language processing}, 2:\penalty0 267--92,
  2010.

\bibitem[B{\"a}r et~al.(2012)B{\"a}r, Zesch, and Gurevych]{bar2012text}
Daniel B{\"a}r, Torsten Zesch, and Iryna Gurevych.
\newblock Text reuse detection using a composition of text similarity measures.
\newblock \emph{Proceedings of COLING 2012}, pages 167--184, 2012.

\bibitem[Bar-Haim et~al.(2006)Bar-Haim, Dagan, Dolan, Ferro, and
  Giampiccolo]{rte2}
Roy Bar-Haim, Ido Dagan, Bill Dolan, Lisa Ferro, and Danilo Giampiccolo.
\newblock The second pascal recognising textual entailment challenge.
\newblock In \emph{Proceedings of the Second PASCAL Challenges Workshop on
  Recognising Textual Entailment}, 01 2006.

\bibitem[Baroni et~al.(2009)Baroni, Bernardini, Ferraresi, and
  Zanchetta]{Wacky}
M.~Baroni, S.~Bernardini, A.~Ferraresi, and E.~Zanchetta.
\newblock The wacky wide web: A collection of very large linguistically
  processed web-crawled corpora.
\newblock \emph{Language Resources and Evaluation}, 43\penalty0 (3):\penalty0
  209--226, 2009.

\bibitem[Baroni(2013)]{Baroni:2013}
Marco Baroni.
\newblock Composition in distributional semantics.
\newblock \emph{Language and Linguistics Compass}, 7\penalty0 (10):\penalty0
  511--22, 2013.

\bibitem[Baroni and Lenci(2010)]{BaroniLenci}
Marco Baroni and Alessandro Lenci.
\newblock Distributional memory: A general framework for corpus-based
  semantics.
\newblock \emph{Comput. Linguist.}, 36\penalty0 (4):\penalty0 673--721,
  December 2010.

\bibitem[Baroni et~al.(2010)Baroni, Murphy, Barbu, and
  Poesio]{baroni2010strudel}
Marco Baroni, Brian Murphy, Eduard Barbu, and Massimo Poesio.
\newblock Strudel: A corpus-based semantic model based on properties and types.
\newblock \emph{Cognitive Science}, 34\penalty0 (2):\penalty0 222--54, 2010.

\bibitem[Barr{\'o}n-Cede{\~n}o et~al.(2013)Barr{\'o}n-Cede{\~n}o, Vila,
  Mart{\'i}, and Rosso]{Barron}
Alberto Barr{\'o}n-Cede{\~n}o, Marta Vila, M.~Ant{\`o}nia Mart{\'i}, and Paolo
  Rosso.
\newblock Plagiarism meets paraphrasing: Insights for the next generation in
  automatic plagiarism detection.
\newblock \emph{Computational Linguistics}, 39\penalty0 (4):\penalty0 917--947,
  2013.

\bibitem[Bartsch(2004)]{bartsch:2004}
S.~Bartsch.
\newblock \emph{{Structural and functional properties of collocations in
  English: A corpus study of lexical and pragmatic constraints on lexical
  co-occurrence}}.
\newblock Gunter Narr Verlag, 2004.

\bibitem[Batanovi{\'c} et~al.(2018)Batanovi{\'c}, Cvetanovi{\'c}, and
  Nikoli{\'c}]{BATAN18}
Vuk Batanovi{\'c}, Milo\v{s} Cvetanovi{\'c}, and Bo\v{s}ko Nikoli{\'c}.
\newblock Fine-grained semantic textual similarity for serbian.
\newblock In \emph{Proceedings of LREC-2018}, 2018.

\bibitem[Benikova and Zesch(2017)]{Benikova}
Darina Benikova and Torsten Zesch.
\newblock Same same, but different: Compositionality of paraphrase granularity
  levels.
\newblock In \emph{Proceedings of RANLP 2017}, 2017.

\bibitem[Bentivogli et~al.(2009)Bentivogli, Magnini, Dagan, Dang, and
  Giampiccolo]{rte5}
Luisa Bentivogli, Bernardo Magnini, Ido Dagan, Hoa~Trang Dang, and Danilo
  Giampiccolo.
\newblock The fifth {PASCAL} recognizing textual entailment challenge.
\newblock In \emph{Proceedings of the Second Text Analysis Conference, {TAC}
  2009, Gaithersburg, Maryland, USA, November 16-17, 2009}, 2009.

\bibitem[Bentivogli et~al.(2010)Bentivogli, Clark, Dagan, and
  Giampiccolo]{rte6}
Luisa Bentivogli, Peter Clark, Ido Dagan, and Danilo Giampiccolo.
\newblock The sixth {PASCAL} recognizing textual entailment challenge.
\newblock In \emph{Proceedings of the Third Text Analysis Conference, {TAC}
  2010, Gaithersburg, Maryland, USA, November 15-16, 2010}, 2010.

\bibitem[Bentivogli et~al.(2011)Bentivogli, Clark, Dagan, and
  Giampiccolo]{rte7}
Luisa Bentivogli, Peter Clark, Ido Dagan, and Danilo Giampiccolo.
\newblock The seventh {PASCAL} recognizing textual entailment challenge.
\newblock In \emph{Proceedings of the Fourth Text Analysis Conference, {TAC}
  2011, Gaithersburg, Maryland, USA, November 14-15, 2011}, 2011.

\bibitem[Bertr{\'{a}}n et~al.(2008)Bertr{\'{a}}n, Borrega, Recasens, and
  Soriano]{ancora}
Manuel Bertr{\'{a}}n, Oriol Borrega, Marta Recasens, and B{\`{a}}rbara Soriano.
\newblock Ancorapipe: {A} tool for multilevel annotation.
\newblock \emph{Procesamiento del Lenguaje Natural}, 41, 2008.
\newblock URL
  \url{http://journal.sepln.org/sepln/ojs/ojs/index.php/pln/article/view/2577/1116}.

\bibitem[Bhagat(2009)]{Bhagat}
Rahul Bhagat.
\newblock \emph{Learning Paraphrases from Text}.
\newblock PhD thesis, Los Angeles, CA, USA, 2009.
\newblock AAI3368694.

\bibitem[Bhagat and Hovy(2013)]{BhagatHovy}
Rahul Bhagat and Eduard~H. Hovy.
\newblock What is a paraphrase?
\newblock \emph{Computational Linguistics}, 39\penalty0 (3):\penalty0 463--472,
  2013.

\bibitem[Biemann and Giesbrecht(2011)]{biemann2011distributional}
Chris Biemann and Eugenie Giesbrecht.
\newblock Distributional semantics and compositionality 2011: Shared task
  description and results.
\newblock In \emph{Proceedings of the workshop on distributional semantics and
  compositionality}, pages 21--8. Association for Computational Linguistics,
  2011.

\bibitem[Bird et~al.(2009)Bird, Klein, and Loper]{nltk}
Steven Bird, Ewan Klein, and Edward Loper.
\newblock \emph{Natural Language Processing with Python}.
\newblock O'Reilly Media, Inc., 1st edition, 2009.
\newblock ISBN 0596516495, 9780596516499.

\bibitem[Bobrow et~al.(2007)Bobrow, Crouch, King, Condoravdi, Karttunen, Nairn,
  de~Paiva, and Zaenen]{bobrow2007precision}
Daniel Bobrow, Dick Crouch, Tracy~Halloway King, Cleo Condoravdi, Lauri
  Karttunen, Rowan Nairn, Valeria de~Paiva, and Annie Zaenen.
\newblock Precision-focused textual inference.
\newblock In \emph{Proceedings of the ACL-PASCAL Workshop on Textual Entailment
  and Paraphrasing}, pages 16--21, 2007.

\bibitem[Boleda and Erk(2015)]{BoledaErk}
Gemma Boleda and Katrin Erk.
\newblock Distributional semantic features as semantic primitives -- or not,
  2015.
\newblock URL
  \url{https://www.aaai.org/ocs/index.php/SSS/SSS15/paper/view/10240/10025}.

\bibitem[Bosma and Callison-Burch(2006)]{bosma2006paraphrase}
Wauter Bosma and Chris Callison-Burch.
\newblock Paraphrase substitution for recognizing textual entailment.
\newblock In \emph{Workshop of the Cross-Language Evaluation Forum for European
  Languages}, pages 502--509. Springer, 2006.

\bibitem[Bowman et~al.(2015)Bowman, Angeli, Potts, and Manning]{snli}
Samuel~R. Bowman, Gabor Angeli, Christopher Potts, and Christopher~D. Manning.
\newblock A large annotated corpus for learning natural language inference.
\newblock In \emph{Proceedings of the 2015 Conference on Empirical Methods in
  Natural Language Processing (EMNLP)}. Association for Computational
  Linguistics, 2015.

\bibitem[Bruni et~al.(2014)Bruni, Tran, and
  Baroni]{DBLP:journals/jair/BruniTB14}
Elia Bruni, Nam{-}Khanh Tran, and Marco Baroni.
\newblock Multimodal distributional semantics.
\newblock \emph{J. Artif. Intell. Res.}, 49:\penalty0 1--47, 2014.
\newblock \doi{10.1613/jair.4135}.
\newblock URL \url{https://doi.org/10.1613/jair.4135}.

\bibitem[Cabrio and Magnini(2014)]{CabrioMagnini}
Elena Cabrio and Bernardo Magnini.
\newblock Decomposing semantic inferences, 2014.

\bibitem[Cali{\'n}ski and Harabasz(1974)]{Calinski:1974}
T.~Cali{\'n}ski and J.~Harabasz.
\newblock A dendrite method for cluster analysis.
\newblock \emph{Communications in Statistics-Simulation and Computation},
  3\penalty0 (1):\penalty0 1--27, 1974.

\bibitem[Castillo and Cardenas(2010)]{castillo2010using}
Julio~J. Castillo and Marina~E. Cardenas.
\newblock {Using sentence semantic similarity based on WordNet in recognizing
  textual entailment}.
\newblock In \emph{Ibero-American Conference on Artificial Intelligence}, pages
  366--375. Springer, 2010.

\bibitem[Cer et~al.(2017)Cer, Diab, Agirre, Lopez-Gazpio, and
  Specia]{cer2017semeval}
Daniel Cer, Mona Diab, Eneko Agirre, Inigo Lopez-Gazpio, and Lucia Specia.
\newblock Semeval-2017 task 1: Semantic textual similarity multilingual and
  crosslingual focused evaluation.
\newblock In \emph{Proceedings of the 11th International Workshop on Semantic
  Evaluation (SemEval-2017)}, pages 1--14, 2017.

\bibitem[Chen and Styler(2013)]{anafora}
Wei-Te Chen and Will Styler.
\newblock Anafora: A web-based general purpose annotation tool.
\newblock In \emph{HLT-NAACL}, 2013.
\newblock URL
  \url{http://dblp.uni-trier.de/db/conf/naacl/naacl2013.html#ChenS13}.

\bibitem[Conneau et~al.(2017)Conneau, Kiela, Schwenk, Barrault, and
  Bordes]{infersent}
Alexis Conneau, Douwe Kiela, Holger Schwenk, Lo{\"{\i}}c Barrault, and Antoine
  Bordes.
\newblock Supervised learning of universal sentence representations from
  natural language inference data.
\newblock \emph{CoRR}, abs/1705.02364, 2017.
\newblock URL \url{http://arxiv.org/abs/1705.02364}.

\bibitem[Conneau et~al.(2018)Conneau, Rinott, Lample, Williams, Bowman,
  Schwenk, and Stoyanov]{conneau2018xnli}
Alexis Conneau, Ruty Rinott, Guillaume Lample, Adina Williams, Samuel~R.
  Bowman, Holger Schwenk, and Veselin Stoyanov.
\newblock Xnli: Evaluating cross-lingual sentence representations.
\newblock In \emph{Proceedings of the 2018 Conference on Empirical Methods in
  Natural Language Processing}. Association for Computational Linguistics,
  2018.

\bibitem[Creutz(2018)]{creutz-2018-open}
Mathias Creutz.
\newblock Open subtitles paraphrase corpus for six languages.
\newblock In \emph{Proceedings of the Eleventh International Conference on
  Language Resources and Evaluation ({LREC} 2018)}, Miyazaki, Japan, May 2018.
  European Language Resources Association (ELRA).
\newblock URL \url{https://www.aclweb.org/anthology/L18-1218}.

\bibitem[Croft and Cruse(2004)]{Croft:2004}
W.~Croft and D.A. Cruse.
\newblock \emph{Cognitive Linguistics}.
\newblock Cambridge Textbooks in Linguistics. Cambridge University Press, 2004.
\newblock ISBN 9780521667708.

\bibitem[Cruse(1977)]{cruse1977pragmatics}
D.~Alan Cruse.
\newblock The pragmatics of lexical specificity.
\newblock \emph{Journal of linguistics}, 13\penalty0 (2):\penalty0 153--164,
  1977.

\bibitem[Cunningham et~al.(2011)Cunningham, Maynard, Bontcheva, Tablan, Aswani,
  Roberts, Gorrell, Funk, Roberts, Damljanovic, Heitz, Greenwood, Saggion,
  Petrak, Li, and Peters]{gate}
Hamish Cunningham, Diana Maynard, Kalina Bontcheva, Valentin Tablan, Niraj
  Aswani, Ian Roberts, Genevieve Gorrell, Adam Funk, Angus Roberts, Danica
  Damljanovic, Thomas Heitz, Mark~A. Greenwood, Horacio Saggion, Johann Petrak,
  Yaoyong Li, and Wim Peters.
\newblock \emph{{Text Processing with GATE (Version 6)}}.
\newblock 2011.
\newblock ISBN 978-0956599315.
\newblock URL \url{http://tinyurl.com/gatebook}.

\bibitem[Dagan et~al.(2006)Dagan, Glickman, and Magnini]{rte}
Ido Dagan, Oren Glickman, and Bernardo Magnini.
\newblock The pascal recognising textual entailment challenge.
\newblock In \emph{Proceedings of the First International Conference on Machine
  Learning Challenges: Evaluating Predictive Uncertainty Visual Object
  Classification, and Recognizing Textual Entailment}, MLCW'05, pages 177--190,
  Berlin, Heidelberg, 2006. Springer-Verlag.
\newblock ISBN 3-540-33427-0, 978-3-540-33427-9.
\newblock \doi{10.1007/11736790_9}.
\newblock URL \url{http://dx.doi.org/10.1007/11736790_9}.

\bibitem[De~Beaugrande and Dressler(1981)]{de1981introduction}
Robert De~Beaugrande and Wolfgang~U Dressler.
\newblock \emph{Introduction to text linguistics}.
\newblock Routledge, 1981.

\bibitem[Dem\v{s}ar(2006)]{Demsar}
Janez Dem\v{s}ar.
\newblock Statistical comparisons of classifiers over multiple data sets.
\newblock \emph{J. Mach. Learn. Res.}, 7:\penalty0 1--30, December 2006.
\newblock ISSN 1532-4435.
\newblock URL \url{http://dl.acm.org/citation.cfm?id=1248547.1248548}.

\bibitem[Devlin et~al.(2019)Devlin, Chang, Lee, and Toutanova]{bert}
Jacob Devlin, Ming-Wei Chang, Kenton Lee, and Kristina Toutanova.
\newblock {BERT}: Pre-training of deep bidirectional transformers for language
  understanding.
\newblock pages 4171--4186, June 2019.
\newblock \doi{10.18653/v1/N19-1423}.
\newblock URL \url{https://www.aclweb.org/anthology/N19-1423}.

\bibitem[Dolan et~al.(2004)Dolan, Quirk, and Brockett]{mrpc}
Bill Dolan, Chris Quirk, and Chris Brockett.
\newblock Unsupervised construction of large paraphrase corpora: Exploiting
  massively parallel news sources.
\newblock In \emph{Proceedings of Coling 2004}, pages 350--356, Geneva,
  Switzerland, Aug 23--Aug 27 2004. COLING.

\bibitem[Dolan and Brockett(2005)]{dolan2005}
William~B. Dolan and Chris Brockett.
\newblock Automatically constructing a corpus of sentential paraphrases.
\newblock In \emph{Proceedings of the Third International Workshop on
  Paraphrasing ({IWP}2005)}, 2005.
\newblock URL \url{https://www.aclweb.org/anthology/I05-5002}.

\bibitem[Dubremetz and Nivre(2014)]{dubremetz2014extraction}
Marie Dubremetz and Joakim Nivre.
\newblock {Extraction of Nominal Multiword Expressions in French}.
\newblock \emph{EACL 2014}, page~72, 2014.

\bibitem[Duffield et~al.(2010)Duffield, Hwang, and
  Michaelis]{duffield2010identifying}
Cecily~Jill Duffield, Jena~D Hwang, and Laura~A Michaelis.
\newblock Identifying assertions in text and discourse: the presentational
  relative clause construction.
\newblock In \emph{{Proceedings of the NAACL HLT Workshop on Extracting and
  Using Constructions in Computational Linguistics}}, pages 17--24. Association
  for Computational Linguistics, 2010.

\bibitem[En{\c{c}}(1991)]{encc1991semantics}
M{\"u}rvet En{\c{c}}.
\newblock The semantics of specificity.
\newblock \emph{Linguistic inquiry}, pages 1--25, 1991.

\bibitem[Erk(2012)]{Erk}
Katrin Erk.
\newblock Vector space models of word meaning and phrase meaning: A survey.
\newblock \emph{Language and Linguistics Compass}, 6\penalty0 (10):\penalty0
  635--653, 2012.

\bibitem[Espa{\~n}a~Bonet et~al.(2009)Espa{\~n}a~Bonet, Vila~Rigat,
  Rodr{\'i}guez, and Mart{\'i}]{coco}
Cristina Espa{\~n}a~Bonet, Marta Vila~Rigat, Horacio Rodr{\'i}guez, and Antonia
  Mart{\'i}.
\newblock Coco, a web interface for corpora compilation.
\newblock 2009.

\bibitem[Evert(2008)]{evert2008corpora}
Stefan Evert.
\newblock Corpora and collocations.
\newblock \emph{Corpus Linguistics. An International Handbook}, 2:\penalty0
  223--33, 2008.

\bibitem[Farahmand and Martins(2014)]{farahmand2014supervised}
Meghdad Farahmand and Ronaldo Martins.
\newblock {A Supervised Model for Extraction of Multiword Expressions Based on
  Statistical Context Features}.
\newblock \emph{EACL 2014}, page~10, 2014.

\bibitem[Farkas(2002)]{farkas2002specificity}
Donka~F. Farkas.
\newblock Specificity distinctions.
\newblock \emph{Journal of semantics}, 19\penalty0 (3):\penalty0 213--243,
  2002.

\bibitem[Fernando and Stevenson(2008)]{fernando2008}
Samuel Fernando and Mark Stevenson.
\newblock {A semantic similarity approach to paraphrase detection}.
\newblock \emph{Computational Linguistics UK (CLUK 2008) 11th Annual Research
  Colloqium}, 2008.

\bibitem[Fillmore et~al.(2012)Fillmore, Lee-Goldman, and
  Rhodes]{fillmore2012framenet}
Charles~J Fillmore, Russell Lee-Goldman, and Russell Rhodes.
\newblock {The Framenet constructicon}.
\newblock \emph{Sign-based Construction Grammar. CSLI, Stanford, CA}, 2012.

\bibitem[Finch et~al.(2005)Finch, Hwang, and Sumita]{finch-etal-2005-using}
Andrew Finch, Young-Sook Hwang, and Eiichiro Sumita.
\newblock Using machine translation evaluation techniques to determine
  sentence-level semantic equivalence.
\newblock In \emph{Proceedings of the Third International Workshop on
  Paraphrasing ({IWP}2005)}, 2005.
\newblock URL \url{https://www.aclweb.org/anthology/I05-5003}.

\bibitem[Firth(1957)]{Firth}
J.~R. Firth.
\newblock A synopsis of linguistic theory 1930-55.
\newblock 1952-59:\penalty0 1--32, 1957.

\bibitem[Forsberg et~al.(2014)Forsberg, Johansson, B\"{a}ckstr\"{o}m, Borin,
  Lyngfelt, Olofsson, and Prentice]{forsberg:2014}
Markus Forsberg, Richard Johansson, Linn\'{e}a B\"{a}ckstr\"{o}m, Lars Borin,
  Benjamin Lyngfelt, Joel Olofsson, and Julia Prentice.
\newblock From construction candidates to constructicon entries. an experiment
  using semi-automatic methods for identifying constructions in corpora.
\newblock \emph{Constructions and Frames}, 6\penalty0 (1):\penalty0 114--35,
  2014.
\newblock ISSN 1876-1933.

\bibitem[Franco-Salvador et~al.(2015)Franco-Salvador, Francisco, Paolo,
  Mariona, and Ant\'{o}nia]{francosalvadoretal:2015}
Marc Franco-Salvador, Rangel Francisco, Rosso Paolo, Taul\'{e} Mariona, and
  Mart\'{i}~M. Ant\'{o}nia.
\newblock Language variety identification using distributed representations of
  words and documents.
\newblock In \emph{{Proceedings of the 6th International Conference of CLEF on
  Experimental IR meets Multilinguality, Multimodality and Interaction}},
  Lectures Notes in Computer Science. Springer Verlag, 2015.

\bibitem[Friedman(1940)]{Friedman:1940}
M.~Friedman.
\newblock A comparison of alternative tests of significance for the problem of
  $m$ rankings.
\newblock \emph{The Annals of Mathematical Statistics}, 11\penalty0
  (1):\penalty0 86--92, March 1940.

\bibitem[Gamallo et~al.(2005)Gamallo, Agustini, and
  Lopes]{gamallo2005clustering}
Pablo Gamallo, Alexandre Agustini, and Gabriel~P Lopes.
\newblock Clustering syntactic positions with similar semantic requirements.
\newblock \emph{Computational Linguistics}, 31\penalty0 (1):\penalty0 107--146,
  2005.

\bibitem[Ganitkevitch et~al.(2013)Ganitkevitch, Durme, and
  Callison-Burch]{ppdb}
Juri Ganitkevitch, Benjamin~Van Durme, and Chris Callison-Burch.
\newblock Ppdb: The paraphrase database.
\newblock In Lucy Vanderwende, Hal~Daum{\'e} III, and Katrin Kirchhoff,
  editors, \emph{HLT-NAACL}, pages 758--764. The Association for Computational
  Linguistics, 2013.

\bibitem[Garoufi(2007)]{Garoufi}
Konstantina Garoufi.
\newblock Towards a better understanding of applied textual entailment:
  Annotation and evaluation of the rte-2 dataset.
\newblock Master's thesis, Saarland University, September 2007.

\bibitem[Giampiccolo et~al.(2007)Giampiccolo, Magnini, Dagan, and Dolan]{rte3}
Danilo Giampiccolo, Bernardo Magnini, Ido Dagan, and Bill Dolan.
\newblock The third {PASCAL} recognizing textual entailment challenge.
\newblock In \emph{Proceedings of the ACL-PASCAL@ACL 2007 Workshop on Textual
  Entailment and Paraphrasing, Prague, Czech Republic, June 28-29, 2007}, pages
  1--9, 2007.

\bibitem[Giampiccolo et~al.(2008)Giampiccolo, Dang, Magnini, Dagan, Cabrio, and
  Dolan]{rte4}
Danilo Giampiccolo, Hoa~Trang Dang, Bernardo Magnini, Ido Dagan, Elena Cabrio,
  and Bill Dolan.
\newblock The fourth {PASCAL} recognizing textual entailment challenge.
\newblock In \emph{Proceedings of the First Text Analysis Conference, {TAC}
  2008, Gaithersburg, Maryland, USA, November 17-19, 2008}, 2008.

\bibitem[Glockner et~al.(2018)Glockner, Shwartz, and
  Goldberg]{glockner-etal-2018-breaking}
Max Glockner, Vered Shwartz, and Yoav Goldberg.
\newblock Breaking {NLI} systems with sentences that require simple lexical
  inferences.
\newblock In \emph{Proceedings of the 56th Annual Meeting of the Association
  for Computational Linguistics (Volume 2: Short Papers)}, pages 650--655,
  Melbourne, Australia, July 2018. Association for Computational Linguistics.
\newblock \doi{10.18653/v1/P18-2103}.
\newblock URL \url{https://www.aclweb.org/anthology/P18-2103}.

\bibitem[Gold et~al.(2019)Gold, Kovatchev, and
  Zesch]{gold-etal-2019-annotating}
Darina Gold, Venelin Kovatchev, and Torsten Zesch.
\newblock Annotating and analyzing the interactions between meaning relations.
\newblock In \emph{Proceedings of the 13th Linguistic Annotation Workshop},
  pages 26--36, Florence, Italy, August 2019. Association for Computational
  Linguistics.
\newblock URL \url{https://www.aclweb.org/anthology/W19-4004}.

\bibitem[Goldberg(1995)]{Goldberg:1995}
A.~E. Goldberg.
\newblock \emph{Constructions: A Construction Grammar Approach to Argument
  Structure}.
\newblock Cognitive Theory of Language and Culture. University of Chicago
  Press, 1995.
\newblock ISBN 9780226300863.

\bibitem[Goldberg(2006)]{Goldberg:2006}
A.~E. Goldberg.
\newblock \emph{Constructions at work}.
\newblock Oxford University Press, 2006.

\bibitem[Goldberg(2013)]{goldberg2013argument}
Adele~E Goldberg.
\newblock Argument structure constructions versus lexical rules or derivational
  verb templates.
\newblock \emph{Mind \& Language}, 28\penalty0 (4):\penalty0 435--65, 2013.

\bibitem[Gries and Ellis(2015)]{griesellis:2015}
Stefan~Th. Gries and Nich~C. Ellis.
\newblock Statistical measures for usage-based linguistics.
\newblock \emph{Language Learning}, \penalty0 (65):\penalty0 1--28, 2015.

\bibitem[Gries et~al.(2005)Gries, Hampe, and
  Sch{\"o}nefeld]{gries2005converging}
Stefan~Th. Gries, Beate Hampe, and Doris Sch{\"o}nefeld.
\newblock Converging evidence: Bringing together experimental and corpus data
  on the association of verbs and constructions.
\newblock \emph{Cognitive Linguistics}, \penalty0 (16):\penalty0 635--76, 2005.

\bibitem[Gururangan et~al.(2018)Gururangan, Swayamdipta, Levy, Schwartz,
  Bowman, and Smith]{gururangan-etal-2018-annotation}
Suchin Gururangan, Swabha Swayamdipta, Omer Levy, Roy Schwartz, Samuel Bowman,
  and Noah~A. Smith.
\newblock Annotation artifacts in natural language inference data.
\newblock In \emph{Proceedings of the 2018 Conference of the North {A}merican
  Chapter of the Association for Computational Linguistics: Human Language
  Technologies, Volume 2 (Short Papers)}, pages 107--112, New Orleans,
  Louisiana, June 2018. Association for Computational Linguistics.
\newblock \doi{10.18653/v1/N18-2017}.
\newblock URL \url{https://www.aclweb.org/anthology/N18-2017}.

\bibitem[Harabagiu and Hickl(2006)]{harabagiu2006methods}
Sanda Harabagiu and Andrew Hickl.
\newblock Methods for using textual entailment in open-domain question
  answering.
\newblock In \emph{Proceedings of the 21st International Conference on
  Computational Linguistics and the 44th annual meeting of the Association for
  Computational Linguistics}, pages 905--912. Association for Computational
  Linguistics, 2006.

\bibitem[Harabagiu and Lacatusu(2010)]{harabagiu2010using}
Sanda Harabagiu and Finley Lacatusu.
\newblock Using topic themes for multi-document summarization.
\newblock \emph{ACM Transactions on Information Systems (TOIS)}, 28\penalty0
  (3):\penalty0 13, 2010.

\bibitem[Harris(1954)]{Harris}
Zellig Harris.
\newblock Distributional structure.
\newblock \emph{Word}, 10\penalty0 (23):\penalty0 146--162, 1954.

\bibitem[He and Lin(2016)]{He2016}
Hua He and Jimmy Lin.
\newblock Pairwise word interaction modeling with deep neural networks for
  semantic similarity measurement.
\newblock In \emph{Proceedings of the 2016 Conference of the North American
  Chapter of the Association for Computational Linguistics: Human Language
  Technologies (NAACL-HLT)}, 2016.

\bibitem[He et~al.(2015)He, Gimpel, and Lin]{He2015}
Hua He, Kevin Gimpel, and Jimmy Lin.
\newblock Multi-perspective sentence similarity modeling with convolutional
  neural networks.
\newblock In \emph{Proceedings of the 2015 Conference on Empirical Methods in
  Natural Language Processing}, pages 1576--1586. Association for Computational
  Linguistics, 2015.
\newblock \doi{10.18653/v1/D15-1181}.
\newblock URL \url{http://aclweb.org/anthology/D15-1181}.

\bibitem[Hendrickx et~al.(2010)Hendrickx, Kim, Kozareva, Nakov, S\'{e}aghdha,
  Pad\'{o}, Pennacchiotti, Romano, and Szpakowicz]{10.5555/1859664.1859670}
Iris Hendrickx, Su~Nam Kim, Zornitsa Kozareva, Preslav Nakov, Diarmuid~\'{O}.
  S\'{e}aghdha, Sebastian Pad\'{o}, Marco Pennacchiotti, Lorenza Romano, and
  Stan Szpakowicz.
\newblock Semeval-2010 task 8: Multi-way classification of semantic relations
  between pairs of nominals.
\newblock In \emph{Proceedings of the 5th International Workshop on Semantic
  Evaluation}, SemEval ’10, page 33–38, USA, 2010. Association for
  Computational Linguistics.

\bibitem[Hill et~al.(2015)Hill, Reichart, and Korhonen]{hill2015simlex999}
Felix Hill, Roi Reichart, and Anna Korhonen.
\newblock Simlex-999: Evaluating semantic models with (genuine) similarity
  estimation.
\newblock \emph{Computational Linguistics}, 2015.

\bibitem[Huang et~al.(2012)Huang, Socher, Manning, and
  Ng]{10.5555/2390524.2390645}
Eric~H. Huang, Richard Socher, Christopher~D. Manning, and Andrew~Y. Ng.
\newblock Improving word representations via global context and multiple word
  prototypes.
\newblock In \emph{Proceedings of the 50th Annual Meeting of the Association
  for Computational Linguistics: Long Papers - Volume 1}, ACL ’12, page
  873–882, USA, 2012. Association for Computational Linguistics.

\bibitem[Hwang et~al.(2010)Hwang, Nielsen, and Palmer]{hwang2010towards}
Jena~D Hwang, Rodney~D Nielsen, and Martha Palmer.
\newblock Towards a domain independent semantics: Enhancing semantic
  representation with construction grammar.
\newblock In \emph{{Proceedings of the NAACL HLT Workshop on Extracting and
  Using Constructions in Computational Linguistics}}, pages 1--8. Association
  for Computational Linguistics, 2010.

\bibitem[Iyer et~al.(2017)Iyer, Dandekar, and Csernai]{quora}
Shankar Iyer, Nikhil Dandekar, and Kornl Csernai.
\newblock First quora dataset release: Question pairs, 2017.

\bibitem[Ji and Eisenstein(2013)]{ji}
Yangfeng Ji and Jacob Eisenstein.
\newblock Discriminative improvements to distributional sentence similarity.
\newblock In \emph{Proceedings of the 2013 Conference on Empirical Methods in
  Natural Language Processing, {EMNLP} 2013, 18-21 October 2013, Grand Hyatt
  Seattle, Seattle, Washington, USA, {A} meeting of SIGDAT, a Special Interest
  Group of the {ACL}}, pages 891--896, 2013.
\newblock URL \url{http://aclweb.org/anthology/D/D13/D13-1090.pdf}.

\bibitem[Joao et~al.(2007)Joao, Ga{\"e}l, and Pavel]{joao2007new}
Cordeiro Joao, Dias Ga{\"e}l, and Brazdil Pavel.
\newblock New functions for unsupervised asymmetrical paraphrase detection.
\newblock \emph{Journal of Software}, 2\penalty0 (4):\penalty0 12--23, 2007.

\bibitem[Karypis(2002)]{Karypis}
George Karypis.
\newblock {CLUTO} a clustering toolkit.
\newblock Technical Report 02-017, Dept. of Computer Science, University of
  Minnesota, 2002.

\bibitem[Kesselmeier et~al.(2009)Kesselmeier, Kiss, M\"{u}ller, Roch, Stadteld,
  and Strunk]{kebelmeier:2009}
K.~Kesselmeier, T.~Kiss, A.~M\"{u}ller, C.~Roch, T.~Stadteld, and J.~Strunk.
\newblock Mining for preposition-noun constructions in german.
\newblock In \emph{{Workshop on Extracting and Using Constructions in Natural
  Language Processing}}, NODALIDA 2009, 2009.

\bibitem[Kiros et~al.(2015)Kiros, Zhu, Salakhutdinov, Zemel, Urtasun, Torralba,
  and Fidler]{skipt}
Ryan Kiros, Yukun Zhu, Ruslan~R Salakhutdinov, Richard Zemel, Raquel Urtasun,
  Antonio Torralba, and Sanja Fidler.
\newblock Skip-thought vectors.
\newblock In C.~Cortes, N.~D. Lawrence, D.~D. Lee, M.~Sugiyama, and R.~Garnett,
  editors, \emph{Advances in Neural Information Processing Systems 28}, pages
  3294--3302. Curran Associates, Inc., 2015.
\newblock URL \url{http://papers.nips.cc/paper/5950-skip-thought-vectors.pdf}.

\bibitem[Ko et~al.(2019)Ko, Durrett, and Li]{ko2019domain}
Wei-Jen Ko, Greg Durrett, and Junyi~Jessy Li.
\newblock Domain agnostic real-valued specificity prediction.
\newblock In \emph{AAAI}, 2019.

\bibitem[Kovatchev et~al.(2016)Kovatchev, Salam{\'o}, and
  Mart{\'i}]{Kovatchev:2016}
Venelin Kovatchev, Maria Salam{\'o}, and M.~Ant{\`o}nia Mart{\'i}.
\newblock Comparing distributional semantics models for identifying groups of
  semantically related words.
\newblock \emph{Procesamiento del Lenguaje Natural}, 57:\penalty0 109--116,
  2016.

\bibitem[Kovatchev et~al.(2018{\natexlab{a}})Kovatchev, Mart{\'i}, and
  Salam{\'o}]{etpc}
Venelin Kovatchev, M.~Ant{\`o}nia Mart{\'i}, and Maria Salam{\'o}.
\newblock Etpc - a paraphrase identification corpus annotated with extended
  paraphrase typology and negation.
\newblock In \emph{Proceedings of LREC-2018}, 2018{\natexlab{a}}.

\bibitem[Kovatchev et~al.(2018{\natexlab{b}})Kovatchev, Mart{\'\i}, and
  Salam{\'o}]{kovatchev-etal-2018-warp}
Venelin Kovatchev, M.~Ant{\`o}nia Mart{\'\i}, and Maria Salam{\'o}.
\newblock {WARP}-text: a web-based tool for annotating relationships between
  pairs of texts.
\newblock In \emph{Proceedings of the 27th International Conference on
  Computational Linguistics: System Demonstrations}, pages 132--136, Santa Fe,
  New Mexico, August 2018{\natexlab{b}}. Association for Computational
  Linguistics.
\newblock URL \url{https://www.aclweb.org/anthology/C18-2029}.

\bibitem[Kovatchev et~al.(2019{\natexlab{a}})Kovatchev, Gold, and
  Zesch]{ws-2019-relations}
Venelin Kovatchev, Darina Gold, and Torsten Zesch, editors.
\newblock \emph{{RELATIONS} - Workshop on meaning relations between phrases and
  sentences}, Gothenburg, Sweden, May 2019{\natexlab{a}}. Association for
  Computational Linguistics.
\newblock URL \url{https://aclanthology.org/W19-0800}.

\bibitem[Kovatchev et~al.(2019{\natexlab{b}})Kovatchev, Marti, Salamo, and
  Beltran]{kovatchev-etal-2019-qualitative}
Venelin Kovatchev, M.~Antonia Marti, Maria Salamo, and Javier Beltran.
\newblock A qualitative evaluation framework for paraphrase identification.
\newblock In \emph{Proceedings of the International Conference on Recent
  Advances in Natural Language Processing (RANLP 2019)}, pages 568--577, Varna,
  Bulgaria, September 2019{\natexlab{b}}. INCOMA Ltd.
\newblock \doi{10.26615/978-954-452-056-4_067}.
\newblock URL \url{https://www.aclweb.org/anthology/R19-1067}.

\bibitem[Kovatchev et~al.(2020)Kovatchev, Gold, Mart{\`i}, Salamo, and
  Zesch]{mtype_lrec}
Venelin Kovatchev, Darina Gold, M.~Ant{\'o}nia Mart{\`i}, Maria Salamo, and
  Torsten Zesch.
\newblock {Decomposing and Comparing Meaning Relations: Paraphrasing, Textual
  Entailment, Contradiction, and Specificity}.
\newblock In \emph{Proceedings of the Twelfth International Conference on
  Language Resources and Evaluation (LREC 2020)}. European Language Resources
  Association (ELRA), 2020.
\newblock ISBN 979-10-95546-00-9.

\bibitem[Kozareva and Montoyo(2006)]{Kozareva}
Zornitsa Kozareva and Andr\'{e}s Montoyo.
\newblock Paraphrase identification on the basis of supervised machine learning
  techniques.
\newblock In \emph{Proceedings of the 5th International Conference on Advances
  in Natural Language Processing}, FinTAL’06, page 524–533, Berlin,
  Heidelberg, 2006. Springer-Verlag.
\newblock ISBN 3540373349.
\newblock \doi{10.1007/11816508_52}.
\newblock URL \url{https://doi.org/10.1007/11816508_52}.

\bibitem[Kremer et~al.(2014)Kremer, Erk, Pad{\'o}, and
  Thater]{kremer-etal-2014-substitutes}
Gerhard Kremer, Katrin Erk, Sebastian Pad{\'o}, and Stefan Thater.
\newblock What substitutes tell us - analysis of an {``}all-words{''} lexical
  substitution corpus.
\newblock In \emph{Proceedings of the 14th Conference of the {E}uropean Chapter
  of the Association for Computational Linguistics}, pages 540--549,
  Gothenburg, Sweden, April 2014. Association for Computational Linguistics.
\newblock \doi{10.3115/v1/E14-1057}.
\newblock URL \url{https://www.aclweb.org/anthology/E14-1057}.

\bibitem[Lan and Xu(2018{\natexlab{a}})]{lan}
Wuwei Lan and Wei Xu.
\newblock Neural network models for paraphrase identification, semantic textual
  similarity, natural language inference, and question answering.
\newblock In \emph{Proceedings of COLING 2018}, 2018{\natexlab{a}}.

\bibitem[Lan and Xu(2018{\natexlab{b}})]{subw}
Wuwei Lan and Wei Xu.
\newblock Character-based neural networks for sentence pair modeling.
\newblock In \emph{Proceedings of the 2018 Conference of the North American
  Chapter of the Association for Computational Linguistics: Human Language
  Technologies (NAACL-HLT)}, 2018{\natexlab{b}}.

\bibitem[Lan et~al.(2017)Lan, Qiu, He, and Xu]{twitt}
Wuwei Lan, Siyu Qiu, Hua He, and Wei Xu.
\newblock A continuously growing dataset of sentential paraphrases.
\newblock In \emph{Proceedings of the 2017 Conference on Empirical Methods in
  Natural Language Processing, {EMNLP} 2017, Copenhagen, Denmark, September
  9-11, 2017}, pages 1224--1234, 2017.

\bibitem[Landauer et~al.(2007)Landauer, McNamara, Dennis, and
  Kintsch]{landauer2007handbook}
T.K. Landauer, D.S. McNamara, S.~Dennis, and W.~Kintsch.
\newblock \emph{{Handbook of Latent Semantic Analysis}}.
\newblock University of Colorado Institute of Cognitive Science Series.
  Lawrence Erlbaum Associates, 2007.
\newblock ISBN 9780805854183.

\bibitem[Lapesa and Evert(2014)]{TACL457}
Gabriella Lapesa and Stefan Evert.
\newblock A large scale evaluation of distributional semantic models:
  Parameters, interactions and model selection.
\newblock \emph{Transactions of the Association for Computational Linguistics},
  2\penalty0 (0):\penalty0 531--545, 2014.
\newblock ISSN 2307-387X.
\newblock URL \url{https://transacl.org/ojs/index.php/tacl/article/view/457}.

\bibitem[Lenci(2008)]{Lenci}
Alessandro Lenci.
\newblock Distributional semantics in linguistic and cognitive research.
\newblock \emph{Rivista di Linguistica}, 20\penalty0 (1):\penalty0 1--31, 2008.

\bibitem[Levin(1993)]{levin1993english}
Beth Levin.
\newblock \emph{English verb classes and alternations: A preliminary
  investigation}.
\newblock University of Chicago press, 1993.

\bibitem[Levy et~al.(2015)Levy, Ein-Dor, Hummel, Rinott, and
  Slonim]{levy-etal-2015-tr9856}
Ran Levy, Liat Ein-Dor, Shay Hummel, Ruty Rinott, and Noam Slonim.
\newblock {TR}9856: A multi-word term relatedness benchmark.
\newblock In \emph{Proceedings of the 53rd Annual Meeting of the Association
  for Computational Linguistics and the 7th International Joint Conference on
  Natural Language Processing (Volume 2: Short Papers)}, pages 419--424,
  Beijing, China, July 2015. Association for Computational Linguistics.
\newblock \doi{10.3115/v1/P15-2069}.
\newblock URL \url{https://www.aclweb.org/anthology/P15-2069}.

\bibitem[Lin and Pantel(2001)]{lin2001dirt}
Dekang Lin and Patrick Pantel.
\newblock Dirt@ sbt@ discovery of inference rules from text.
\newblock In \emph{{Proceedings of the seventh ACM SIGKDD international
  conference on Knowledge discovery and data mining}}, pages 323--8. ACM, 2001.

\bibitem[Linzen et~al.(2016)Linzen, Dupoux, and Goldberg]{Linzen}
Tal Linzen, Emmanuel Dupoux, and Yoav Goldberg.
\newblock Assessing the ability of lstms to learn syntax-sensitive
  dependencies.
\newblock \emph{Transactions of the Association for Computational Linguistics},
  4:\penalty0 521--535, 2016.
\newblock URL \url{http://aclweb.org/anthology/Q16-1037}.

\bibitem[Lloret et~al.(2008)Lloret, Ferr{\'a}ndez, Munoz, and
  Palomar]{lloret2008text}
Elena Lloret, Oscar Ferr{\'a}ndez, Rafael Munoz, and Manuel Palomar.
\newblock {A Text Summarization Approach under the Influence of Textual
  Entailment.}
\newblock In \emph{NLPCS}, pages 22--31, 2008.

\bibitem[LoBue and Yates(2011)]{LoBue}
Peter LoBue and Alexander Yates.
\newblock Types of common-sense knowledge needed for recognizing textual
  entailment.
\newblock In \emph{Proceedings of the 49th Annual Meeting of the Association
  for Computational Linguistics: Human Language Technologies: Short Papers -
  Volume 2}, HLT '11, pages 329--334, Stroudsburg, PA, USA, 2011. Association
  for Computational Linguistics.
\newblock ISBN 978-1-932432-88-6.
\newblock URL \url{http://dl.acm.org/citation.cfm?id=2002736.2002805}.

\bibitem[Louis and Nenkova(2012)]{louis2012corpus}
Annie Louis and Ani Nenkova.
\newblock A corpus of general and specific sentences from news.
\newblock In \emph{LREC}, pages 1818--1821, 2012.

\bibitem[Madnani and Dorr(2010)]{madnani2010generating}
Nitin Madnani and Bonnie~J Dorr.
\newblock Generating phrasal and sentential paraphrases: A survey of
  data-driven methods.
\newblock \emph{Computational Linguistics}, 36\penalty0 (3):\penalty0 341--387,
  2010.

\bibitem[Madnani et~al.(2012)Madnani, Tetreault, and Chodorow]{Madnani}
Nitin Madnani, Joel Tetreault, and Martin Chodorow.
\newblock Re-examining machine translation metrics for paraphrase
  identification.
\newblock In \emph{Proceedings of the 2012 Conference of the North American
  Chapter of the Association for Computational Linguistics: Human Language
  Technologies}, NAACL HLT '12, pages 182--190, Stroudsburg, PA, USA, 2012.
  Association for Computational Linguistics.
\newblock ISBN 978-1-937284-20-6.
\newblock URL \url{http://dl.acm.org/citation.cfm?id=2382029.2382055}.

\bibitem[Mann and Whitney(1947)]{mann1947}
H.~B. Mann and D.~R. Whitney.
\newblock On a test of whether one of two random variables is stochastically
  larger than the other.
\newblock \emph{Ann. Math. Statist.}, 18\penalty0 (1):\penalty0 50--60, 03
  1947.
\newblock \doi{10.1214/aoms/1177730491}.
\newblock URL \url{https://doi.org/10.1214/aoms/1177730491}.

\bibitem[Marci\'{n}czuk et~al.(2017)Marci\'{n}czuk, Oleksy, and
  Koco\'{n}]{inforex}
Micha{\l} Marci\'{n}czuk, Marcin Oleksy, and Jan Koco\'{n}.
\newblock Inforex - a collaborative system for text corpora annotation and
  analysis.
\newblock In \emph{Proceedings of RANLP-2017}, September 2017.
\newblock URL \url{https://doi.org/10.26615/978-954-452-049-6_063}.

\bibitem[Marcus et~al.(1993)Marcus, Marcinkiewicz, and
  Santorini]{marcus1993building}
Mitchell~P Marcus, Mary~Ann Marcinkiewicz, and Beatrice Santorini.
\newblock Building a large annotated corpus of english: The penn treebank.
\newblock \emph{Computational Linguistics}, 19\penalty0 (2):\penalty0 313--30,
  1993.

\bibitem[Marelli et~al.(2014)Marelli, Menini, Baroni, Bentivogli, Bernardi,
  Zamparelli, et~al.]{marelli2014sick}
Marco Marelli, Stefano Menini, Marco Baroni, Luisa Bentivogli, Raffaella
  Bernardi, Roberto Zamparelli, et~al.
\newblock {A SICK cure for the evaluation of compositional distributional
  semantic models.}
\newblock In \emph{LREC}, pages 216--223, 2014.

\bibitem[Mart{\'i} et~al.(2019)Mart{\'i}, Taul{\'e}, Kovatchev, and
  Salam{\'o}]{DISCOver}
Maria~Ant{\`o}nia Mart{\'i}, Mariona Taul{\'e}, Venelin Kovatchev, and Maria
  Salam{\'o}.
\newblock Discover: Distributional approach based on syntactic dependencies for
  discovering constructions.
\newblock \emph{Corpus Linguistics and Linguistic Theory}, 2019.

\bibitem[Mihalcea et~al.(2006)Mihalcea, Corley, and Strapparava]{mihalcea}
Rada Mihalcea, Courtney Corley, and Carlo Strapparava.
\newblock Corpus-based and knowledge-based measures of text semantic
  similarity.
\newblock In \emph{Proceedings of the 21st National Conference on Artificial
  Intelligence - Volume 1}, AAAI’06, page 775–780. AAAI Press, 2006.
\newblock ISBN 9781577352815.

\bibitem[Mikolov et~al.(2013{\natexlab{a}})Mikolov, Chen, Corrado, and
  Dean]{Mikolov}
Tomas Mikolov, Kai Chen, Greg Corrado, and Jeffrey Dean.
\newblock Efficient estimation of word representations in vector space.
\newblock \emph{CoRR}, abs/1301.3781, 2013{\natexlab{a}}.

\bibitem[Mikolov et~al.(2013{\natexlab{b}})Mikolov, Sutskever, Chen, Corrado,
  and Dean]{word2vec}
Tomas Mikolov, Ilya Sutskever, Kai Chen, Greg Corrado, and Jeffrey Dean.
\newblock Distributed representations of words and phrases and their
  compositionality.
\newblock In \emph{Proceedings of the 26th International Conference on Neural
  Information Processing Systems - Volume 2}, NIPS'13, pages 3111--3119, USA,
  2013{\natexlab{b}}. Curran Associates Inc.
\newblock URL \url{http://dl.acm.org/citation.cfm?id=2999792.2999959}.

\bibitem[Mikolov et~al.(2013{\natexlab{c}})Mikolov, Yih, and
  Zweig]{mikolov2013linguistic}
Tomas Mikolov, Wen-tau Yih, and Geoffrey Zweig.
\newblock Linguistic regularities in continuous space word representations.
\newblock In \emph{{HLT-NAACL}}, pages 746--51, 2013{\natexlab{c}}.

\bibitem[Miller(1998)]{miller1998wordnet}
George Miller.
\newblock \emph{{WordNet: An electronic lexical database}}.
\newblock MIT press, 1998.

\bibitem[Miller(1995)]{wordnet}
George~A. Miller.
\newblock Wordnet: A lexical database for english.
\newblock \emph{Commun. ACM}, 38\penalty0 (11):\penalty0 39--41, November 1995.
\newblock ISSN 0001-0782.

\bibitem[Mitchell and Lapata(2010)]{Mitchell:Lapata:2010}
Jeff Mitchell and Mirella Lapata.
\newblock Composition in distributional models of semantics.
\newblock \emph{Cognitive Science}, 34\penalty0 (8):\penalty0 1388--1439, 2010.

\bibitem[Moisl(2015)]{Moisl}
Hermann Moisl.
\newblock \emph{Cluster Analisys for Corpus Linguistics}.
\newblock De Gruyter Mouton, 2015.

\bibitem[Muischnek and Sajkan(2009)]{Muischnek:2009}
K.~Muischnek and H.~Sajkan.
\newblock Using collocation-finding methods to extract constructions and
  estimate their productivity.
\newblock In \emph{{Workshop on Extracting and Using Constructions in Natural
  Language Processing}}, NODALIDA 2009, 2009.

\bibitem[Murphy et~al.(2012)Murphy, Talukdar, and Mitchell]{murphy2012learning}
Brian Murphy, Partha~Pratim Talukdar, and Tom~M Mitchell.
\newblock Learning effective and interpretable semantic models using
  non-negative sparse embedding.
\newblock In \emph{{COLING}}, pages 1933--50, 2012.

\bibitem[Naik et~al.(2018)Naik, Ravichander, Sadeh, Rose, and
  Neubig]{naik-etal-2018-stress}
Aakanksha Naik, Abhilasha Ravichander, Norman Sadeh, Carolyn Rose, and Graham
  Neubig.
\newblock Stress test evaluation for natural language inference.
\newblock In \emph{Proceedings of the 27th International Conference on
  Computational Linguistics}, pages 2340--2353, Santa Fe, New Mexico, USA,
  August 2018. Association for Computational Linguistics.
\newblock URL \url{https://www.aclweb.org/anthology/C18-1198}.

\bibitem[Nastase et~al.(2018)Nastase, Fritz, and Frank]{NASTASE18}
Vivi Nastase, Devon Fritz, and Anette Frank.
\newblock Demodify: A dataset for analyzing contextual constraints on modifier
  deletion.
\newblock In \emph{Proceedings of LREC-2018}, 2018.

\bibitem[Navigli and Ponzetto(2012)]{babelnet}
Roberto Navigli and Simone~Paolo Ponzetto.
\newblock Babelnet: The automatic construction, evaluation and application of a
  wide-coverage multilingual semantic network.
\newblock \emph{Artif. Intell.}, 193:\penalty0 217--250, December 2012.
\newblock ISSN 0004-3702.

\bibitem[Nemenyi(1963)]{Nemenyi:1963}
P.B. Nemenyi.
\newblock \emph{{Distribution-free Multiple Comparisons}}.
\newblock PhD thesis, Princeton University, 1963.

\bibitem[Niwa and Nitta(1994)]{Niwa:1994}
Yoshiki Niwa and Yoshihiko Nitta.
\newblock Co-occurrence vectors from corpora vs. distance vectors from
  dictionaries.
\newblock In \emph{{Proceedings of the 15th Conference on Computational
  Linguistics}}, volume~1 of \emph{COLING '94}, pages 304--309, Stroudsburg,
  PA, USA, 1994. Association for Computational Linguistics.

\bibitem[Nunberg et~al.(1994)Nunberg, Sag, and Wasow]{nunberg1994idioms}
Geoffrey Nunberg, Ivan~A Sag, and Thomas Wasow.
\newblock Idioms.
\newblock \emph{Language}, pages 491--538, 1994.

\bibitem[O'Donnell and Ellis(2010)]{ODonnell:2010}
Matthew~Brook O'Donnell and Nick Ellis.
\newblock Towards an inventory of english verb argument constructions.
\newblock In \emph{{Proceedings of the NAACL HLT Workshop on Extracting and
  Using Constructions in Computational Linguistics}}, EUCCL '10, pages 9--16,
  Stroudsburg, PA, USA, 2010. Association for Computational Linguistics.

\bibitem[Pad{\'o} et~al.(2009)Pad{\'o}, Galley, Jurafsky, and
  Manning]{pado2009textual}
Sebastian Pad{\'o}, Michel Galley, Dan Jurafsky, and Christopher~D Manning.
\newblock Textual entailment features for machine translation evaluation.
\newblock In \emph{Proceedings of the Fourth Workshop on Statistical Machine
  Translation}, pages 37--41. Association for Computational Linguistics, 2009.

\bibitem[Padr\'{o} and Stanilovsky(2012)]{Padro:2012}
Llu\`{i}s Padr\'{o} and Evgeny Stanilovsky.
\newblock Freeling 3.0: Towards wider multilinguality.
\newblock In Nicoletta Calzolari, Khalid Choukri, Thierry Declerck, Mehmet~Ugur
  Dogan, Bente Maegaard, Joseph Mariani, Jan Odijk, and Stelios Piperidis,
  editors, \emph{{LREC}}, pages 2473--9. {European Language Resources
  Association (ELRA)}, 2012.
\newblock ISBN 978-2-9517408-7-7.

\bibitem[Pavlick et~al.(2015)Pavlick, Rastogi, Ganitkevitch, Van~Durme, and
  Callison-Burch]{ppdb2}
Ellie Pavlick, Pushpendre Rastogi, Juri Ganitkevitch, Benjamin Van~Durme, and
  Chris Callison-Burch.
\newblock {PPDB} 2.0: Better paraphrase ranking, fine-grained entailment
  relations, word embeddings, and style classification.
\newblock In \emph{Proceedings of the 53rd Annual Meeting of the Association
  for Computational Linguistics and the 7th International Joint Conference on
  Natural Language Processing (Volume 2: Short Papers)}, pages 425--430,
  Beijing, China, July 2015. Association for Computational Linguistics.
\newblock \doi{10.3115/v1/P15-2070}.
\newblock URL \url{https://www.aclweb.org/anthology/P15-2070}.

\bibitem[Pecina(2010)]{pecina:2010}
Pavel Pecina.
\newblock Lexical association measures and collocation extraction.
\newblock \emph{Language Resources and Evaluation}, 44:\penalty0 137--58, 2010.
\newblock ISSN 1574-020X.

\bibitem[Pe{\~{n}}as et~al.(2006)Pe{\~{n}}as, Rodrigo, and Verdejo]{sparte}
Anselmo Pe{\~{n}}as, {\'A}lvaro Rodrigo, and Felisa Verdejo.
\newblock Sparte, a test suite for recognising textual entailment in spanish.
\newblock In Alexander Gelbukh, editor, \emph{Computational Linguistics and
  Intelligent Text Processing}, pages 275--286, Berlin, Heidelberg, 2006.
  Springer Berlin Heidelberg.
\newblock ISBN 978-3-540-32206-1.

\bibitem[Pennington et~al.(2014)Pennington, Socher, and Manning]{glove}
Jeffrey Pennington, Richard Socher, and Christopher~D. Manning.
\newblock Glove: Global vectors for word representation.
\newblock In \emph{In EMNLP}, 2014.

\bibitem[Peters et~al.(2018)Peters, Neumann, Iyyer, Gardner, Clark, Lee, and
  Zettlemoyer]{peters2018contextualized}
Matthew~E. Peters, Mark Neumann, Mohit Iyyer, Matt Gardner, Christopher Clark,
  Kenton Lee, and Luke Zettlemoyer.
\newblock Deep contextualized word representations, 2018.
\newblock URL \url{http://arxiv.org/abs/1802.05365}.
\newblock cite arxiv:1802.05365Comment: NAACL 2018. Originally posted to
  openreview 27 Oct 2017. v2 updated for NAACL camera ready.

\bibitem[Ramisch et~al.(2010)Ramisch, Villavicencio, and
  Boitet]{ramisch2010multiword}
Carlos Ramisch, Aline Villavicencio, and Christian Boitet.
\newblock Multiword expressions in the wild?: the mwetoolkit comes in handy.
\newblock In \emph{Proceedings of the 23rd International Conference on
  Computational Linguistics: Demonstrations}, pages 57--60. Association for
  Computational Linguistics, 2010.

\bibitem[Sag et~al.(2002)Sag, Baldwin, Bond, Copestake, and
  Flickinger]{sag:2002multiword}
Ivan~A Sag, Timothy Baldwin, Francis Bond, Ann Copestake, and Dan Flickinger.
\newblock Multiword expressions: A pain in the neck for nlp.
\newblock In \emph{Computational Linguistics and Intelligent Text Processing},
  pages 1--15. Springer Berlin Heidelberg, 2002.

\bibitem[Sammons et~al.(2010)Sammons, Vydiswaran, and Roth]{Sammons}
Mark Sammons, V.~G.~Vinod Vydiswaran, and Dan Roth.
\newblock "ask not what textual entailment can do for you...".
\newblock In \emph{{ACL} 2010, Proceedings of the 48th Annual Meeting of the
  Association for Computational Linguistics, July 11-16, 2010, Uppsala,
  Sweden}, pages 1199--1208, 2010.

\bibitem[Sangati and van Cranenburgh(2015)]{sangati2015multiword}
Federico Sangati and Andreas van Cranenburgh.
\newblock Multiword expression identification with recurring tree fragments and
  association measures.
\newblock In \emph{{Proceedings of NAACL-HLT}}, pages 10--18, 2015.

\bibitem[Shutova et~al.(2010)Shutova, Sun, and Korhonen]{shutova2010metaphor}
Ekaterina Shutova, Lin Sun, and Anna Korhonen.
\newblock Metaphor identification using verb and noun clustering.
\newblock In \emph{{Proceedings of the 23rd International Conference on
  Computational Linguistics}}, pages 1002--1010. Association for Computational
  Linguistics, 2010.

\bibitem[Shutova et~al.(2017)Shutova, Sun, Guti{\'e}rrez, Lichtenstein, and
  Narayanan]{shutova2016multilingual}
Ekaterina Shutova, Lin Sun, Elkin~Dar{\'\i}o Guti{\'e}rrez, Patricia
  Lichtenstein, and Srini Narayanan.
\newblock Multilingual metaphor processing: Experiments with semi-supervised
  and unsupervised learning.
\newblock \emph{Computational Linguistics}, 43\penalty0 (1):\penalty0 71--123,
  2017.

\bibitem[Shwartz and Dagan(2016)]{shwartz}
Vered Shwartz and Ido Dagan.
\newblock Adding context to semantic data-driven paraphrasing.
\newblock In \emph{Proceedings of the Fifth Joint Conference on Lexical and
  Computational Semantics}, pages 108--113, Berlin, Germany, August 2016.
  Association for Computational Linguistics.

\bibitem[Socher et~al.(2011)Socher, Huang, Pennington, Ng, and Manning]{socher}
Richard Socher, Eric~H. Huang, Jeffrey Pennington, Andrew~Y. Ng, and
  Christopher~D. Manning.
\newblock Dynamic pooling and unfolding recursive autoencoders for paraphrase
  detection.
\newblock In \emph{Proceedings of the 24th International Conference on Neural
  Information Processing Systems}, NIPS’11, page 801–809, Red Hook, NY,
  USA, 2011. Curran Associates Inc.
\newblock ISBN 9781618395993.

\bibitem[Speer and Havasi(2012)]{speer-havasi-2012-representing}
Robyn Speer and Catherine Havasi.
\newblock Representing general relational knowledge in {C}oncept{N}et 5.
\newblock In \emph{Proceedings of the Eighth International Conference on
  Language Resources and Evaluation ({LREC}'12)}, pages 3679--3686, Istanbul,
  Turkey, May 2012. European Language Resources Association (ELRA).
\newblock URL
  \url{http://www.lrec-conf.org/proceedings/lrec2012/pdf/1072_Paper.pdf}.

\bibitem[Stefanowitsch and Gries(2003)]{StefanowitschGries:2003}
Anatol Stefanowitsch and Stefan~Th. Gries.
\newblock Collostructions: Investigating the interaction between words and
  constructions.
\newblock \emph{International Journal of Corpus Linguistics}, 8\penalty0
  (2):\penalty0 209 -- 43, 2003.

\bibitem[Stefanowitsch and Gries(2008)]{StefanowitschGries:2008}
Anatol Stefanowitsch and Stefan~Th. Gries.
\newblock Corpora and grammar.
\newblock \emph{Corpus Linguistics}, 2008.

\bibitem[Sukhareva et~al.(2016)Sukhareva, Eckle-Kohler, Habernal, and
  Gurevych]{sukhareva2016crowdsourcing}
Maria Sukhareva, Judith Eckle-Kohler, Ivan Habernal, and Iryna Gurevych.
\newblock {Crowdsourcing a Large Dataset of Domain-Specific Context-Sensitive
  Semantic Verb Relations.}
\newblock In \emph{LREC}, 2016.

\bibitem[Toledo et~al.(2014)Toledo, Alexandropoupou, Chesney, Katrenko,
  Klockmann, Kokke, Kruit, and Winter]{Toledo}
Assaf Toledo, Stavroula Alexandropoupou, Sophie Chesney, Sophia Katrenko, Heidi
  Klockmann, Pepijn Kokke, Benno Kruit, and Yoad Winter.
\newblock Towards a semantic model for textual entailment.
\newblock In Cleo Condoravdi, Valeria de~Paiva, and Annie Zaenen, editors,
  \emph{Linguistic Issues in Language Technology vol. 9}. 2014.

\bibitem[Tomasello(2000)]{Tomasello:2000}
Michael Tomasello.
\newblock First steps toward a usage-based theory of language acquisition.
\newblock \emph{Cognitive Linguistics}, 11\penalty0 (1-2):\penalty0 61--82,
  2000.

\bibitem[Turney(2008)]{turney2008latent}
Peter~D Turney.
\newblock The latent relation mapping engine: Algorithm and experiments.
\newblock \emph{Journal of Artificial Intelligence Research (JAIR)},
  33:\penalty0 615--55, 2008.

\bibitem[Turney and Pantel(2010)]{TurneyPantel}
Peter~D. Turney and Patrick Pantel.
\newblock From frequency to meaning: Vector space models of semantics.
\newblock \emph{J. Artif. Int. Res.}, 37\penalty0 (1):\penalty0 141--188,
  January 2010.

\bibitem[Tutubalina(2015)]{tutubalina2015clustering}
Elena Tutubalina.
\newblock Clustering-based approach to multiword expression extraction and
  ranking.
\newblock In \emph{{Proceedings of NAACL-HLT}}, pages 39--43, 2015.

\bibitem[Vila et~al.(2014)Vila, Mart{\'i}, and Rodr{\'i}guez]{Vila}
M.~Vila, M.~A. Mart{\'i}, and H.~Rodr{\'i}guez.
\newblock "is this a paraphrase? what kind? paraphrase boundaries and typology.
  ".
\newblock pages 205--218, 2014.

\bibitem[Vila et~al.(2015)Vila, Bertran, Mart{\'i}, and
  Rodr{\'i}guez]{Vila2015}
Marta Vila, Manuel Bertran, M.~Ant{\`o}nia Mart{\'i}, and Horacio
  Rodr{\'i}guez.
\newblock Corpus annotation with paraphrase types: new annotation scheme and
  inter-annotator agreement measures.
\newblock \emph{Language Resources and Evaluation}, 49\penalty0 (1):\penalty0
  77--105, 2015.
\newblock ISSN 1574-0218.

\bibitem[Vrande\v{c}iundefined(2012)]{10.1145/2187980.2188242}
Denny Vrande\v{c}iundefined.
\newblock Wikidata: A new platform for collaborative data collection.
\newblock In \emph{Proceedings of the 21st International Conference on World
  Wide Web}, WWW ’12 Companion, page 1063–1064, New York, NY, USA, 2012.
  Association for Computing Machinery.
\newblock ISBN 9781450312301.
\newblock \doi{10.1145/2187980.2188242}.
\newblock URL \url{https://doi.org/10.1145/2187980.2188242}.

\bibitem[Wallace et~al.(2019)Wallace, Feng, Kandpal, Gardner, and
  Singh]{wallace-etal-2019-universal}
Eric Wallace, Shi Feng, Nikhil Kandpal, Matt Gardner, and Sameer Singh.
\newblock Universal adversarial triggers for attacking and analyzing {NLP}.
\newblock In \emph{Proceedings of the 2019 Conference on Empirical Methods in
  Natural Language Processing and the 9th International Joint Conference on
  Natural Language Processing (EMNLP-IJCNLP)}, pages 2153--2162, Hong Kong,
  China, November 2019. Association for Computational Linguistics.
\newblock \doi{10.18653/v1/D19-1221}.
\newblock URL \url{https://www.aclweb.org/anthology/D19-1221}.

\bibitem[Wang et~al.(2018)Wang, Singh, Michael, Hill, Levy, and
  Bowman]{wang-etal-2018-glue}
Alex Wang, Amanpreet Singh, Julian Michael, Felix Hill, Omer Levy, and Samuel
  Bowman.
\newblock {GLUE}: A multi-task benchmark and analysis platform for natural
  language understanding.
\newblock In \emph{Proceedings of the 2018 {EMNLP} Workshop {B}lackbox{NLP}:
  Analyzing and Interpreting Neural Networks for {NLP}}, pages 353--355,
  Brussels, Belgium, November 2018. Association for Computational Linguistics.
\newblock \doi{10.18653/v1/W18-5446}.
\newblock URL \url{https://www.aclweb.org/anthology/W18-5446}.

\bibitem[Wang et~al.(2019)Wang, Pruksachatkun, Nangia, Singh, Michael, Hill,
  Levy, and Bowman]{NIPS2019_8589}
Alex Wang, Yada Pruksachatkun, Nikita Nangia, Amanpreet Singh, Julian Michael,
  Felix Hill, Omer Levy, and Samuel Bowman.
\newblock Superglue: A stickier benchmark for general-purpose language
  understanding systems.
\newblock In \emph{Advances in Neural Information Processing Systems 32}, pages
  3261--3275. Curran Associates, Inc., 2019.
\newblock URL
  \url{http://papers.nips.cc/paper/8589-superglue-a-stickier-benchmark-for-general-purpose-language-understanding-systems.pdf}.

\bibitem[Wang et~al.(2016)Wang, Mi, and Ittycheriah]{wang}
Zhiguo Wang, Haitao Mi, and Abraham Ittycheriah.
\newblock Sentence similarity learning by lexical decomposition and
  composition.
\newblock \emph{CoRR}, abs/1602.07019, 2016.
\newblock URL \url{http://arxiv.org/abs/1602.07019}.

\bibitem[Wible and Tsao(2010)]{Wible:2010}
David Wible and Nai-Lung Tsao.
\newblock {StringNet As a Computational Resource for Discovering and
  Investigating Linguistic Constructions}.
\newblock In \emph{{Proceedings of the NAACL HLT Workshop on Extracting and
  Using Constructions in Computational Linguistics}}, EUCCL '10, pages 25--31,
  Stroudsburg, PA, USA, 2010. Association for Computational Linguistics.

\bibitem[Williams et~al.(2018)Williams, Nangia, and Bowman]{mnli}
Adina Williams, Nikita Nangia, and Samuel Bowman.
\newblock A broad-coverage challenge corpus for sentence understanding through
  inference.
\newblock In \emph{Proceedings of the 2018 Conference of the North {A}merican
  Chapter of the Association for Computational Linguistics: Human Language
  Technologies, Volume 1 (Long Papers)}, pages 1112--1122, New Orleans,
  Louisiana, June 2018. Association for Computational Linguistics.
\newblock \doi{10.18653/v1/N18-1101}.
\newblock URL \url{https://www.aclweb.org/anthology/N18-1101}.

\bibitem[Wittgenstein(1953)]{wittgenstien53english}
Ludwig Wittgenstein.
\newblock \emph{Philosophical Investigations. (Translated by Anscombe,
  G.E.M.)}.
\newblock Basil Blackwell, 1953.

\bibitem[Wray and Perkins(2000)]{Wray:2000}
Alison Wray and Mick Perkins.
\newblock The functions of formulaic language: an integrated model.
\newblock \emph{Language and Communication}, 20\penalty0 (1):\penalty0 1--28,
  2000.

\bibitem[Yager(1992)]{yager92}
Ronald Yager.
\newblock Default knowledge and measures of specificity.
\newblock 61:\penalty0 1--44, 04 1992.

\bibitem[Yimam and Biemann(2018)]{yimam-biemann-2018-par4sim}
Seid~Muhie Yimam and Chris Biemann.
\newblock {P}ar4{S}im {--} adaptive paraphrasing for text simplification.
\newblock In \emph{Proceedings of the 27th International Conference on
  Computational Linguistics}, pages 331--342, Santa Fe, New Mexico, USA, August
  2018. Association for Computational Linguistics.
\newblock URL \url{https://www.aclweb.org/anthology/C18-1028}.

\bibitem[Yimam and Gurevych(2013)]{webanno}
Seid~Muhie Yimam and Iryna Gurevych.
\newblock Webanno: A flexible, web-based and visually supported system for
  distributed annotations.
\newblock In \emph{In Proceedings of ACL-2013 System Demonstrations}, pages
  1--6, 2013.

\bibitem[Yokote et~al.(2011)Yokote, Tanaka, and Ishizuka]{yokote2011effects}
Ken-ichi Yokote, Shohei Tanaka, and Mitsuru Ishizuka.
\newblock {Effects of Using Simple Semantic Similarity on Textual Entailment
  Recognition.}
\newblock In \emph{TAC}, 2011.

\bibitem[Zuidema(2006)]{zuidema2006productive}
Willem Zuidema.
\newblock What are the productive units of natural language grammar?: a dop
  approach to the automatic identification of constructions.
\newblock In \emph{{Proceedings of the Tenth Conference on Computational
  Natural Language Learning}}, pages 29--36. Association for Computational
  Linguistics, 2006.

\end{thebibliography}

\appendix

\chapter{Annotation Guidelines for ETPC}\label{a_etpc}

\section{Presentation}\label{a:etpc:p}

This document sets out the guidelines for the paraphrase typology annotation task, using the Extended Paraphrase Typology. 
The task consists of annotating candidate paraphrase pairs (including positive and negative examples of paraphrasing) with a 
textual paraphrase label, the paraphrase types they contain, and negation. These guidelines have been used to annotate the 
Microsoft Research Paraphrase Corpus (MRPC), thus giving raise to the Extended Typology Paraphrase Corpus (ETPC). For 
the purpose of the annotation, we have developed a web based annotation tool, the WARP-Text interface.

This document is divided in five blocks: general considerations about the task and theoretical definitions (Section \ref{a:etpc:t});
tagset definition (section \ref{a:etpc:td}); guidelines for annotating non-paraphrases (section \ref{a:etpc:np}); annotating negation (section \ref{a:etpc:neg}).

Marks and symbols used in this document:
\begin{itemize}

	\item Fagments in the examples that should be annotated are \underline{underlined}.\\
	When no fragment is underlined, it means that it is the whole example that should be tagged.
	\item The so-called ``key elements'' are in \textbf{bold}.
\end{itemize}

\subsection{Credits}

This document has been adapted and extended from the paraphrase typology
annotation guidelines of Vila and Marti (2012).

\section{The task}\label{a:etpc:t}

\textbf{Paraphrasing} stands for sameness of meaning between different wordings. For
example, the pair of sentences in (a) are different in form but have the same
meaning. Our \textbf{paraphrase typology (ETPC)} classifies paraphrases according to the
linguistic nature of this difference in wording.

\begin{itemize}
\item[a)] John said ``I like candies''/John said that he liked sweets.
\item[b)] John said ``I like candies''/John said that he liked onion.
\end{itemize}

The task described in these guidelines consists of annotating  a Paraphrase Identification 
corpus (MRPC) with the Extended Paraphrase Typology (EPT). A Paraphrase Identification
corpus contains textual paraphrase pairs (ex.: (a)), as well as textual non-paraphrase 
pairs (ex.: (b)). Our annotation task consists of two sub-tasks:

\textbf{Annotating atomic paraphrases} within textual paraphrase pairs (a) and textual
non-paraphrase pairs (b). The textual pairs are generally complex in the sense that they 
contain multiple atomic paraphrases. We call these atomic paraphrases paraphrase
phenomena and they are what should be annotated with the typology. The
paraphrase pair in (a) contains two paraphrase phenomena: the direct/indirect
style alternation and a synonymy substitution.

\textbf{Annotating atomic non-paraphrases} within textual non-paraphrase pairs (b).
The non-paraphrase pair in (b) contains one atomic non-paraphrase: the substitution
of ``candies'' with ``onion''.

In the annotation process, three main decisions should be made: 
\begin{itemize}
\item[1)]determine whether a candidate pair is a textual paraphrase (Section \ref{a:etpc:t:binary})
\item[2)]If \textbf{non-paraphrase}, determine the key differences between the two texts: \\
	- choose the tag that best describes the phenomenon behind each difference (Section \ref{a:etpc:t:tag})\\
	- determine the scope of every atomic non-paraphrase (Section \ref{a:etpc:t:scope})
\item[3)]Determine the similarities between the two texts:\\
	- choose the tag that best describes the phenomenon behind each similarity  (Section \ref{a:etpc:t:tag})\\
	- determine the scope of every atomic paraphrase (Section \ref{a:etpc:t:scope})
\end{itemize}

\subsection{Is This a Paraphrase Pair}\label{a:etpc:t:binary}

	The first step in the annotation process is determining whether a candidate
	paraphrase pair is actually a paraphrase. We consider paraphrases those pairs
	having the same or an equivalent propositional content. We consider non-paraphrases
	those pairs that have substantial difference in the propositional content. For example, 
	a) will be annotated as ``paraphrases'', while b) will be annotated as ``non-paraphrases''.

\begin{itemize}
\item[a)] Amrozi accused his brother, whom he called "the witness", of deliberately distorting his
evidence. \\ Referring to him as only "the witness", Amrozi accused his brother of
deliberately distorting his evidence.
\item[b)] Yucaipa owned Dominick's before selling the chain to Safeway in 1998 for \$2.5 billion. \\
Yucaipa bought Dominick's in 1995 for \$693 million and sold it to Safeway for \$1.8 billion
in 1998.

\end{itemize}

	Since the Extended Paraphrase Typology (ETP) can annotate atomic paraphrases (similarities)
	as well as atomic non-paraphrases (dissimilarities), both textual paraphrases and textual 
	non-paraphrases will be subsequently annotated with the paraphrase typology. The subsequent 
	annotation with paraphrase types will allow for distinguishing between paraphrase and 
	non-paraphrase fragments within these sentences.

\subsection{The Tagset}\label{a:etpc:t:tag}

	Our tagset is based on the Extended Paraphrase Typology shown in Table \ref{et:tab}. It is organized in seven meta 
	categories: ``Morphology'', ``Lexicon'', ``Lexico-syntax'', ``Syntax'', ``Discourse'', ``Other'', and 
	``Extremes''. Sense Preserving (Sens Pres.) shows whether a certain type can give raise 
	to textual paraphrases (+), to textual non-paraphrases (-), or to both (+ / -). The typology 
	contains 25 atomic paraphrase types (+) and 13 atomic non-paraphrase types (-).

The subclasses (morphology, lexicon, syntax and discourse based changes) follow
the classical organisation in formal linguistic levels from morphology to discourse.
Our paraphrase types are grouped in classes according to the nature of the
underlying linguistic mechanism: (i) those types where the paraphrase arises at the
morpho-lexicon level, (ii) those that are the result of a different structural
organization and (iii) those types arising at the semantics level. Although the class
stands for the trigger change, paraphrase phenomena in each class can entail
changes in other parts of the sentence.
For instance, a morpho-lexicon based change (derivational) like the one in (a),
where the verb \textit{failed} is exchanged for its nominal form \textit{failure}, has obvious
syntactic implications; however, the paraphrase is triggered by the morphological
change. A structure based change (diathesis) like the one in (b) entails an
inflectional change in \textit{hear/was heard} among others. Finally, paraphrases in
semantics are based on a different distribution of semantic content across the
lexical units with, on many occasions, a complete change in the form (c).

		\begin{table}[H]
		\begin{center}
		\footnotesize	
		\begin{tabular}{| l | l | c |}
		\hline
		\textbf{ID} & \textbf{Type} & \textbf{\begin{tabular}[x]{@{}c@{}}Sense\\Pres.\end{tabular}} \\
		\hline \hline
		\multicolumn{3}{|c|}{Morphology-based changes} \\
		\hline \hline
		1 & Inflectional changes & + / - \\
		\hline
		2 & Modal verb changes & + \\
		\hline
		3 & Derivational changes & + \\
		\hline \hline
		\multicolumn{3}{|c|}{Lexicon-based changes} \\
		\hline \hline
		4 & Spelling changes & + \\
		\hline
		5 & Same polarity substitution (habitual) & + \\
		\hline
		6 & Same polarity substitution (contextual) & + / - \\
		\hline
		7 & Same polarity sub. (named entity) & + / - \\
		\hline
		8 & Change of format & + \\
		\hline \hline
		\multicolumn{3}{|c|}{Lexico-syntactic based changes} \\
		\hline \hline
		9 & Opposite polarity sub. (habitual) & + / - \\
		\hline
		10 & Opposite polarity sub. (contextual) & + / - \\
		\hline
		11 & Synthetic/analytic substitution & + \\
		\hline
		12 & Converse substitution & + / - \\
		\hline \hline
		\multicolumn{3}{|c|}{Syntax-based changes} \\
		\hline \hline
		13 & Diathesis alternation & + / - \\
		\hline
		14 & Negation switching & + / - \\
		\hline
		15 & Ellipsis & + \\
		\hline
		16 & Coordination changes & + \\
		\hline
		17 & Subordination and nesting changes & + \\
		\hline \hline
		\multicolumn{3}{|c|}{Discourse-based changes} \\
		\hline \hline
		18 & Punctuation changes & + \\
		\hline
		19 & Direct/indirect style alternations & + / - \\
		\hline
		20 & Sentence modality changes & + \\
		\hline
		21 & Syntax/discourse structure changes & + \\
		\hline \hline
		\multicolumn{3}{|c|}{Other changes} \\
		\hline \hline
		22 & Addition/Deletion & + / - \\
		\hline
		23 & Change of order & + \\
		\hline
		24 & Semantic (General Inferences) & + / - \\
		\hline \hline
		\multicolumn{3}{|c|}{Extremes} \\
		\hline \hline
		25 & Identity & + \\
		\hline
		26 & Non-Paraphrase & - \\
		\hline
		27 & Entailment & - \\
		\hline
		
		\end{tabular}

		\end{center}
		\caption{Extended Paraphrase Typology} \label{et:tab}
		\end{table}

\begin{itemize}
\item[a)] how the headmaster \underline{failed} / the \underline{failure} of the headmaster
\item[b)] We were able to hear the report of a gun on shore intermittently / the report
of a gun on shore was still heard at intervals
\item[c)] I'm guessing we \underline{won't be done for some time} / I've got a hunch that we \underline{'re
not} \underline{through with that game yet}
\end{itemize}

Miscellaneous changes comprise types not directly related to one single class.
Finally, in paraphrase extremes, two special cases of paraphrase phenomena
should be considered: they consist of the extremes of the paraphrase continuum,
which goes from the highest level of paraphrasability (identity) to the lowest limits
of the paraphrase phenomenon (entailment). Non-paraphrase fragments within
paraphrase pairs are also part of the class paraphrase extremes.

As some of the names of our types explicitly reflect (e.g. ADDITION / DELETION), they
are \textbf{bidirectional}: in a paraphrase pair, they can be applied from the first member
of the pair to the second and vice versa.

	ETP contains both "sense preserving" atomic phenomena (atomic paraphrases)
	and "non sense preserving" atomic phenomena (atomic non-paraphrases). While
	some phenomena are considered to (almost) always preserve the meaning (ex.: abbreviation,
	habitual same polarity substitution), other phenomena are not innately preserving the
	meaning and can lead both to paraphrasing and to non-paraphrasing at the textual level (ex.:
	In (d) and (e) the involved phenomena is the same - "inflectional change", however in (d)
	the two texts are paraphrases, while in (e) they are not). The ``sense preserving'' feature is
	required for the annotation of the ``non-paraphrases''. 

\begin{itemize}
\item[d]It was with difficulty that the course of \underline{streets} could
be followed. \\You couldn't even follow the path of the \underline{street}.
\item[e]You can't travel from Barcelona to Mallorca with the
\underline{boat}. \\underline{Boats} can't travel from Barcelona to Mallorca.
\end{itemize}

\subsection{The Scope}\label{a:etpc:t:scope}

The scope refers to the selection of the tokens to be annotated within each tag. In
what follows, we first define the type of units we are willing to annotate (Section 
\ref{a:etpc:t:s:units}), the criteria followed in the scope selection (Section \ref{a:etpc:t:s:criteria}) and when the
punctuation marks should be included (Section \ref{a:etpc:t:s:punct}).

\subsubsection{Kind of Units to Be Annotated}\label{a:etpc:t:s:units}

We annotate \textbf{linguistic units}, not strings that do not correspond to a full linguistic
unit. These linguistic units can go from the word to the (multiple-)sentence level.

In the paraphrase pair in (a), although a change takes place between the snippets
\textit{here by} and \textit{it is there in}, two paraphrase mappings have to be established
between \textit{here} and \textit{there} (1), and \textit{by virtue of} and \textit{in virtue of} (2), two different
pairs of linguistic units.

\begin{itemize}
\item[a)] \underline{Here} \textsubscript{1} \underline{by virtue of} \textsubscript{2} humanity's vestures. \\ 
It is \underline{there} \textsubscript{1} \underline{in virtue of} \textsubscript{2} the vesture of humanity in which it is clothed.
\end{itemize}

However, selecting full linguistic units is not always possible or adequate from the
paraphrase annotation point of view. In the following, we set out some exceptions
to the above rule:

\textbf{1.} Cases in which \textbf{only one member of the paraphrase pair corresponds to a
linguistic unit}. In (b), a \textsc{SEMANTICS BASED CHANGE} occurs between the underlined
fragments. In the first sentence, it consists in a full linguistic unit, namely a causal
clause; in the second sentence, the semantic content in the first appears divided
into a nominal phrase and part of a verbal phrase, i.e., the verb \textit{has impressed}.
This nominal phrase plus the verb, although they do not constitute a full linguistic
unit, are the scope of the phenomenon in the second sentence

\begin{itemize}
\item[b)] There is a pattern of regularity and order in the entire cosmos, \underline{due to
some} \underline{hints that science provides us.}\\
\underline{A presiding mind has impressed} the stamp of order and regularity upon
the whole cosmos.
\end{itemize}

\textbf{2.} Cases in which \textbf{none of the members of the paraphrase pair correspond to
a linguistic unit}. The prototypical example of this situation are contractions,
within the \textsc{SPELLING} tag. In (c), \textit{I} constitutes a nominal phrase and \textit{will} is part of a
verbal phrase. As the contraction is produced between these two pieces, they and
only they constitute the scope of the phenomenon.

\begin{itemize}
\item[c)] \underline{I will} go to the cinema. \\ 
\underline{I'll} go to the cinema.
\end{itemize}

\textbf{3.} Cases of \textbf{identical} (see Section \ref{a:etpc:t:s:criteria})

\subsubsection{Scope Annotation Criteria}\label{a:etpc:t:s:criteria}

The way the scope should be annotated depends on the class of the tag. Three
criteria should be followed:

\textbf{1. Morpho-lexicon based changes, semantics based changes and
miscellaneous changes:} only the linguistic units affected by the trigger change
are tagged.

\begin{itemize}
\item[a)] I dislike rash \underline{motorists} . \\
I dislike rash \underline{drivers} . 
\item[b)] He \underline{rarely} makes us smile . \\
He \underline{has little to do with} making us smile .
\end{itemize}

\textbf{2. Structure based changes:} the whole linguistic unit suffering the syntactic or
discourse reorganization is tagged (light green rectangle in Figure 2). If the
reorganization takes place within a phrase, the phrase is tagged. If the
reorganization takes place within a clause, the clause is tagged. If the
reorganization takes place within a sentence, the sentence is tagged. If the
reorganization takes place between different phrases/clauses/sentences (mainly
coordination and subordination phenomena), all and only the
phrases/clauses/sentences affected are tagged. In the case of clause changes, if
the reorganizations takes place within the subordinate clause, only this one is
annotated (not the main clause) and vice versa.

Moreover, all structure based changes (except from diathesis alternations) have a
\textbf{key element} that gives rise to the change and/or distinguishes it from others. 
This key element is also annotated. First, the
whole linguistic unit (including the key element) is tagged, and then the key
element is annotated independently.

In (d), an active/passive alternation takes place (DIATHESIS tag). As the change
takes place within the subordinate clause, only this clause is tagged. In (e), a
change in the subordination form takes place (SUBORDINATION \& NESTIG tag). As the
change affects the way the two clauses (the main and the subordinate) are
connected, the whole sentence is tagged. The connective mechanisms (the
conjunction and the gerund clause) are annotated as key elements.

\begin{itemize}
\item[d)] \underline{When she sings that song}, everything seems possible. \\
\underline{When that song is sang}, everything seems possible.
\item[e)] \textbf{When} we hear that song, everything seems possible. \\
\textbf{Hearing that song}, everything seems possible.
\end{itemize}

\textbf{3. Entailment and non-paraphrase tags:} the affected linguistic unit is tagged.
The example in (f) is a case of ENTAILMENT; the example in (g) is a NON-PARAPHRASE.

\begin{itemize}
\item[f)] Google \underline{was in talks to buy} YouTube. \\
Google \underline{bought} YouTube.
\item[g)] Mary and Wendy \underline{went to the cinema} .\\
Mary and Wendy \underline{like each other} .
\end{itemize}

\textbf{4. Identical tag:}
Once all other phenomena are annotated, snippets which are identical in
both sentences may remain. We should annotate as IDENTICAL these \textbf{snippet
(not linguistic unit)} residues (h). In this case, we do not follow the
linguistic unit criteria (Section \ref{a:etpc:t:s:units}).

\textbf{Only one (discontinuous) identical tag} will be used in each pair of
sentences.

\textbf{Punctuation marks} will also be annotated as IDENTICAL if they effectively
are.

\begin{itemize}
\item[h)] \underline{The two} argued \underline{that only a new board would have had the credibility
to}\\ \underline{restore El Paso to health.} \\
\underline{The two} believed \underline{that only a new board would have had the
credibility to}\\ \underline{restore El Paso to health.}
\end{itemize}

Finally, it should be noted that tags overlap on many occasions. In (i), a SAMEPOLARITY
tag overlaps an ORDER one.

\begin{itemize}
\item[i)] shaking his head \underline{wisely} .\\
\underline{sagely} shaking his head.
\end{itemize}

\subsubsection{Should Punctuation Marks Be Included?}\label{a:etpc:t:s:punct}

When a whole phrase/cause/sentence is annotated, \textbf{the closing (and opening)
punctuation mark (if any) is(are) included}. Some examples are (a) and (b), which
are cases of DIATHESIS and ADDITION/ELETION, respectively. In contrast, in (c) and (d),
the commas are not included as they are not the opening and closing punctuation
marks of the paraphrase phenomenon tagged (SAME-POLARITY), but of a bigger unit.

\begin{itemize}
\item[a)] This song \underline{(John sang it last year in the festival)} will be a great success.\\
This song \underline{(it was sung by John last year in the festival)} will be a great success.
\item[b)] His judgment have kept equal pace in that conclusion. \\ 
His judgment and interest may \underline{, however ,} have kept equal pace in that conclusion.
\item[c)] Before leaving and before \underline{saying goodbye} , I looked around. \\
Before leaving and before \underline{the bye bye moment} , I looked around.
\item[d)] My sisters, \underline{lovely} girls, live in Melbourne. \\
My sisters, \underline{nice} girls, live in Melbourne.
\end{itemize}

\section{Tagset Definition}\label{a:etpc:td}

In the following, the annotation specifics are presented. For each tag, we provide
(1) the definition and (2) examples both for ``positive sense preserving'' and ``negative 
sense preserving'' instances, where applicable. 

\subsection{Morphology based changes}

Morphology based changes stand for those paraphrases that take place at the
morphology level of language. Some changes in this class arise at the morphology
level, but entail significant structural implications in the sentence. Only the linguistic 
unit affected by the trigger morphology change is annotated. 

\subsubsection{Inflectional changes}

\textbf{Definition:} 
Inflectional changes consist in changing inflectional affixes of words. In the case
of verbs, this type includes all changes within the verbal paradigm.
\textbf{Negative sense preserving} inflectional changes lead to significant changes in 
the meaning of the whole text, thus giving raise to non-paraphrases.

\begin{itemize}

\item Positive sense preserving:\\
	It was with difficulty that the course of \textbf{streets} could be followed.\\
	You couldn't even follow the path of the \textbf{street}.
\item Negative sense preserving:\\
	You can't travel from Barcelona to Mallorca with the \textbf{boat}.\\
	Boats can't travel from Barcelona to Mallorca.

\end{itemize}

\subsubsection{Modal verb changes}

\textbf{Definition:} The MODAL VERB tag stands for changes of modality using modal verbs.

\begin{itemize}
\item Positive sense preserving:\\
	I was still lost in conjectures who they \textbf{might} be.\\
	I was pondering who they \textbf{could} be.
\end{itemize}

\subsubsection{Derivational changes}

\textbf{Definition:} The DERIVATIONAL tag stands for changes of category by adding derivational affixes
to words. These changes comprise a syntactic reorganization in the sentence where
they occur.

\begin{itemize}
\item Positive sense preserving:\\
	I have heard many accounts of him all \textbf{differing} from each other.\\
	I have heard many \textbf{different} things about him.
\end{itemize}

Although drivers and driving are linked by a derivational process, in the following example 
this type is classified as SAME-POLARITY, and not as a DERIVATIONAL, because there is 
not an actual change of category, both are acting as nouns. 
\begin{itemize}
\item{} I dislike rash \underline{drivers}. \\ 
I dislike rash \underline{driving}.
\end{itemize}

\subsection{Lexicon based changes}

Lexicon based change tags stand for those paraphrases that arise at the lexical
level. 

Always the \textbf{smallest} possible lexical unit has to be annotated. In (a), we should not
consider one single paraphrase phenomenon because it can be divided into two
lexical units pairs: often-debated/much-disputed (1) and issue/question (2). These
SAME-POLARITY substitutions are independent paraphrase phenomena, as we could
substitute often-debated by much-disputed, leaving issue unchanged (much-disputed
issue). Thus, two different SAME-POLARITY tags
should be used. In contrast, in (b), lies and is revealed should not be tagged on
their own as SAME-POLARITY substitutions, as they are semantically embedded in the
wider lexical units lies its appeal and its appeal is revealed, respectively. The tag
used in this case is, again, SAME-POLARITY.

\begin{itemize}
\item[a)] \underline{often-debated}\textsubscript{1} \underline{issue}\textsubscript{2} \\ 
\underline{much-disputed}\textsubscript{1} \underline{question}\textsubscript{2}
\item[b)] Here by virtue of humanity's vestures, \underline{lies its appeal} . \\
Here by virtue of humanity's vestures, \underline{its appeal is revealed} .
\end{itemize}

Auxiliaries and infinitive marks are not tagged within the lexical unit in question.
Only the verb to be, when it is part of a passive voice, should be included in the
scope (c).

\begin{itemize}
\item[c)] The viewpoint of these lands had \underline{been altered} . \\ 
The whole aspect of the land had \underline{changed}.
\end{itemize}

\subsubsection{Spelling changes}

\textbf{Definition:} This type comprises spelling changes and changes in the lexical form in general.
Spelling is always sense preserving. Some examples:

\textbf{1. Spelling}

\begin{itemize}
\item[a)] color / colour
\end{itemize}

\textbf{2. Acronyms}

\begin{itemize}
\item[b)] North Atlantic Treaty Organization / NATO
\end{itemize}

\textbf{3. Abbreviations}

\begin{itemize}
\item[c)] Mister / Mr.
\end{itemize}

\textbf{4. Contractions}

\begin{itemize}
\item[d)] you have / you've
\end{itemize}

\textbf{5. Hyphenation}

\begin{itemize}
\item[e)] flow-accretive / flow accretive
\end{itemize}

\subsubsection{Same Polarity Substitution}

\textbf{Definition: }The SAME-POLARITY tag is used when a lexical unit is changed for another one with
approximately the same meaning. Both lexical (a) and functional (b) units are
considered within this type. Sameness of category is not a requisite to belong to
this type (c).

\begin{itemize}
\item[a)] The \underline{pilot} took off despite the stormy weather . \\
The \underline{plane} took off despite the stormy weather .
\item[b)] \underline{Despite} the stormy weather \\ 
\underline{In spite of} the stormy weather
\item[c)] He \underline{rarely} makes us smile . \\
He \underline{has little to do with} making us smile.
\end{itemize}

When prepositions are part of a larger lexical unit, changes or deletions of
these prepositions are tagged as SAME-POLARITY and annotated together with the
lexical unit where they are embedded (d).

\begin{itemize}
\item[d)] do away / do away with
\end{itemize}

SAME-POLARITY may be used to tag several linguistic mechanisms, the following
among them:

\textbf{1. Synonymy}
\begin{itemize}
\item[e)] I like your \underline{house} .\\ 
I like your \underline{place} .
\end{itemize}

\textbf{2. General/specific}
\begin{itemize}
\item[f)] I dislike rash \underline{motorists} . \\
I dislike rash \underline{drivers} .
\end{itemize}

\textbf{3. Exact/approximate}
\begin{itemize}
\item[g)] They were \underline{9} . \\
They were \underline{around 10} .
\end{itemize}

\textbf{4. Metaphor}
\begin{itemize}
\item[h)] I was staring at her shinning \underline{teeth} . \\
I was staring at her shinning \underline{pearls} .
\end{itemize}

\textbf{5. Metonymy}
\begin{itemize}
\item[i)] I read \underline{a book written by Shakespeare} . \\
I read \underline{a Shakespeare}
\end{itemize}

\textbf{6. Expansion/compression:} expressing the same content with multiple pieces
and/or in a more detailed way.
\begin{itemize}
\item[j)] Ended up causing a \underline{calm} aura \\
Caused a \underline{rather sober and subdued} air
\end{itemize}

\textbf{7. Word/definition}
\begin{itemize}
\item[k)] \underline{Heart attacks} have experienced an increase in the last decades.\\
\underline{Sudden coronary thromboses} have experiences an increase in the last decades.
\end{itemize}

\textbf{8. Translation}
\begin{itemize}
\item[l)] Jean-Francois Revel, in \underline{History of the Western Philosophy} \\
Jean-Francois Revel, in \underline{Histoire de la philosophie occidentale}
\end{itemize}

\textbf{9. Idiomatic expressions}
\begin{itemize}
\item[m)] It is \underline{raining cats and dogs} . \\
It is \underline{raining a lot} .
\end{itemize}

\textbf{10. Part/whole}
\begin{itemize}
\item[n)] Yesterday I cut my \underline{finger} . \\
 Yesterday I cut my \underline{hand}
\end{itemize}

In the EPT, we distinguish between \textbf{three different kinds same-polarity substitution}:
habitual, contextual, and named entity. The kind of same-polarity substitution depends on
the nature of the relation between the substituted text.

\myparagraph{Same Polarity Substitution (habitual)}

The SAME-POLARITY (HABITUAL) tag is used when a lexical unit is changed for another one 
with approximately the same \textbf{dictionary} meaning. The substituted units have a similar
meaning outside of the particular context as well as within the context. Same-polarity (habitual)
is always \textbf{sense preserving}:

\begin{itemize}
\item Positive sense preserving:\\
A federal \underline{magistrate} in Fort Lauderdale ordered him held without bail.\\
Zuccarini was ordered held without bail Wednesday by a federal \underline{judge} in Fort
Lauderdale, Fla.
\end{itemize}

\myparagraph{Same Polarity Substitution (contextual)}

The SAME-POLARITY (CONTEXT) tag is used when a lexical unit is changed for another one 
with approximately the same meaning \textbf{within the given context}. The substituted units
have different out-of-context meaning. The negative sense preserving SAME-POLARITY is
always contextual (unless it requires named entity reasoning).
In the case of \textbf{negative sense preserving} same polarity substitution, the meaning of the 
units is similar, but not the same - it includes key differenced and/or incompatibilities that give 
raise to non-paraphrasing at the level of the two texts. 

\begin{itemize}
\item Positive sense preserving:\\
Meanwhile, the global death toll \underline{approached} 770 with more than 8,300 people
sickened since the severe acute respiratory syndrome virus first appeared in southern
China in November.\\
The global death toll from SARS \underline{was} at least 767, with more than 8,300 people
sickened since the virus first appeared in southern China in November.
\item Negative sense preserving: \\
The loonie, meanwhile, continued to \underline{slip} in early trading Friday. \\
The loonie, meanwhile, \underline{was on the rise} again early Thursday. 
\end{itemize}

\myparagraph{Same Polarity Substitution (Named Entity)}

The SAME-POLARITY (NE) tag is used when a lexical unit is changed for another one 
with approximately the same meaning. Both replaced units are \textbf{named entities or
properties of named entities}. Some degree of world knowledge and named entity reasoning
is required to correctly determine whether the substitution is sense preserving or not.
In the case of \textbf{negative sense preserving} same polarity substitution, the meaning of the 
units is similar, but not the same - it includes key differenced and/or incompatibilities that give 
raise to non-paraphrasing at the level of the two texts.

\begin{itemize}
\item Positive sense preserving:\\
He told The Sun newspaper that \underline{Mr. Hussein}'s daughters had British schools and
hospitals in mind when they decided to ask for asylum. \\
\underline{Saddam} 's daughters had British schools and hospitals in mind when they decided to
ask for asylum -- especially the schools, he told The Sun.
\item Negative sense preserving: \\
Yucaipa owned Dominick's before selling the chain to Safeway in 1998 for \underline{\$2.5 billion} .\\
Yucaipa bought Dominick's in 1995 for \$693 million and sold it to Safeway for \underline{\$1.8 billion} in 1998.
\end{itemize}

\subsubsection{Change of Format}

\textbf{Definition:} This tag stands for changes in the format. Format is always sense preserving.
Some examples:

\textbf{1. Digits/in letters}
\begin{itemize}
\item[a)] 12 / twelve
\end{itemize}
\textbf{2. Case changes}
\begin{itemize}
\item[b)] Chapter 3 / CHAPTER 3
\end{itemize}
\textbf{3. Format changes}
\begin{itemize}
\item[c)] 03/08/1984 / Aug 3 1984
\end{itemize}

\subsection{Lexico-syntactic based changes}

Lexico-syntactic based change tags stand for those paraphrases that arise at the lexical
level but are also entailing significant structural implications in the sentence. 
Similar to lexicon changes always the \textbf{smallest} possible lexical unit has to be 
annotated.

\subsubsection{Opposite polarity substitution}

\textbf{Definition:} OPPOSITE-POLARITY stands for changes of one lexical unit for another one with
opposite polarity. In order to maintain the same meaning, other changes have to
occur. Two phenomena are considered within this type:

\textbf{1. Double change of polarity}
A lexical unit is changed for its antonym or complementary. In order to maintain
the same meaning, a double change of polarity has to occur within the same
sentence: another antonym (a) or complementary substitution (b), or a negation
(c). In the case of double change of polarity, the two changes of polarity have to be
tagged as a single (and possibly discontinuous, like in b) phenomenon and using a
single tag.

\begin{itemize}
\item[a)] John \underline{lost interest in} the endeavor . \\
John \underline{developed disinterest} in the endeavor .
\item[b)] \underline{Only 20\%} of the students were \underline{late} . \\
\underline{Most} of the students were \underline{on time} .
\item[c)] He \underline{did not succeed} in either case . \\
He \underline{failed} in both enterprises .
\end{itemize}

\textbf{2. Change of polarity and argument inversion}
An adjective is changed for its antonym in comparative structures. In order to
maintain the same meaning, an argument inversion has to occur (d).
In the case of change of polarity and argument inversion, only the antonym
adjectives are tagged.

\begin{itemize}
\item[d)] The neighboring town is \underline{poorer} in forest resources than our town. \\
Our town is \underline{richer} in forest resources than the neighboring town.
\end{itemize}

In the EPT, we distinguish between \textbf{two different kinds opposite-polarity substitution}:
habitual and contextual. The kind of opposite-polarity substitution depends on
the nature of the relation between the substituted text.

\myparagraph{Opposite polarity substitution (habitual)}

The OPP-POLARITY (HABITUAL) tag is used when a lexical unit is changed for another one 
with approximately the opposite \textbf{dictionary} meaning. The substituted units have an opposite
meaning outside of the particular context as well as within the context. 
The \textbf{negative sense preserving} Opposite Polarity Substitution appears in two different situations. 
First, the case where the meaning of the two units is not completely opposite - it includes key 
differences and/or incompatibilities that give raise to non-paraphrasing at the level of the two texts. 
Second, the case where the meaning of the two units are the same, but the other changes 
(double change of polarity or argument inversion) are not found.

\begin{itemize}
\item Positive sense preserving:\\
Leicester \underline{failed} in both enterprises.\\
He \underline{did not succeed} in either case.
\item Negative sense preserving:\\
John \underline{loved} his new car.\\
He \underline{hated} that car.
\end{itemize}

\myparagraph{Opposite polarity substitution (contextual)}
The OPP-POLARITY (CONTEXT) tag is used when a lexical unit is changed for another one 
with approximately the opposite meaning \textbf{within the given context}. The substituted units
have different out-of-context meaning.
The \textbf{negative sense preserving} Opposite Polarity Substitution appears in two different situations. 
First, the case where the meaning of the two units is not completely opposite - it includes key 
differences and/or incompatibilities that give raise to non-paraphrasing at the level of the two texts. 
Second, the case where the meaning of the two units are the same, but the other changes 
(double change of polarity or argument inversion) are not found.

\begin{itemize}
\item Positive sense preserving:\\
A big surge in consumer confidence has \underline{provided} the only positive economic news in
recent weeks.\\
Only a big surge in consumer confidence has \underline{interrupted} the bleak economic news.
\item Negative sense preserving:\\
Johnson \underline{welcomed} the new proposal.\\
Johnson \underline{did not approve of} the new proposal.
\end{itemize}

\subsubsection{Synthetic/Analytic substitution}

\textbf{Definition:}SYNTHETIC/ANALYTIC stands for those changes of synthetic structures to analytic
structures and vice versa. It should be noted, however, that sometimes
``syntheticity'' or ``analyticity'' is a matter of degree. Consider examples (a) and (b).
In (a), we would probably consider as analytic the genitive structure. In (b), in
contrast, the genitive structure would probably be the synthetic one. Genitive
structures are not synthetic or analytic by definition, but more or less
synthetic/analytic compared to other structures. Thus, we could redefine this group
as a change in the degree of syntheticity/analyticity.

\begin{itemize}

\item[a)] the Met show / the Met's show
\item[b)] Tina's birthday / The birthday of Tina
\end{itemize}

SYNTHETIC/ANALYTIC is always \textbf{sense preserving} and comprises phenomena such as:

\textbf{1. Compounding/decomposition}
A compound is decomposed through the use of a prepositional phrase (a). The
alternation adjectival/prepositional phrase (b) and single word/adjective+noun
alternations (c) are also considered here.

\begin{itemize}
\item[a)] The gamekeeper preferred to make \underline{wildlife television documentaries} .\\
The gamekeeper preferred to make \underline{television documentaries about wildlife} .
\item[b)] \underline{Chemical life-cycles} of the sexes \\
\underline{Life-cycles for chemistry} for genders
\item[c)] One of his works holding the title "Liber Cosmographicus De Natura Locorum" belongs to a category of \underline{physiography} . \\
One of his works bearing the title of "Liber Cosmographicus De Natura Locorum" is a species of \underline{physical geography} .
\end{itemize}

\textbf{2. Alternations affecting genitives and possessives}
Alternations between genitive/prepositional phrases (d), possessive/prepositional
phrases (e), genitive/ nominal phrases (f), genitive/adjectival phrases (g), etc.
\begin{itemize}
\item[d)] Tina's birthday / The birthday of Tina
\item[e)] His reflection / The reflection of his own features
\item[f)] the Met show / the Met's show
\item[g)] Russia's Foreign Ministry / the Russian Foreign Ministry
\end{itemize}

\textbf{N.B.:} A distinction has to be established between this type and DERIVATIONAL. Some
DERIVATIONAL cases also contain genitive alternations (h), but these alternations are
part of a wider derivational change. In the cases of genitive alternations classified
as SYNTHETIC/ANALYTIC, the alternation is an isolated and independent phenomenon.

\begin{itemize}
\item[h)] Mary \underline{teaches} John . \\
 Mary is John's \underline{teacher} .
\end{itemize}

\textbf{N.B.:} Cases of 1 (compounding/decomposition) and 2 (alternation involving genitives
and possessives) in which the alternation takes place with a clause (with a verb)
are not considered here but in SUBORDINATION \& NESTING (i)
\begin{itemize}
\item[i)] Volcanoes \textbf{which} are now extinct / extinct volcanoes
\end{itemize}

\textbf{3. Synthetic/analytic superlative}
\begin{itemize}
\item[j)] He's \underline{smarter than everybody else} .\\
He's \underline{the smartest} .
\end{itemize}

\textbf{4. Light/generic element addition:} Changing a synthetic form A for an analytic
form BA by adding a more generic element (B is more generic than A). A has to
have the same lemma/stem in both member of the pair as in (k). Moreover,
although the category of the phrase A and the phrase BA may differ, the change
does not have structural consequences outside A or BA. In (l), although the
adverbial phrase \textit{cheerfully} is changed to the prepositional phrase \textit{in a cheerful
way}, the rest of the sentence remains unchanged. Finally, the order of the A and B
units can be BA (k) or AB (l).
\begin{itemize}
\item[k)] John \underline{boasted} about his work. \\ 
John \underline{spoke boastfully} about his work.
\item[l)] Marilyn carried on with her life \underline{cheerfully} . \\
Marilyn carried on with her life \underline{in a cheerful way} .
\end{itemize}

\textbf{N.B.: }When B is the verb to be and there is a change of category of A through a
derivational process, the phenomenon is tagged as DERIVATIONAL (m)
\begin{itemize}
\item[m)] Sister Mary was \underline{helpful} to Darrell . \\
Sister Mary \underline{helped} Darrell .
\end{itemize}

\textbf{5. Specifier addition:} This type is parallel to the previous one, but the added
element B is not more generic, but focuses on one of the components or
characteristics of A (n), emphasises A (o) or determines A (p).

\begin{itemize}
\item[n)] I had to drive through \underline{fog} to get there . \\
I had to drive through \underline{a wall of fog} to get there .
\item[o)] We are meeting at \underline{5} .
We are meeting at \underline{5 o'clock} .
\item[p)] \underline{Translation} is what they need . \\
\underline{The translation} is what they need .
\end{itemize}

\textbf{N.B.:} Contrary to SAME-POLARITY or SEMANTICS BASED CHANGES, where words vary from
one member of the paraphrase pair to the other, in synthetic/analytic substitutions
\begin{itemize}
\item although a change of category may take place, lexical word stems are
the same (1 and 2) or
\item a support element is added, but other lexical word stems are the same(4 and 5).
\end{itemize}

\subsubsection{Converse substitution}

\textbf{Definition:} A lexical unit is changed for its converse. In order to maintain the same meaning,
an argument inversion has to occur. The \textbf{negative sense preserving} converse substitution 
occurs when the arguments are not inverted.

\begin{itemize}
\item Positive sense preserving:\\
The Geological society of London in 1855 \underline{awarded to him} the Wollaston medal. \\
Resulted in him \underline{receiving} the Wollaston medal \underline{from} the Geological society in London
in 1855.
\item Negative sense preserving:\\
Last Monday, John \underline{bought} the new black car from his friend Sam. \\
Last week, John \underline{sold} his black car to Sam, his friend from high school.
\end{itemize}

\subsection{Syntax based changes}

Syntax based change tags stand for those changes that involve a syntactic
reorganization in the sentence. This type basically comprises changes within a
single sentence; and changes in the way sentences, clauses or phrases are
connected. 
The phrase/clause/sentence(s) suffering the modification is(are) tagged. All syntax
tags but DIATHESIS have key elements that should be annotated as well.

\subsubsection{Diathesis alternation}

\textbf{Definition: }DIATHESIS gathers the diathesis alternations in which verbs can participate.
The whole linguistic unit suffering the syntactic reorganization is tagged. The \textbf negative 
sense preserving} diathesis alternation occurs when the arguments are not properly changed or 
inverted.

\begin{itemize}
\item Positive sense preserving:\\
The guide \underline{drew our attention to} a gloomy little dungeon.\\
Our attention \underline{was drawn by} our guide \underline{to} a little dungeon.
\item Negative sense preserving:\\
The president \underline{gave} a speech about his plan to change the Constitution.\\
The president \underline{was given} a speech about his plan to change the Constitution.
\end{itemize}

\subsubsection{Negation switching}
\textbf{Definition: }Changing the position of the negation within a sentence.
The whole linguistic unit suffering the modification is tagged (not only the negation
scope). Negation marks are tagged as key elements. The \textbf{negative sense 
preserving} negation switching occurs when the scope of negation in the two texts is 
significantly different and that changes the overall meaning. A special case of negative
sense preserving negation switching is when one of the texts (sentences) is affirmative, 
and the other is negative.

\begin{itemize}
\item Positive sense preserving:\\
In order to move us, \underline{it needs \textbf{no} reference to any recognized original}. \\
\underline{One does \textbf{not} need to recognize a tangible object} to be moved by its artistic representation.
\item Negative sense preserving:\\
Frege did \textbf{not} say that Hesperus is Phosphorus.\\
Frege said that Hesperus is \textbf{not} Phosphorus.
\end{itemize}

\subsubsection{Ellipsis}
\textbf{Definition: }This tag includes linguistic ellipsis, i.e., those cases in which the 
elided snippets can be recovered through linguistic mechanisms. In (a), in the first 
member of the pair the idea of ``being able to change to'' is expressed twice; in the 
second member of the pair it is only expressed once due to elision.
The whole linguistic unit suffering the modification is tagged (not only the elided
snippets). All appearances of the elided snippet in both sentences are tagged as
key elements: the idea of ``being able to change to'' in (a).
Ellipsis is always \textbf{sense preserving}.

\begin{itemize}
\item[a)] - Thus, chemical force \textbf{can become} electrical current and that current \textbf{can
change back} into chemical being. \\
- So we \textbf{can change} chemical force into the electric current, or the current
into chemical force.
\end{itemize}

\textbf{N.B.:} When the elided snippets cannot be recovered solely through linguistic
mechanisms, they must be considered DELETIONS.

\subsubsection{Coordination changes}
\textbf{Definition: } Changes in which one of the members of the pair contains coordinated snippets.
This coordination is not present (in (a) it changes to a juxtaposition) or changes its
position and/or form (b) in the other member of the pair. Only the coordinated or juxtaposed linguistic 
units are tagged. Only the coordination (not juxtaposition) marks are tagged as key elements.
Coordination changes are always \textbf{sense preserving}.

\begin{itemize}
\item[a)] I like pears \textbf{and} apples. \\ 
I like pears. I like apples
\item[b)] \underline{Older plans \textbf{and} contemporary ones} \\ 
\underline{Old \textbf{and} contemporary} plans
\end{itemize}

\textbf{N.B.: }When the alternation takes place between, on the one hand, coordinated or
juxtaposed units and, on the other hand, subordinated or nested units, the
phenomenon is tagged as SUBORDINATION \& NESTING.

\subsubsection{Subordination and Nesting changes}

\textbf{Definition: }Changes in which one of the members of the pair contains a subordination (a) or a
nesting (b). This subordination or nesting is not present (in (a) and (b) it changes
to a juxtaposition) or changes the position and/or form (c) in the other member of
the pair. Nesting is understood as a general term meaning that something is
embedded in a bigger unit. Only the linguistic units involved in the subordination or nesting, as well as the
coordinated and juxtaposed units, are tagged.
In case a conjunction, a relative pronoun or a preposition are present, they are
tagged as the key elements (a and c). In case they are not present, the whole
subordinated o nested snippet is tagged (b). Juxtaposition or coordination elements
are not tagged as key elements.
Subordination and Nesting changes are always \textbf{sense preserving}.

\begin{itemize}
\item[a)] A building, \textbf{which} was devastated by the bomb, was completely destroyed.\\
A building was devastated by the bomb. It was completely destroyed.
\item[b)] Patrick Ewing scored \textbf{a personal season high} of 41 points. \\
Patrick Ewing scored 41 points. It was a personal season high
\item[c)] The conference venue is in \underline{the building \textbf{whose} roof is red} . \\
The conference venue is in \underline{the building with red roof} .
\end{itemize}

\subsection{Discourse based changes}

These tags stand for those changes that take place at the discourse level of
language. This type gathers phenomena that are very different in nature, though all
having in common that consist in structural changes not affecting the argumental
elements in the sentence.
The phrase/clause/sentence(s) suffering the modification is(are) tagged. Moreover,
a key element should be tagged in all discourse based tags.

\subsubsection{Punctuation changes}

\textbf{Definition: } Changes in the punctuation (a). Cases consisting of linguistic mechanisms parallel
to punctuation like (b) are also considered here. 
The changing punctuation signs are tagged as key elements.
The whole linguistic unit(s) suffering the modification is(are) tagged. 
Punctuation is always \textbf{sense preserving}.

\begin{itemize}
\item[a)] This\textbf{,} as I see it\textbf{,} is wrong . \\
This \textbf{--} as I see it \textbf{--} is wrong.
\item[b)] - You will purchase a return ticket to Streatham Common and a platform
ticket at Victoria station . \\
- At Victoria Station you will purchase \textbf{(1)} a return ticket to Streatham
Common and \textbf{(2)} a platform ticket
\end{itemize}

Sometimes occurs that several changes in the punctuation take place at the same
time. These multiple changes are considered as a single phenomenon if they take
place at the same level (between phrase, between clause or between sentence),
like in (c). If they belong to different levels, they are tagged as separate
phenomena: two changes in the punctuation take place in (d), repeated in (e), but
they are annotated as independent paraphrase phenomena: one of them is tagged
in (d) and the other in (e).

\begin{itemize}
\item[c)] I know she is coming\textbf{.} She will be fine\textbf{;} I know it . \\
I know she is coming\textbf{;} she will be fine\textbf{.} I know it .
\item[d)] I need to buy a couple of things\textbf{.} Then, I will come . \\
I need to buy a couple of things\textbf{;} then I will come . 
\item[e)] I need to buy a couple of things. \underline{Then \textbf{,} I will come .} \\
I need to buy a couple of things. \underline{then I will come .} 
\end{itemize}

\subsubsection{Direct/Indirect style alternations}

\textbf{Definition: }Changing direct style for indirect style, and vice versa.
The whole linguistic unit suffering the modification is tagged. The conjunction in
the indirect style is tagged as key element. If no conjunction is present, the whole
subordinate clause is tagged. The \textbf{negative sense preserving} Direct/Indirect 
Style alternations do not trigger the appropriate changes for pronoun resolution.

\begin{itemize}
\item Positive sense preserving:\\
She is mine, said the Great Spirit. \\
The Great Spirit said \textbf{that} she is hers. 
\item Negative sense preserving:\\
\underline{I'm on my way!, said Peter} and hung up his phone .\\
Peter called Ana to \underline{tell her \textbf{that} she is on her way} .
\end{itemize} 

\subsubsection{Sentence modality changes}

\textbf{Definition: } Cases in which there is a change of modality (a). We are 
referring strictly to changes between affirmative, interrogative, exclamatory and 
imperative sentences. The whole unit suffering the modification is tagged. The 
elements that change are tagged as key elements. Modality change is always
\textbf{sense preserving}.

\begin{itemize}
\item[a)] \textbf{Can} I make a reservation? \\ 
I'\textbf{d like to} make a reservation.
\end{itemize}

\textbf{N.B.:} In MODAL VERB tags, in contrast, only modal verb alternations are involved.

\subsubsection{Syntax/Discourse Structure}

\textbf{Definition: } This tag is used to annotate other changes in the structure of the sentences not
considered in the syntax and discourse based tags above: (a), (b) and (c).
The linguistic unit(s) suffering the modification is(are) tagged. The elements that
change are tagged as key elements.

\begin{itemize}
\item[a)] John wore his best suit to the dance last night . \\
\textbf{It was} John \textbf{who} wore his best suit to the dance last night .
\item[b)] He wanted to eat \textbf{nothing but} apples . \\
\textbf{All} he wanted to eat \textbf{were} apples.
\item[c)] \textbf{You are very} courageous . \\
\textbf{You have shown how} courageous \textbf{you are} .
\end{itemize}

\subsection{Other changes}

This class gathers those changes that are related to more than one of the classes
and subclasses in our typology, as they can take place in any of them.

\subsubsection{Addition/Deletion}

\textbf{Definition: }Deletion of lexical and functional units. In the \textbf{negative sense preserving} 
case, the deletion leads to a significant modification of the meaning. Only the linguistic unit deleted 
is tagged. When a functional unit is deleted together with a lexical unit, this functional unit is included 
in the scope. Normally, the scope of Addition/Deletion is only in one of the two texts, as opposed to 
the other types, which are pairwise.

\begin{itemize}
\item Positive sense preserving: \\
\underline{One day,} she took a hot flat-iron, removed my clothes, and held it on my naked back until I howled with pain. \\
As a proof of bad treatment, she took a hot flat-iron and put it on my back after removing my clothes.
\item Negative sense preserving: \\
Legislation making it harder for consumers to erase their debts in bankruptcy court won overwhelming House approval in March. \\
Legislation making it harder for consumers to erase their debts in bankruptcy court
won speedy, House approval in March \underline{and was endorsed by the White} \\  \underline{House}.
\end{itemize}

\subsubsection{Change of order}

\textbf{Definition: }This tag includes any type of change of order from the word 
level to the sentence level: (a), (b) and (c). Change of order is always \textbf{sense preserving}.

\begin{itemize}
\item[a)] She used to \underline{only} eat hot dishes. \\
She used to eat \underline{only} hot dishes.
\item[b)] ``I want a beer'', \underline{he} said. \\
``I want a beer'', said \underline{he}.
\item[c)] They said : \underline{``We believe that the time has come for legislation to make} \\ \underline{public places smoke-free''} . \\ 
\underline{``The time has come to make public places smoke-free,''} they wrote in a letter to the Times newspaper. 
\end{itemize}

\subsubsection{Semantic (General Inferences)}

\textbf{Definition: }SEMANTICS BASED CHANGES tag stands for changes that imply a 
different lexicalisation pattern of the same content units.
Typically the semantic relation between the two can only be determined through (common 
sense) reasoning. In the \textbf{negative sense preserving} case, the reasoning identifies 
contradiction and/or incompatibility.

\begin{itemize}
\item Positive sense preserving: \\
Uncle Tarek was born Aribert Ferdinand Heim.\\
The real name of Tarek Hussein Farid was Aribert Ferdinand Heim.
\item Negative sense preserving: \\
Children were among the victims of a plane crash that killed as many as 17 people
Sunday in Butte, Montana. \\
17 adults died in a plane crash in Montana.
\end{itemize}

\subsection{Extremes}

The following types stand for the extremes of the paraphrase continuum: identity
on the one hand, and entailment and non-paraphrase on the other.

\subsubsection{Identity}

\textbf{Definition:} We annotate as IDENTICAL those linguistic units that are exactly the 
same in wording. Identical is always \textbf{sense preserving}.

\begin{itemize}
\item \underline{The two} argued \underline{that only a new board would have had the credibility} \\
\underline{to restore El Paso to health.} \\
\underline{The two} believed \underline{that only a new board would have had the credibility} \\
\underline{to restore El Paso to health.}
\end{itemize}

\subsubsection{Non-paraphrase}

\textbf{Definition: } Non-paraphrase includes fragments which do not have the same meaning (a), as
well as cases in which we need extralinguistic information in order to establish a
link between the members of the paraphrase pair: cases of same ilocutive value but
different meaning (b), cases of subjectivity (c), cases of potential coreference (d),
(e) and (f), etc.

\begin{itemize}
\item[a)]The two had argued that \underline{you shouldn't go there} .\\
He and Zilkha believed that \underline{this is unfair} .
\item[b)] I want some fresh air. \\
Could you open the window?
\item[c)] The U.S.-led \underline{invasion} of Iraq . \\
The U.S.-led \underline{liberation} of Iraq.
\item[d)] They got married \underline{last year} . \\
They got married \underline{in 2004} .
\item[e)] I live \underline{here} .
I live \underline{in Barcelona} .
\item[f)] They will come \underline{later} .\\
They will come \underline{this afternoon}
\end{itemize}

\textbf{N.B.: }Paraphrase and coreference overlap considerably. Those cases that
may corefer, but at the same time are paraphrases, should be annotated as
paraphrases.

In cases (d), (e) and (f), the linguistic information is not enough to link the
two members of the pair, we need to know which point in the time or in the
space are we taking as reference. Thus, they are annotated as nonparaphrases.
Cases in (g), (h) and (i) can be linked only through linguistic information (a
year in the past, a 'city' type of entity, a masculine singular entity,
respectively). Thus, they are annotated as paraphrases.

\begin{itemize}
\item[g)] They got married \underline{last year} .\\ 
They got married \underline{a year ago} .
\item[h)] I live in \underline{Barcelona} . \\
I live \underline{in a city} .
\item[i)] I love \underline{John} . \\
I love \underline{him} .
\end{itemize}

\textbf{N.B.: } Although sometimes a non-paraphrase fragment may actually affect the
meaning of the full sentence, only the fragment in question will be tagged as
NON-PARAPHRASE (j) and the rest of the sentence will be annotated
independently of this fact.

\begin{itemize}
\item[j)] \underline{Mike and Lucy} decided to leave . \\
\underline{Mark} decided to leave .
\end{itemize}

\textbf{N.B.:} When two linguistic units having a different meaning are not aligned
formally nor informatively, they should be tagged as two different
ADDITION/ DELETION cases (1 and 2 in k), not as NON-PARAPHRASES.

\begin{itemize}
\item[k)] \underline{Yesterday,}\textsubscript{1} Google failed . \\
Google failed \underline{because of the crisis}\textsubscript{2}.
\end{itemize}

\subsubsection{Entailment}

\textbf{Definition: } Fragments having an entailment relation. \textbf{N.B.:} It should be noted that 
entailment relations are present in many paraphrase types (e.g. general/specific in SAME-POLARITY 
or ADDITION/DELETION). We will only use the ENTAILMENT tag when there is a substantial difference
in the information content. Entailment is always \textbf{negative sense preserving}.

\begin{itemize}
\item Google \underline{was in talks to buy} Youtube . \\ 
Google \underline{bought} Youtube
\end{itemize}

\section{Annotating non-paraphrases}\label{a:etpc:np}

Annotating non-paraphrases (negative examples of paraphrasing in the MRPC corpus) is a non-trivial 
task that has not been carried out for other paraphrase typology corpora. The non-paraphrases in the 
MRPC corpus have many of the properties of paraphrases, they have a very high degree of lexical and 
syntactic similarity. In a) we can see an example of a non-paraphrase pair. The two sentences talk about 
the same NEs (Yucaipa and Dominick) in the same syntactic-semantic roles of the same actions (buying, 
selling, owning). At the same time, there are key differences between the two sentences -- the price of 
the sale in the first sentence is \$2.5 billion, while in the second it is \$1.8 billion. 

\begin{itemize}
\item[a)] Yucaipa owned Dominick's before selling the chain to Safeway in 1998 for \$2.5 billion. \\
Yucaipa bought Dominick's in 1995 for \$693 million and sold it to Safeway for \$1.8 billion in 1998.
\end{itemize}

Due to the complex nature of the non-paraphrasing, the annotation of these pairs goes in three steps

\begin{itemize}
\item[1)] (Re)evaluation of the paraphrasing or non-paraphrasing relation between the two sentences 
as a whole (this is the first step for both paraphrases and non-paraphrases).
\item[2)] (After the pair has been annotated as non-paraphrases) Annotation of the non-sense-preserving 
phenomena, responsible for the non-paraphrasing label of the pair.
\item[3)] Annotation of the sense-preserving phenomena, responsible for the high degree of similarity 
between the two sentences.
\end{itemize}

An example annotation of the pair in a) follows:

\begin{itemize}
\item[1)] The relation between the two sentences is non-paraphrases
\item[2)] The non-sense-preserving phenomena responsible for the ``non-paraphrase'' label of the pair 
is ``Lexical Substitution (Named Entities)'': \\
Yucaipa owned Dominick's before selling the chain to Safeway in 1998 for \underline{\$2.5 billion} .\\
Yucaipa bought Dominick's in 1995 for \$693 million and sold it to Safeway for \underline{\$1.8 billion} in 1998.

\item[3)] The sense-preserving phenomena, responsible for the high degree of similarity are:
\begin{itemize}
\item[a.] Same polarity substitution (contextual) \\
Yucaipa owned Dominick's before selling \underline{the chain} to Safeway in 1998 for \$2.5 billion. \\
Yucaipa bought Dominick's in 1995 for \$693 million and sold \underline{it} to Safeway for \$1.8 billion in 1998. 
\item[b.] Entailment\\
Yucaipa \underline{owned} Dominick's before selling the chain to Safeway in 1998 for \$2.5 billion. \\
Yucaipa \underline{bought} Dominick's in 1995 for \$693 million and sold it to Safeway for \$1.8 billion in 1998.

\item[c.] Inflectional changes \\
Yucaipa owned Dominick's before \underline{selling} the chain to Safeway in 1998 for \$2.5 billion. \\
Yucaipa bought Dominick's in 1995 for \$693 million and \underline{sold} it to Safeway for \$1.8 billion in 1998.

\item[d.] Order \\
Yucaipa owned Dominick's before selling the chain to Safeway \underline{in 1998} for \$2.5 billion. \\
Yucaipa bought Dominick's in 1995 for \$693 million and sold it to Safeway for \$1.8 billion \underline{in 1998}.

\item[e.] Addition/Deletion \\
Yucaipa owned Dominick's \underline{before}\textsubscript{1} selling the chain to Safeway in 1998 for \$2.5 billion. \\
Yucaipa bought Dominick's \underline{in 1995 for \$693 million and}\textsubscript{2} sold it to Safeway for \$1.8 billion in 1998.

\item[f.] Identity \\
\underline{Yucaipa} owned \underline{Dominick's} before selling the chain \underline{to Safeway} in 1998 \underline{for} \$2.5 billion \underline{.}
\underline{Yucaipa} bought \underline{Dominick's} in 1995 for \$693 million and sold it \underline{to Safeway for} \$1.8 billion in 1998 \underline{.}

\end{itemize}

\end{itemize}

\section{Annotating negation}\label{a:etpc:neg}

Annotating negation within paraphrases is a novel approach. For the pilot annotation we will mark
the scope as negation and the negation cue as a ``key''.

\begin{itemize}
\item \underline{We did \textbf{not} drive up to the door} but got down near the gate of the avenue .
\end{itemize}

\chapter{Annotation Guidelines for \citet{gold-etal-2019-annotating}}\label{a_law}

\chaptermark{Annotation Guidelines (Gold et al., 2019)}

In this task each text pair is annotated independently for Paraphrasing, Textual Entailment, Contradiction,
Textual Specificity, and Textual Similarity. At each annotation step, annotators are asked
to determine the presense or absense of a single textual meaning relation. For Textual 
Entailment and Textual Specificity, each pair is shown twice, with the order of the texts 
changed to address the directionality of the relations. The instructions provided to the
annotators are the following.

\section{Paraphrasing}

\textbf{Background:} We want to study the meaning relation between two texts. Thus you 
are asked to determine whether the two sentences mean (approximately) the same or not.

\textbf{Task:} In this task you are presented with \textbf{two sentences}. You are required to decide whether the two
sentences \textbf{have approximately the same meaning or not}.

In the case of pronouns (he, she, it, mine, his, our, ...) being used, you can assume they reference
proper names, if your common sense does not suggest otherwise (e.g. ``Linda'' is a female name and
can be referenced by ``she, her, ...'', but not ``he, his, ...'').

\textbf{Examples of the choce ``approximately the same meaning'':}

\begin{itemize}
	\item John goes to work every day with the metro.
	\item He takes the metro to work every day.
	\item [] \textit{In the content of the task, we assume that ``He'' and ``John'' are the same person.}

	\item Mary sold her Toyota to Jeanne.
	\item Jeanne bought her Toyota from Mary Smith.
	\item [] \textit{In the content of the task, we assume that ``Mary Smith'' and ``Mary'' are the same person.}
\end{itemize}

\textbf{Examples of the choice of ``not the same meaning'':}

\begin{itemize}
	\item Mary sold her Toyota to Jeanne.
	\item Mary had a blue Toyota.
	\item [] \textit{The two texts are related, but are not the same.}
	\item John Smith takes the metro to work every day.
	\item John works from home every Tuesday.
	\item [] \textit{The two texts are not closely related except for the person (John).}
\end{itemize}

\section{Textual Entailment}

\textbf{Background:} We want to research causal relationships between sentences, which will help in information retrieval
or summarization tasks. Thus, you are asked to determine whether given that the first sentence is
true, the second sentence is also true.

\textbf{Task:} In this task, you are presented with \textbf{two sentences}. You are required to decide whether \textbf{if Sentence
1 is true, this also makes Sentence 2 true}.

In the case of pronouns (he, she, it, mine, his, our, ...) being used, you can assume they reference
proper names, if your common sense does not suggest otherwise (e.g. ``Linda'' is a female name and
can be referenced by ``she, her, ...'', but not ``he, his, ...'').

\textbf{Examples for the option ``Sentence 1 causes Sentence 2 to be true'':}

\begin{itemize}
	\item[] \textit{In that case, the first sentence causes the second sentence to be true, as assuming that John bought
a car, it means that he has a car now.}
	\item John bought a car from Mike.
	\item John has a car.
	\item[] \textit{In that case, the first sentence causes the second sentence to be true, as the first sentence says
that both boys and girls play games, it also contains the information that boys play games.}
	\item Boys and girls play games.
	\item Boys play games.
\end{itemize}

\textbf{Examples for the option ``Sentence 1 does not cause Sentence 2 to be true'':}

\begin{itemize}
	\item[] \textit{If the second sentence makes the first sentence true (but the first doesn't make the second
true), choose the option ``Sentence 1 \textbf{does not} cause Sentence 2 to be true'':}
	\item John has a car.
	\item John bought a car from Mike.
	\item[] \textit{If you cannot tell if the first sentence causes the second sentence to be true, choose the option
``Sentence 1 \textbf{does not} cause Sentence 2 to be true'':}
	\item He works as a teacher in Peru.
	\item He is an English teacher.
\end{itemize}

\section{Contradiction}

\textbf{Background:} We want to study the meaning relation between two texts. Thus you are asked to determine whether
the two sentences contradict each other.

\textbf{Task: }In this task you are presented with \textbf{two sentences}. You are required to decide whether the \textbf{two
sentences contradict each other}. Two contradicting sentences can't be true at the same time.

In the case of pronouns (he, she, it, mine, his, our, ...) being used, you can assume they reference
proper names, if your common sense does not suggest otherwise (e.g. ``Linda'' is a female name and
can be referenced by ``she, her, ...'', but not ``he, his, ...'').

\textbf{Examples for the option ``the sentences contradict each other'':}

\begin{itemize}
	\item John bought a new house near the beach.
	\item John didn't buy the house near the beach.
	\item[] \textit{The second sentence directly contradicts the first one they can't both be true.}
	\item Mary is on a vacation in Florida.
	\item Mary is at the office, working.
	\item[] \textit{The two sentences can't be true at the same time Mary is either on vacation in Florida, or at the office. She can't be in two places.}
\end{itemize}

\textbf{Examples for the option ``the sentences do not contradict each other'':}

\begin{itemize}
	\item Mary is on vacation in Florida.
	\item John is at the office.
	\item[] \textit{John and Mary are two different persons. There is no contradiction. Both statements can be true.}
\end{itemize}

\section{Similarity}

\textbf{Background:} We want to study the meaning relation between two texts. Thus you are asked to determine how
similar two texts are.

\textbf{Task: }In this task you are presented with \textbf{two sentences}. You are required to decide \textbf{how similar the
two sentences are on a scale from 0 (completely dissimilar) to 5 (identical)}.

In the case of pronouns (he, she, it, mine, his, our, ...) being used, you can assume they reference
proper names, if your common sense does not suggest otherwise (e.g. ``Linda'' is a female name and
can be referenced by ``she, her, ...'', but not ``he, his, ...'').

\textbf{Example for Similarity 0:}

\begin{itemize}
	\item John goes to work every day with the metro.
	\item The kids are playing baseball on the field.
	\item[] \textit{The two texts are completely dissimilar.}
\end{itemize}

\textbf{Example for Similarity 1-2:}

\begin{itemize}
	\item John goes to work every day with the metro.
	\item John sold his Toyota to Sam.
	\item[] \textit{The two texts have some common elements, but are overall not very similar.}
\end{itemize}

\textbf{Example for Similarity 3-4:}

\begin{itemize}
	\item Mary is writing the report on her Lenovo laptop.
	\item Mary has a Lenovo laptop.
	\item[] \textit{The two texts have a lot in common, but also have differences.}
\end{itemize}

\textbf{Example for Similarity 5:}

\begin{itemize}
	\item Mary was feeling blue.
	\item Mary was sad.
	\item[] \textit{The two texts are (almost) identical.}
\end{itemize}

\section{Specificity}

\textbf{Background:} We want to research whether displaying more specific sentences is helpful in information retrieval or
summarization tasks. Thus, you are asked to determine whether the 1st sentence is more specific
than the 2nd. The specificity of sentence is defined as a measure of how broad or specific its
information level is.

\textbf{Task: }In this task, you are presented with \textbf{two sentences}. You are required to decide whether \textbf{the 1st
sentence IS more specific than the 2nd}. If this is not the case, choose the option \textbf{the 1st
sentence IS NOT more specific than the 2nd}.

\textbf{Examples for the option ``Sentence 1 IS more specific''}

\begin{itemize}
	\item I like cats.
	\item I like animals.
	\item[] \textit{As the 1st sentence gives the more specific information on which animal is liked, it is more specific.
	Hence, you have to choose the option that the 1st sentence is more specific.}
	\item The cute cafe has great coffee.
	\item The cafe sells coffee.
	\item[] \textit{As the 1st sentence gives the more specific information on both the cafe and the coffee, it is more 
	specific. Hence, you have to choose the option that the 1st sentence is more specific.}
\end{itemize}

\textbf{Examples for the option ``Sentence 1 IS NOT more specific''}

\begin{itemize}
	\item I like animals.
	\item I like cats.
	\item[] \textit{As the 2nd sentence gives the more specific information on which animal is liked, it is more specific.
	Hence, you have to choose the option that the 1st sentence is not more specific.}
	\item I like dogs.
	\item I like cats.
	\item[] \textit{Now, as in both cases the liked animal is mentioned, they have the same level of specificity. Hence,
	you have to choose the option that the 1st sentence is not more specific.}
	\item I like black dogs.
	\item He saw a blind cat.
	\item[] \textit{Now, as the information is very diverse, it is impossible to say which sentence is more specific.
	Hence, you have to choose the option that the 1st sentence is not more specific.}
\end{itemize}

\chapter{Annotation Guidelines for \citet{mtype_lrec}}\label{a_sharel}

\chaptermark{Annotation Guidelines (SHARel)}

\section{Presentation}

This document sets out the guidelines for the annotation of atomic types using the Extended Typology for Relations. 
The task consists of annotating pairs of text that hold a textual semantic relation (paraphrasing, entailment, contradiction,
similarity) with a textual label, and the atomic phenomena they contain. These guidelines have been used to annotate 
the ETRC corpus. For the purpose of the annotation, the WARP-Text annotation tool has been used. 

\textbf{N.B.:} The task definition, tagset definition and annotation of linguistic phenomena in these Guidelines overlap
with those for the ETPC corpus. The reader is encouraged to consult the ETPC guidelines presented in Appendix \ref{a_etpc}
or the full SHARel guidelines available online. Here I only provide the guidelines for the reason-based types.

\section{Annotating reason-based Phenomena}

Reason-based phenomena account for relations that cannot be expressed and processed using only linguistic knowledge. 
Like the linguistic phenomena, the reason-based phenomena can be sense-preserving or non-sense preserving. Our goals 
with the annotation of reason-based phenomena are twofold: 

\begin{itemize}
\item[1)] we want to make a precise and explicit annotation of the units involved in the inference
\item[2)] we want to determine the kind of reason-based and background knowledge required. 
\end{itemize}

1a and 1b show an example of an ``existential'' reason-based -- ``speaking X'' entails ``X exists''. 
2a and 2b show an example of ``causal'' reason-based -- ``X is written in Y (language)'' entails 
``reading X requires Y (language)'' .

\begin{itemize}
\item[1a] Speaking more than one language is imperative today.	
\item[1b] There is more than one language.
\item[2a] Reading the Bible requires studying Latin.	
\item[2b] The Bible is written in Latin.
\end{itemize}

When annotating reason-based phenomena, there are several important things:

\begin{itemize}
\item Annotate all possible phenomena separately. The aim is to annotate every token that is not already annotated 
as linguistic or addition-deletion.
\item The scope of some phenomena can overlap. That means some tokens may be part of multiple scopes.
\item When choosing the scope, we choose the smallest scope possible. Unlike the sense-preserving, in this part of 
the annotation, the goal is to choose the most specific scope possible. For example, in 1a and 1b we could annotate 
``Speaking more than one language'' and ``There is more than one language'', but in order to be as specific as 
possible, we choose to only annotate ``Speaking'' and ``There is''.
\item When choosing the scope, if possible, try to annotate whole linguistic units without breaking them. For example 
in 2a and 2b, we could only annotate ``Reading requires'' and ``is written in''
\item Like in the linguistic phenomena -- the sense preserving reason-based phenomena need not relate units that 
have similar syntactic or semantic role; however, the non-sense preserving reason-based phenomena must relate 
units that have similar syntactic or semantic role.
\end{itemize}

\section{List of reason-based phenomena}

\textbf{1. Cause and Effect:} T causes H to be true [neg. sense preserving: T causes H to be FALSE]

\begin{itemize}
\item[a.] ``Once a person is welcomed into an organization, they belong to that organization''
\end{itemize}

\textbf{2. Conditions and Properties:} A very general type where H containing facts (and properties) 
implied by T [neg. sense preserving: H contains facts and properties that contradict the implied 
from T (ex.: ``There is only one language'')]

\begin{itemize}
\item[a.] Existential -- T entails H exists (pre-requirement)
\item[b.] ``To become a naturalized citizen, one must not have been born there'' (pre-requirement)
\item[c.] ``The type of thing that adopts children is person'' (argument type)
\item[d.] ``When a person is an employee, that organization pays his salary'' (simultaneous conditions)
\end{itemize}

\textbf{3. Functionality:} Relationships which are functional [neg. sense preserving: mutual exclusivity 
-- types of things that do not participate in the same relationship]

\begin{itemize}
\item[a.] ``A person can only have one father (or two arms)'' (+)
\item[b.] ``Government and media sectors usually do not employ the same person'' (-)
\end{itemize}

\textbf{4. Transitivity:} If R is transitive and R(a,b) and R(b,c) are true, then R(a,c)

\begin{itemize}
\item[a.] ``The ``support'' is transitive. If Putin supports United Russia party, and United Russia party supports Medvedev, then Putin supports Medvedev''
\end{itemize}

\textbf{5. Numerical Reasoning}

\textbf{6. Named Entity Reasoning:} reasoning that goes beyond substitution; relations between multiple 
entities (i.e. not just Trump -- president, but rather Trump -- Clinton)

\textbf{7. Temporal and Spatial Reasoning}

\textbf{8. Other (World Knowledge)}

\end{document}